\documentclass{article}

\usepackage[table]{xcolor}
\usepackage{tabularx}
\usepackage{float}
\usepackage{graphicx}
\usepackage{soul}
\usepackage{listings}
\usepackage{subfigure}

\definecolor{myyellow}{rgb}{1,0.6,0}
\definecolor{myblue}{rgb}{0.0,0.2,0.8}
\definecolor{mygreen}{rgb}{0.2,0.8,0.4}
\definecolor{mymagenta}{rgb}{0.3,0.1,0.4}
\definecolor{mygrey}{rgb}{0.75,0.75,0.75}
\definecolor{myred}{rgb}{1,0.2,0.2}
\definecolor{myhotpink}{rgb}{1, 0.66, 0.867} 

\usepackage[final]{neurips_2021}

\usepackage[utf8]{inputenc} 
\usepackage[T1]{fontenc}    
\usepackage{hyperref}       
\usepackage{url}            
\usepackage{booktabs}       
\usepackage{amsfonts}       
\usepackage{nicefrac}       
\usepackage{microtype}      
\usepackage{xcolor}         

\newcommand{\nummodels}{38 }

\title{Progress and limitations of deep networks to recognize objects in unusual poses}

\author{Amro Abbas\\
  The African Institute For Mathematical Sciences\\
  \texttt{afagiri@aimsammi.org} \\
 \And
 Stéphane Deny \\
 Aalto University\\
  \texttt{stephane.deny@aalto.fi} \\
}

\begin{document}

\maketitle


\begin{abstract}
Deep networks should be robust to rare events if they are to be successfully deployed in high-stakes real-world applications (e.g., self-driving cars). Here we study the capability of deep networks to recognize objects in unusual poses. We create a synthetic dataset of images of objects in unusual orientations, and evaluate the robustness of a collection of \nummodels recent and competitive deep networks for image classification. We show that classifying these images is still a challenge for all networks tested, with an average accuracy drop of 29.5\% compared to when the objects are presented upright. This brittleness is largely unaffected by various network design choices, such as training losses (e.g., supervised vs. self-supervised), architectures (e.g., convolutional networks vs. transformers), dataset modalities (e.g., images vs. image-text pairs), and data-augmentation schemes. However, networks trained on very large datasets substantially outperform others, with the best network tested---Noisy Student EfficentNet-L2 trained on JFT-300M---showing a relatively small accuracy drop of only 14.5\% on unusual poses. Nevertheless, a visual inspection of the failures of Noisy Student reveals a remaining gap in robustness with the human visual system. Furthermore, combining multiple object transformations---3D-rotations and scaling---further degrades the performance of all networks. Altogether, our results provide another measurement of the robustness of deep networks that is important to consider when using them in the real world. Code and datasets are available at https://github.com/amro-kamal/ObjectPose.

\end{abstract}


\section{Introduction}

A considerable effort has been made in the past decade to develop deep neural networks that could see as well as humans. These networks, when they are tested on in-distribution test sets, achieve or even exceed human-level performance. For example, FlorenceNet \citep{yuan2021florence} and Meta Pseudo Labels \citep{pham2021metapseudolabels} achieve less than 2\% top-5 error on ImageNet, outperforming an expert human with 5.1\% top-5 error \citep{russakovsky2015imagenet}. However, in real-world applications, deep networks are often deployed and used to detect harder examples than those seen in the development test set. This has led researchers to investigate the performance of networks using more challenging test examples, in the so-called out-of-distribution (OOD) regime \citep{naturaladvexamples,imagenetc,imagenetr,imagenetsketch,dodge2017study,imagnetv2,GeneralizationAcrossTime,geirhos2018generalisation,closingthegap}. Many of the previous studies on out-of-distribution generalization have focused on measuring the generalization capabilities of networks to distorted images. Notably, \cite{closingthegap} show that the newest generation of very large deep networks is closing the human-machine robustness gap on 17 different out-of-distribution image distortion types. However, most of the transformations used to generate these datasets are local distortions that only affect the texture of objects, but do not change the global structure of the image. \cite{taori2020measuring} shows that robustness to these kinds of distortions does not transfer entirely to natural shifts and does not represent a comprehensive measure of the network's robustness. 

\begin{figure}[t]
    \centering
    \includegraphics[width=0.90\textwidth]{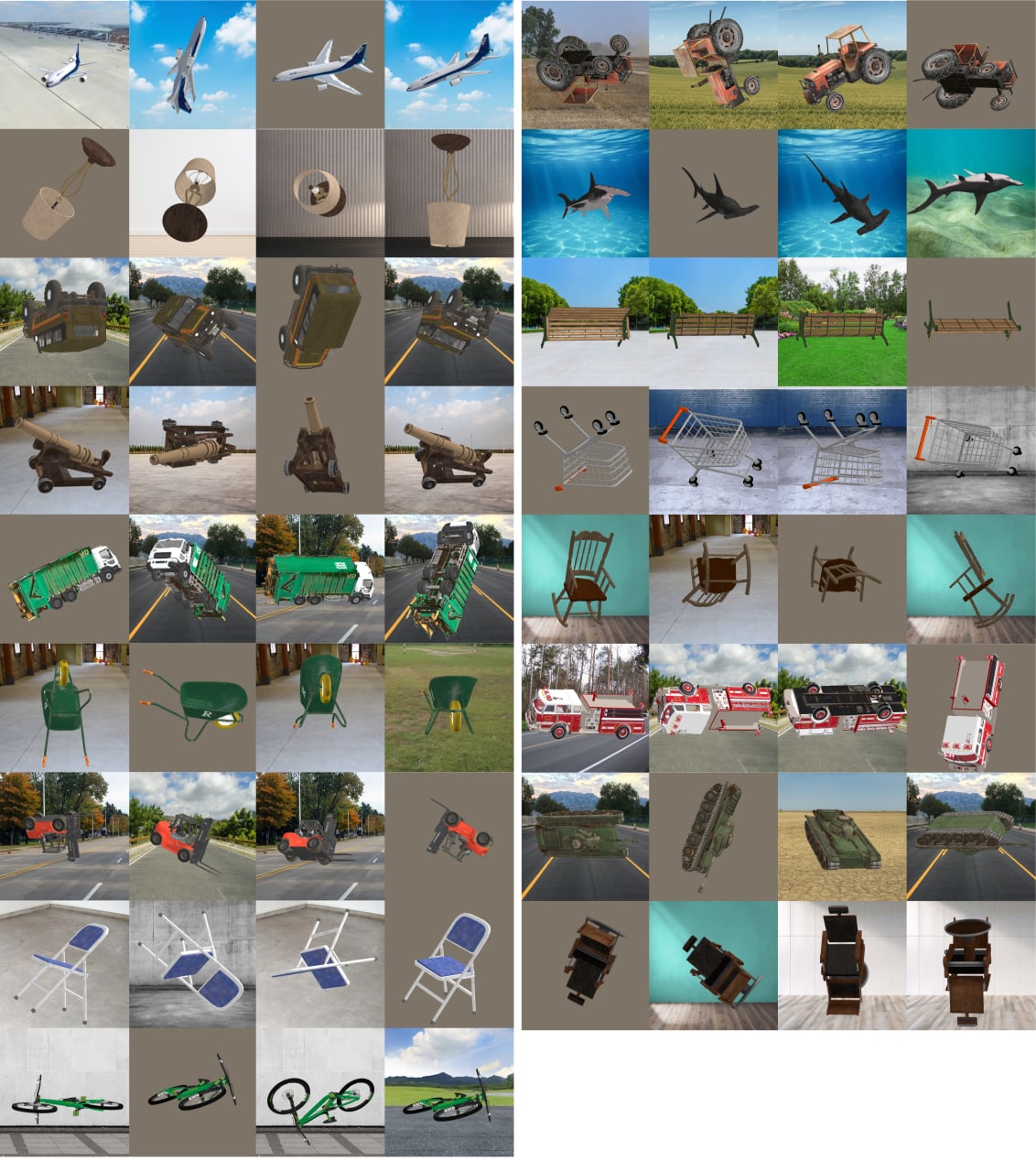}
    \caption{Random samples from our custom dataset ObjectPose, designed to probe deep networks' robustness to unusual poses. For each of the 17 objects composing the dataset, background and orientation are systematically varied to generate the full dataset. The 17 object categories in ObjectPose are `airliner', `barber chair', `cannon', `fire engine', `folding chair', `forklift', `garbage truck', `hammerhead shark', `jeep', `mountain bike', `park bench', `rocking chair', `shopping cart', `table lamp', `tank', `tractor', and `wheel barrow'.}
    \label{fig:OP17}
\end{figure}

Pose transformations (e.g., a bus seen upside-down) represent an interesting case-study for networks' robustness, as (1) unlike simple distortions, these transformations affect the global structure of the image, and (2) it would be technically challenging to augment an image dataset at scale with this type of 3D transformations. In a rare study of its kind, \cite{strikewithapose} show that two deep networks, Inception-V3 \citep{inceptionv3} and ResNet50 \citep{resnet}, drastically fail to recognize objects in unusual poses, incorrectly classifying most of the poses explored. Later, \cite{madan_small_2021} also reveal a brittleness of ResNet18 and CLIP \citep{clip} to small changes in pose, in an adversarial setting.\\

In this study, we revisit the question of networks' robustness to unusual poses using a diverse set of the latest and best publicly available networks for image classification. We test \nummodels networks with a variety of different architectures, sizes, training datasets, and training objectives on a custom dataset of images of objects in unusual poses. Our contributions are:
\begin{itemize}
    \item We observe that, on unusual poses, the networks of our collection suffer from a 14.5\% to 45.5\% accuracy drop compared to usual poses. A visual inspection of networks' failures reveal that, even for the best models, a robustness gap remains with the human visual system.
    \item We provide a detailed study of the effect of in-(image)-plane vs. out-of-plane object rotations, background-foreground congruency, and rotation angle on performance. We show that the best networks rely on a different strategy than weaker networks when it comes to incorporating information from the background.
    \item By combining multiple unusual transformations---such as object rotations and scaling---we show that such combinations lead to further performance degradation for all networks, as predicted by a combinatorial model of error.
    \item In an effort to go beyond synthetic datasets, we test the networks on images from the \emph{Common Objects in 3D Dataset} (CO3D), a dataset of objects seen in various poses, and show a generalization gap of 5.2\% on average across networks compared to a benchmark of objects presented in their usual pose, ImageNetV2. 
    \item We make our custom datasets, deep network collection and code available publicly on Github\footnote{https://github.com/amro-kamal/ObjectPose} for future investigations by the community.
\end{itemize}


\section{Dataset and Networks}

\paragraph{ObjectPose Dataset} We generated a synthetic dataset of objects in unusual poses, ObjectPose (Fig. \ref{fig:OP17}, all details in Appendix \ref{sec:datasetdistcription}). The dataset contains 27,540 images of 17 high-quality 3D objects---each belonging to one of 1000 ImageNet \citep{russakovsky2015imagenet} classes---rendered in a range of different orientations and over different background images, following the pipeline of \citep{strikewithapose}. Briefly, the object is first placed in an initial upright position. We then choose one of the \emph{YAW}, \emph{ROLL}, or \emph{PITCH} axes to rotate the object along it, and render it on top of a background image. We used three different background images with each object. Two of the backgrounds are images chosen manually from the internet to match the object's usual context and not to contain any other ImageNet object. We chose the third background to be grey with all its RGB pixel values equal to (0.485, 0.456, 0,406), corresponding to the average pixel color of ImageNet images. In order to focus on robustness to unusual poses, each object is chosen carefully so that the resulting images for that object are correctly classified with >90\% accuracy by a ResNet-50 when the object's orientation is less than 10° apart from its upright pose. We gather these images where the object is rotated by -10° to 10° only in the  ObjectPose +-10 dataset (1,683 images in total), and exclude them from ObjectPose, such that ObjectPose only contains unusal poses (11° to 349° from the upright pose).

\textbf{Deep Networks} We tested a collection of \nummodels networks on ObjectPose (all details in Appendix \ref{appendix:modelsdescription} and suppl. table \ref{table:modelstable}). We chose a diverse set of networks with different architectures, training datasets sizes, number of parameters, and training objectives. The networks have varying number of parameters (from 22M up to 645M), and different architectures including convolutional neural networks (CNNs) \citep{resnet,noisystudent,bit,swsl,convnext}, Vision Transformers (ViTs) \citep{vit,vitsam,swin} \citep{beit,deit}, and MLP-Mixers \citep{mlpmixer}. We also include ConViT \citep{convit}, a hybrid CNN-ViT architecture that adds a convolutional inductive bias to the Vision Transformer. \\
We chose networks trained under different objectives, including:
\begin{itemize}
\item Supervised learning, including convolutional architectures, such as \citep{resnet,noisystudent,convnext,bit}, Vision Transformers, such as \citep{vit,vitsam,beit,deit,swin,convnext}, and MLP-mixer\citep{mlpmixer},
\item Self-supervised learning, such as SimCLR \citep{simclr}, and BEiT \citep{beit}, 
\item Semi-weakly supervised learning, such as SWSL-ResNet50 and SWSL-ResNeXt101 \citep{swsl}, and weakly supervised learning, such as SWAG \citep{swagnet},
\item Text supervision, such as CLIP \citep{clip}. 
\end{itemize}

The networks were trained on datasets with different sizes ranging from 1M to 3.6B images. Among the networks we use, Noisy Student EfficientNet \citep{noisystudent} (300M images) and SWAG \citep{swagnet} (3.6B images) are the only networks pretrained on extremely large datasets and fine-tuned on ImageNet. Although CLIP \citep{clip} was pretrained on a very large dataset (400M image-caption examples), it was not fine-tuned on ImageNet.\\
The best network of our collection according to its performance on ImageNet is the SWAG-RegNetY-128GF-384 model \citep{swagnet}, pretrained with a weakly-supervised learning approach on 3.6B Instagram images (IG). It achieves 88.55\% ImageNet top-1 accuracy. \\
We also include networks trained using special optimizers such as the ViT-B16-SAM model \citep{vitsam}, a vision transformer trained with the recently proposed sharpness-aware minimization optimizer (SAM) \citep{samoptimizer}, which is designed to improve generalization and robustness. 

\section{Results} \label{sec:results}

\begin{figure}[t]

    \centering
    \subfigure{
    \hspace*{0.1cm}
    \includegraphics[width=0.44\textwidth]{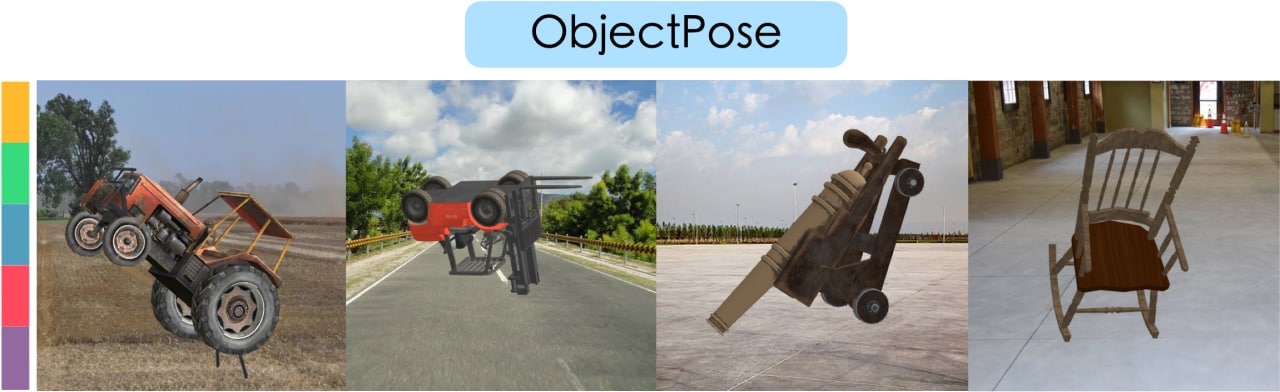}}
    \subfigure{
    \hspace*{0.5cm} \includegraphics[width=0.44\textwidth]{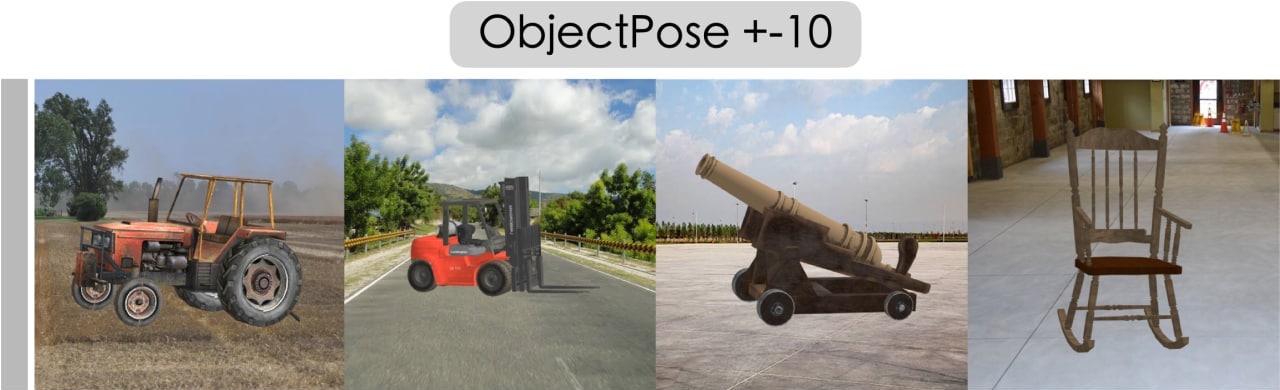}}
    \hspace*{-0.5cm}
    $\vcenter{\hbox{\includegraphics[width=.64\textwidth]{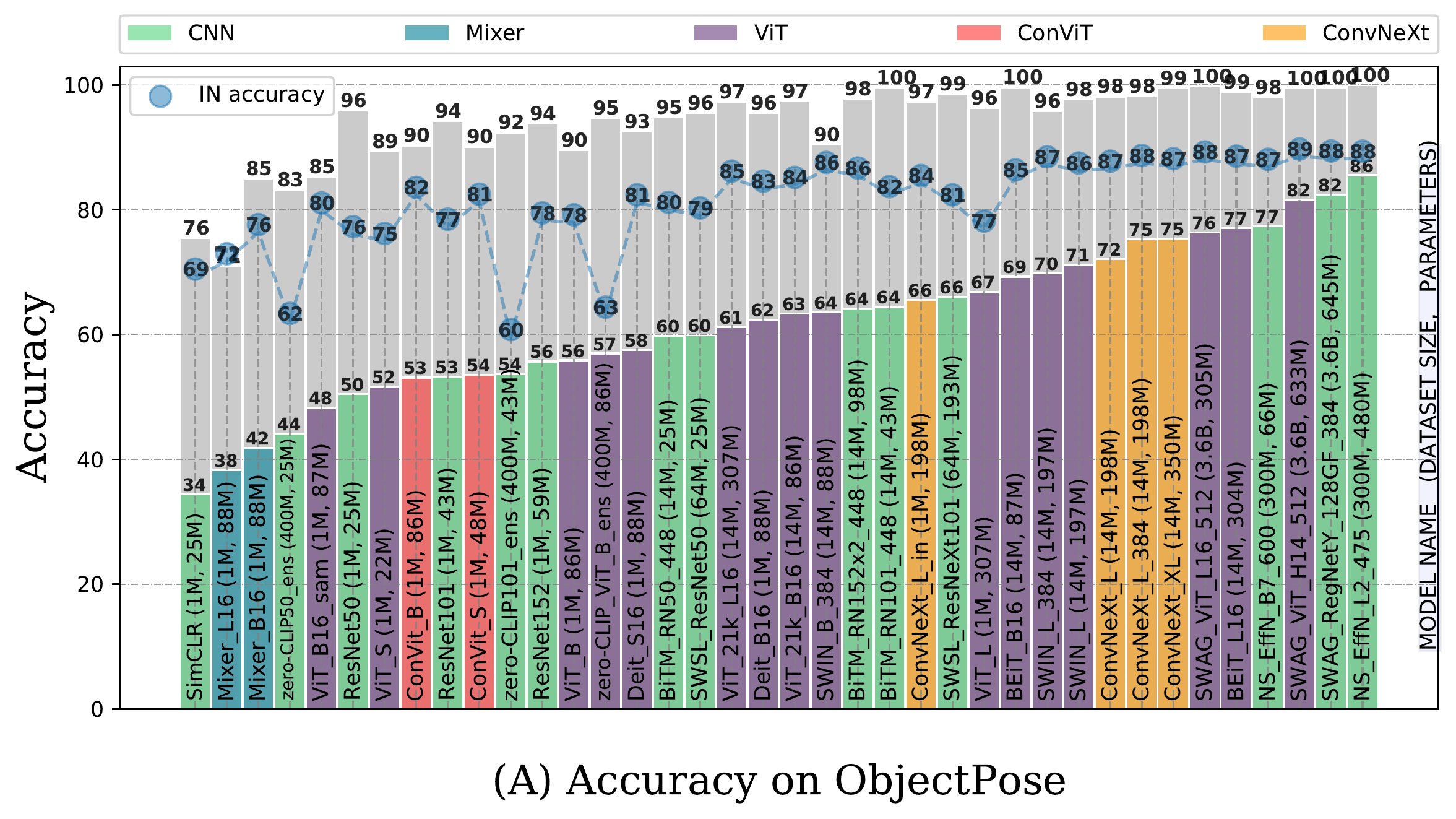}}}$
    $\vcenter{\hbox{\includegraphics[width=.341\textwidth]{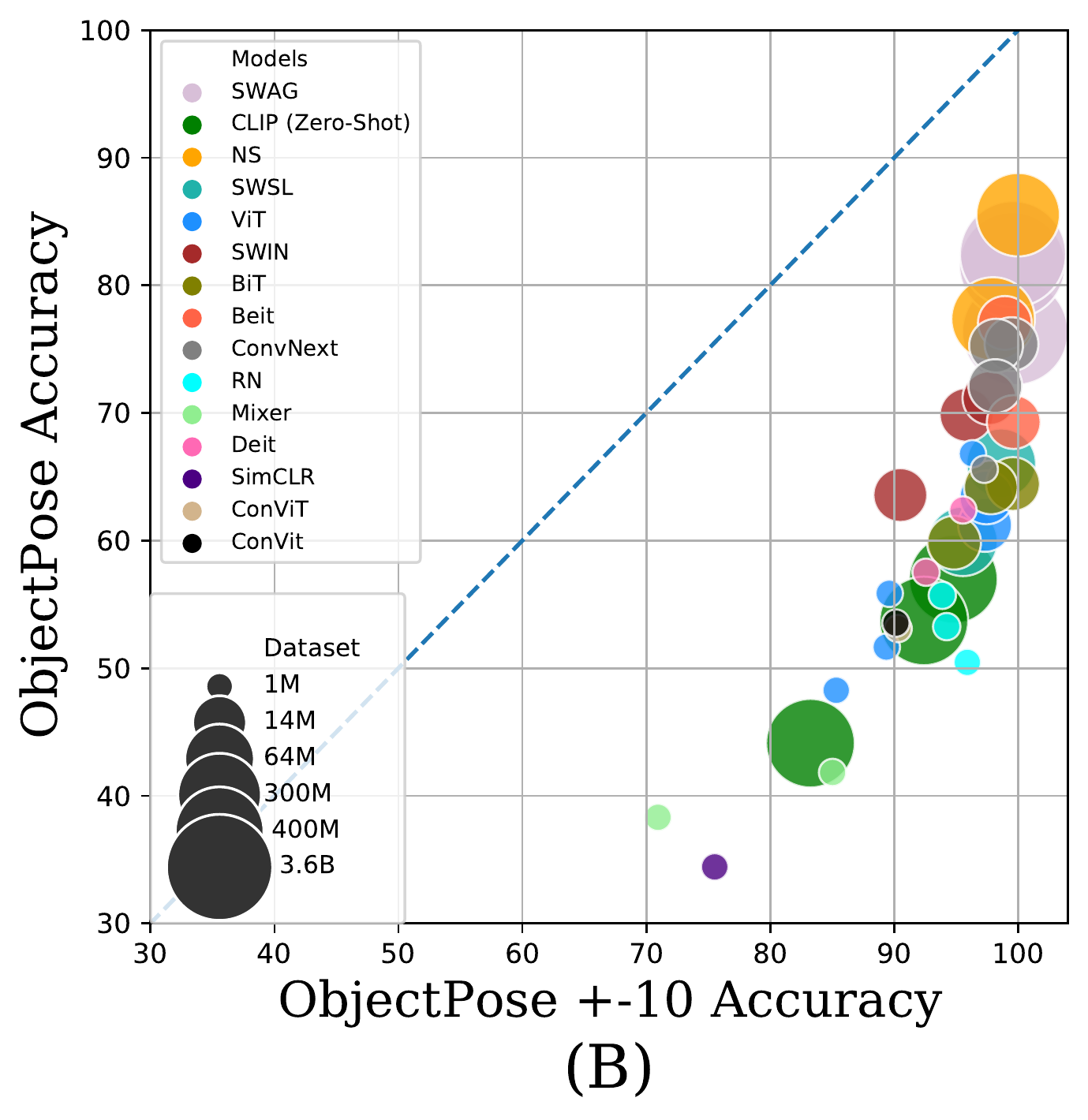}}}$
    \caption{\textbf{Performance of all networks on ObjectPose. (A)} Rotating the objects in unusual poses (ObjectPose, colored bars) induces a top-1 accuracy drop of 14.5\%-45.5\% compared to when objects are presented upright (ObjectPose +-10, grey bars). Bar colors indicate different architectures. Blue dots: Top-1 accuracy on ImageNet (as reported in the papers). \textbf{(B) Accuracy on usual vs. unusual poses}. Networks trained on ImageNet1k cluster together with low accuracy on ObjectPose. Networks trained on ImageNet21k perform better, and networks trained on extremely large datasets---Noisy Student models (300M images) and SWAG (3.6B images)---perform the best, with the exception of CLIP models (400M images), which were not fine-tuned on ImageNet categories.}
    \label{fig:ObjectPose_accuracy}
\end{figure}

\begin{figure}[t]
    \centering
    \includegraphics[width=0.245\textwidth]{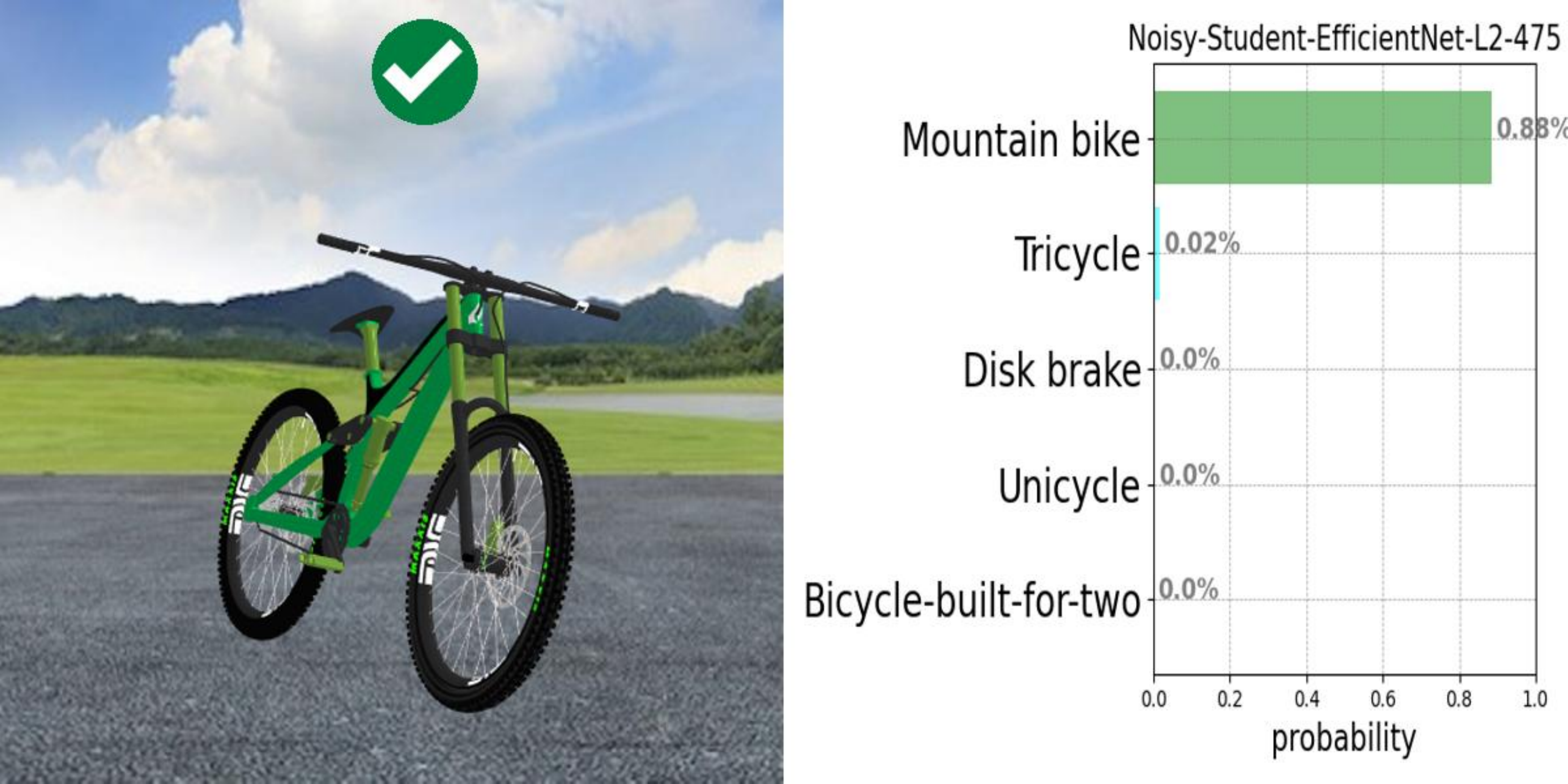}
    \includegraphics[width=0.245\textwidth]{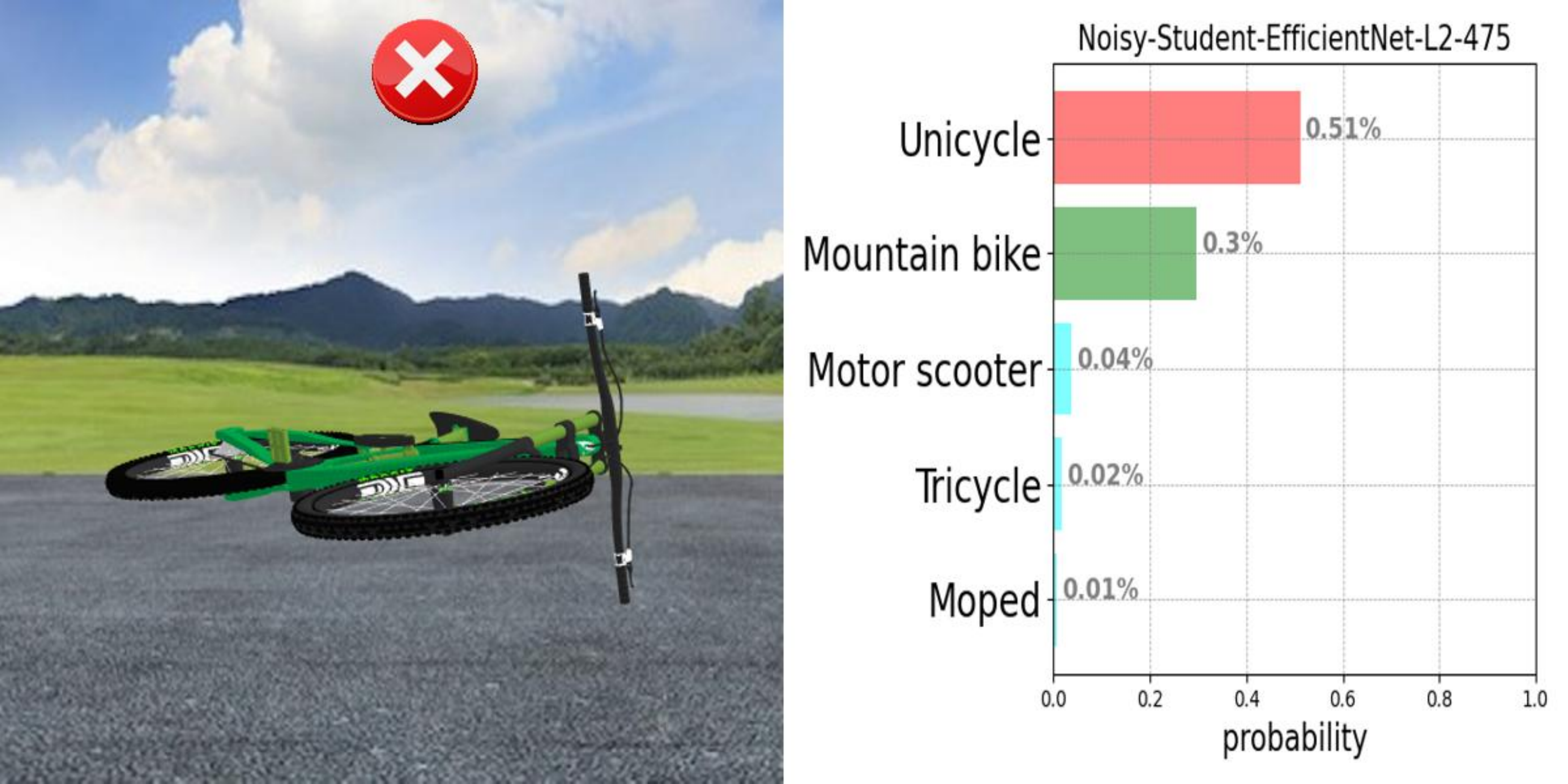}
    \includegraphics[width=0.245\textwidth]{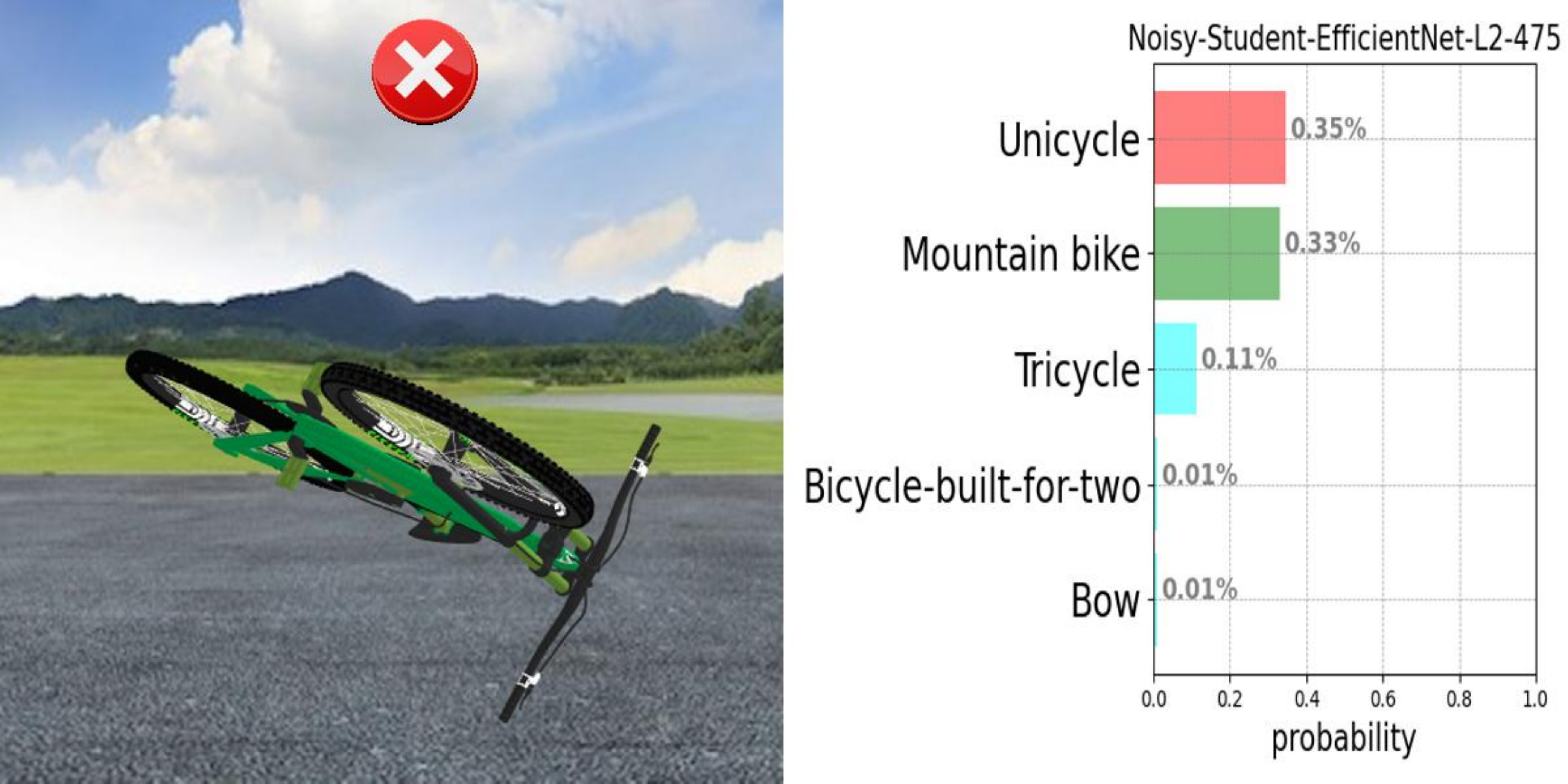}
    \includegraphics[width=0.245\textwidth]{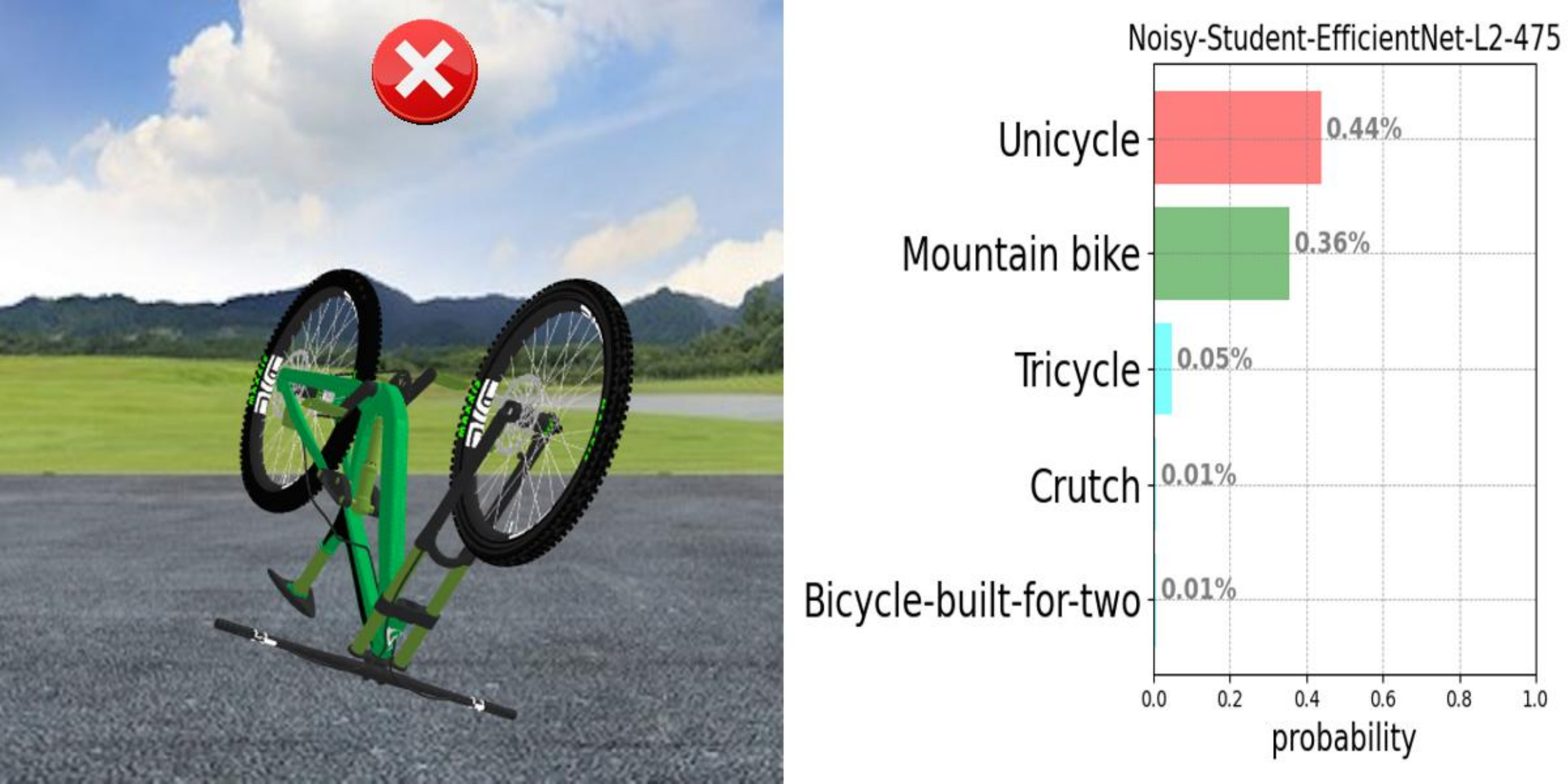}

    \includegraphics[width=0.245\textwidth]{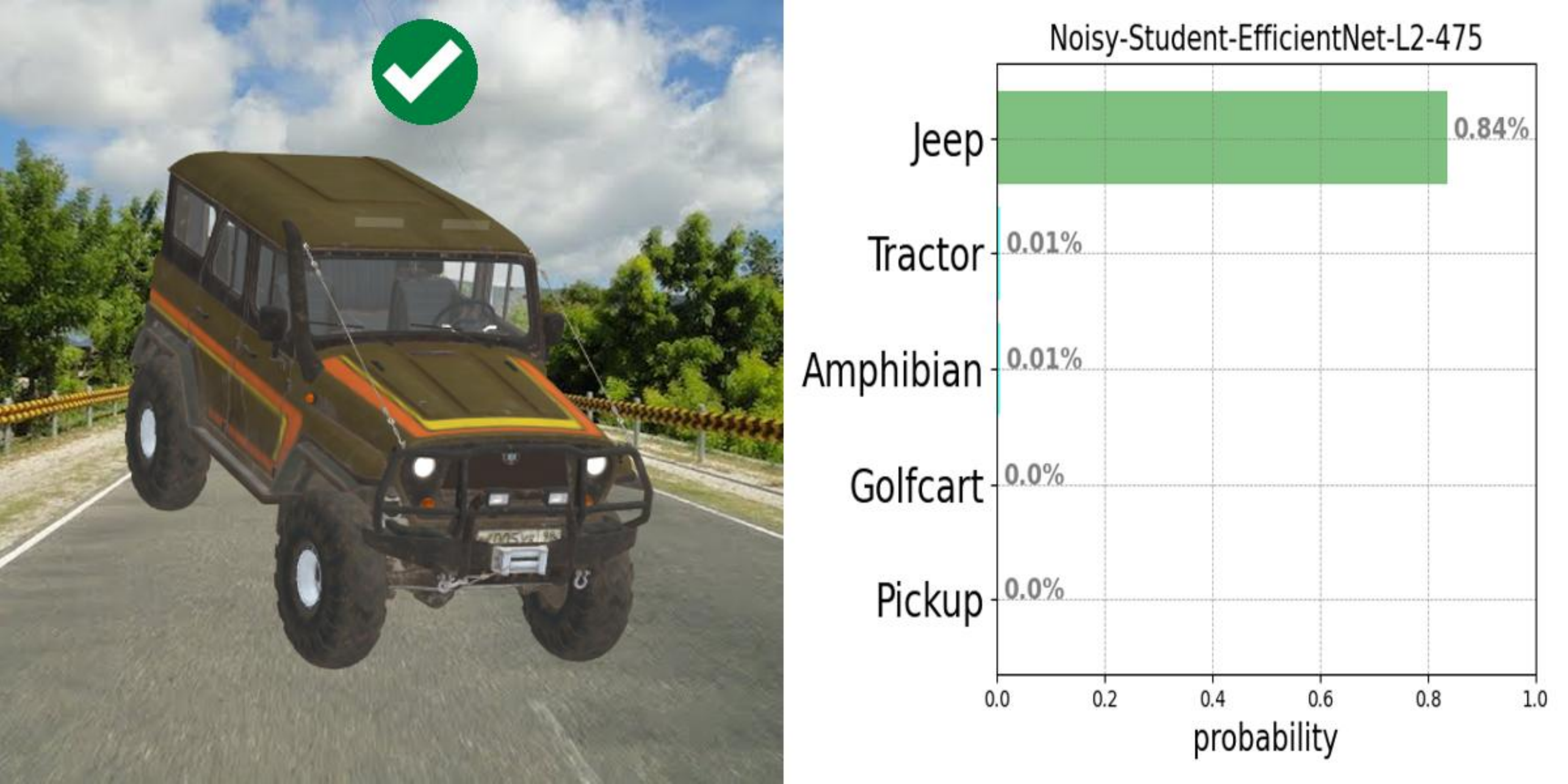}
    \includegraphics[width=0.245\textwidth]{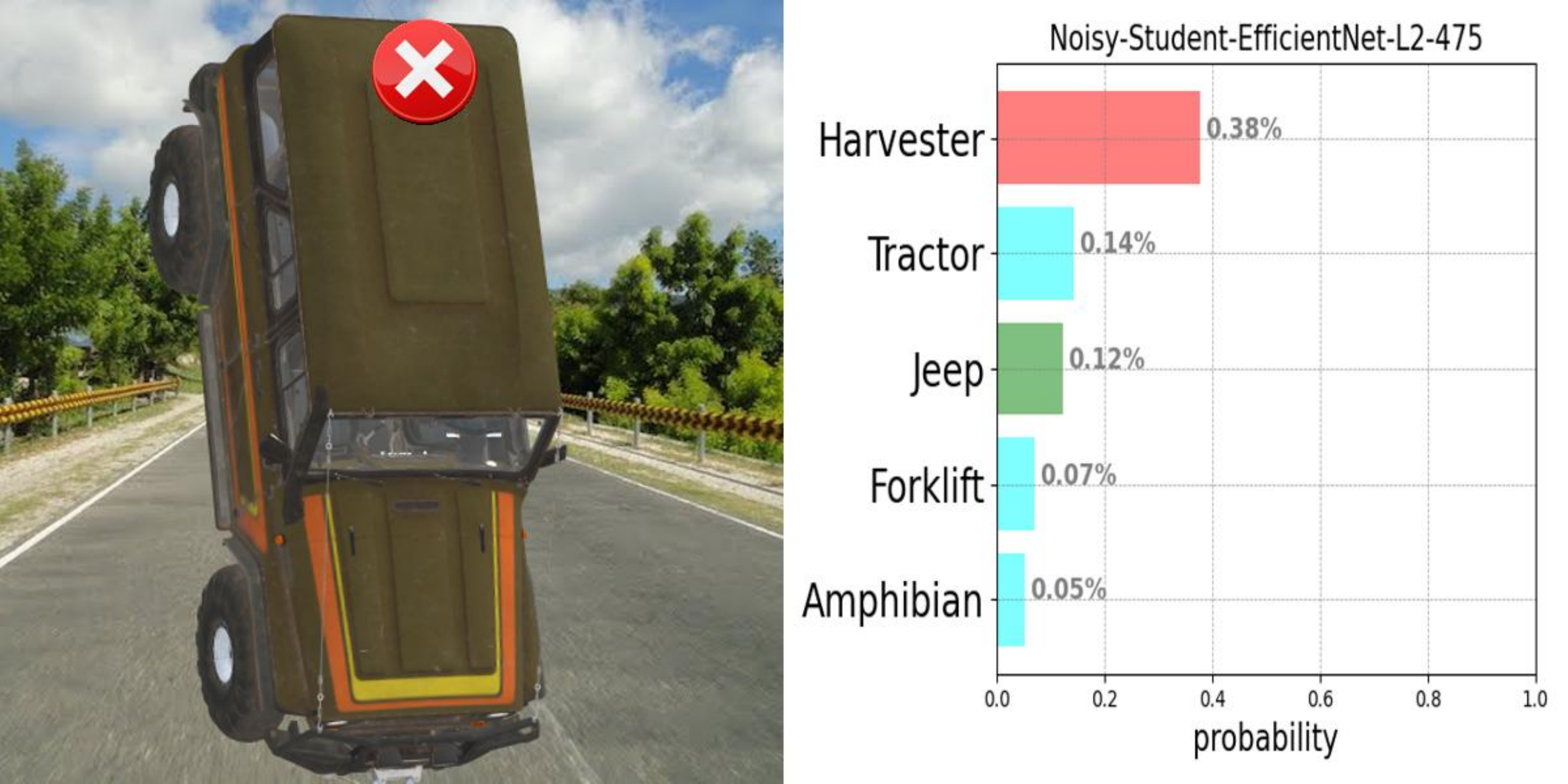}
    \includegraphics[width=0.245\textwidth]{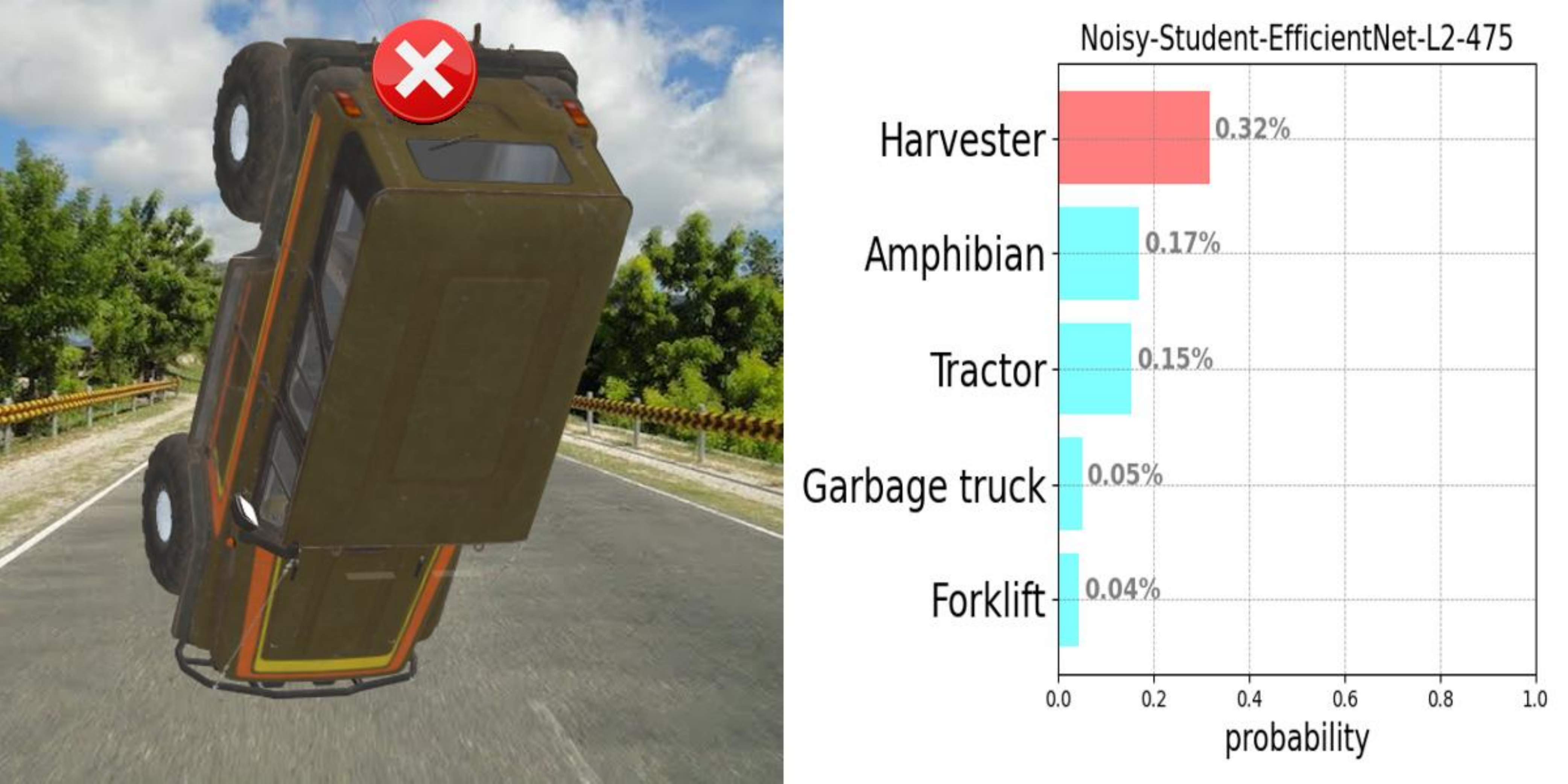}
    \includegraphics[width=0.245\textwidth]{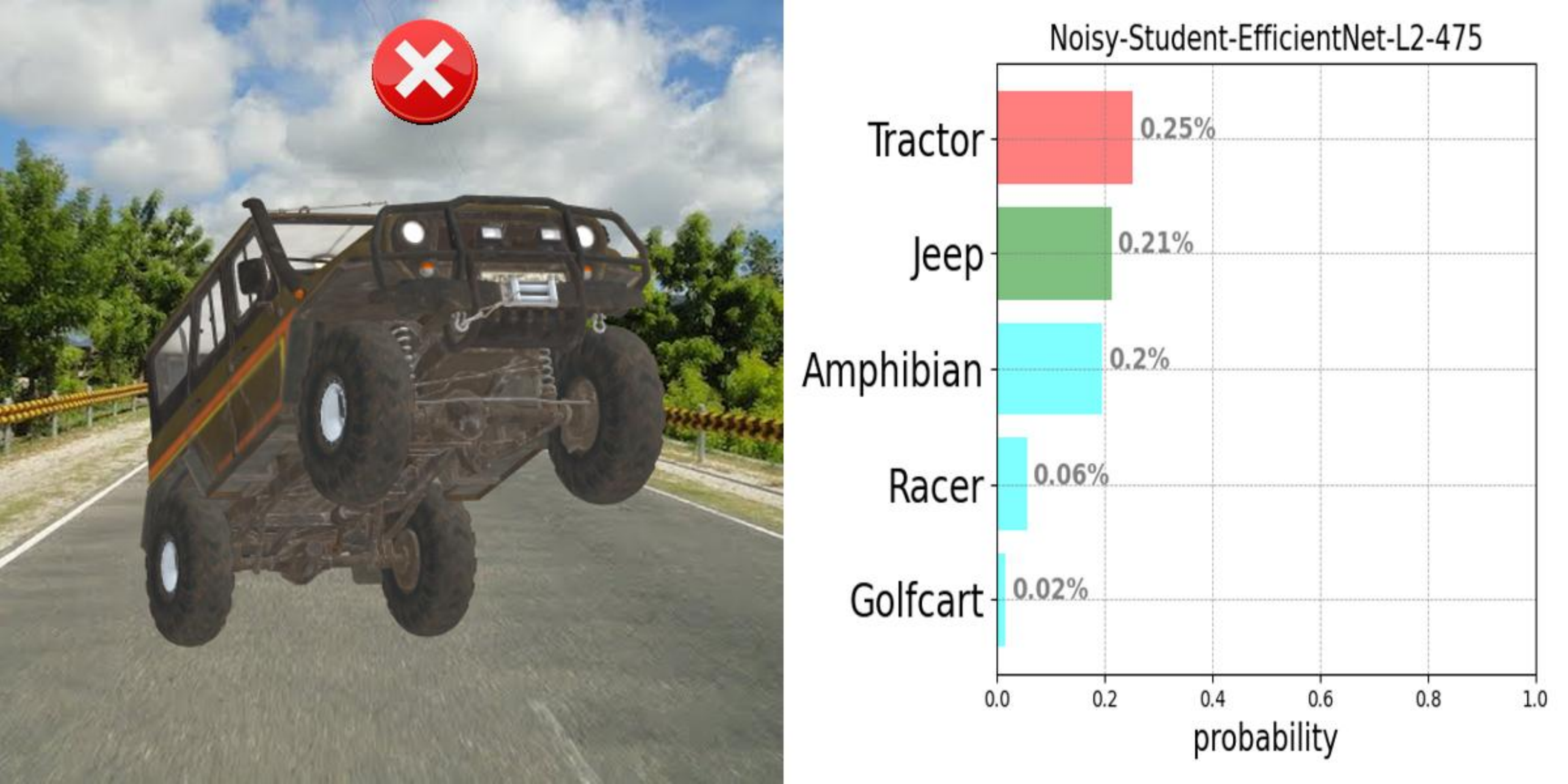}
    
    \includegraphics[width=0.245\textwidth]{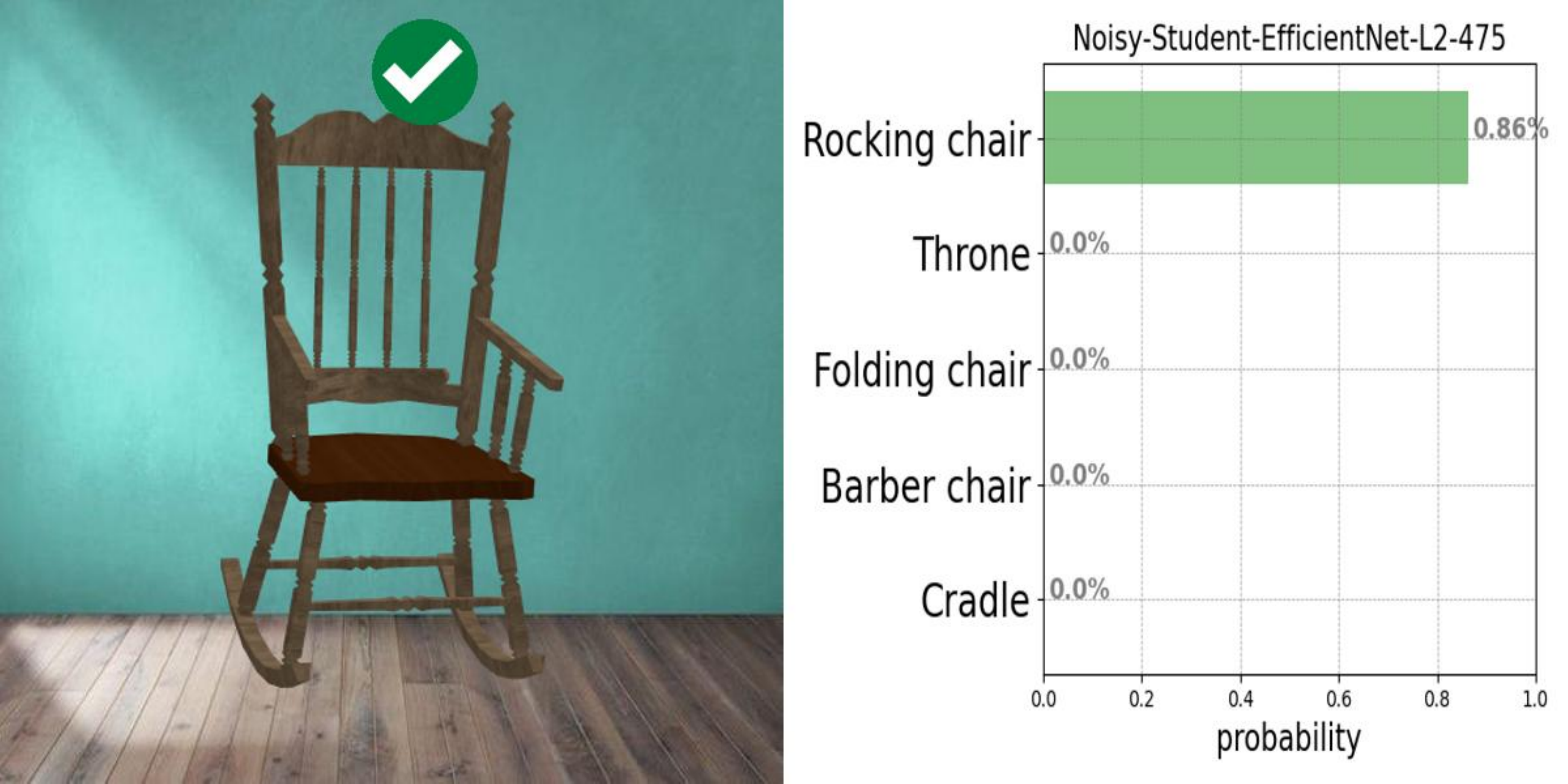}
    \includegraphics[width=0.245\textwidth]{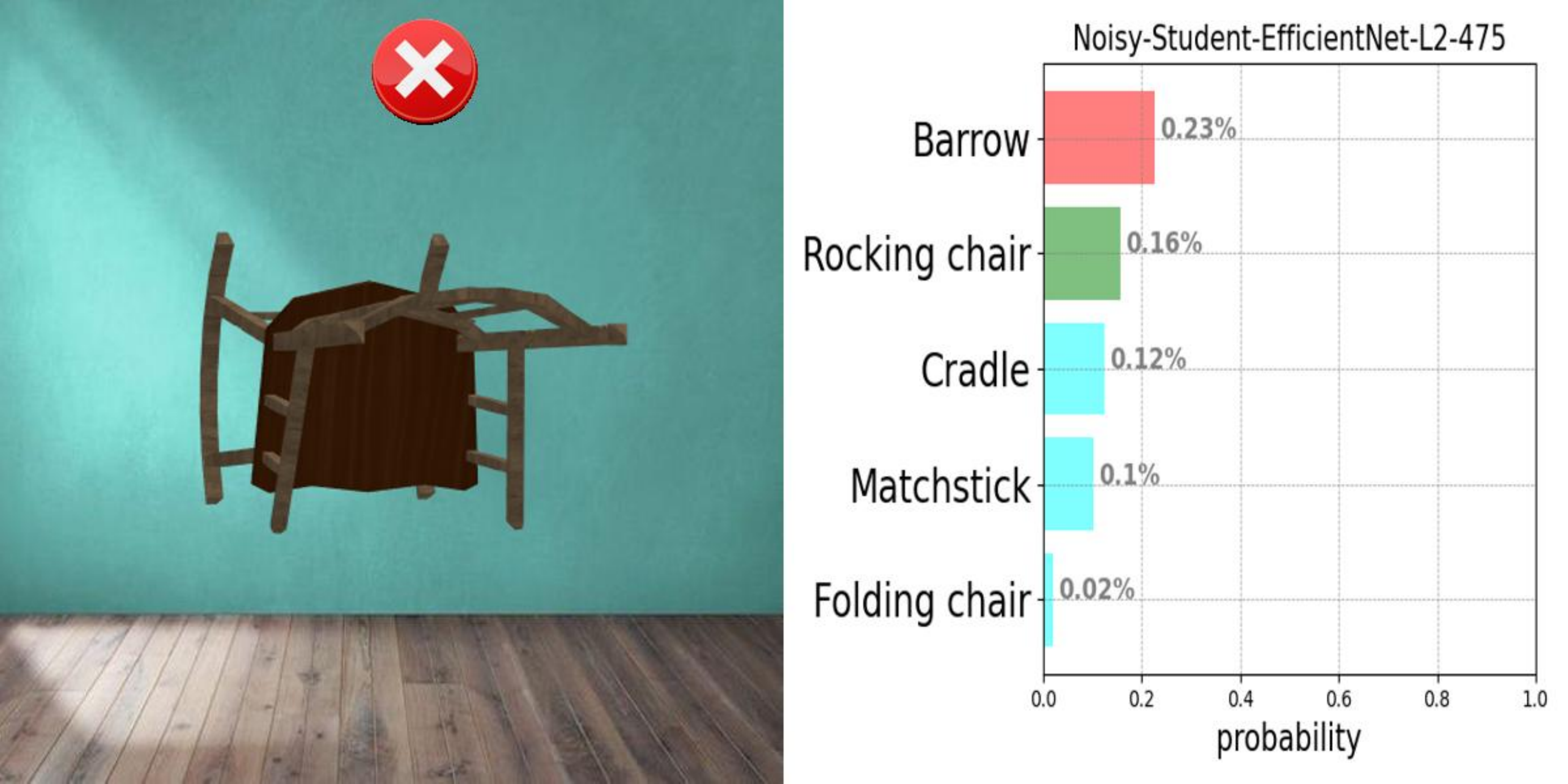}
    \includegraphics[width=0.245\textwidth]{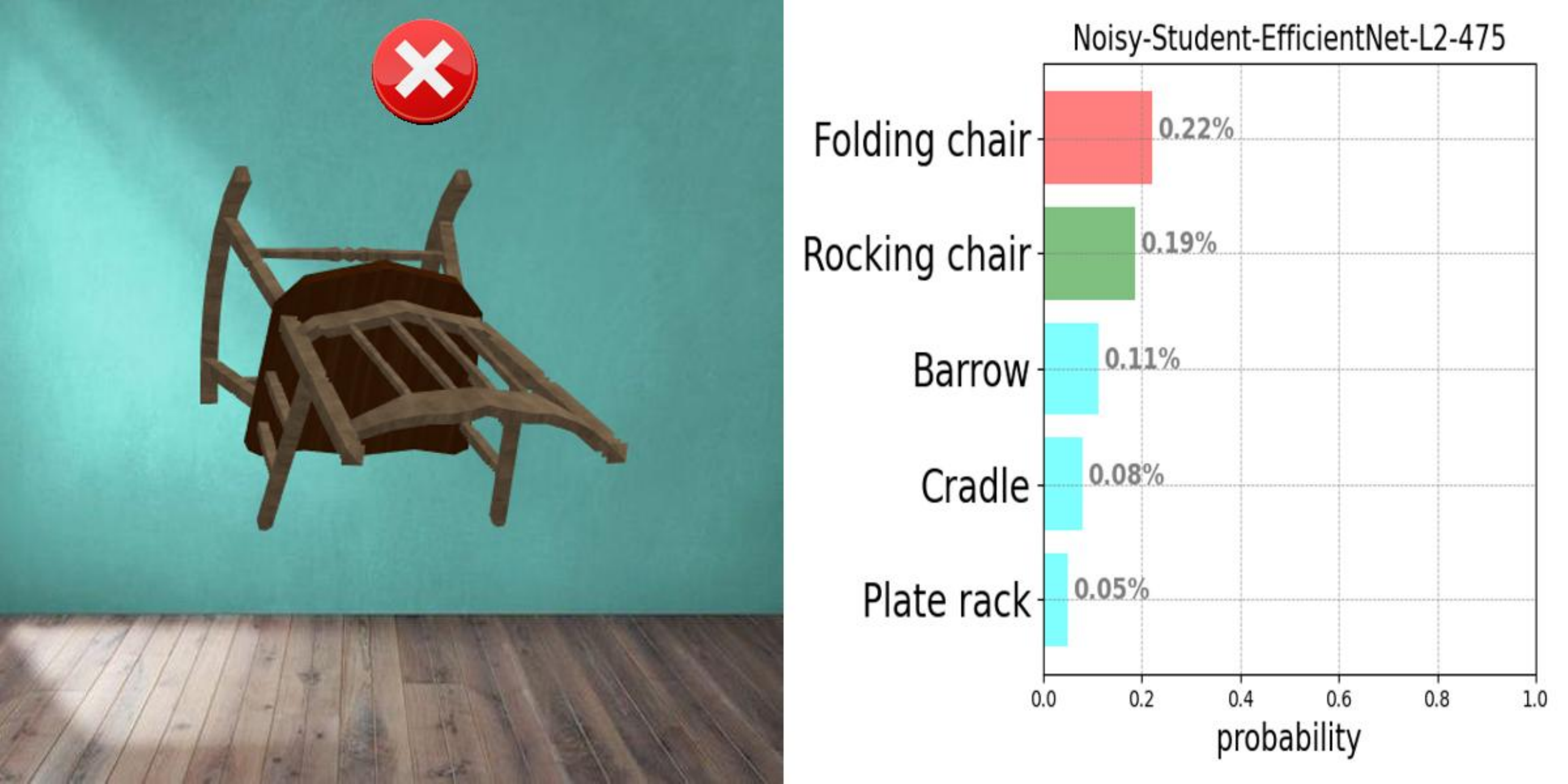}
    \includegraphics[width=0.245\textwidth]{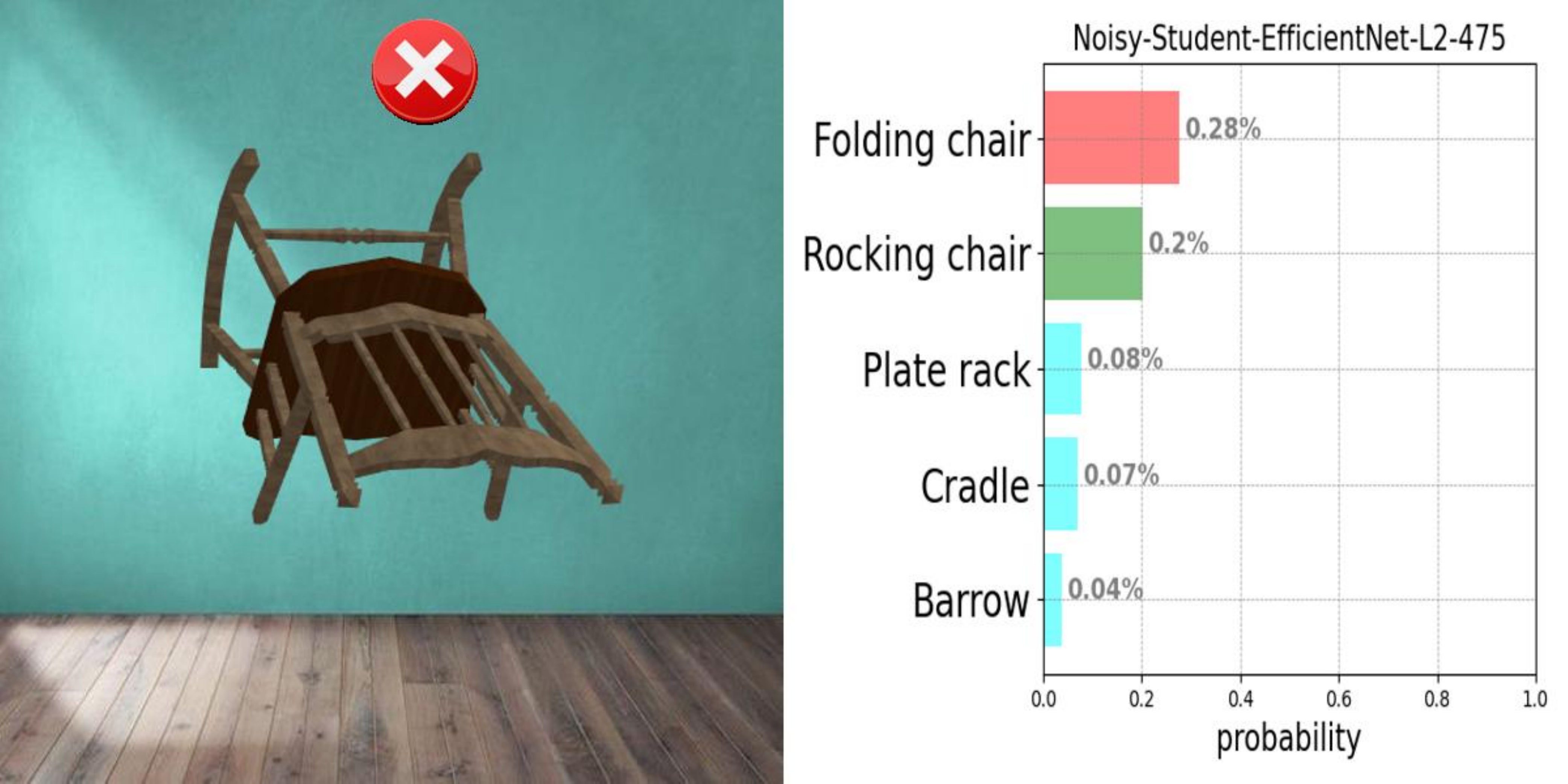}
    
    \includegraphics[width=0.245\textwidth]{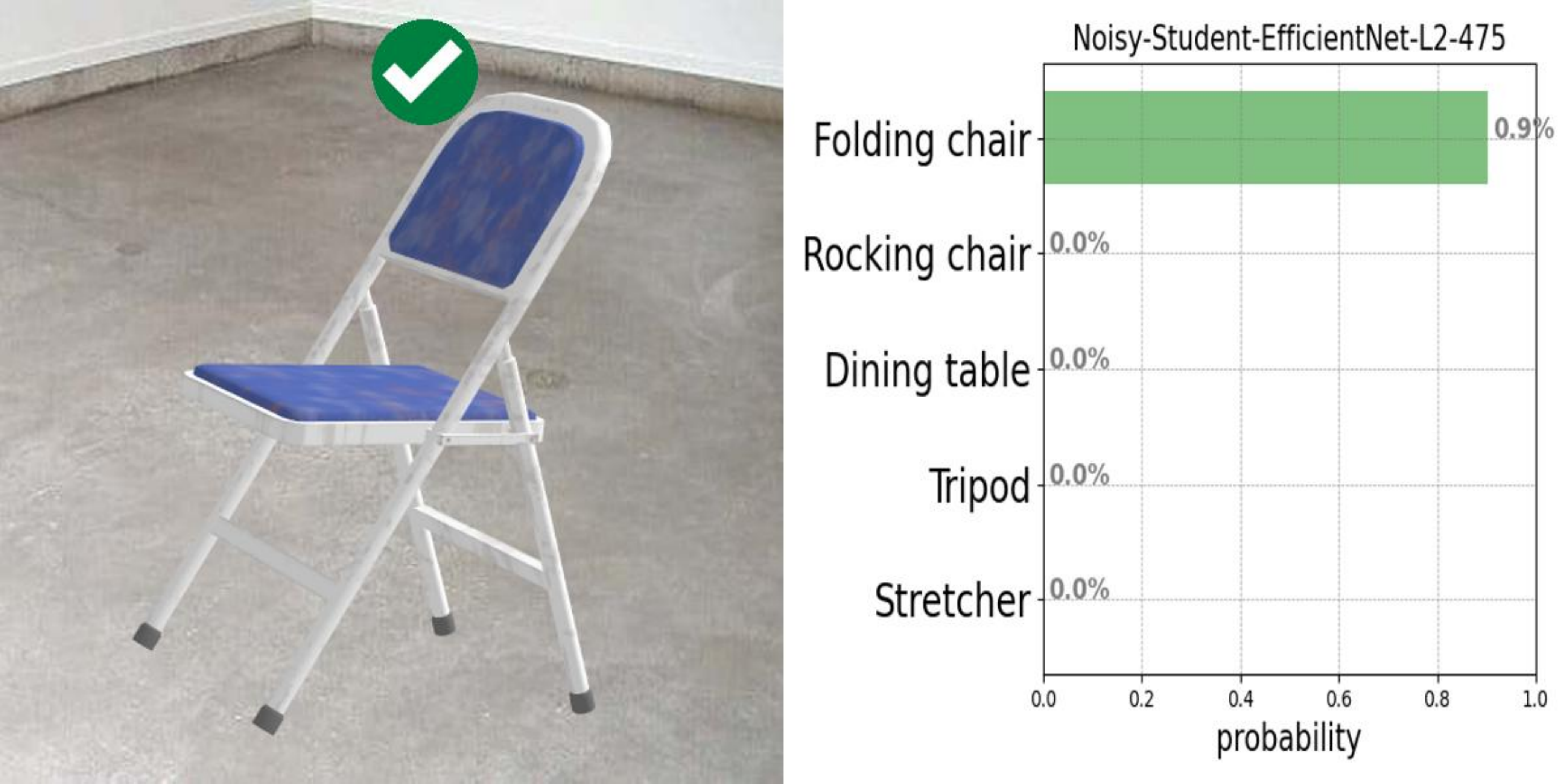}
    \includegraphics[width=0.245\textwidth]{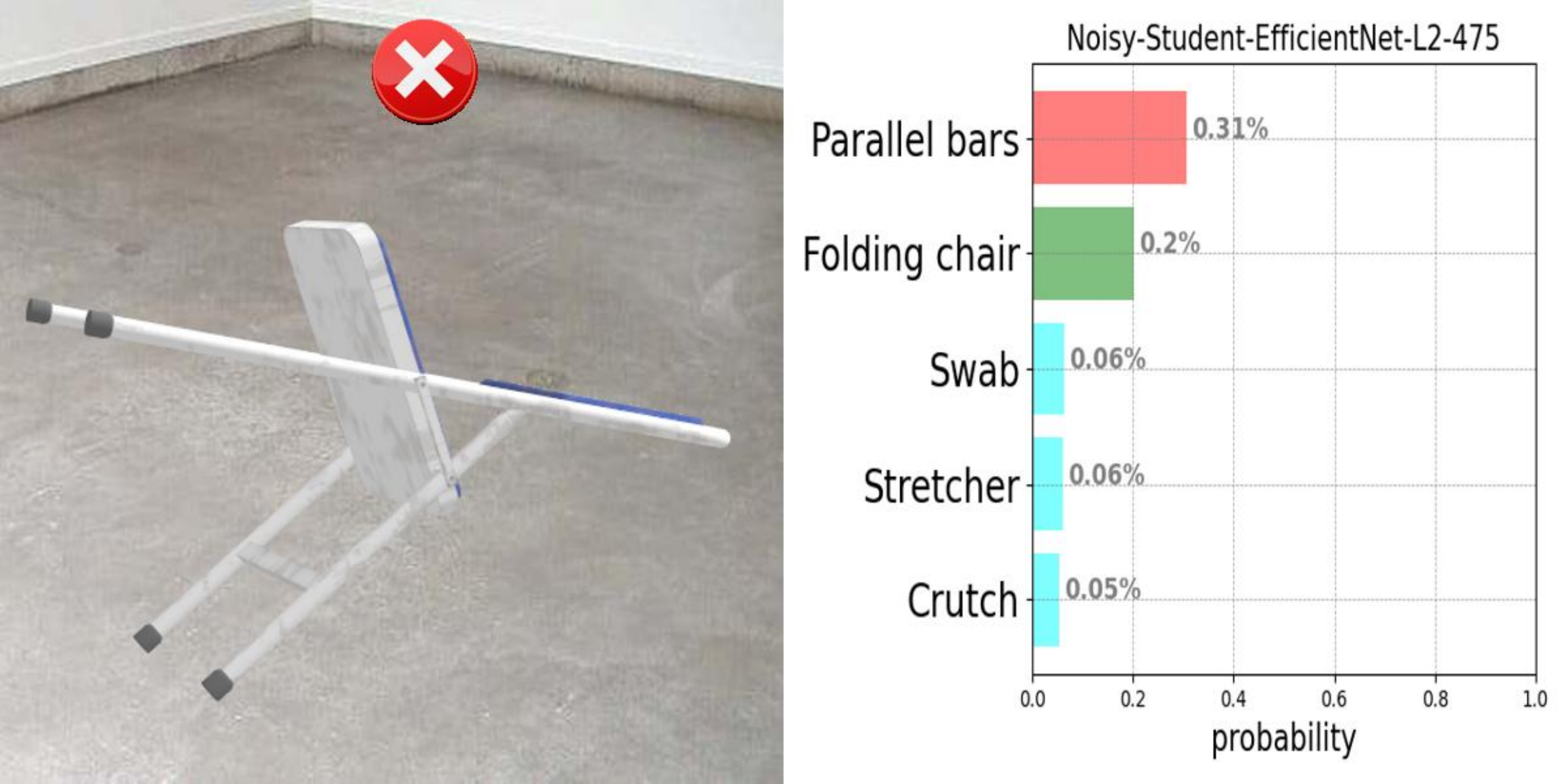}
    \includegraphics[width=0.245\textwidth]{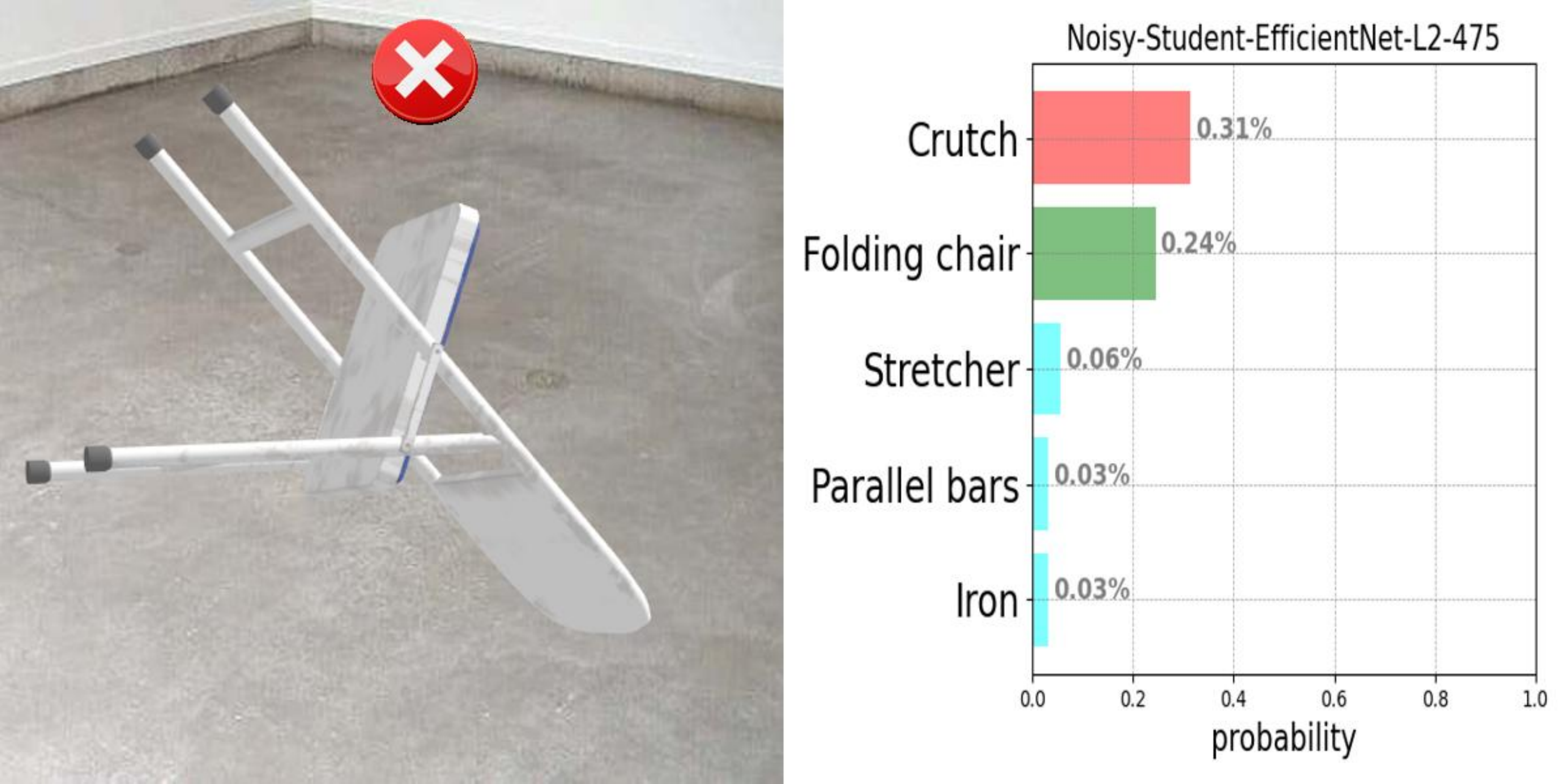}
    \includegraphics[width=0.245\textwidth]{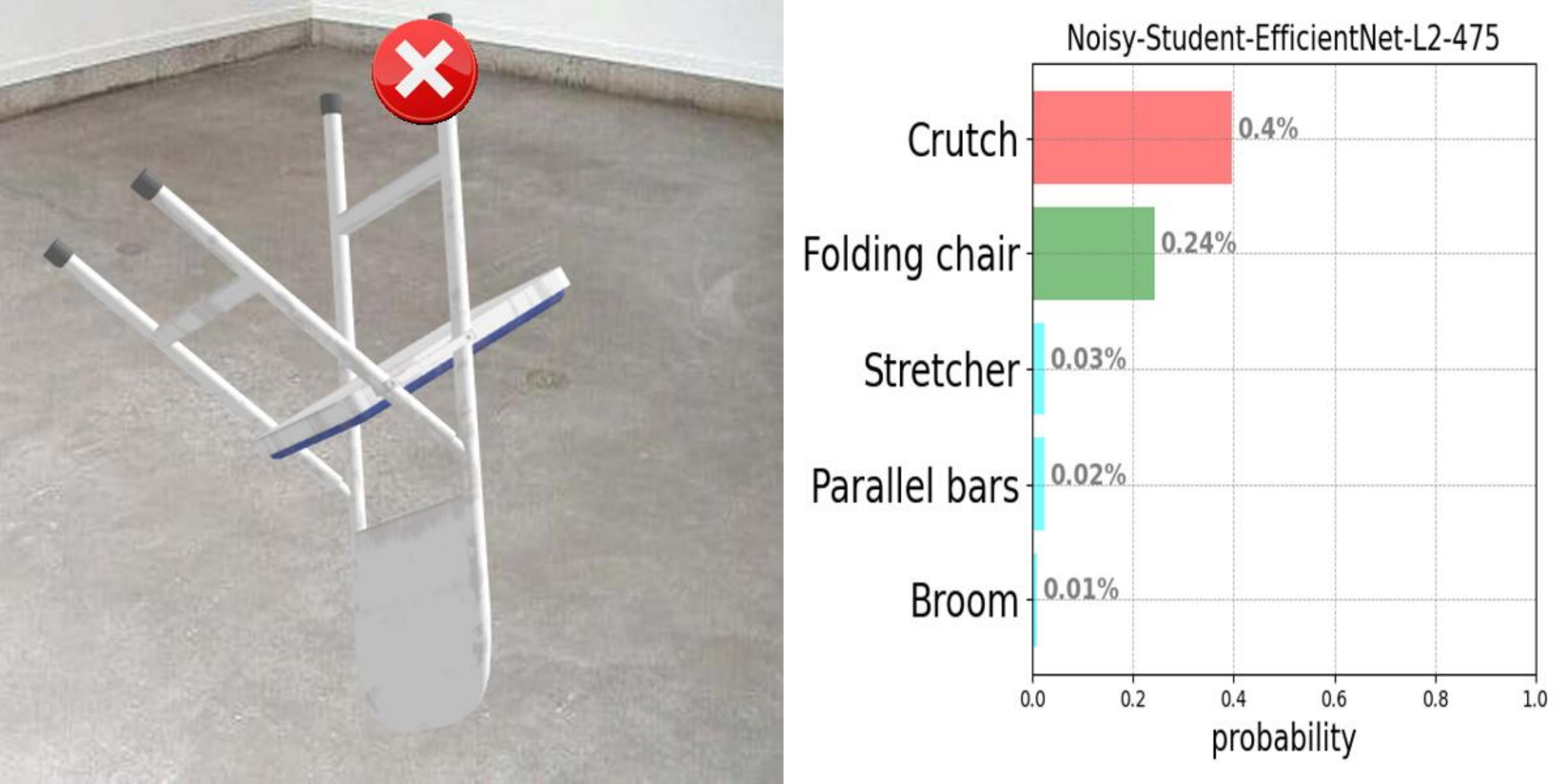}
    
    \includegraphics[width=0.245\textwidth]{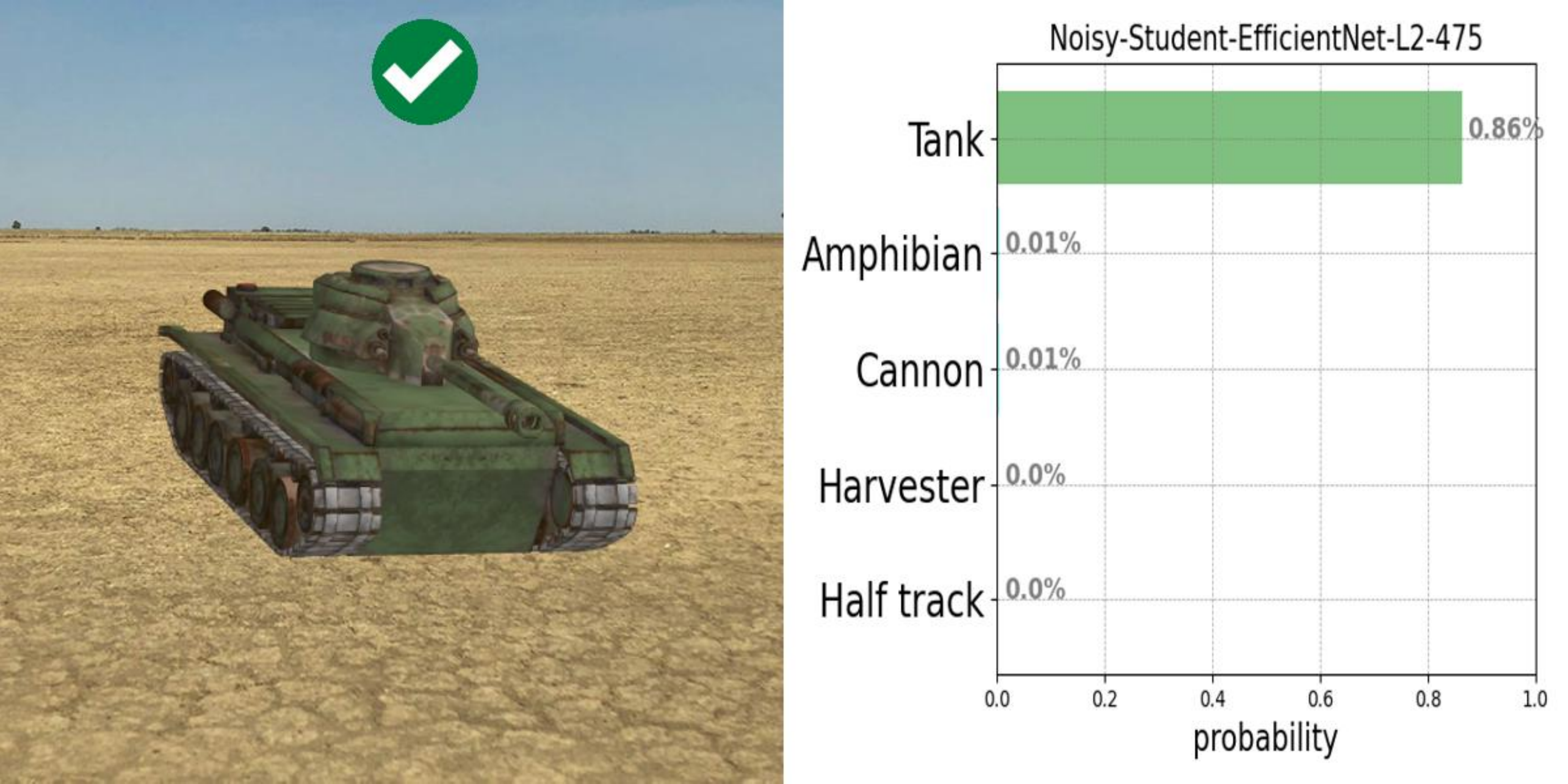}
    \includegraphics[width=0.245\textwidth]{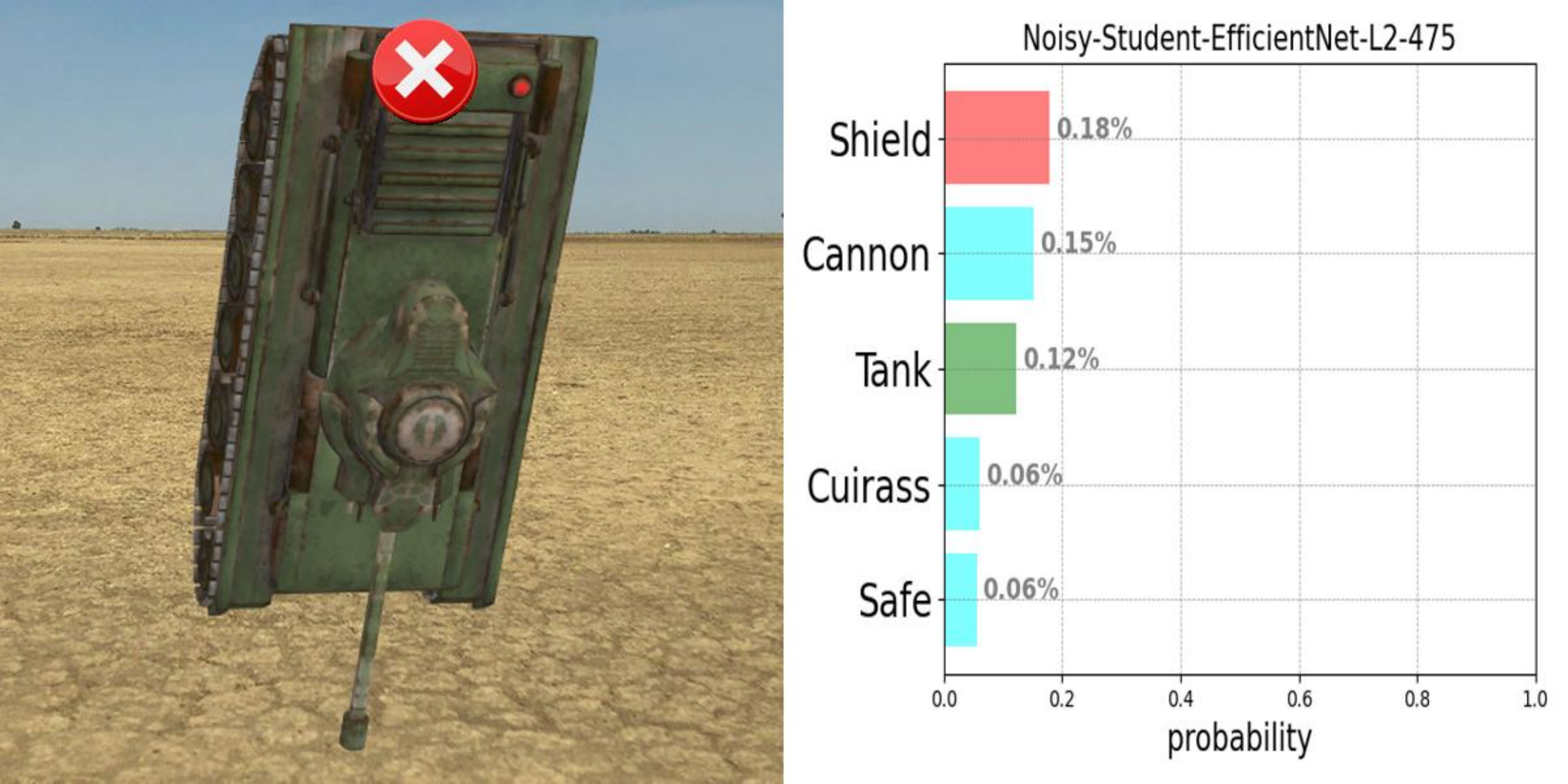}
    \includegraphics[width=0.245\textwidth]{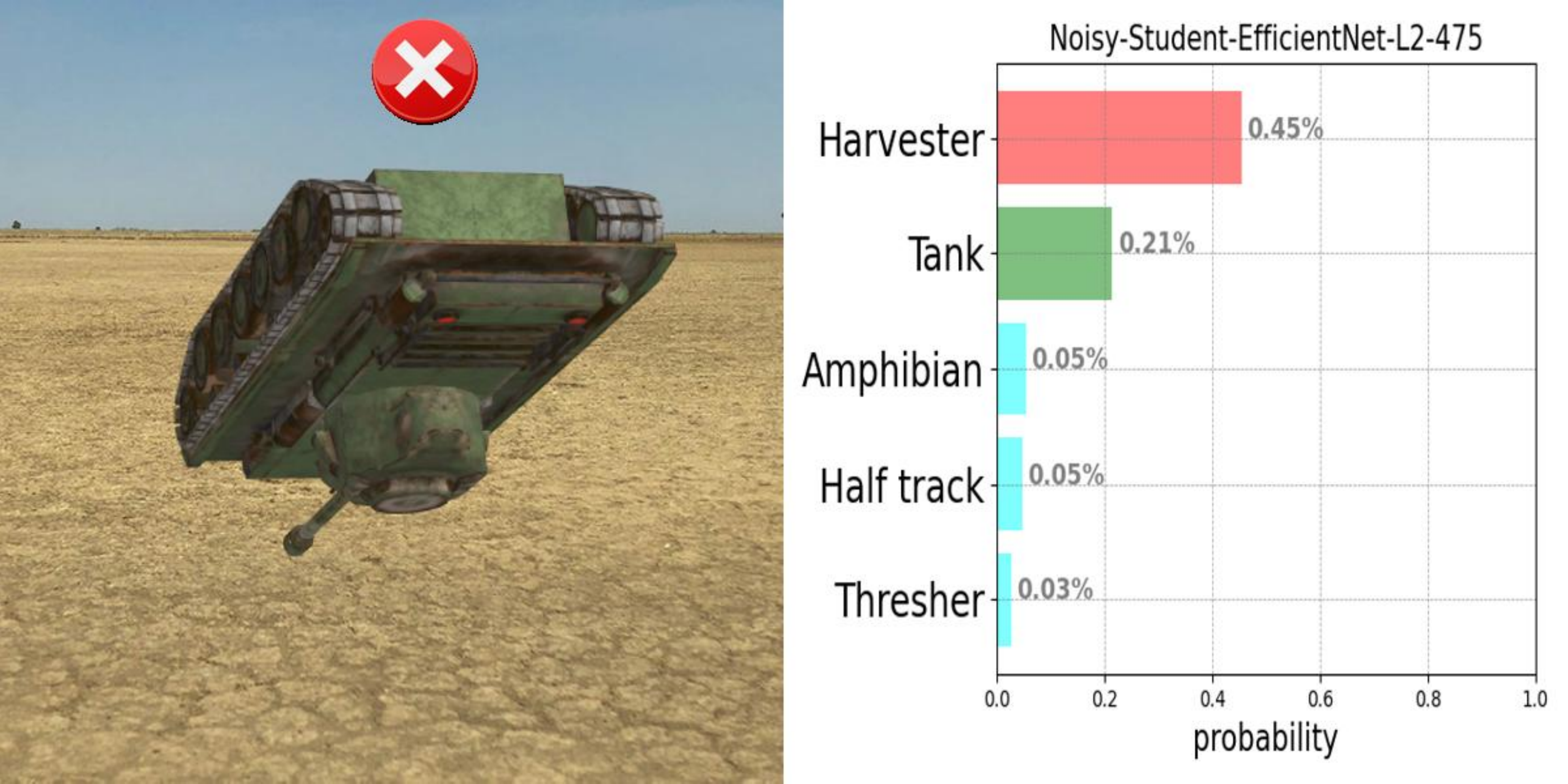}
    \includegraphics[width=0.245\textwidth]{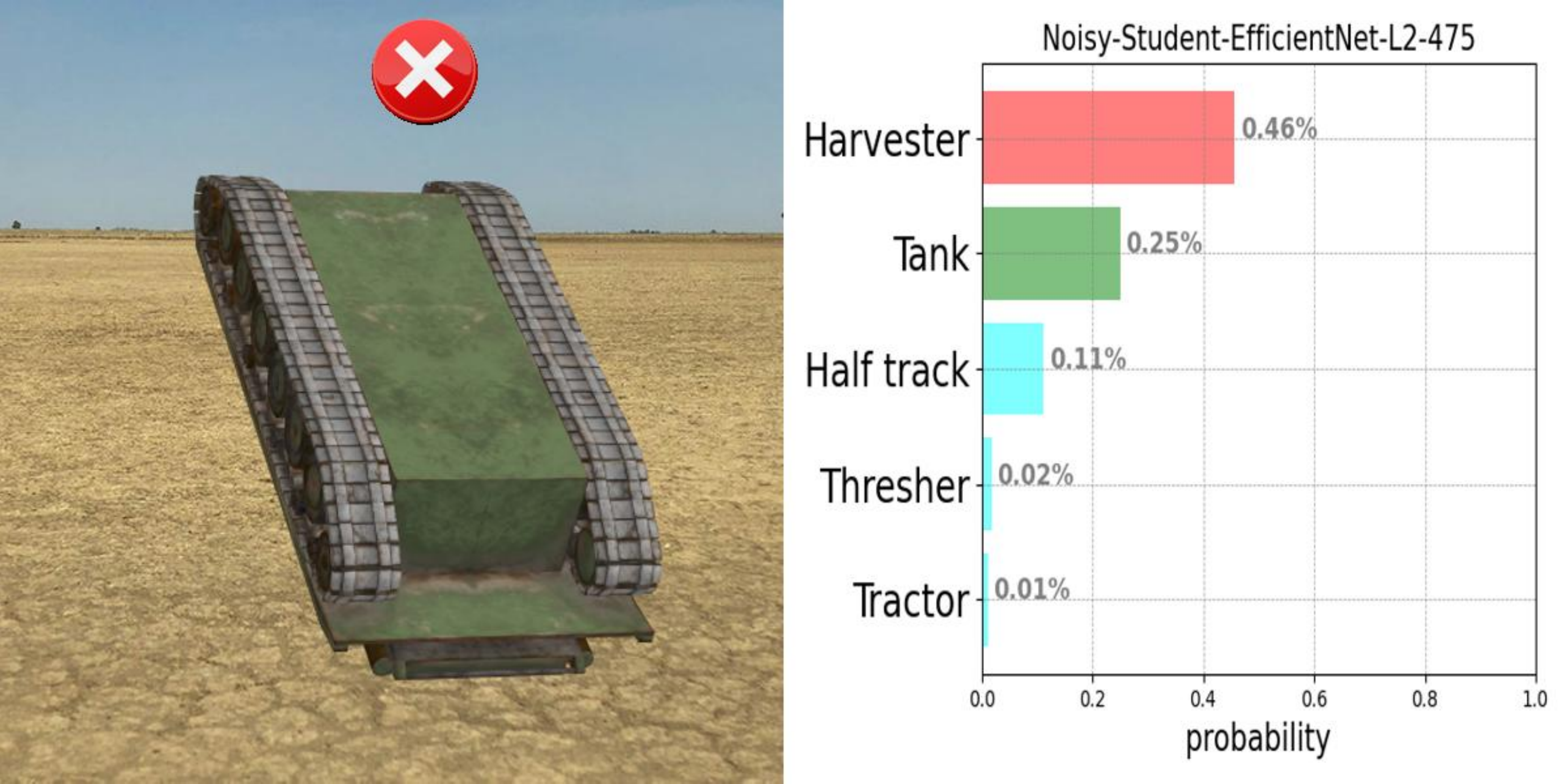}

    \includegraphics[width=0.245\textwidth]{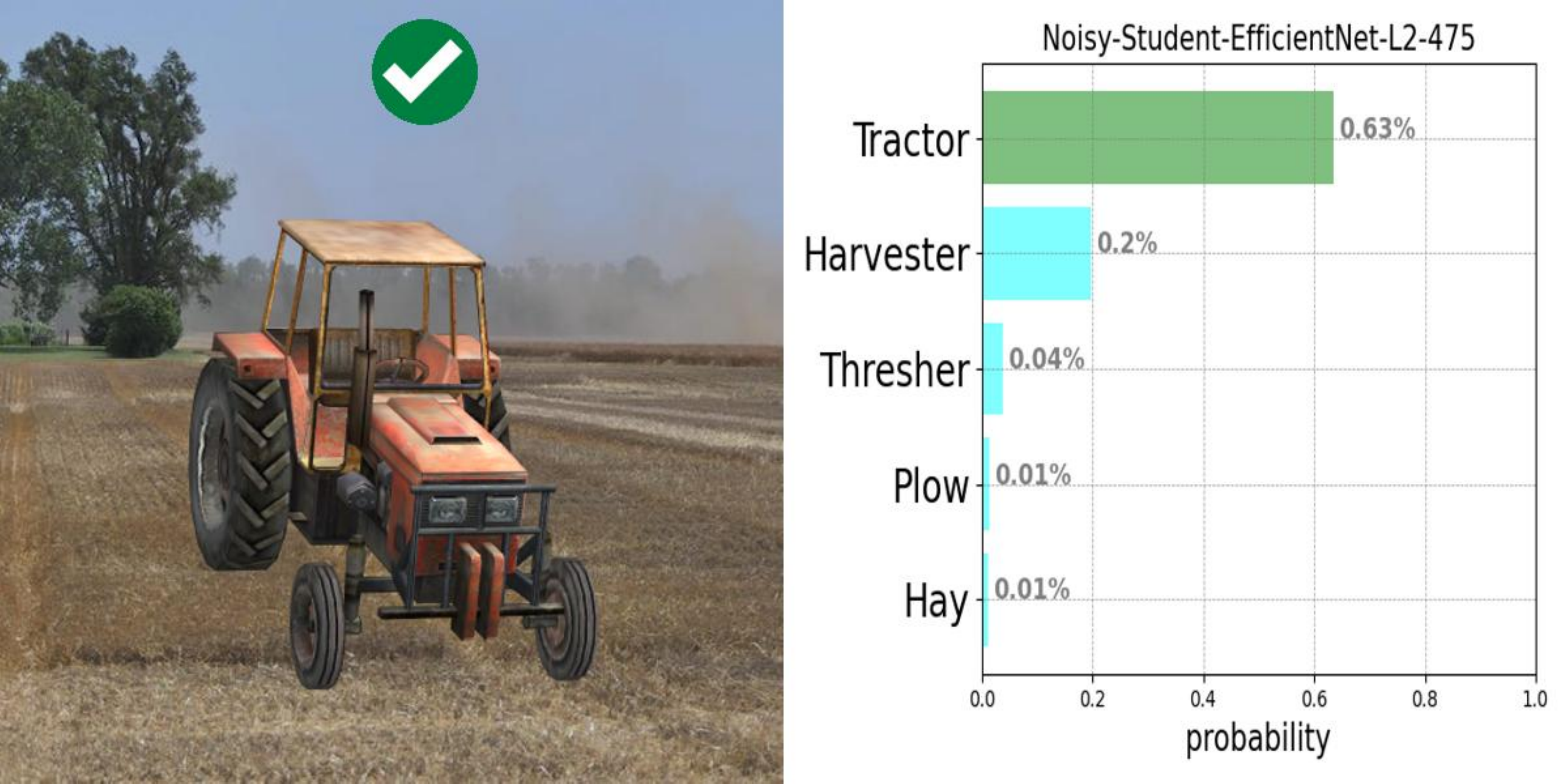}
    \includegraphics[width=0.245\textwidth]{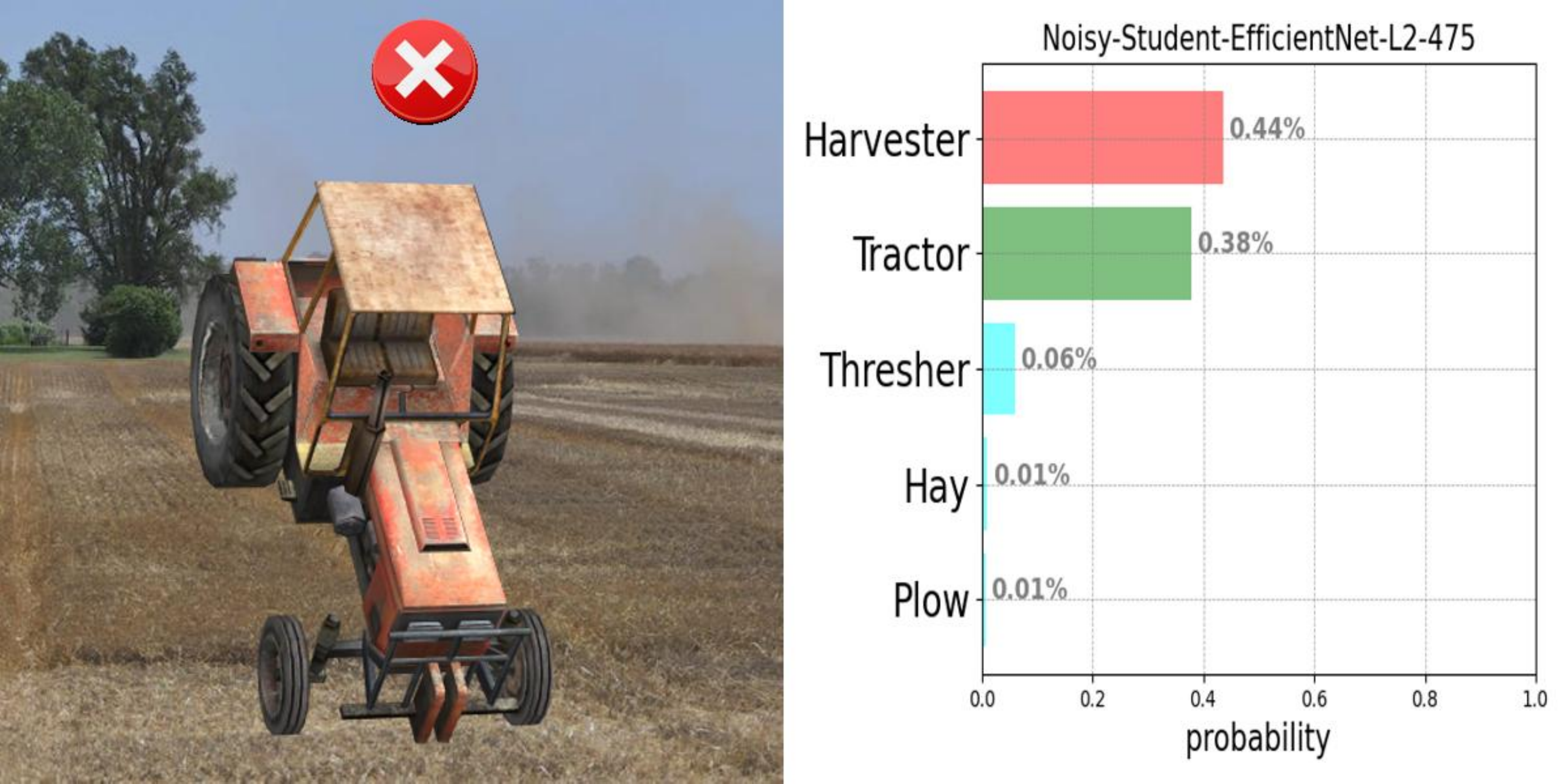}
    \includegraphics[width=0.245\textwidth]{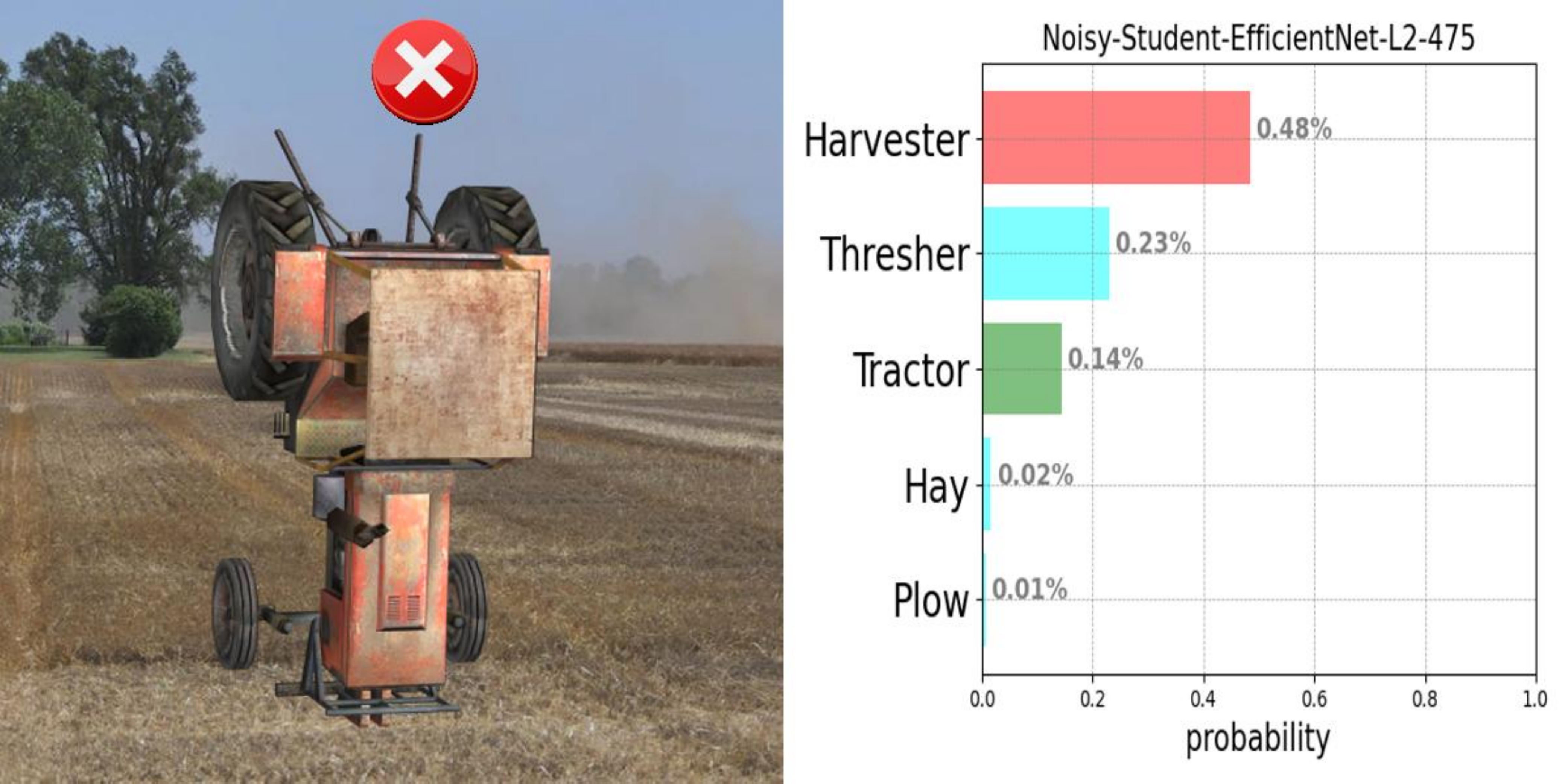}
    \includegraphics[width=0.245\textwidth]{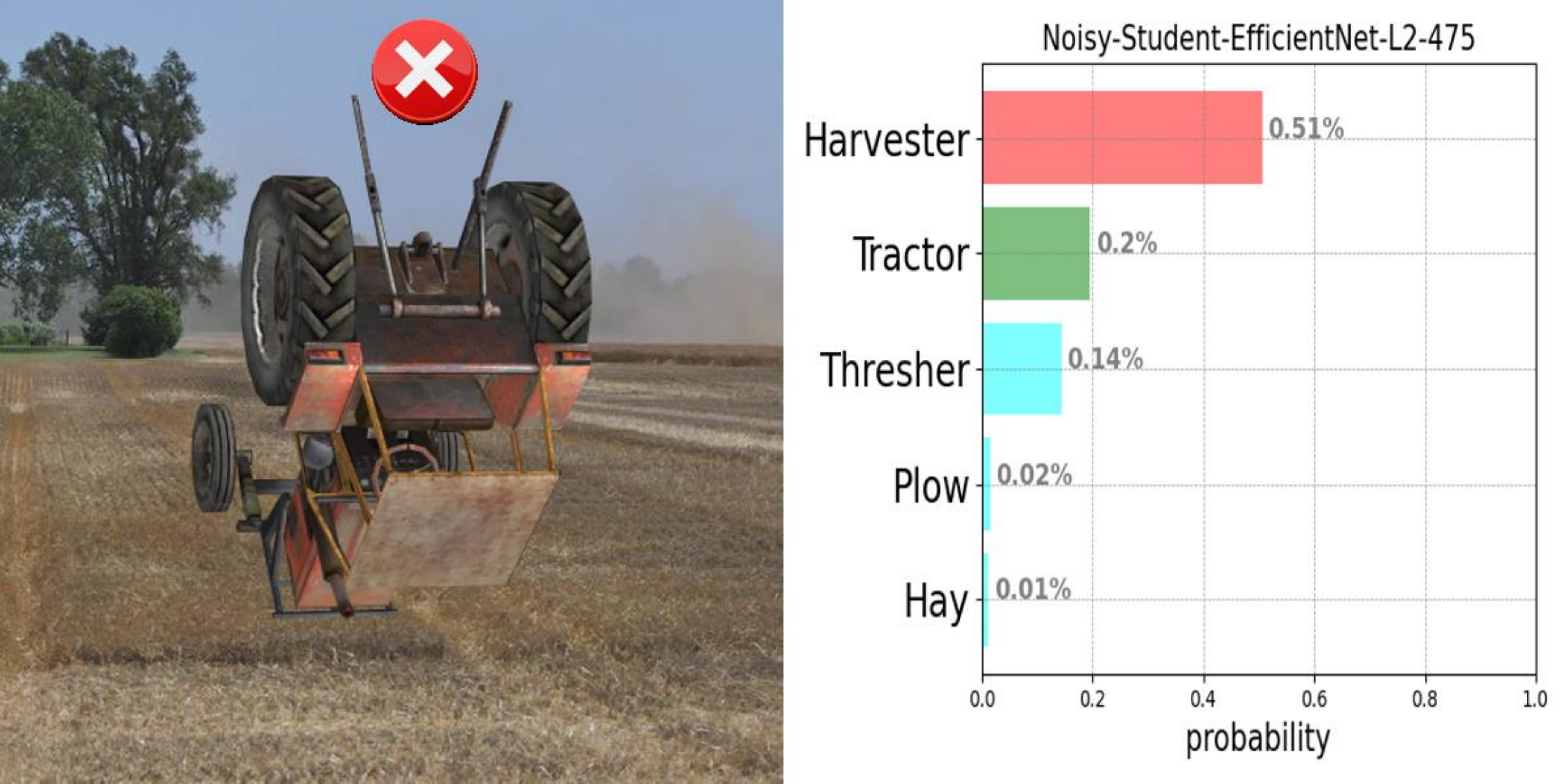}
 
    \caption{\textbf{Selected failures of Noisy Student EfficientNet-L2---the best-performing network of our collection---on ObjectPose.} Left column: Objects presented upright and top-5 predictions from the network (as measured by the softmax layer activations). Other columns: Objects presented in incorrectly classified poses and top-5 predictions from the network. Some of these errors reveal a brittleness compared to the human visual system (e.g., a tank at 90° is confused with a shield).}
    \label{fig:nsfailureexamples}
\end{figure}

\paragraph{All networks exhibit a performance drop on unusual poses compared to usual poses (Fig. \ref{fig:ObjectPose_accuracy}).} We measure networks' robustness to unusual object poses by testing their accuracy on our synthetic dataset, ObjectPose. Our collection of networks show a top-1 accuracy drop in a range 14.5\%-45.5\% on unusual poses compared to usual poses (ObjectPose +-10). Examples of network failures are shown in Fig. \ref{fig:nsfailureexamples} for the best model tested Noisy Student EfficientNet-L2, and in suppl. fig. \ref{fig:beitfailureexamples} for another high-performing model, BEiT-B/16.

\paragraph{Scaling both training dataset size and network capacity is helpful on ObjectPose.} Networks trained on larger datasets show a narrower performance gap than networks trained on smaller datasets (Fig. \ref{fig:ObjectPose_accuracy}B), with the exception of CLIP models which were not fine-tuned on ImageNet categories. It is also important for the network to have a sufficient size (i.e. number of parameters) to benefit from the large dataset. This is deduced by the performance of the EfficientNet-L2 model (480M parameters) compared to EfficientNet-B7 (66M parameters), both trained using the Noisy Student method on the JFT-300M dataset. While EfficientNet-L2 outperforms the rest of the networks on ObjectPose, EfficientNet-B7 performs on par with many networks pretrained on smaller datasets (such as ImageNet21k). Although SWAG models are pretrained on more images (3.6B) and have more parameters, they do not outperform Noisy Student EfficientNet-L2. This also shows that scaling the dataset is not the only factor that matters, here the training procedure and architecture also play an important role (see appendix \ref{appendix:modelsdescription} for individual networks' descriptions). In particular, Noisy student was trained using the RangAugment data augmentation method \citep{randaugment} which applies intensive data augmentation including rotations and shears. In contrast, SWAG only used a limited augmentation strategy of cropping and flipping and not including rotations and shears.  Lastly, in Appendix \ref{sec:clipexperiments}, we show that carefully engineering the textual prompts to CLIP to suggest that the object is in an unusual pose can slightly improve this model's performance. 

\paragraph{A visual inspection of networks' failures reveals room for improvement even for the best networks tested (Fig. \ref{fig:nsfailureexamples}).} By visually inspecting the errors made by the networks (Fig. \ref{fig:nsfailureexamples} and suppl. figures \ref{fig:beitfailureexamples}-\ref{fig:bitgrid2} at the end of the appendix), we find that even the best model tested, Noisy Student EfficientNet-L2 (NS), makes errors that a human observer would not. On some objects, such as the forklift in suppl. fig. \ref{fig:engrid1}, we see that NS performs amazingly well in all orientations, unlike models trained on smaller datasets such as BiTM (suppl. fig. \ref{fig:bitgrid1}). On other examples, such as the hammerhead shark in poses where we cannot see the hammerhead (suppl. fig. \ref{fig:engrid2}), NS makes errors that a human observer would certainly make as well, confusing the hammerhead shark with another type of shark. But some errors made by NS reveal a brittleness compared to our visual system. For example, in suppl. fig. \ref{fig:engrid3}, NS confuses a bicycle with a unicycle in many poses; in suppl. fig. \ref{fig:engrid4}, NS confuses a canon at $90^{\circ}$ with a harp or a guillotine, and in suppl. fig. \ref{fig:engrid5} a chair with a plow. In conclusion, none of these networks performs perfectly, and even the best networks tested make errors that would probably be avoided by a human. 

\begin{figure}[t]
    \centering

    $\vcenter{\hbox{\includegraphics[width=.325\textwidth]{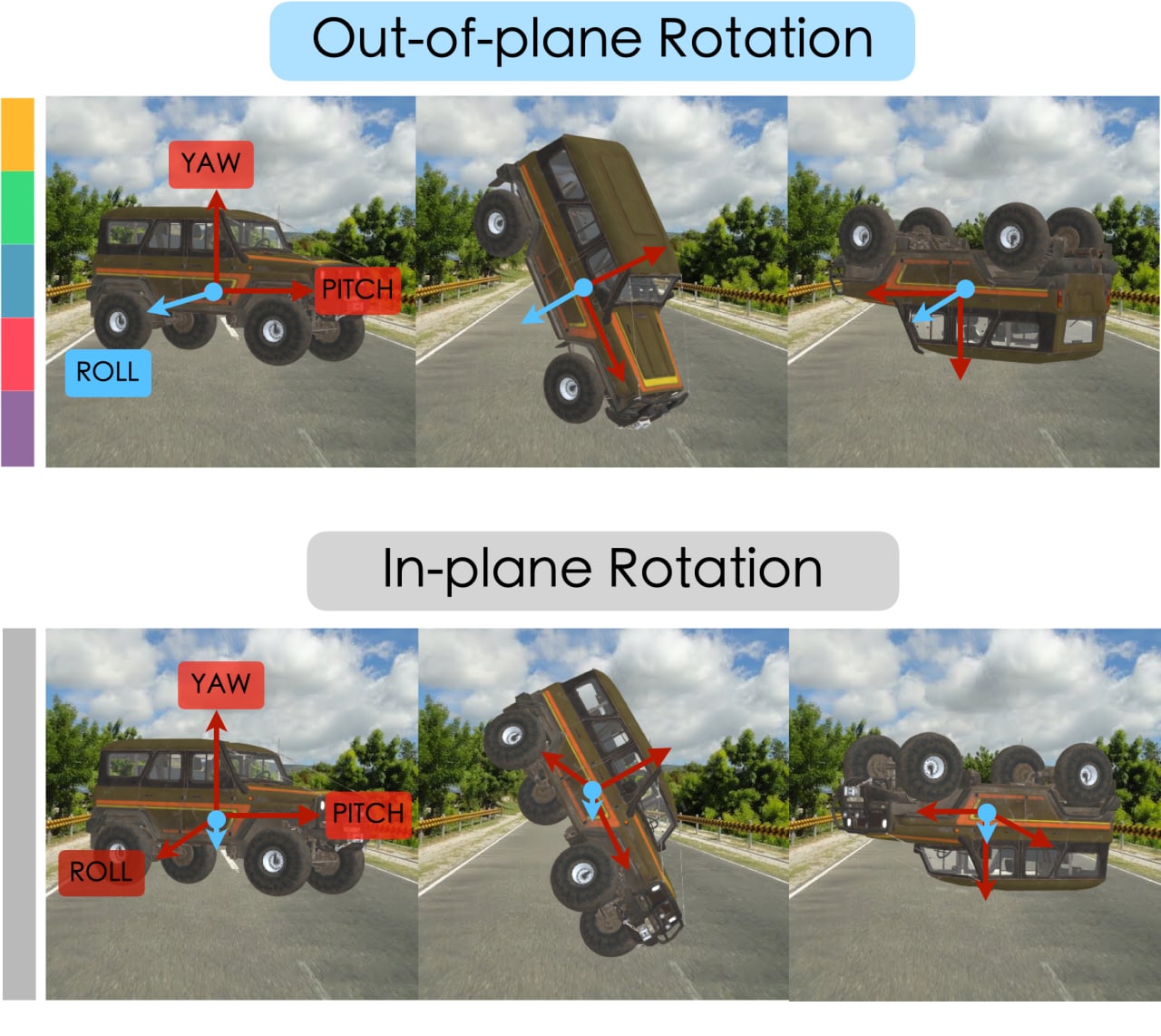}}}$
    $\vcenter{\hbox{\includegraphics[width=.66\textwidth]{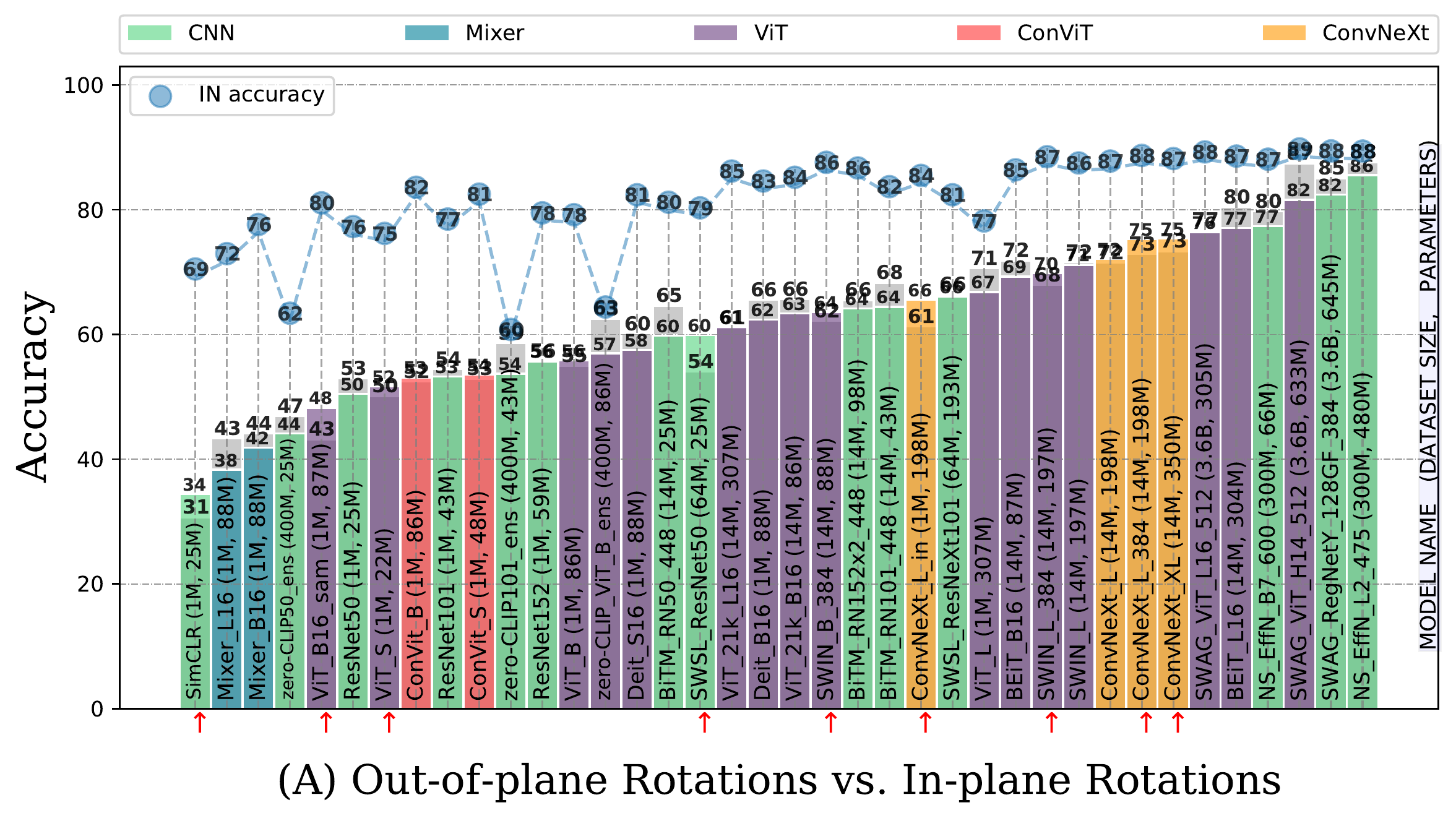}}}$
    
    $\vcenter{\hbox{\includegraphics[width=.325\textwidth]{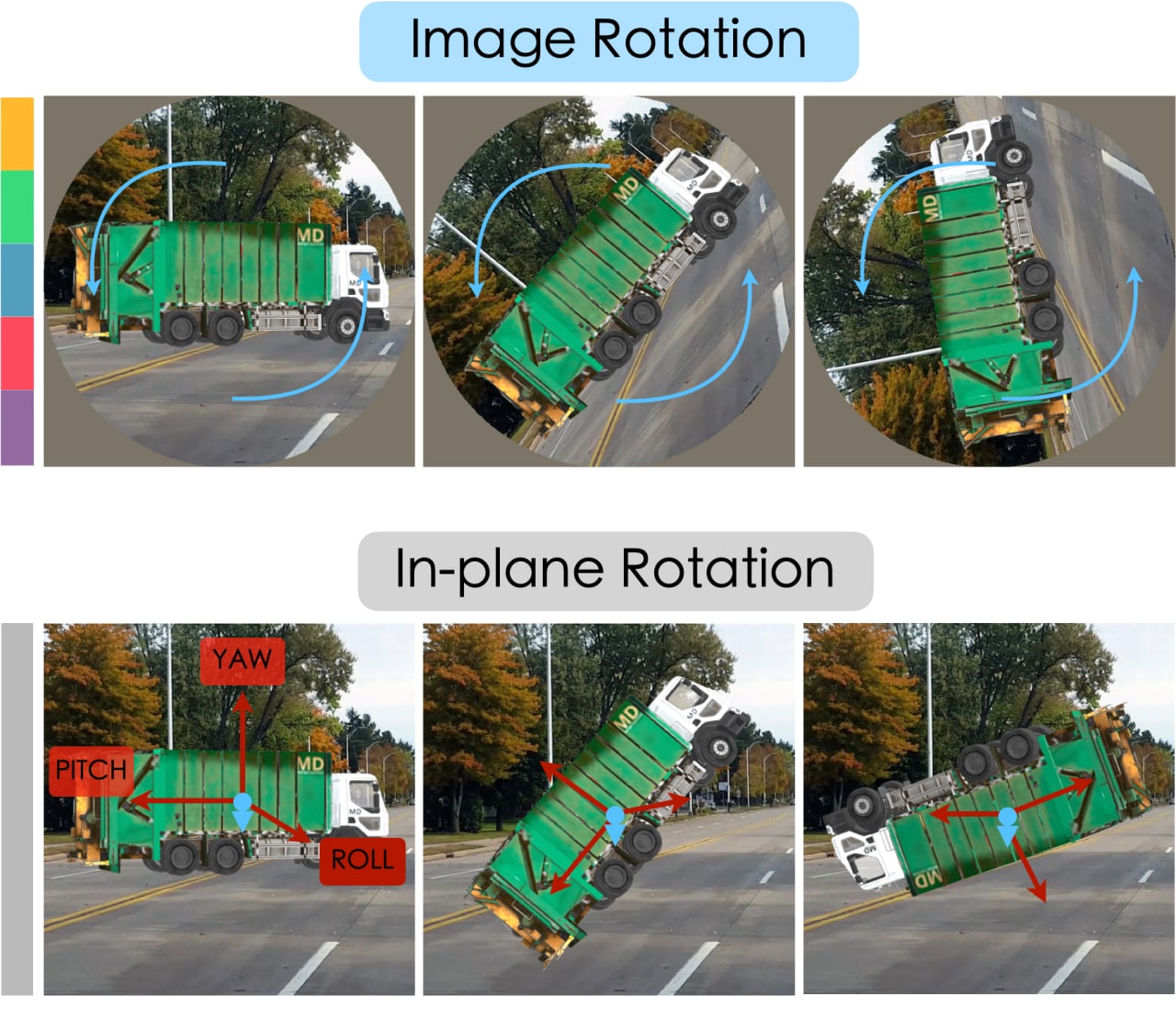}}}$
    $\vcenter{\hbox{\includegraphics[width=.66\textwidth]{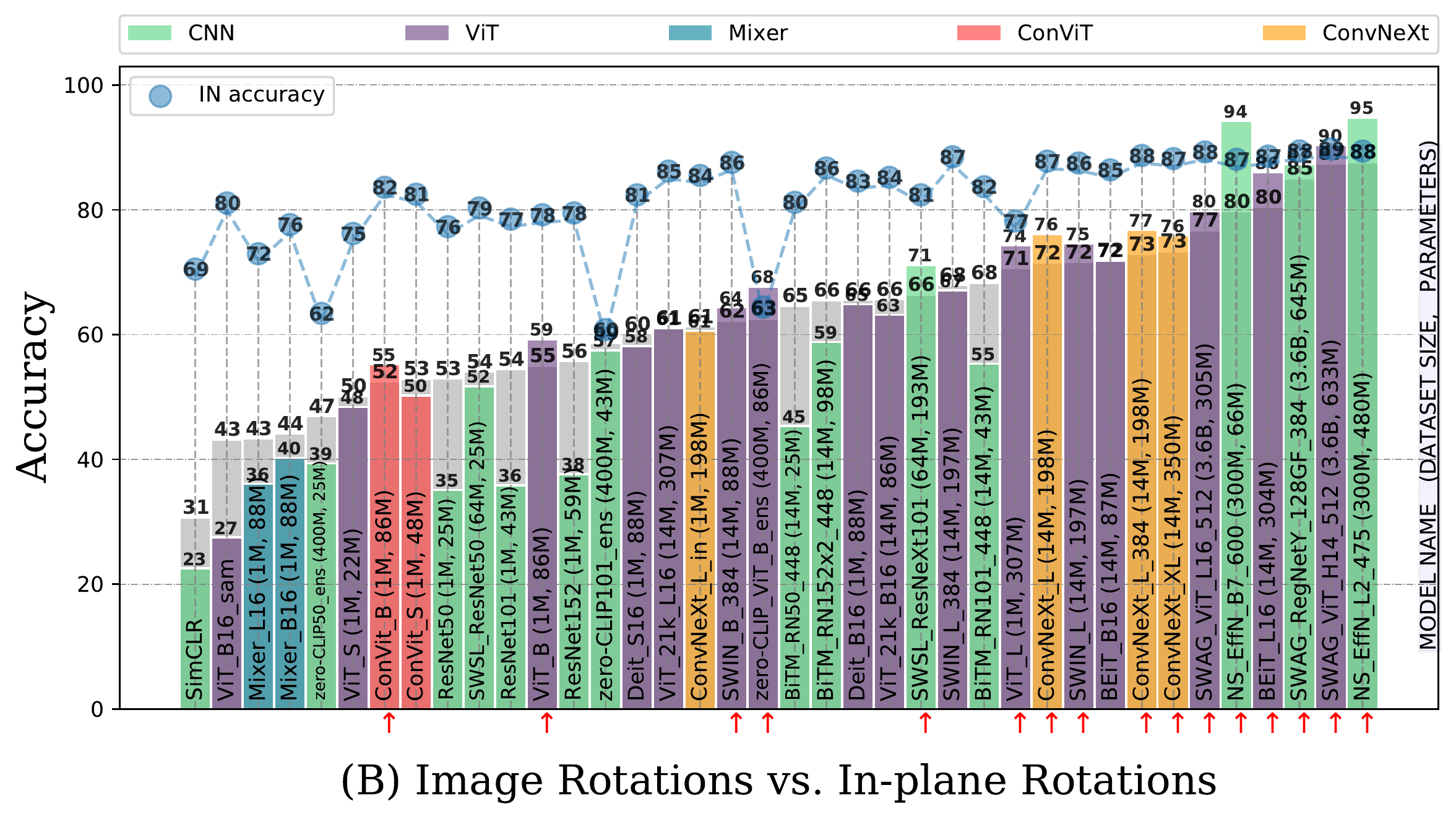}}}$
    
    \caption{\textbf{(A) In-plane rotation accuracy.} Most networks are slightly more robust to in-plane rotations (colored bars) than out-of-plane rotations (grey bars). Exceptions are marked by red arrows. \textbf{(B) Image Rotation.} The best networks benefit from the background image being rotated with the object (colored bars), unlike most weaker networks which prefer an upright background (in-plane condition, grey bars), revealing a difference in strategy between these two groups of networks.}
    \label{fig:in_plane_rotation}
\end{figure}

\paragraph{Which types of rotation are most problematic for deep networks (Fig. \ref{fig:in_plane_rotation})?} We compare the effect of object rotations in the plane of the image (in-plane condition, Appendix \ref{sec:ObjectPoseDataset}), vs. out-of-plane rotations seen in ObjectPose (Fig. \ref{fig:in_plane_rotation}A). We find that both conditions are problematic for all networks, with the out-of-plane condition only slightly worse than the in-plane condition for most networks. We then compare the in-plane condition with simple image rotations, where the background is rotated with the foreground object (Fig. \ref{fig:in_plane_rotation}B). By comparing these two conditions, we find that the best networks on ObjectPose rely on a different strategy than the weaker networks when it comes to incorporating background information. Indeed, weak networks perform better on in-plane rotations than on image rotations, in contrast to the best networks (e.g., Noisy Student EfficientNets and SWAG) which perform better on image rotations. Our interpretation is that weaker networks benefit from seeing features of the background upright, whereas the best networks suffer from the incongruency between the upright background and the rotated foreground object. The interpretation that the best networks suffer from the incongruency of the background is reinforced by the observation that they are the only ones to see an increase in accuracy on ObjectPose when the natural backgrounds are removed altogether (suppl. fig. \ref{fig:bg_rot}A). The interpretation that the weaker networks benefit from seeing the background features upright is confirmed by the observation that their accuracy drops when the background alone is rotated (suppl. fig. \ref{fig:bg_rot}B).\\ 

\begin{figure}[t]
    \centering
        \hspace*{0.5cm}
        \subfigure{\includegraphics[width=0.27\textwidth]{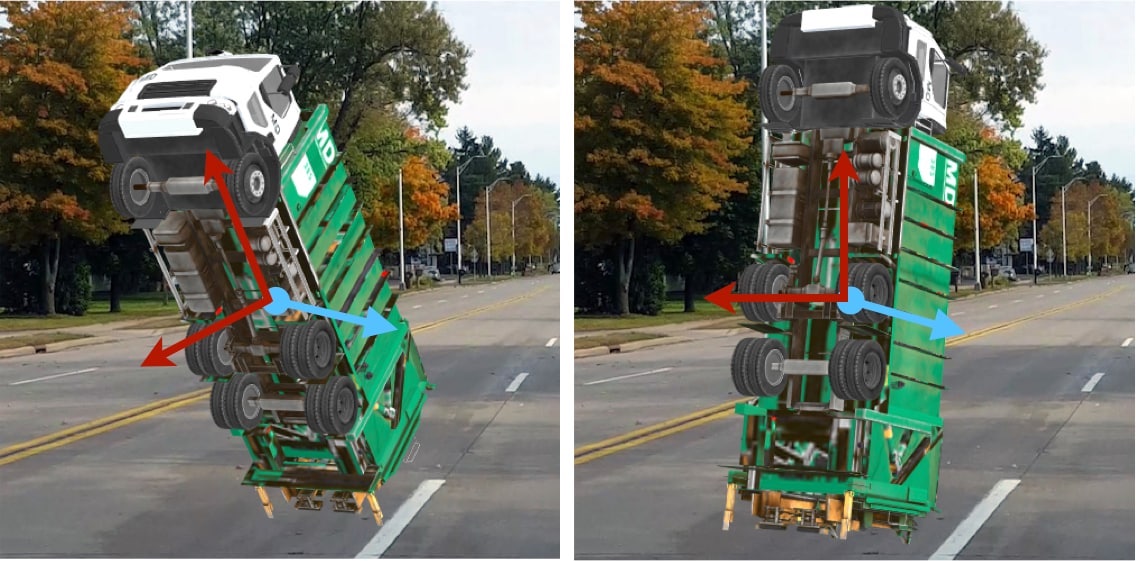}}
        \hspace*{0.7cm}
        \subfigure{\includegraphics[width=0.27\textwidth]{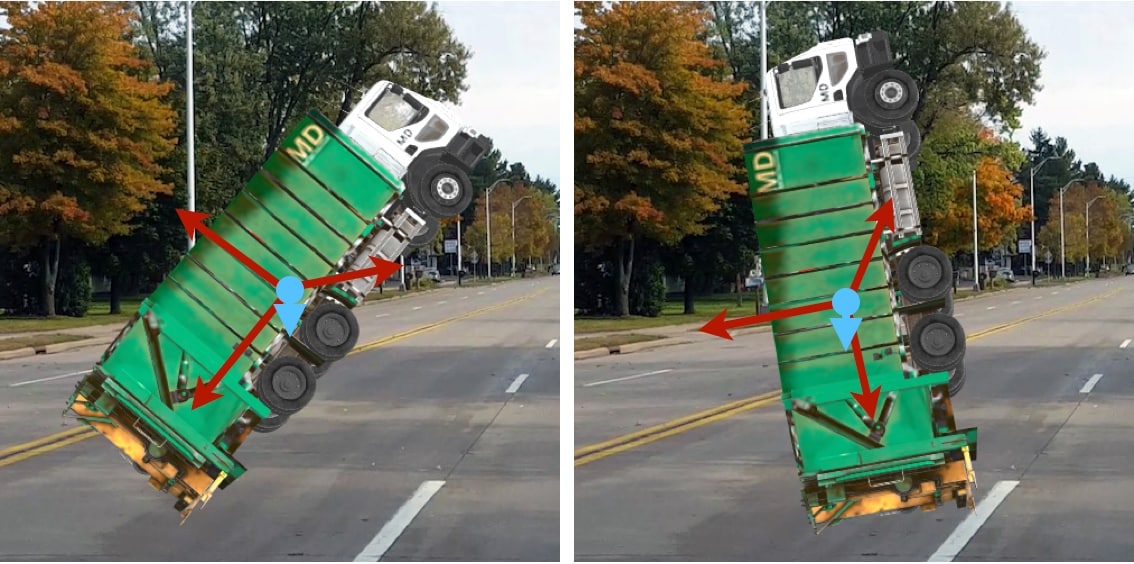}}
        \hspace*{0.7cm}
        \subfigure{\includegraphics[width=0.27\textwidth]{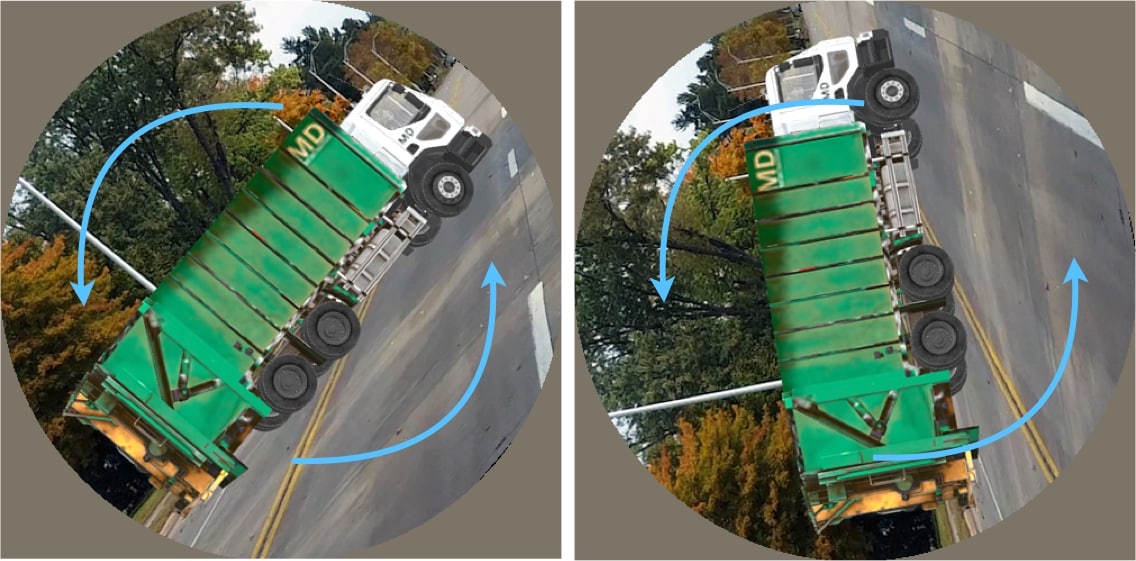}}
    \subfigure{\includegraphics[width=0.325\textwidth]{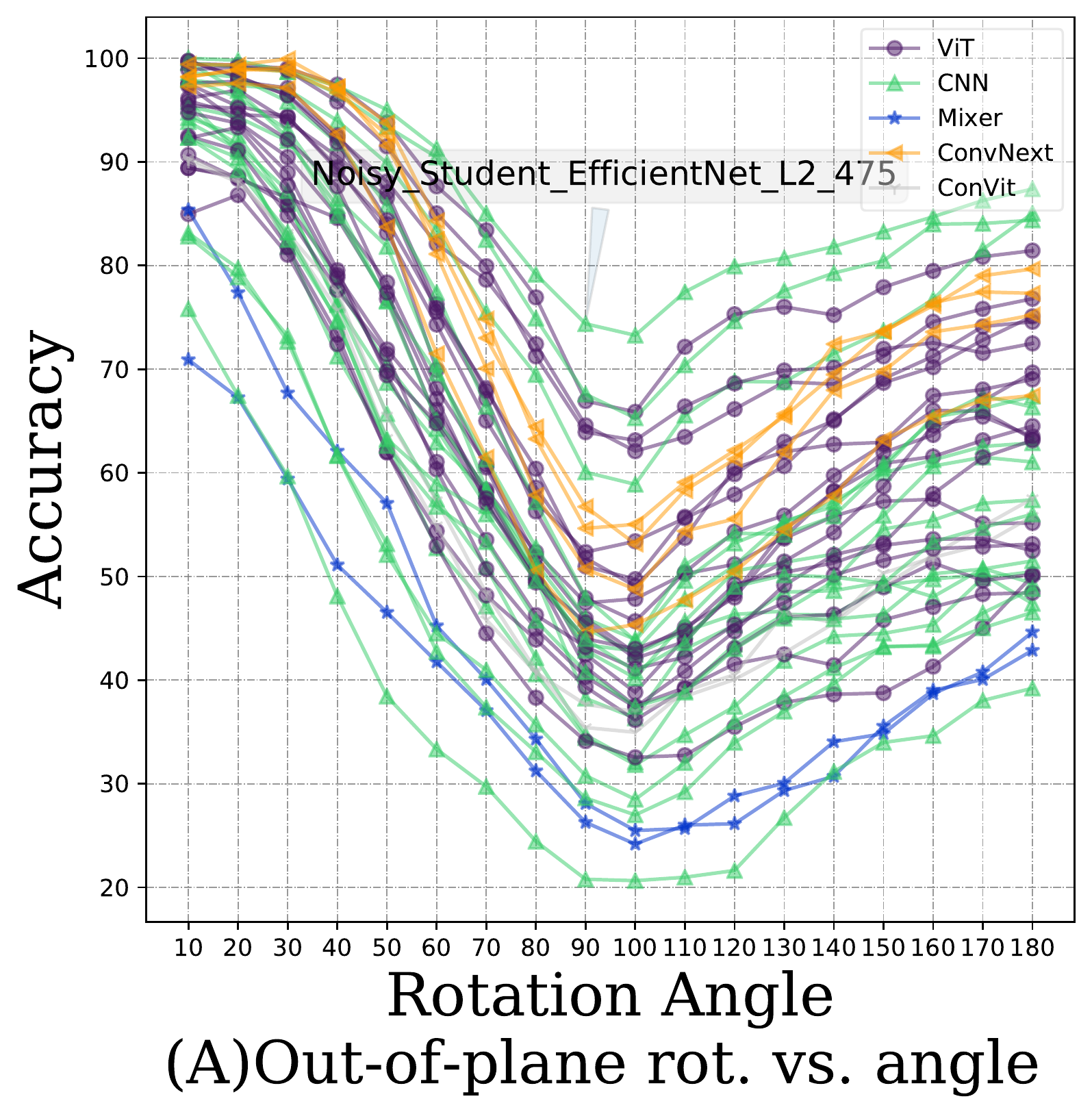}}
    \subfigure{\includegraphics[width=0.325\textwidth]{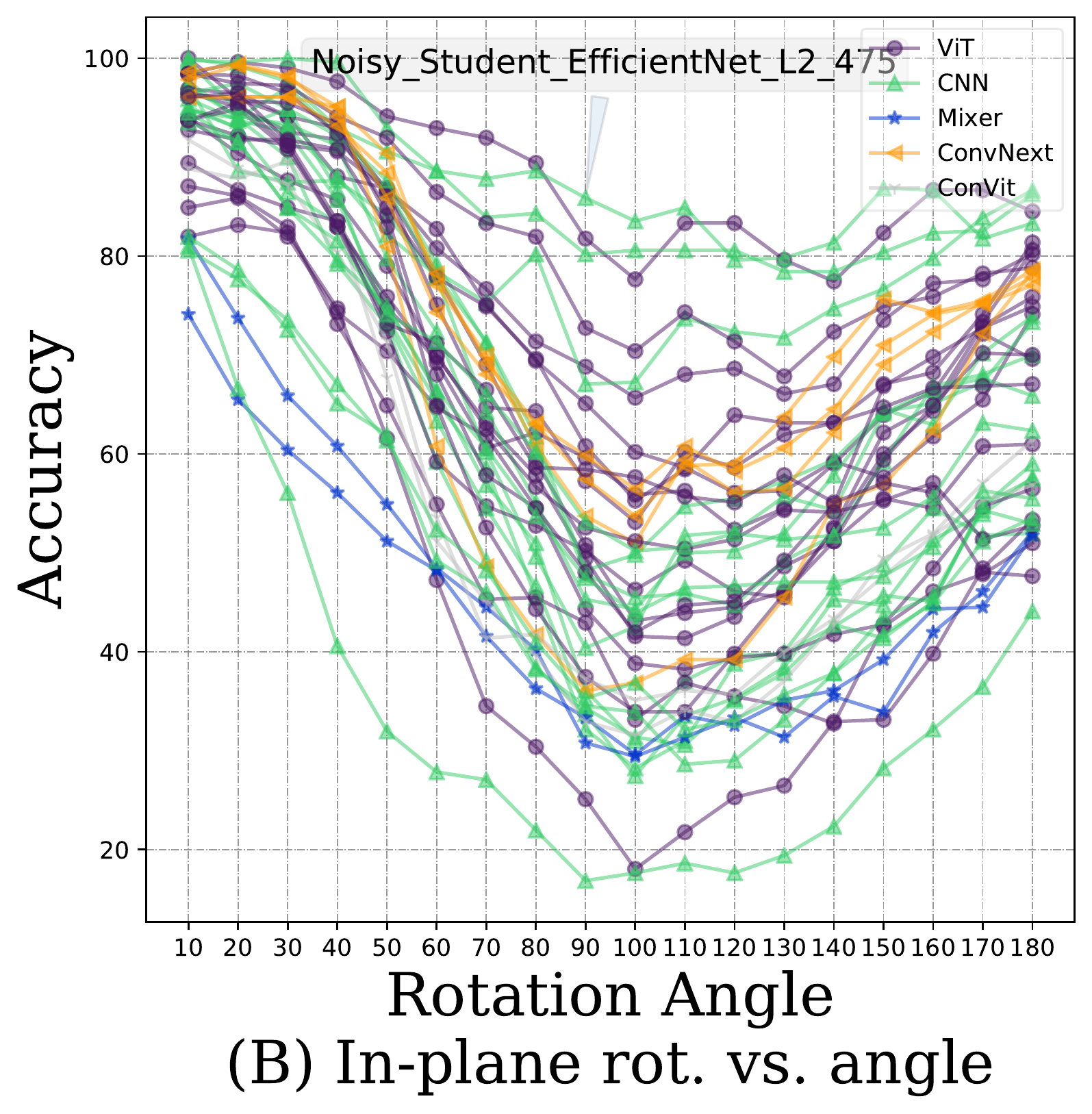}}
    \subfigure{\includegraphics[width=0.325\textwidth]{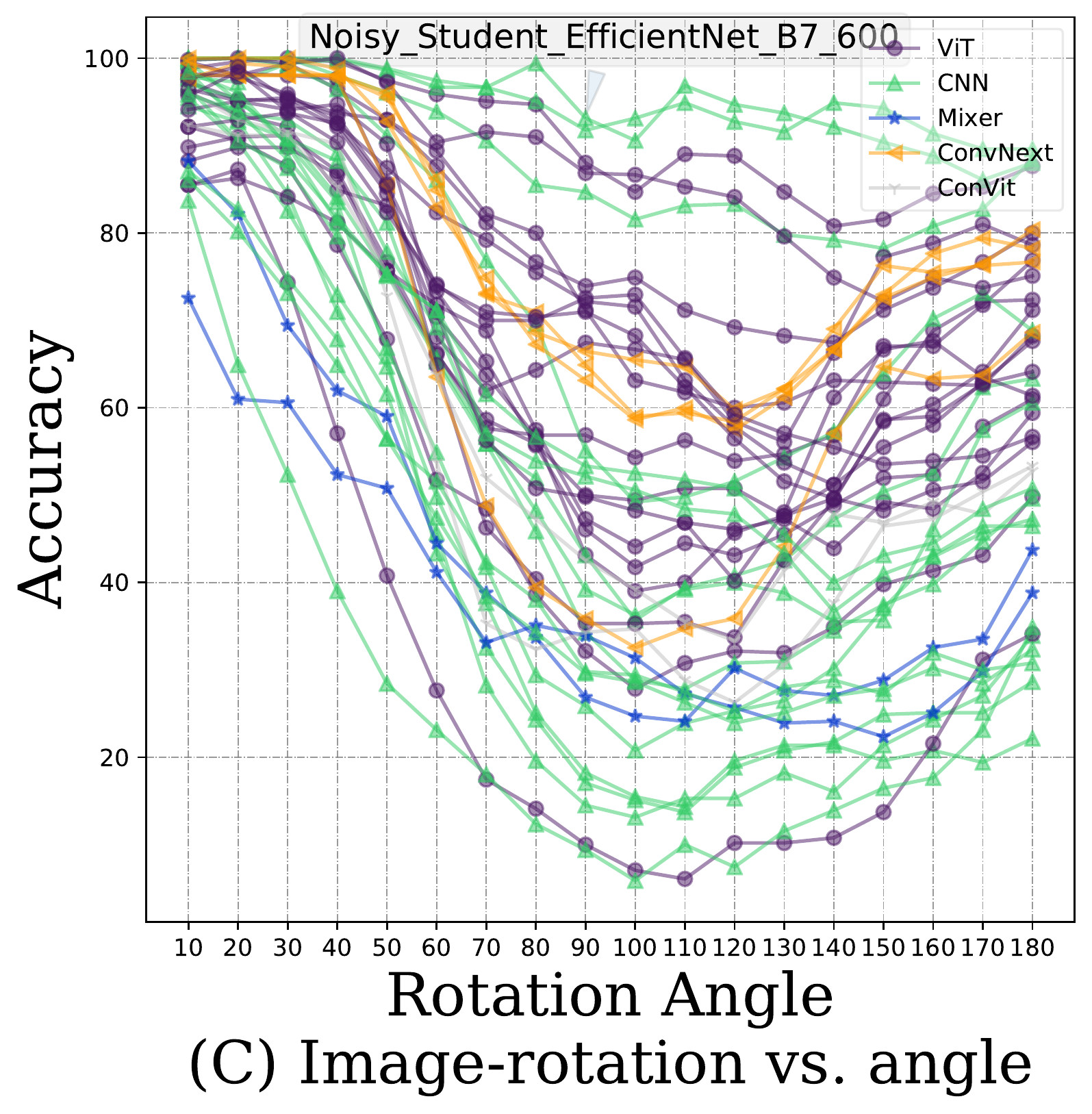}}
    \caption{Accuracy decreases as we increase rotation angle, with a low point at 90°. Networks are less robust to (A) out-of-plane rotations than (B) in-plane rotations and (C) image rotations. Noisy Student EfficientNet models are very robust to image rotations, perhaps for the reason that they were trained with this type of augmentation (RandAugment). Yet they are not completely robust to the two other types of rotations.}
        \label{fig:angle_rotation}
\end{figure}

\paragraph{We next study the relation between network accuracy and rotation angle for different rotation types (Fig. \ref{fig:angle_rotation}).} We find that all networks are most fragile when the object is rotated by 90° in the out-of-plane condition (ObjectPose). We also find that the best networks are more robust across the full range of rotation angles in the in-plane and image-rotation conditions than in the out-of-plane condition. Noisy Student EfficientNets are especially robust to image rotations, which might be explained by the fact that they are trained with image-rotation augmentations. Yet, we see that this type of data augmentation is not enough to guarantee full robustness to in-plane and out-of-plane rotations.

\begin{figure}[t]
    \centering
    
    $\vcenter{\hbox{\includegraphics[width=.325\textwidth]{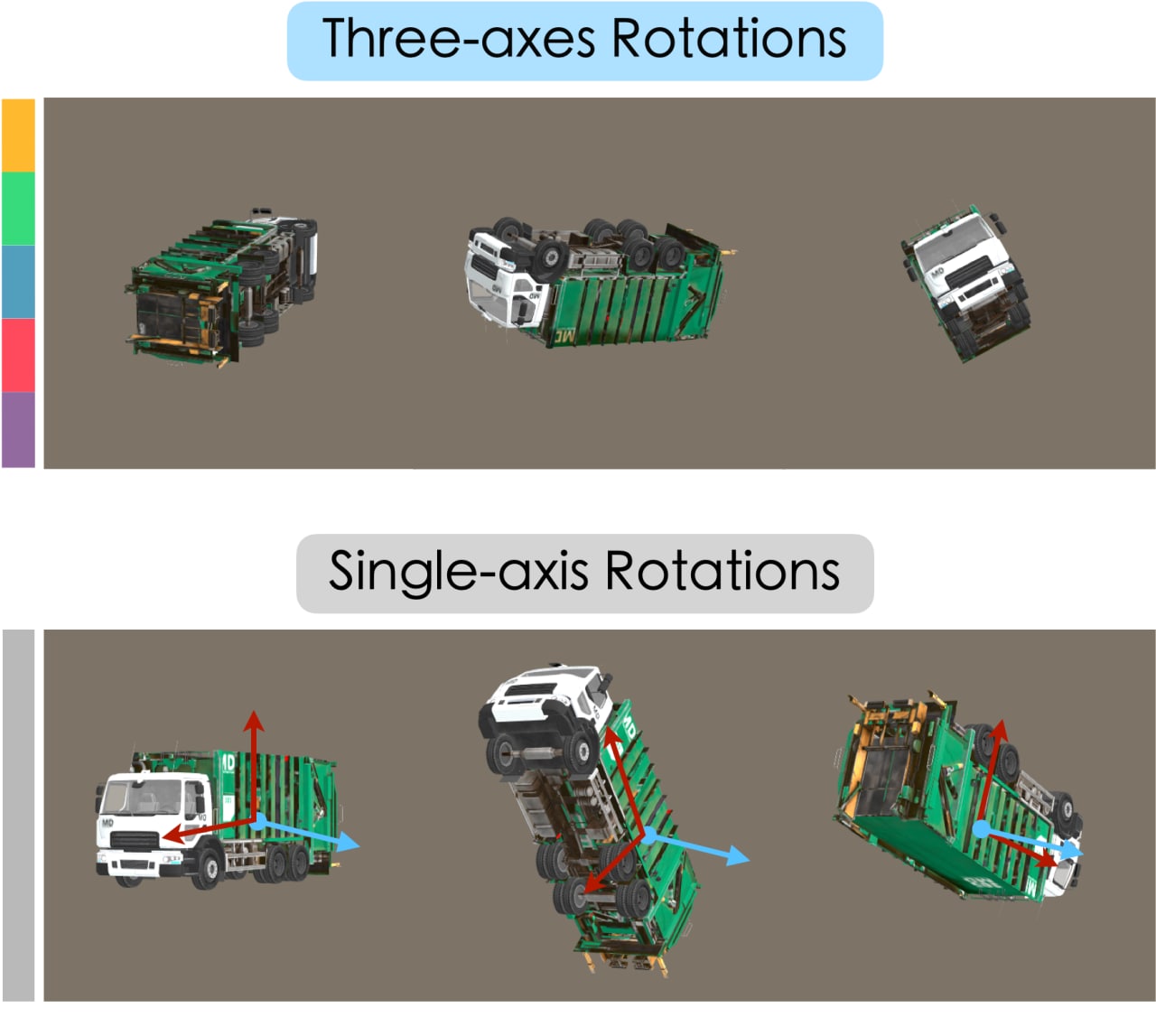}}}$
    $\vcenter{\hbox{\includegraphics[width=.66\textwidth]{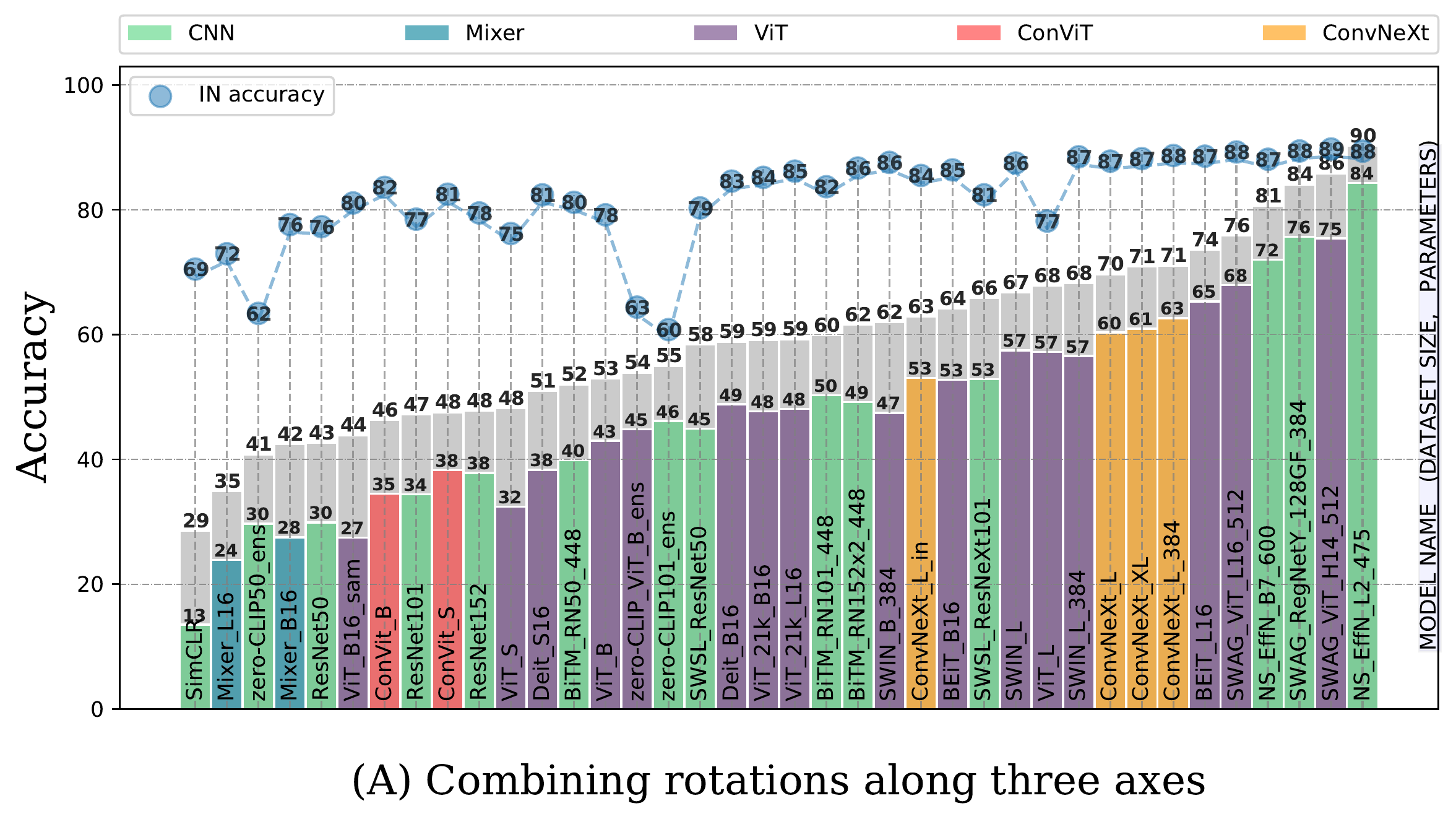}}}$
    
    $\vcenter{\hbox{\includegraphics[width=.325\textwidth]{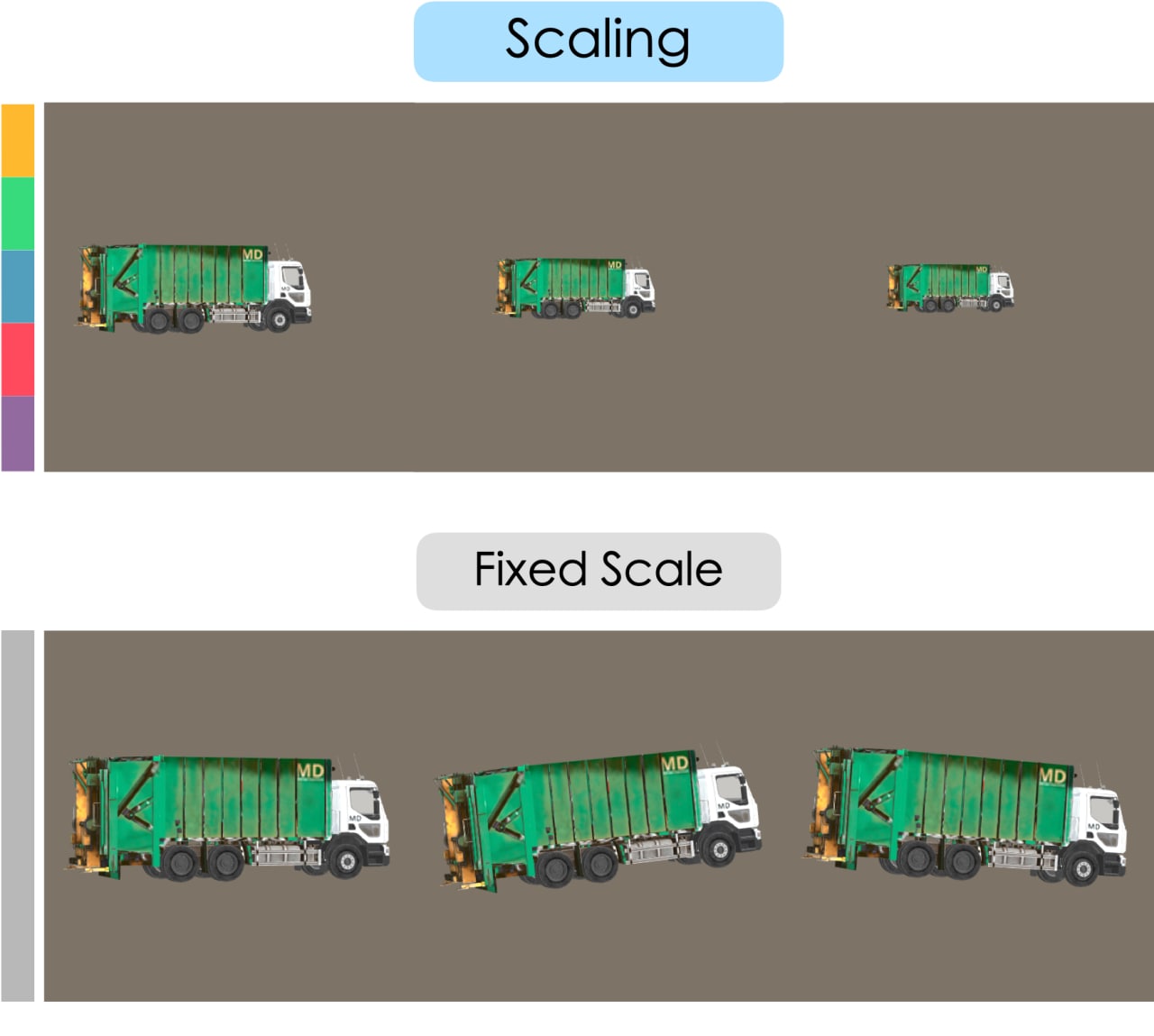}}}$
    $\vcenter{\hbox{\includegraphics[width=.66\textwidth]{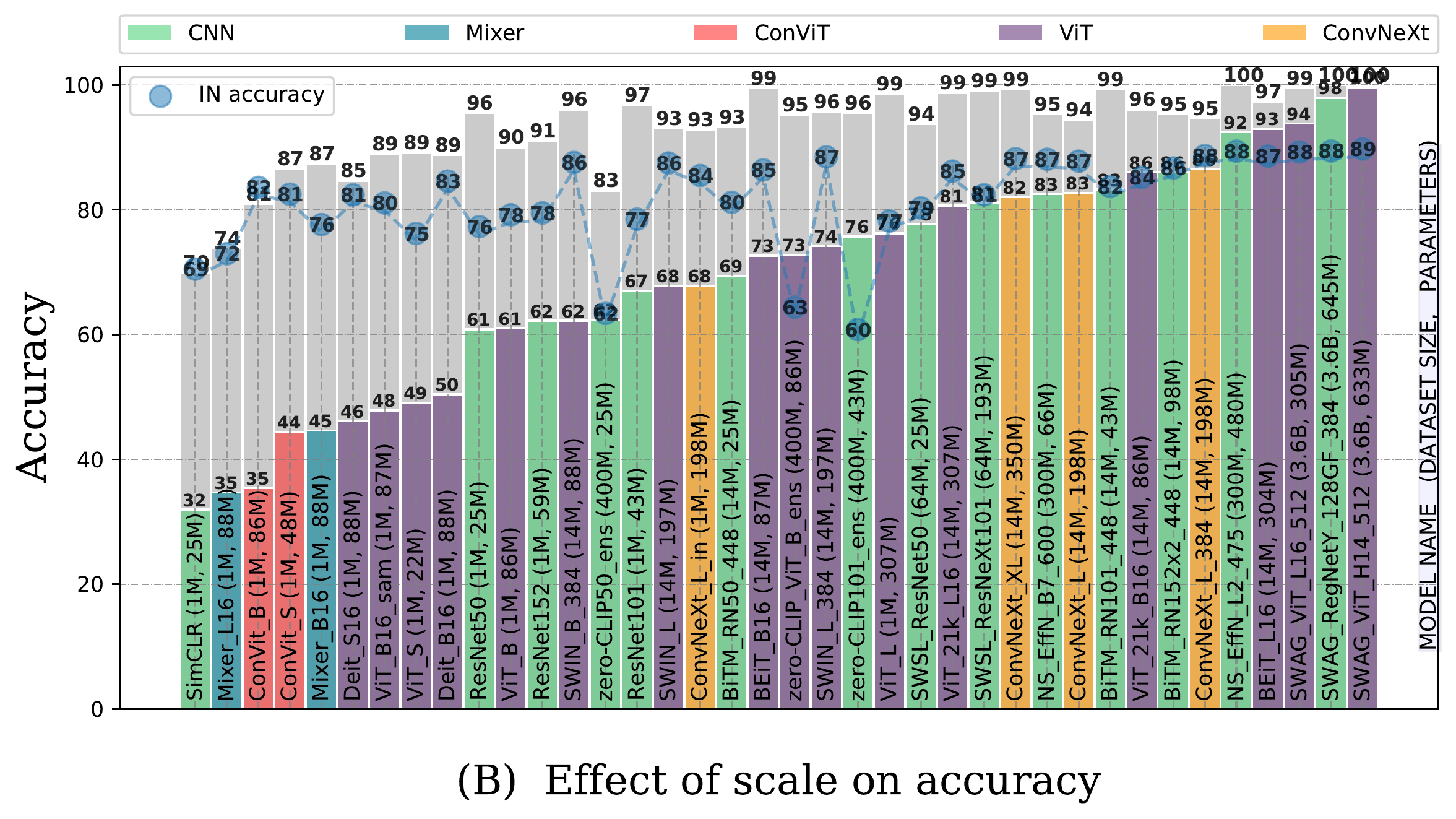}}}$
    
    $\vcenter{\hbox{\includegraphics[width=.325\textwidth]{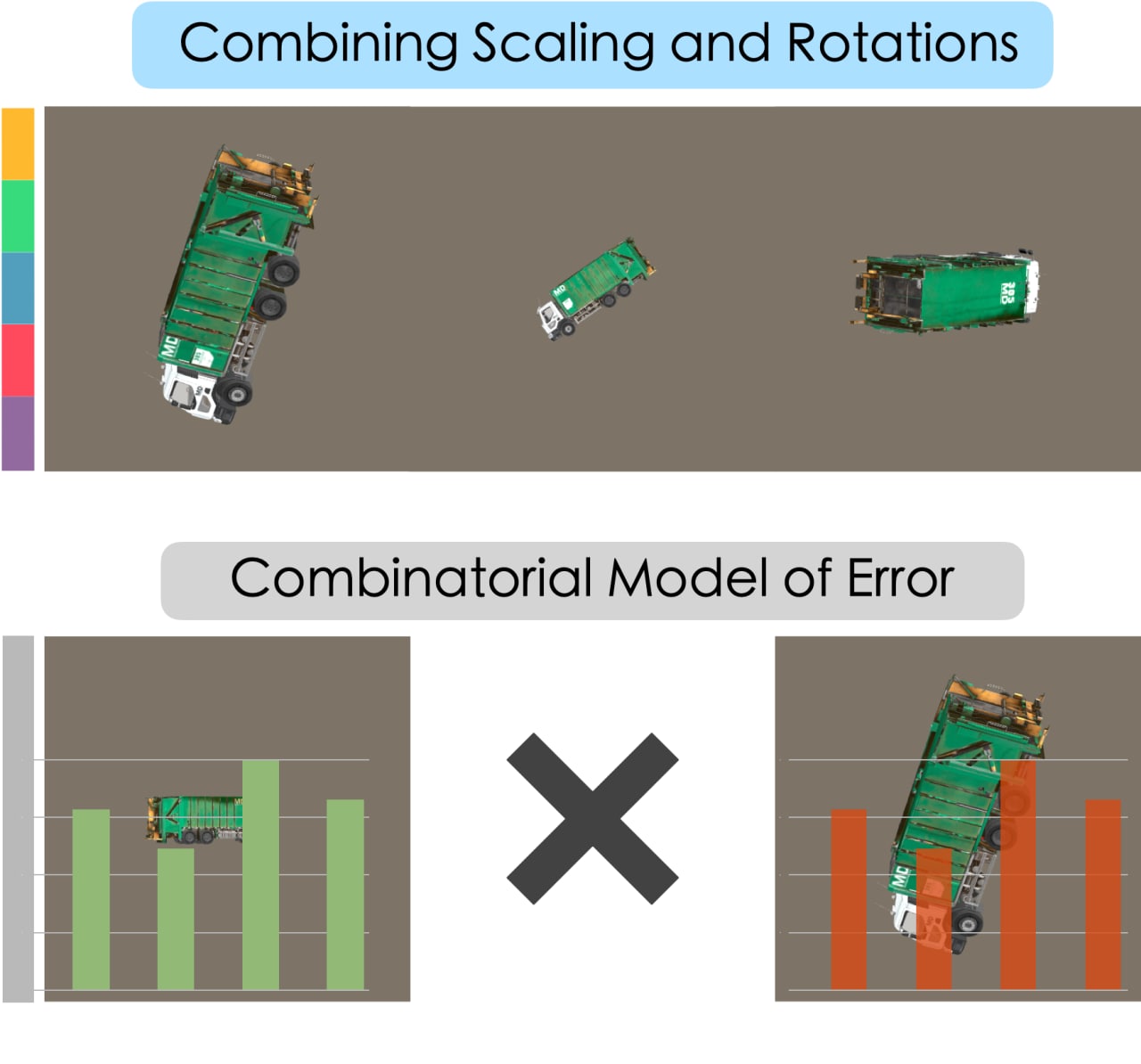}}}$
    $\vcenter{\hbox{\includegraphics[width=.66\textwidth]{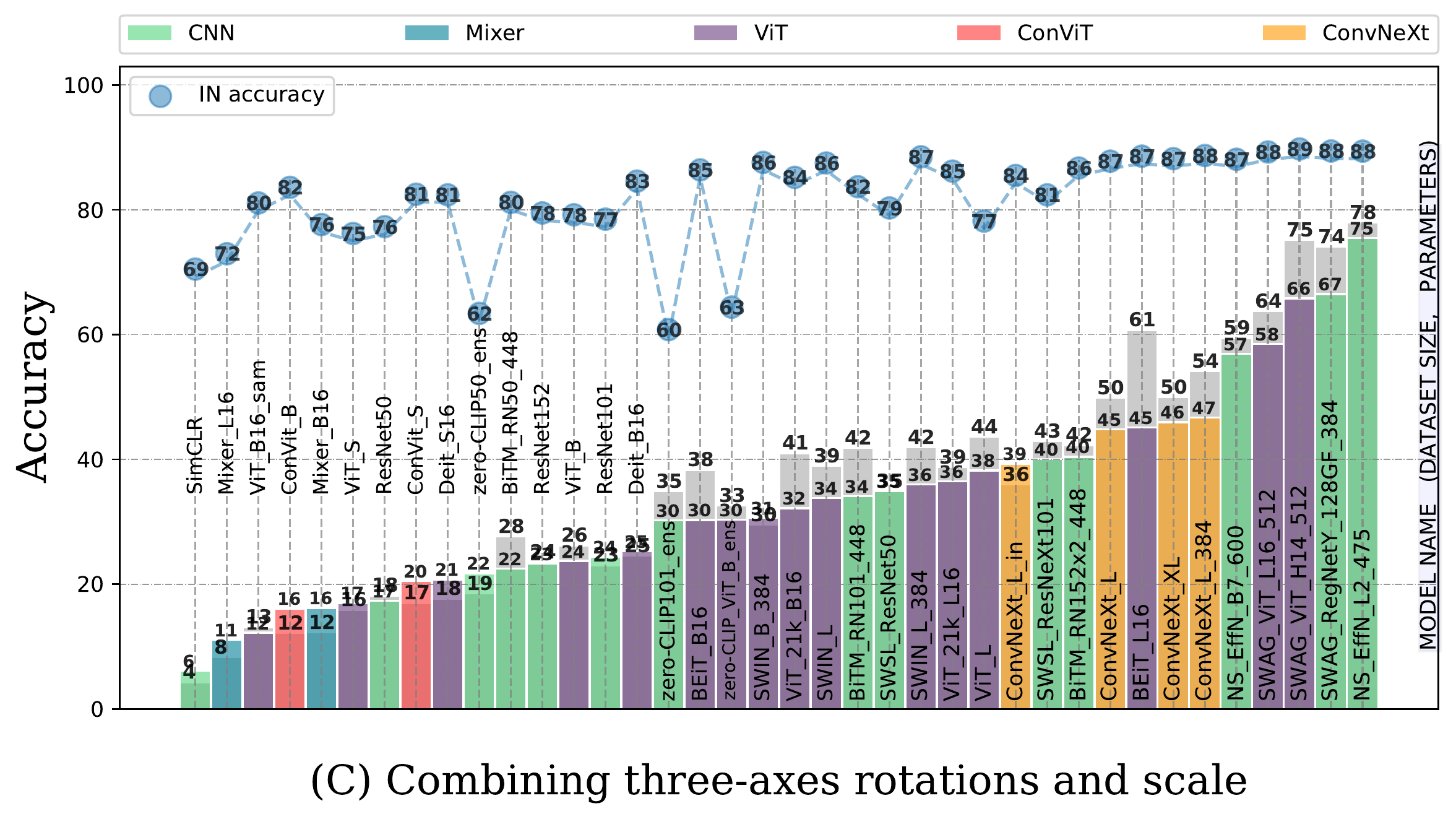}}}$
  
    \caption{\textbf{Effect of combining multiple transformations on performance.} \textbf{(A)} Combining rotations along the three axes ROLL, PITCH and YAW  decreases the accuracy of all networks (colored bars) compared to the single-axis rotation condition (grey bars). \textbf{(B)} Scaling the object size in the image decreases the accuracy severely for most networks (colored bars) compared to the fixed-scaled upright condition (grey bars), but only slightly for the best networks. \textbf{(C)} Combining three-axes rotations and scaling strongly affects the accuracy of all networks (colored bars), with the best network (Noisy Student) at 75\% accuracy only in this condition. The grey bars represent a combinatorial model of error which predicts performance degradation well (see text).}

    \label{fig:scalingandrotation}
\end{figure}

\paragraph{Combining more than one transformation degrades performance of all networks further, as predicted by a combinatorial model of error (Fig. \ref{fig:scalingandrotation}).} We next sought to study how the combination of multiple transformations would affect the performance of the networks. For this set of experiments, we use a grey background for all images in order to avoid complex interferences between foreground and background. First, we try combining rotations along the three axes of rotations, YAW, PITCH, and ROLL together (Appendix \ref{sec:ObjectPose Dataset Generation}). We find that this combination leads to a degradation of performance for all networks (Fig. \ref{fig:scalingandrotation}A) in  a range 6\%-16.5\% compared to the condition where only one axis is rotated at a time. Next we investigate the effect of combining three-axes rotations with scaling (Fig. \ref{fig:scalingandrotation}C). We find that this combination of transformation further degrades the performance of all networks, with an accuracy drop in the range 24.5\%-78.2\% compared to usual poses, larger than the accuracy drop seen for ObjectPose in the range 14.5\%-45\%. We find that a simple combinatorial model of errors (grey bars in Fig. \ref{fig:scalingandrotation}C) recapitulates the accuracy drop well for most of the networks: this model assumes that the probability of the network being correct in the scaled-rotated condition (panel C) is simply the product of the probabilities of being correct in the scaled-only (panel B) and rotated-only (panel A) conditions respectively (see Discussion for implications). 

\begin{figure}[t]
    \centering
    \includegraphics[width=0.95\textwidth]{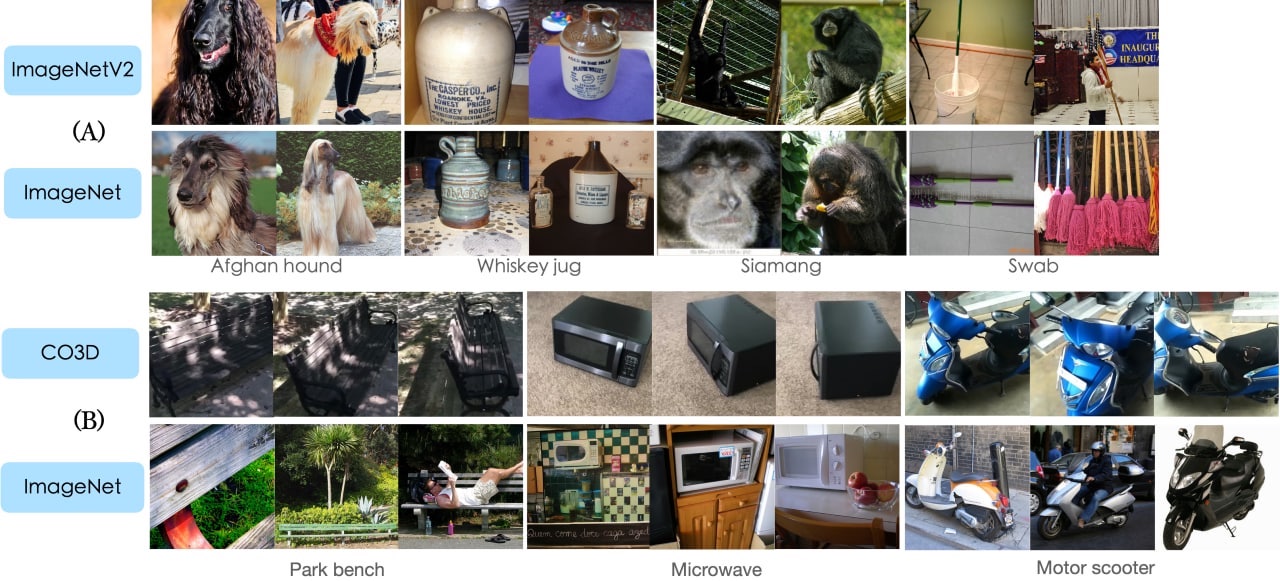}

    $\vcenter{\hbox{\includegraphics[width=.638\textwidth]{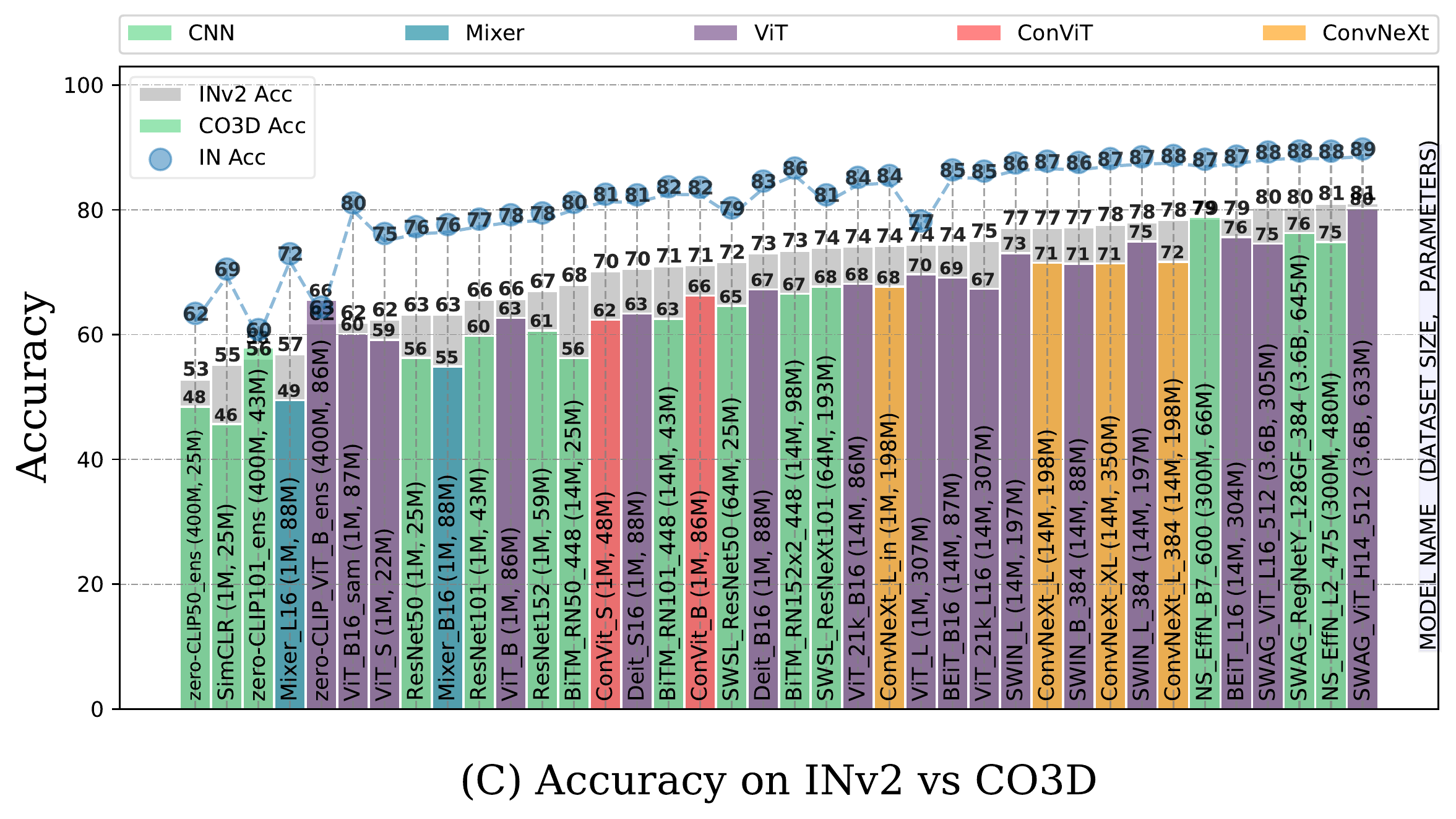}}}$
    $\vcenter{\hbox{\includegraphics[width=.345\textwidth]{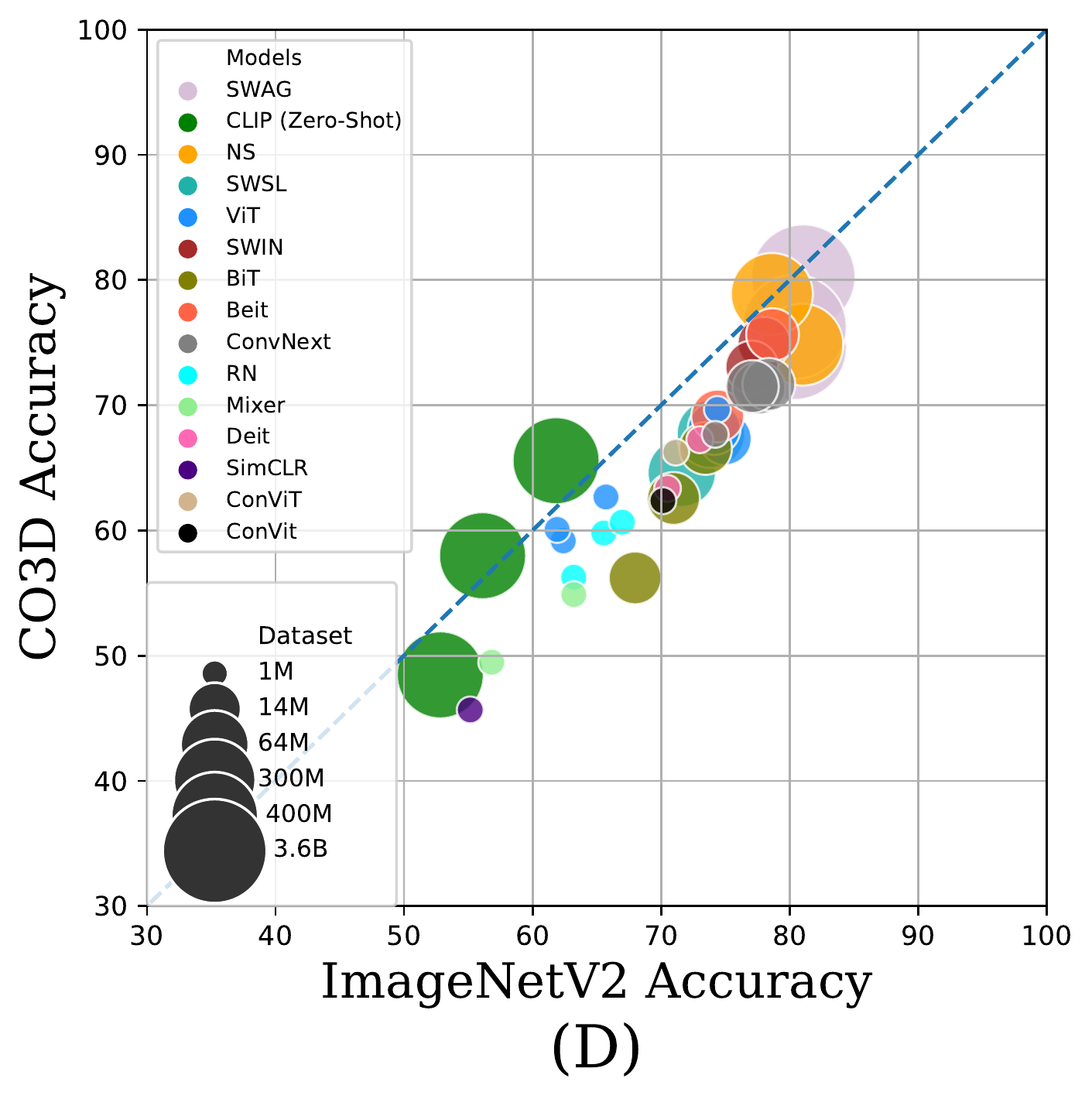}}}$
     \caption{\textbf{Going beyond synthetic datasets.} \textbf{(A)} Comparison of images from ImageNet and ImagenetV2, a near-distribution dataset with mostly canonical views. \textbf{(B)} Comparison of images from ImageNet and CO3D, a dataset of objects in unusual views. \textbf{(C-D)} Most networks perform worse on CO3D (colored bars), a dataset exploring various object views, than on ImageNetV2 (grey bars), which mostly presents objects in their usual canonical views (average accuracy drop of 5.2\% between the two datasets). ImageNetV2 is a fairer comparison benchmark to CO3D than ImageNet as there was no overfitting it.
     }
    \label{fig:co3d_results}
\end{figure}

\paragraph{How well do our findings on synthetics datasets transfer to real-world datasets (Fig. \ref{fig:co3d_results})?} In an attempt to go beyond synthetic datasets, we explore the robustness of our collection of networks to a dataset of real objects filmed from various points of view, the \emph{Common Objects in 3D Dataset} (CO3D) \citep{co3ddataset} (App. \ref{sec:co3d}). Originally designed for 3D reconstruction and new-view synthesis tasks, this dataset was collected by workers turning around and filming common object from their environment. We sample 1000 images from each of 10 categories common to CO3D and ImageNet, to get a total of about 10,000 images. We then estimate the performance of our collection of networks on these images (Fig. \ref{fig:co3d_results}). As a point of comparison, we measure the accuracy of networks on images of the same object categories taken from ImageNetV2 (App. \ref{sec:imagenetv2}), a datatset where objects are mostly presented in their usual canonical view. ImageNetV2 can be seen as a fairer comparison benchmark to CO3D than ImageNet, as networks were not overfitted to the the exact statistics of ImageNetV2 \citep{recht_imagenet_2019}. We observe an average accuracy drop across networks of 10.4\% on ImageNetV2 over ImageNet, and an average accuracy drop of 5.2\% on CO3D over ImageNetV2. However, four networks trained on very large datasets perform nearly as well on both datasets: CLIP-RN-101, CLIP-ViT-B/16, Noisy Student EfficientNet-B7, and SWAG-ViT (Fig. \ref{fig:co3d_results}B). In summary, the variety of object views seen in CO3D seems less problematic for our collection of networks than the views from of our synthetic dataset ObjectPose. This discrepancy could be due to the fact that although the workers turn around the objects in CO3D, they do not necessarily explore all the unusual views that we can explore in our synthetic dataset.

\section{Discussion}

Probing the robustness of deep networks to objects in unusual poses is interesting for multiple reasons. First, this type transformation changes the global structure of the image, which might represent a bigger challenge for deep networks than local image distortions. Second, data-augmentation strategies do not provide an easy fix to the problem of generalization to unusual poses. Indeed, a classical mitigation strategy in deep learning consists in augmenting the dataset with the out-of-distribution case that one would like to become robust to, effectively making that case in-distribution. For most image distortions, this augmentation strategy is easy to implement. But for objects in unusual poses, one would need to collect a very large number of images of objects in unusual poses, which is practically challenging, especially at the scale needed to compete with the best current networks (e.g., Noisy Student is trained on JFT-300M comprising 300 million images). Alternatively, one could synthetically generate such a dataset with 3D models of objects, but this would require a very large database of high-quality 3D models which is currently lacking. 

A striking result of our study is that very large networks trained on very large datasets (e.g., Noisy Student trained JFT-300M, SWAG trained on 3.6B Instagram images) are quite robust to unusual poses. However, a careful visual inspection of the errors made by these networks reveal that they do not yet meet the robustness of the human visual system. A thorough comparison to human would be useful to estimate the fraction of the errors that are due to the images themselves not containing enough information about the class (e.g., a jeep seen from the bottom could be any sort of car) vs. out-of-distribution generalization errors that a human would not fall into (e.g., a cannon seen at a 90° angle is confused by the network with a harp). In future work, we plan to establish this human benchmark on our custom dataset ObjectPose in order to precisely quantify the gap in robustness between humans and networks.

Having the possibility to feed language qualifiers to deep networks is a promising avenue to allow them to adapt to all sorts of out-of-distribution situations. For example in our case, one could indicate to the network to look for an object in an unusual pose. In our experiments with CLIP (Appendix \ref{sec:clipexperiments}), we do see a slight improvement in performance when prompting the network with textual indications that the object to detect is in an unusual pose (e.g., "an upside-down forklift"). Although promising, the improvement in accuracy we observe remains marginal for this model. 

When combining object rotations and scaling, we find that a combinatorial model of error---which assumes that the probability of the network being correct on the combination of transformations is equal to the product of the probabilities of it being correct on each respective transformation---accounts well for the degradation of performance of most networks. It would be interesting to study whether networks' performance degrade according to this combinatorial model when combining even more transformations, such as translations, texture-removals etc. Indeed, an implication of this combinatorial model of error is that networks should be brittle in the face of a large combination of transformations, as each factor of variation adds its own source for potential errors. 

It is an open question how the human visual system builds robustness to unusual poses. When performing mental rotation, a task consisting in comparing two 3D shapes in different orientations, human subjects take a time to respond that scales linearly with the angle of rotation between the two shapes \citep{shepard_mental_1971}. A similar phenomenon happens when recognizing every-day-life objects in unusual orientations \citep{jolicoeur_time_1985}: the recognition time is again linear with the angle of the object with respect to its upright pose. These observations are striking as they suggest that different mechanisms take place in the brain compared to feed-forward deep networks, where the processing time is fixed and does not depend on the complexity of the input image. We hypothesize that recurrent mechanisms may play a key role in recognizing objects in unusual poses in the brain, as there is mounting evidence that such mechanisms are critical for recognizing images that are challenging to deep networks \citep{kar_evidence_2019, bonnen_when_2021}.

\section*{Acknowledgements and Author Contributions}
We thank \cite{strikewithapose} and \cite{closingthegap} for making their codebases publicly available. We thank Robert Geirhos, Michael Alcorn, Anh Nguyen, and the OOD group at Meta AI (led by Pascal Vincent) for their kind feedback and suggestions to improve this work. We also thank Google for providing GCP credits for the project. Author contributions: analyses and visualizations performed by A.A.; conceptualization, interpretation and writing by A.A. and S.D.


\cleardoublepage

\vspace{4mm}
\appendix{\Large{\textbf{Appendix}}}
\vspace{3mm}

\section{Datasets} \label{sec:datasetdistcription}

In this section, we describe in detail the  three datasets we use in our experiments: (1) our custom dataset ObjectPose, which measures the robustness of networks' to unusual object poses, (2) our adaptation of the CO3D dataset \citep{co3ddataset}, which measures the generalization capability of networks' on various natural poses, (3) ImageNetV2 \citep{imagnetv2}, which measures the generalization capability of networks' to a similar distribution than ImageNet while preventing overfitting effects to ImageNet's exact statistics.


\subsection{ObjectPose Dataset} \label{sec:ObjectPoseDataset}
To measure the robustness of the models to  unusual object poses, we generated a synthetic dataset of objects in unusual poses, ObjectPose. The dataset contains around 27,540 images of objects in unusual poses. It covers 17 of the 1000 ImageNet ILSVRC-2012 \citep{russakovsky2015imagenet} image classification task classes. Each image in the dataset contains only one ImageNet object centered in the image. The dataset was created by rendering high-quality 3D objects on top of a background image following the pipeline introduced in \citep{strikewithapose} and building on their codebase made available online. The images are generated by rotating the 3D object along one axis to put it in an unusual pose and then rendering the object on top of the background. The 17 object categories in ObjectPose are `airliner', `barber chair', `cannon', `fire engine', `folding chair', `forklift', `garbage truck', `hammerhead shark', `jeep', `mountain bike', `park bench', `rocking chair', `shopping cart', `table lamp/ lampshade', `tank', `tractor', and `wheel barrow'. To focus more on the effect of the poses we consider both `table lamp' and `lampshade' as correct labels for table lamp images in ObjectPose. The dataset is designed carefully so that the images are easy to classify when the object is upright. 

\textbf{3D objects}: We picked the 3D objects very carefully so they all have high quality. To make sure that the 2D images generated from the 3D objects have high quality and look realistic at the same time, we first manually excluded all the 3D objects that don’t give realistic 2D images, then we run the evaluation experiment explained in section (\ref{sec:ObjectPoseevaluation}) to make sure that the images have sufficient quality. We excluded all the 3D objects that do not pass this evaluation. Section (\ref{appendix:3dobjects}) provides more details about the 3D objects we used.\\

\textbf{Background images}: To increase the diversity of the dataset we used three different backgrounds with each object. Each time, we render the 3D objects on top of one of these backgrounds. Two of the backgrounds are natural background images chosen manually from the internet. We chose each background image carefully so that it matches the object and does not contain any of the ImageNet ILSVRC-2012 dataset objects. Following \citep{strikewithapose}, we chose the third background to be grey background with all its RGB pixel values equal to (0.485, 0.456, 0,406). These values correspond to the average pixel color of ImageNet images.


\subsubsection{ObjectPose Dataset Generation}\label{sec:ObjectPose Dataset Generation}
The ObjectPose dataset was generated using a 3D renderer following the pipeline introduced in \citep{strikewithapose}. For each 3D object we choose three rotation axes. These axes are orthogonal to the 3D object and defined relative to it. For each of the three axes, we create 180 images that cover the 360 rotation degrees by rotating the 3D object along this axis.\\

\begin{figure}[t]
    \centering
    \includegraphics[width=1.0\textwidth]{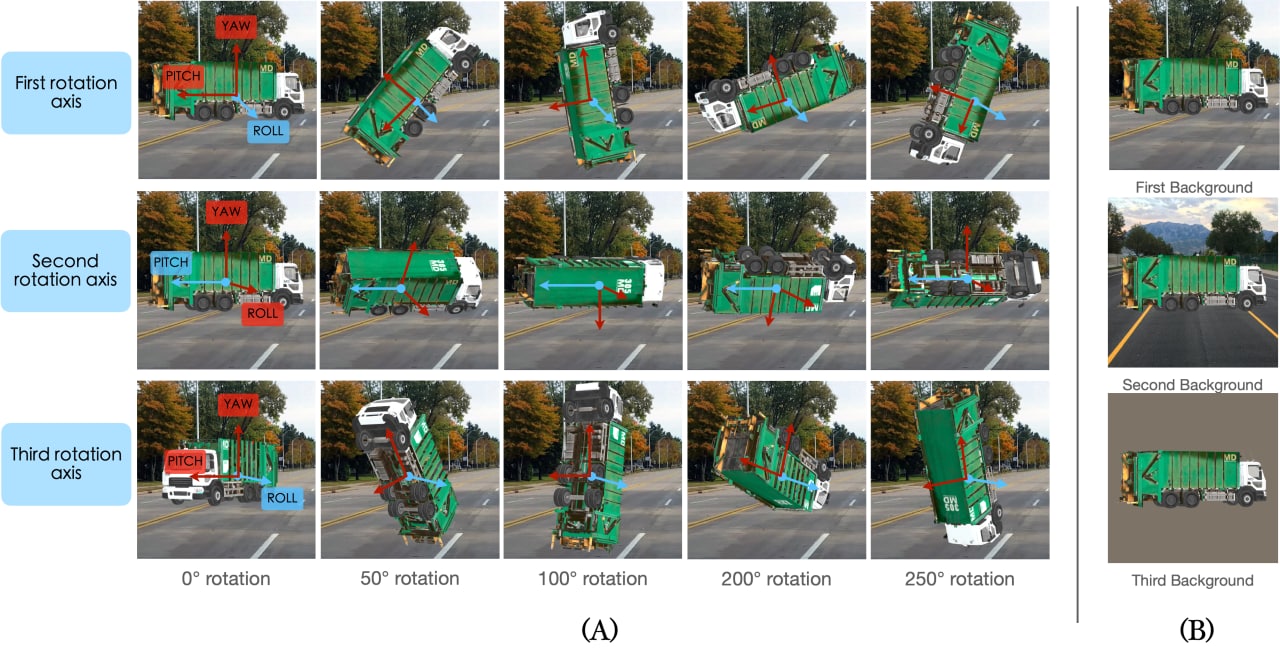}

    \caption{\textbf{ObjectPose dataset generation. (A)} The figure shows how the object is rotated around the three different axes. First, we manually align the object to be in a natural pose  with the background (\emph{First Column}). This defines the relative zero rotation angle with respect to the initial position. \emph{Rows}: Each row shows the rotation of the object along a different axis. The rotation axes are orthogonal to the 3D object as shown with the arrows. The blue arrow corresponds to the rotation axis in each row. \textbf{(B)} Three different backgrounds used with the truck object, two are natural images, and the third one is grey.
    }
    \label{fig:ObjectPosedescription}
\end{figure}

Rotating the object is done by using three 3x3 rotation matrices $R_{r}$, $R_{y}$, $R_{p}$. Each matrix is defined by one rotation angle $\phi_{y}$, $\phi_{r}$, or $\phi_{p}$ respectively. These matrices correspond to the rotation around the \emph{YAW}, \emph{ROLL}, and \emph{PITCH} axes respectively. These are the axes orthogonal to the 3D object as shown in Fig. \ref{fig:ObjectPosedescription}. To generate the images, first, we manually choose the initial rotation angles $\phi_{y}^{\circ}$, $\phi_{r}^{\circ}$, and $\phi_{p}^{\circ}$ that place the object in a usual pose, then each time we increase the rotation angle by two degrees along one axis and render the 3D object on top of the background to get a 2D image. In order to better blend the object in its initial pose, we also perform 3D translation. We adjust the three translation parameters $x$, $y$, and $z$ manually for each 3D object. We choose $x$ and $y$ to be near zero so that the object is centered and reasonably placed in the 2D image. The z parameter controls how close the object is to the camera. We set the value of $z$ manually for each object.\\

Let R be a 3D renderer that takes as input a set of parameters $\Theta = \{\phi_{y}, \phi_{r}, \phi_{p}, x, y, z, Obj, C, BG, L $\} and outputs 2D image following the pipeline in \citep{strikewithapose} defined by equation (\ref{eq:redererequation}). 

\begin{equation}\label{eq:redererequation}
I=R(\phi_{y}, \phi_{r}, \phi_{p}, x, y, z, Obj, L, BG)
\end{equation}

Where $I \in \mathbb{R}^{HxWx3} $ is the 2D image. $H$ and $W$ represent the dimensions of the image. Here $Obj$, $BG$, $L$ are the 3D object, the background image, and the lighting parameters respectively. The lighting parameter $L$ is used to control the ambient light and the directional light in the image. We set the ambient light intensity manually for each object separately, while keeping the directional light oriented at (0.3, 1.0, 1.0) with its intensity fixed at 0.9 according to the settings defined in \citep{strikewithapose}.\\

For each of the 17 3D objects there are 180*9 images (9 because we have three backgrounds and three rotation axes). This gives a balanced dataset with a total of 27,540 images. In order to discard usual poses, we then exclude all images with an angle smaller than 10° with respect to the upright pose, keeping only the range [11°, 349°]. All the images have a size of 500x500 pixels. Samples from the dataset are shown in Fig. \ref{fig:ObjectPosesamples} in the appendix.\\

\textbf{Choosing the rotation axes:} After placing the object in the initial position (the first column in Fig. \ref{fig:ObjectPosedescription}), we choose one of the  \emph{YAW}, \emph{ROLL}, or \emph{PITCH} axes to rotate the object along it. When choosing the rotation axis, we consider only the axes that result in unusual poses. For example, rotating the truck in Fig. \ref{fig:ObjectPosedescription} around the \emph{YAW} axis does not result in the unusual poses we want, that is why we do not rotate the truck around the \emph{YAW} axis for this specific object. The  truck object is rotated along the \emph{ROLL} axis (with two different initialization), and along the \emph{PITCH} axis. This gives us three different rotations for this object (Fig. \ref{fig:ObjectPosedescription}). We follow the same process for generating the images from all the 3D objects. The kind of rotation we follow to generate the dataset is considered as out-of-plane rotation. 

In addition to the main ObjectPose dataset, we created five additional datasets. Each one measures the robustness of networks to different conditions:\\

\begin{figure}[h]
    \centering
    \includegraphics[width=0.9\textwidth]{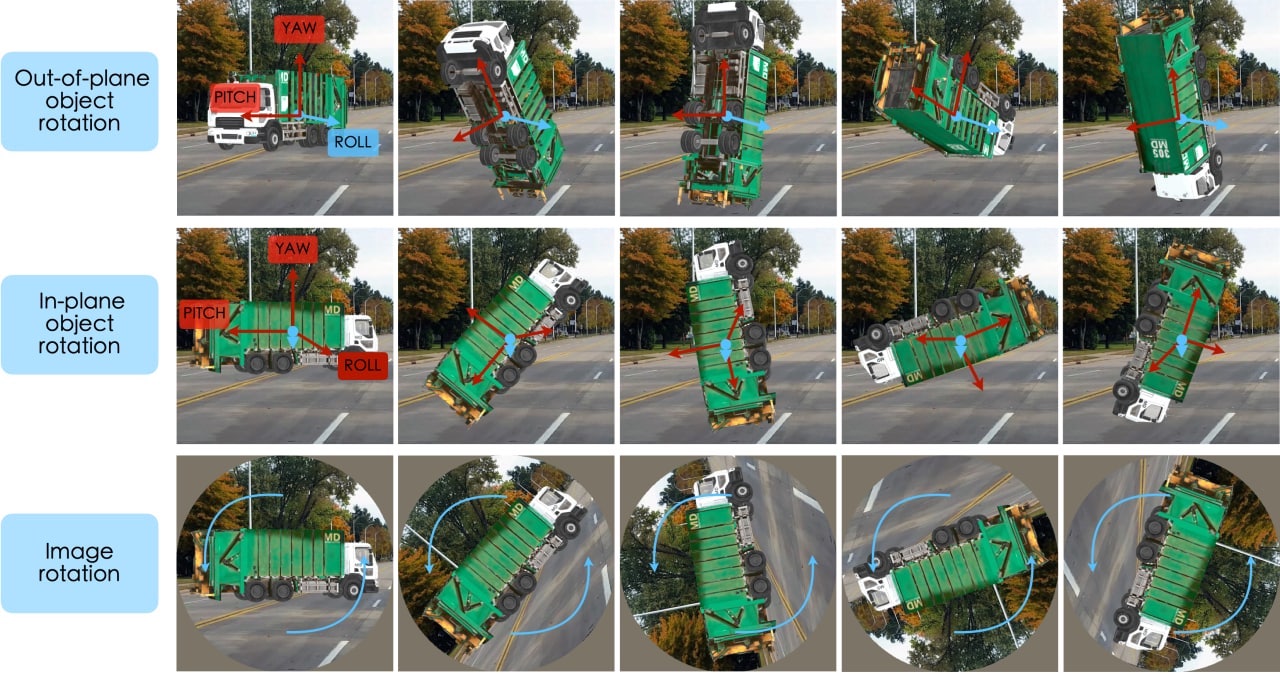}
    \caption{\textbf{Out-of-plane vs In-plane vs 2D image-rotation}: For out-of-plane rotations (ObjectPose), the rotation axis is orthogonal to the object. The \textbf{in-plane rotation} is similar to rotating the 2D image (image rotation) while fixing the background, such that we always see the same view of the object. For the \textbf{image rotation}, we rotate the 2D image itself, and the corners are filled with grey to avoid border-cropping effects.
    }
    \label{fig:ObjectPoserotation}
\end{figure}

\textbf{Image-rotation dataset:} Here we take one 2D image from the ObjectPose dataset with the object on the manually chosen usual pose, then we generate 180 images from the image by rotating it by 360° with a 2° step. To avoid border-cropping effects while rotating the image, we filled the image corners with grey color using a circular crop. Fig. \ref{fig:ObjectPoserotation} shows how images are generated.\\

\textbf{In-plane rotation vs out-of-plane rotation:} Another kind of rotation that we are testing is the in-plane rotation. The in-plane rotation is similar to the image rotation by the fact that for both of them the object is rotated around the same axis---the axis orthogonal to the screen. The difference is that for in-plane rotations, we fix the background in its upright position, in contrast to the 2D-image rotations. Comparing the out-of-plane rotation (ObjectPose) to the in-plane rotation, we notice that the background is fixed for both, while the rotation axis is different. For the out-of-plane case, the rotation axis is orthogonal to the 3D object, but for the in-plane rotation case, the rotation axis is orthogonal to the 2D image (the screen). Fig. \ref{fig:ObjectPoserotation} describes the in-plane rotation visually. \\

\begin{figure}[h]
    \centering
    \includegraphics[width=0.9\textwidth]{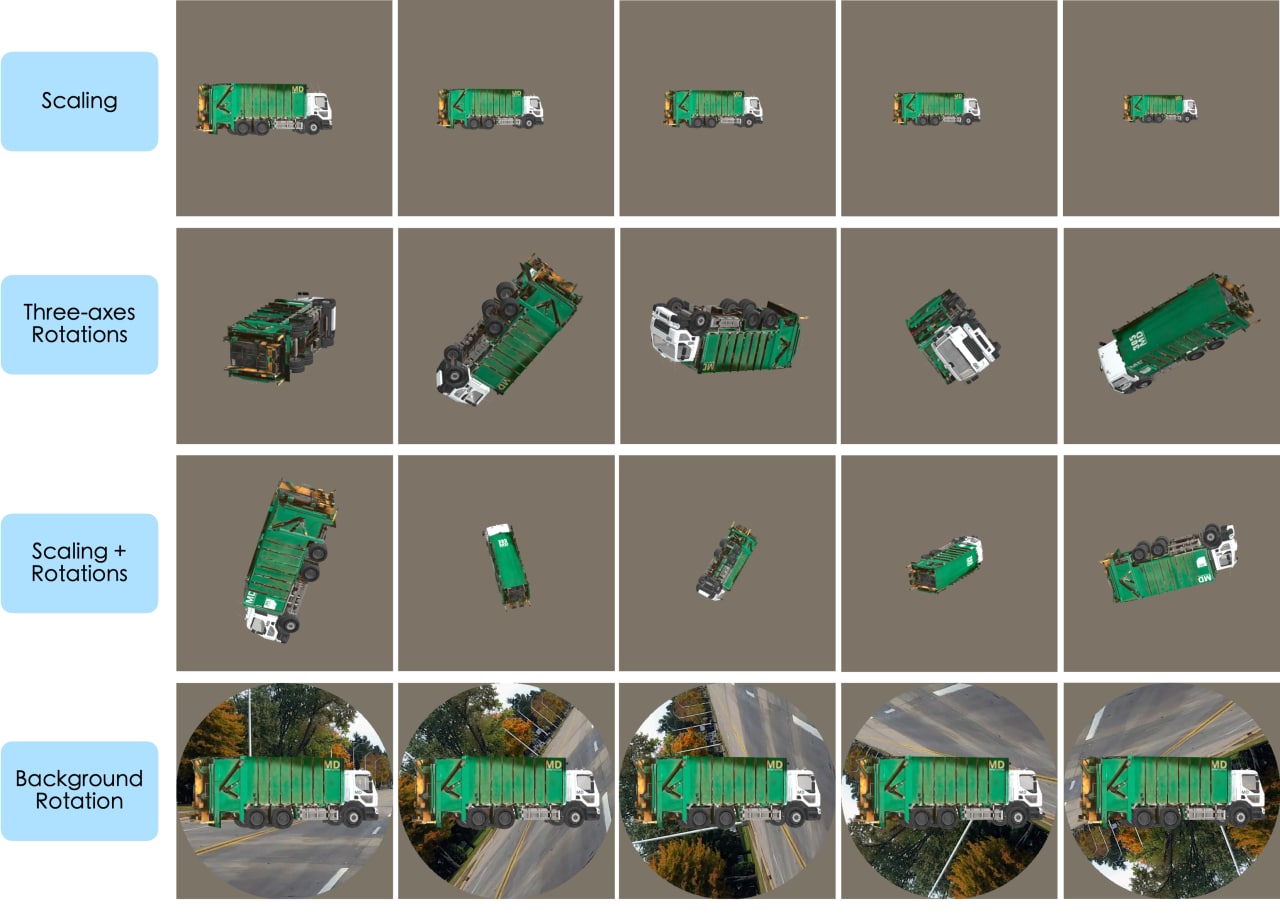}
    \caption{Different datasets to test different kinds of transformations.}
    \label{fig:scaling+three-axes+compined+bg_rot_examples}
\end{figure}

\textbf{ObjectScale dataset:} To study the effect of reducing the object scale, we created a dataset of images showing objects at different scales. Starting form the scale used for ObjectPose, we scale down the object by increasing the value of the $z$ parameter of the 3D renderer. The object is scaled down until its resolution approaches the limit of what can be recognized by (our) eye. Using two different initializations, 35 images were created for each object on average. The dataset contains 593 images in total. To avoid any mismatch between the object size and background we used grey background images only. See Fig. \ref{fig:scaling+three-axes+compined+bg_rot_examples} for samples from the dataset.

\textbf{Three-axes rotation dataset:} This time we rotate the object randomly along the axes \emph{ROLL}, \emph{YAW}, and \emph{PITCH} at the same time. For each 3D object, 180 images were created by sampling the three rotation angles randomly each time. For fair comparison, only grey background is used for creating this dataset. Fig. \ref{fig:scaling+three-axes+compined+bg_rot_examples} shows some samples from the dataset.

\textbf{Combining scale and rotation dataset:} The dataset is created by combining two types of transformations, three-axes rotation and scaling. The rotation angles are samples randomly from the range (0, 360). For fair comparison to using single transformations, the scale is sampled randomly from the same range used to create ObjectScale dataset. For each object, 180 images with grey background were created. Samples from the dataset are shown in Fig. \ref{fig:scaling+three-axes+compined+bg_rot_examples}.

\textbf{Background Rotation dataset:} Using two different initializations the object is fixed and 180 images are created for each initialization by rotating the background by 360° with a step of 2°. The image corners are filled with grey color. See Fig. \ref{fig:scaling+three-axes+compined+bg_rot_examples} for samples from the dataset.


\subsubsection{ObjectPose Dataset Evaluation} \label{sec:ObjectPoseevaluation}

Since we do not want anything else than the object pose to affect the model decision, we performed a simple experiment to evaluate the ObjectPose dataset quality. The experiment evaluates the dataset by testing the models on a subset of images with the objects in usual poses (upright). These are the images from ObjectPose with the object rotated by -10 to 10 degrees only. We call this dataset ObjectPose +-10. It consists of 1,683 images in total. We consider ObjectPose +-10 as usual-pose images and ObjectPose as unusual-pose images.

While we are generating the ObjectPose dataset, we excluded all the 3D objects with a ResNet50 \citep{resnet} top1 accuracy of less than 90\% on ObjectPose +-10. Among approximately 30 3D objects evaluated, we selected 17 of them for the ObjectPose dataset. Following \citep{naturaladvexamples}, we chose ResNet50 because it has a moderate ImageNet top1 accuracy of 76.1\%.  

\begin{figure}[h]
    \centering
    \includegraphics[width=0.7\textwidth]{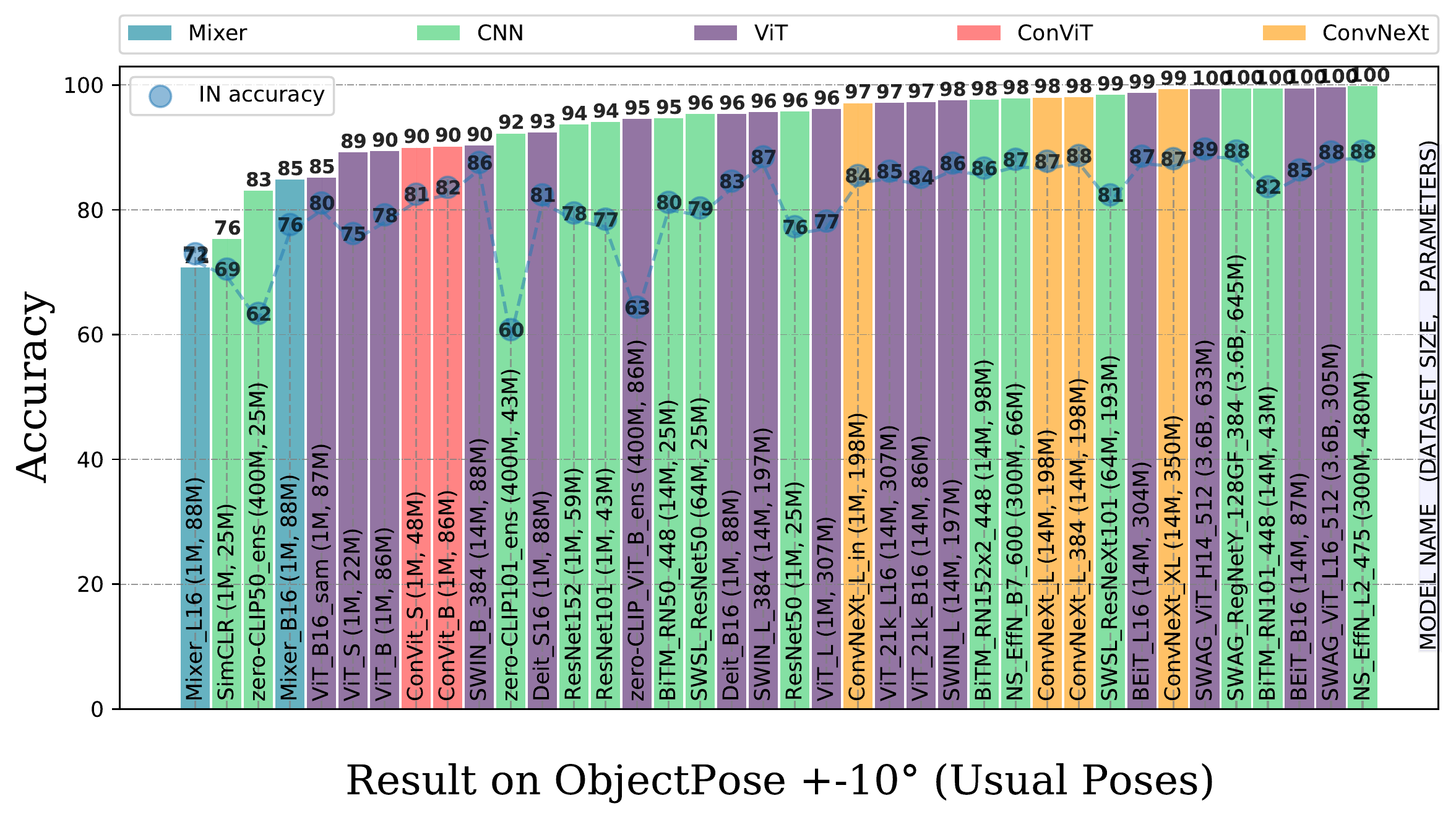}
    \caption{The performance on ObjectPose +-10. Most of the models can detect the objects in the usual poses. Note that CLIP models are trained on the WebImageText dataset and tested here without fine-tuning on ImageNet.}
    \label{fig:ObjectPosevalidationresult}
\end{figure}

\textbf{The objects are very easy to detect in usual poses}: After generating the ObjectPose dataset, we tested all the \nummodels models on ObjectPose +-10. The results are shown in Fig. \ref{fig:ObjectPosevalidationresult}. We see from the results that on ObjectPose +-10, ResNet-50 achieves 93\% average accuracy, which is higher than ResNet-50 top1 accuracy on ImageNet (76.1\%). Clearly, we see that most of the models obtain accuracy higher than 80\%. All the models perform on ObjectPose +-10 better than on ImageNet test dataset. This indicates that the method we follow to generate ObjectPose dataset gives high-quality images that look realistic for the models.\\


\begin{figure}[h]
    \centering
    \includegraphics[width=0.85\textwidth, height=0.45\textwidth]{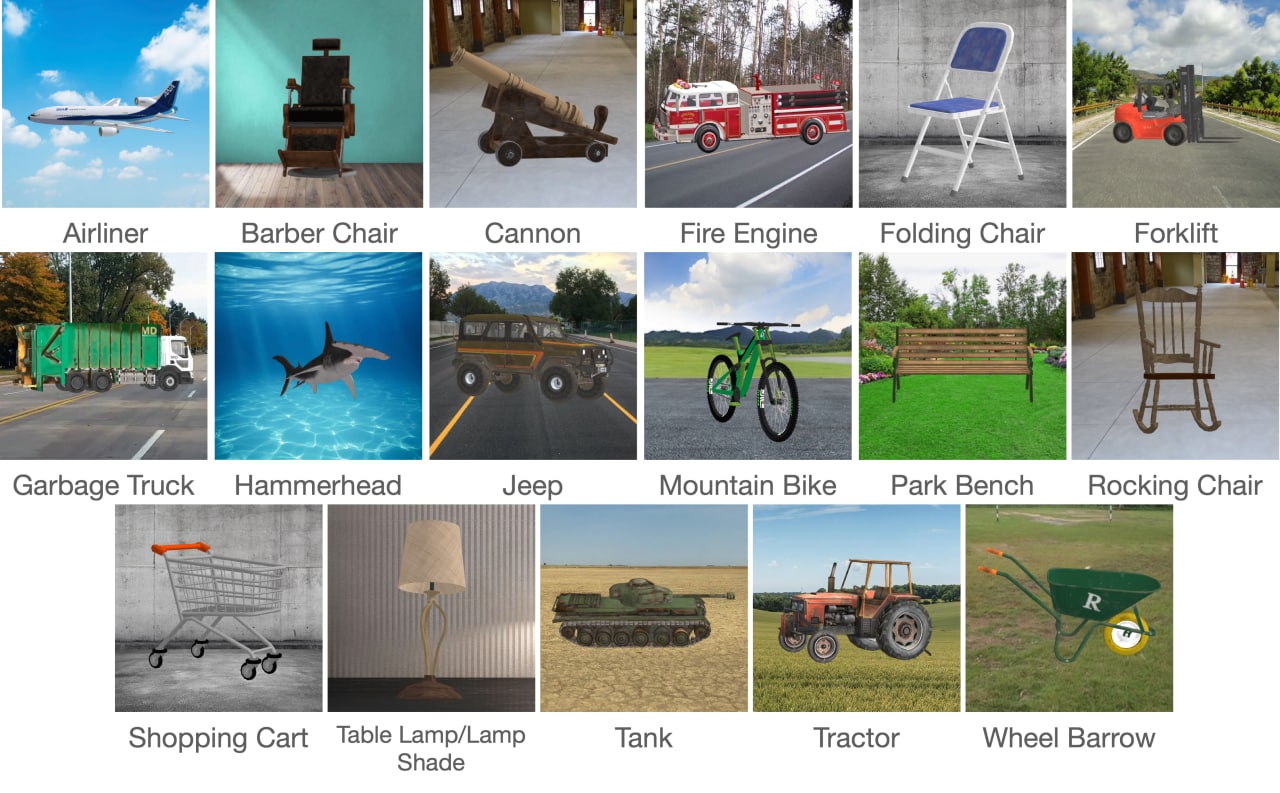}
    \caption{Presentation of the 17 3D objects composing ObjectPose. These objects were carefully selected from https://sketchfab.com/ for their visual quality as well as the ability of a ResNet-50 to correctly classify them in their upright pose. They were then rendered on a background image which we selected to correspond semantically to the object (e.g., a jeep is on a road).}
    \label{fig:ObjectPosesamples}
\end{figure}

\subsubsection{Background Images for ObjectPose Dataset}
The background images used for ObjectPose dataset were downloaded from the internet. Some of the background images are shared between more than one 3D object. In total, we use 24 different background images. Each background was chosen to logically match the object (i.e. a road background image for the Garbage Truck object). Fig. \ref{fig:ObjectPosesamples} shows some sample images from ObjectPose dataset with different backgrounds.

\subsubsection{The 3D Objects} \label{appendix:3dobjects}
All the 17 3D objects we selected were downloaded from \url{https://sketchfab.com}. All the objects are in \emph{.obj} format.

\begin{itemize}

    \item Barber chair: \href{https://sketchfab.com/3d-models/louis-hanson-barber-chair-453e164d571b4ed58712b4515da37e46}{link}
    \item Shopping cart: \href{    https://sketchfab.com/3d-models/shopping-cart-b96f896453b240ae804d0399f1faf027}{link}
    \item Rocking chair:
    \href{https://sketchfab.com/3d-models/rocking-chair-aebc93f3a0c443bd94a08f934323982d}{link}
    \item Mountain Bike: (edited: the texture was changed)
    \href{https://sketchfab.com/3d-models/afterburner-downhill-mountainbike-e3cd36c670cd4d19ab630a8468906667#download}{link} 
    \item Forklift:
    \href{https://sketchfab.com/3d-models/forklift-d40cae50e04145dd997cdca415cd72ad}{link}
    \item Wheel barrow:
    \href{https://sketchfab.com/3d-models/brouette-08c0739e36a74cbc9742125eb1199435}{link}
    \item Folding Chair: \href{https://sketchfab.com/3d-models/metal-folding-chair-c4428a7f4a2f472689a914a3373befc3}{link}
    \item Fire engine:
    \href{https://sketchfab.com/3d-models/american-lafrance-34d395bfe3c04dbdb5c78f8197af94e2}{link} 
    \item Garbage Truck:
    \href{https://sketchfab.com/3d-models/garbage-truck-01bbd89644ae4fb4acbffe5d6c345fa6}{link}
    \item Tractor: 
    \href{https://sketchfab.com/3d-models/tractor-1b258bcc01bf4ed0935ef73e80442c30}{link}
    \item Tank: 
    \href{https://sketchfab.com/3d-models/tank-e60aee73a34c4e4ab2d10e0e71763935}{link} 
    \item  Jeep (edited: some of the object parts were removed and some parts were scaled):
    \href{ https://sketchfab.com/3d-models/jeep-hunter-d2348066cbc34b0baabbaf42cb55c8c7}{link}
    \item Cannon:
    \href{https://sketchfab.com/3d-models/cannon-d8258903a23e4a368ede8fd584919b68}{link}
    \item Hammerhead Shark:
    \href{https://sketchfab.com/3d-models/hiu-palu-9a1b82b78cdb4c6a8d560b91073fbd65}{link}
    \item Park Bench:
    \href{https://sketchfab.com/3d-models/fancy-bench-82cadef8dfb24c0784eefff7be71610c}{link}
    \item Table Lamp:
    \href{https://sketchfab.com/3d-models/table-lamp-b140f762f97d43f5b48d13997e0d997b}{link}
    \item Airliner: \href{https://sketchfab.com/3d-models/l1011-tristar-ana-241e729d2f77438ea1bb2b3079abde81}{link} 
    
\end{itemize}

\subsection{CO3D Dataset} \label{sec:co3d}
The Common Objects in 3D (CO3D) dataset \citep{co3ddataset} was originally designed for 3D reconstruction and new-view synthesis tasks. The dataset contains real-world multi-view videos of 50 common object categories. The videos were collected by workers from their own environment using their own devices. The dataset shares some of its 50 object categories with the 1000-classes ImageNet ILSVRC-2012 dataset \citep{russakovsky2015imagenet}. In our experiments, we use a subset from CO3D videos with the following 10 object categories: `bench', `motorscooter', `toilet', `toaster', `parking meter', `broccoli', `microwave', `banana', `hotdog', and `bicycle'. We sampled 1000 images from videos of each of these 10 categories to get a total of about 10,000 images. In the rest of the paper, we will refer to this subset of 10,000 images as CO3D. 

\textbf{Selecting Subset From CO3D:} For each category, the CO3D dataset provides a set of folders, each one containing a set of images for one object. These images are sampled from a video taken for the object by walking in a complete circle around the object. In this way, we get images from different views and poses. In all the images, there is only one object centered in the image. Each category has a different number of folders available. When selecting a subset from CO3D, we sampled $k$ images from each folder so that the total number of images we have for the category is 1000 images (ie. we sample 1000/$k$ images from each folder). When reviewing CO3D dataset, we found that for some folders, the kinds of objects in CO3D don’t always perfectly match the kind of objects in ImageNet. As an example, the bench category in CO3D dataset includes images of objects such as park benches, sofas, and other kinds of benches, while ImageNet includes images of park benches only. To solve this, we manually reviewed the images of the 10 object categories in CO3D and compared them to the images of the same categories in the ImageNet validation dataset, and then we manually excluded all the folders with images that don’t match ImageNet. After filtering the dataset, we sampled 1000 images from each category. This gives us a balanced dataset of 10 classes and 10,000 images in total. Fig. \ref{fig:co3d_results} shows samples from the dataset.\\ 

\subsection{ImageNetV2 Dataset} \label{sec:imagenetv2}
ImageNetV2 was introduced in \citep{imagnetv2} as a new testset to measure the generalization of ImageNet classifiers. The dataset was collected from the internet following the same dataset creation process as the original ImageNet dataset \citep{russakovsky2015imagenet}. It mimics ImageNet by the fact that both of them are clean internet images collected from the Flickr hosting service and both of them contain images uploaded to the internet during the same period of time. The dataset helps in measuring the performance of networks on a distribution as close as possible to the training distribution (near-distribution), while avoiding overfitting effects to ImageNet's exact statistics. It consists of 10,009 images with fewer examples per class than ImageNet. Fig. \ref{fig:co3d_results} shows some examples sampled randomly from the same classes from both ImageNet and ImageNetV2. The figure shows how the two datasets are similar. In our experiments, we use the "Matched Frequency" version of ImageNetV2 dataset.


\section{Deep Networks details} \label{appendix:modelsdescription}
\subsection{Datasets and Training Objectives} The models were trained on datasets with different sizes ranging from 1 million to 3.6 billions images. Among the models we use, Noisy Student EfficientNet \citep{noisystudent} (300M images) and SWAG \citep{swagnet} (3.6B images) are the only models pretrained on extremely large dataset and fine-tuned on ImageNet. Although CLIP \citep{clip} is also pretrained on a very large dataset (400M image-text pairs), it is not fine-tuned on ImageNet. We also chose models trained under different objectives, including:
\begin{itemize}
\item Supervised learning, including convolutional architectures, such as \citep{resnet} \citep{noisystudent} \citep{convnext} \citep{bit}, Vision Transformers, such as \citep{vit} \citep{vitsam}\citep{beit} \citep{deit} \citep{swin}, \citep{convnext}, and MLP-mixer\citep{mlpmixer},
\item Self-supervised learning, such as SimCLR \citep{simclr}, and BEiT \citep{beit}, 
\item Semi-weakly supervised learning, such as SWSL-ResNet50 and SWSL-ResNeXt101 \citep{swsl}, and weakly supervised learning, such as SWAG \citep{swagnet},
\item Text supervision, such as CLIP \citep{clip}. 
\end{itemize}
We also included models trained using special optimizers such as the ViT-B16-SAM model \citep{vitsam}, a vision transformer trained with the recently proposed optimizer SAM \citep{samoptimizer}, which is designed to improve generalization and robustness. 

\subsection{Model descriptions}
In this section, we describe in detail the networks tested in this study. Table (\ref{table:modelstable}) shows the full list of models.

\textbf{CLIP:} CLIP was introduced in \citep{clip} as a model trained using a unique text supervision objective that works as an alternative to the (image, label) supervision. The model is pretrained using an image-caption matching task where we pass the image and its caption to the model. The task is simply to predict which caption matches which image. The model is supervised by the image caption text, then it is asked to maximize the similarity between the image representation and the caption representation of the paired inputs and to minimize the similarity between the representations of the unpaired inputs using a contrastive loss. The model consists of an image encoder that processes the image and outputs the image representation, and a text encoder that processes the text and outputs the text representation. The model is pretrained on a dataset of 400 million (image, caption) pairs. textbf{Zero-shot prediction:} After pretraining the model we can use it for image classification without funetuning on any dataset. This is called zero-shot prediction. The way CLIP is trained makes it easy to perform zero-shot prediction on any image classification dataset. To classify an image for a $k$-classes classification task, we get the image representation from the pretrained image encoder. We add the name of each of the $k$ classes to a sentence (prompt) and pass all the $k$ sentences to the pretrained text encoder to get the representation of each one of them. Then we compute the cosine similarity between the image representation and all the sentence (prompt) representations. The predicted class is the one associated with the input sentence with the highest similarity to the image. The paper proposed ensembling the predictions over different input sentences to improve the performance. This is done by averaging the representations of more than one text input. In our experiments, we ensemble over 80 input prompts proposed by the authors for ImageNet. We use three zero-shot CLIP models, CLIP-ResNet-50, CLIP-ResNet-101, and CLIP-ViT-B/16.\\

\textbf{Standard Vision Transformer:} We include six standard Vision Transformers (ViTs) \citep{vit}, four of them (ViT-S/16, ViT-B/16, ViT-B16-SAM/16, ViT-L16) were pretrained on ImageNet with input size 224x224. The ViT-B16-SAM/16 model was pretrained using SAM optimizer introduced in \citep{samoptimizer}. The last two models are ViT-21k-B/16, and ViT-21k-L/16 pretrained on ImageNet-21k and fine-tuned on ImageNet with the input size of 384x384.\\

\textbf{BEiT:} The model is a Vision Transformer trained using a self-supervised training method introduced with the model \citep{beit} known as Masked Image Modeling (MIM). MIM is inspired by the Masked Language Modeling (MLM) task used for training BERT \citep{bert} text transformer. BEiT outperforms standard ViT even when using a smaller pretraining dataset. In our experiments we use two BEiT models (BEiT-B/16 and BEiT-L/16). Both of them were pretrained with the self-supervised MIM task on ImageNet22k and fine-tuned on ImageNet.\\

\textbf{ConvNext:} We also use four ConvNeXt \citep{convnext} models. ConNeXT was introduced recently as a pure convolutional architecture that competes with the ViT. The convolutional layers of ConvNeXt are designed carefully by choosing a good collection of architecture hyperparameters after running many experiments. The model is trained by following techniques used to train ViTs that were proved to improve the performance of ConvNeXt also.\\

\textbf{ConViT:} We also use two ConVit models (ConViT-B and ConViT-L)\citep{convit}. ConVit replaces some of the ViT self-attention layers with a new kind of attention layer called gated positional self-attention (GPSA) layers introduced with the model.\\ 

\textbf{Deit:} We also use two Data-efficient Vision Transformers (DeiT) \citep{deit}. DeiT was introduced to reduce the need for large pretraining dataset for Vision Transformers. The model relies on a teacher-student knowledge distillation \citep{knowledgedistilling} strategy designed for transformers. It can achieve the performance of ViTs with much less data (ImageNet only) while having the same number of parameters. In our experiments, we use DeiT-B/16 and DeiT-S/16.\\

\textbf{SWIN Transformer:} We also use three SWIN transformers (SWIN-L-384,  SWIN-L, and SWIN-B-384). SWIN transformer \citep{swin} is a hierarchical Vision Transformer that replaces the self-attention layer of the original vision transformer with a Shifted Window Self-attention layer (SWSA). The SWSA layer partitions the image into windows and computes the self-attention locally inside each window. This operation results in reducing the attention layer computational time to be linear to the input size instead of quadratic, while improving the performance over the original ViT.\\

\textbf{SimCLR:} We use one SimCLR model \citep{simclr} with ResNet50 backbone. The backbone was trained using a self-supervised contrastive loss without fine-tuning on ImageNet. A linear classifier was trained on top of the backbone architecture to perform classification.\\

\textbf{ResNet:} We use three ResNets \citep{resnet} trained on ImageNet. These are ResNet50, ResNet101, and ResNet152.\\

\textbf{BiT:} Big Transfer was introduced in \citep{bit} as a recipe for transfer learning from a large dataset to the target dataset. We use three medium Big Transfer models pretrained on ImageNet21k dataset and fine-tuned on ImageNet. Namely, these are the models BiTM-ResNetv2-50x1, BiTM-ResNetv2-101x1, and BiTM-ResNetv2-152x1.\\

\textbf{SWSL:} We use two Semi-weakly supervised models \citep{swsl} (SWSL-ResNeXt101, and SWSL-ResNeXt101-32x16d). The models were pretrained on 64M images extracted from a larger dataset of 940 million using a teacher model. The labels for the 64M images are predicted by the teacher models. The teacher model is a ResNet-101-32x48 pretrained on ImageNet. The models are then fine-tuned on ImageNet\\

\textbf{Noisy Student:} We use two EfficientNets pretrained using the Noisy Student paradigm introduced in \citep{noisystudent}. The models are trained on a combination of a noisily labeled JFT-300M dataset and ImageNet dataset. The two models are Noisy Student EfficientNet-L2 with an input size of 475x475, and Noisy Student EfficientNet-B7 with an input size of 600x600. These models were pretrained with the RandAugment \citep{randaugment} data augmentation method.
The following pseudo-code shows data preprocessing for Noisy Student:

\begin{lstlisting}
image = random_flip_left_right(image)
image = random_crop(image)
image = randaugment(image) 
\end{lstlisting}
The randaugment(.) method applies a set of 14 augmentations including [
      `AutoContrast', `Equalize', `Invert', `Rotate', `Posterize',
      `Solarize', `Color', `Contrast', `Brightness', `Sharpness',
      `ShearX', `ShearY', `TranslateX', `TranslateY', `SolarizeAdd']\\

\textbf{SWAG:} We use three SWAG models \citep{swagnet}. These model were petrained using weakly supervised pretraining on more than 3.6 billions Instagram images. The pretraining dataset was labeled with hashtags and have 27K classes. We use SWAG-ViT-L, SWAG-ViT-H, and SWAG-RegNetY.\\

\textbf{MLP-Mixer:} We use two models with MLP-like architectures. These are MLP-Mixer-B/16 and MLP-Mixer-L/16 introduced in \citep{mlpmixer}.\\

\subsection{Model Sources}
All the models checkpoints we use are in PyTorch
format. The models are from seven main sources. These sources are Pytorch Image Model library (timm), the offical \href{https://github.com/openai/CLIP}{CLIP repository}, \href{https://github.com/lukemelas/PyTorch-Pretrained-ViT}{PyTorch Pretrained ViT repository}, \href{https://github.com/sacadena/ptrnets}{ptrnets} repository, Hugging Face Transformers library, Torchvision, and Torch Hub. Table (\ref{table:modelstable}) shows the source of each model we use.

\cleardoublepage
\begingroup 
\setlength{\tabcolsep}{10pt} 
\renewcommand{\arraystretch}{1.5} 
\begin{table}[H]
\centering
\caption{ \textbf{Models details:} The full list of models. Here we report the ImageNet accuracy mentioned in the papers, unless a different accuracy score is provided with the implementation we use. Some model were tested with input resolution higher than the standard 224x224 resolution, in this case the input resolution is added after the model name.}
\begin{small}
\begin{tabular}{ p{5cm} p{2cm} p{3cm} p{0.5cm} p{0.5cm} }
\hline
Model & Source & Data & Size(M) & IN  \\
\hline
Resnet50  & torchvision & ImageNet(1M) & 25M & 76.1\% \\
Resnet152 & torchvision & ImageNet(1M) & 69M & 77.3\% \\
Resnet101 & torchvision  & ImageNet(1M) & 43M & 78.3\% \\
CLIP-ViT-B16 & CLIP &  WT(400M) & 86M & 63.2\% \\
CLIP-RN50     & CLIP &  WT(400M) & 25M  & 62.2\% \\
CLIP-RN101     & CLIP &  WT(400M) & 43M & 59.6\% \\
ViT-L16 & timm &  ImageNet(1M) & 307M & 77\% \\
ViT-B16 & timm &  ImageNet(1M) & 86M  & 78\% \\
ViT-S16 & timm &  ImageNet(1M) & 22M  & 75\%  \\
ViT-B16-sam & timm &  ImageNet(1M) & 86M & 79.9\% \\
ViT-21k-B16 & pretrained-vit & ImageNet21k(14M) & 86M & 84\% \\
ViT-21k-L16 & pretrained-vit & ImageNet21k(14M) & 307M & 85\% \\
SWIN-B-384 & timm & ImageNet21k(14M)  & 88M  & 86.4\% \\
SWIN-L     & timm & ImageNet21k(14M) & 197M & 86.3\% \\
SWIN-L-384 & timm & ImageNet21k(14M)  & 197M & 87.3\% \\
Simclr & timm & ImageNet(1M) & 25M & 68.9\% \\
BiTM-RN50  & timm  & ImageNet21k(14M) & 25M & 80.0\% \\
BiTM-RN101 & timm  & ImageNet21k(14M) & 43M & 82.5\% \\
BiTM-RN152x2 & timm  & ImageNet21k(14M) & 98M & 85.5\% \\
SWSL-ResNet50  & torchhub  & (64M) & 25M & 79.1\% \\
SWSL-ResNeXt101 & torchhub  & (64M) & 193M & 81.2\% \\
Mixer-B16 & timm & ImageNet(1M) & 59M & 76.44\% \\
Mixer-L16 & timm &  ImageNet(1M) & 207M & 71.76\% \\
BEiT-B16 & transformers & ImageNet21k(14M) & 87M & 85.2\%  \\
BEiT-L16 & transformers &  ImageNet21k(14M) & 304M & 87.4\% \\
Deit-B16 & timm &  ImageNet(1M) & 86M & 83.4\% \\
Deit-S16 & timm &  ImageNet(1M) & 22M & 81.2\%  \\
EffN-B7-NS & torchhub & JFT(300M) & 66M & 86.9\% \\
EffN-L2-NS & torchhub &   JFT(300M) & 480M&  88.4\% \\ 
ConvNeXt-XL & timm &  ImageNet21k(14M) & 350M & 87.0\% \\
ConvNeXt-L-384 & timm & ImageNet21k(14M) & 198M & 87.5\%  \\
ConvNeXt-L & timm &  ImageNet21k(14M) & 198M & 86.6\% \\
ConvNeXt-L & timm &  ImageNet21k(1M) & 306M & 82.6\% \\
ConVit-B & timm & ImageNet(1M) & 86M & 82.4\% \\
ConVit-S & timm & ImageNet(1M) & 27M & 81.3\% \\
SWAG-ViT-L16-512 & torchhub & IG(3.6B) & 305M & 88.07\% \\
SWAG-ViT-H14-512 & torchhub & IG(3.6B) & 633M & 88.55\% \\
SWAG-RegNetY-128GF-384 & torchhub & IG(3.6B) & 645M & 88.23\%
\\
\hline
\end{tabular}
\end{small}
\label{table:modelstable}
\end{table}
\endgroup

\section{Additional Experiments}

\subsection{Effects of the Background Image}
\begin{figure}[h]
    \centering
    
    $\vcenter{\hbox{\includegraphics[width=.325\textwidth]{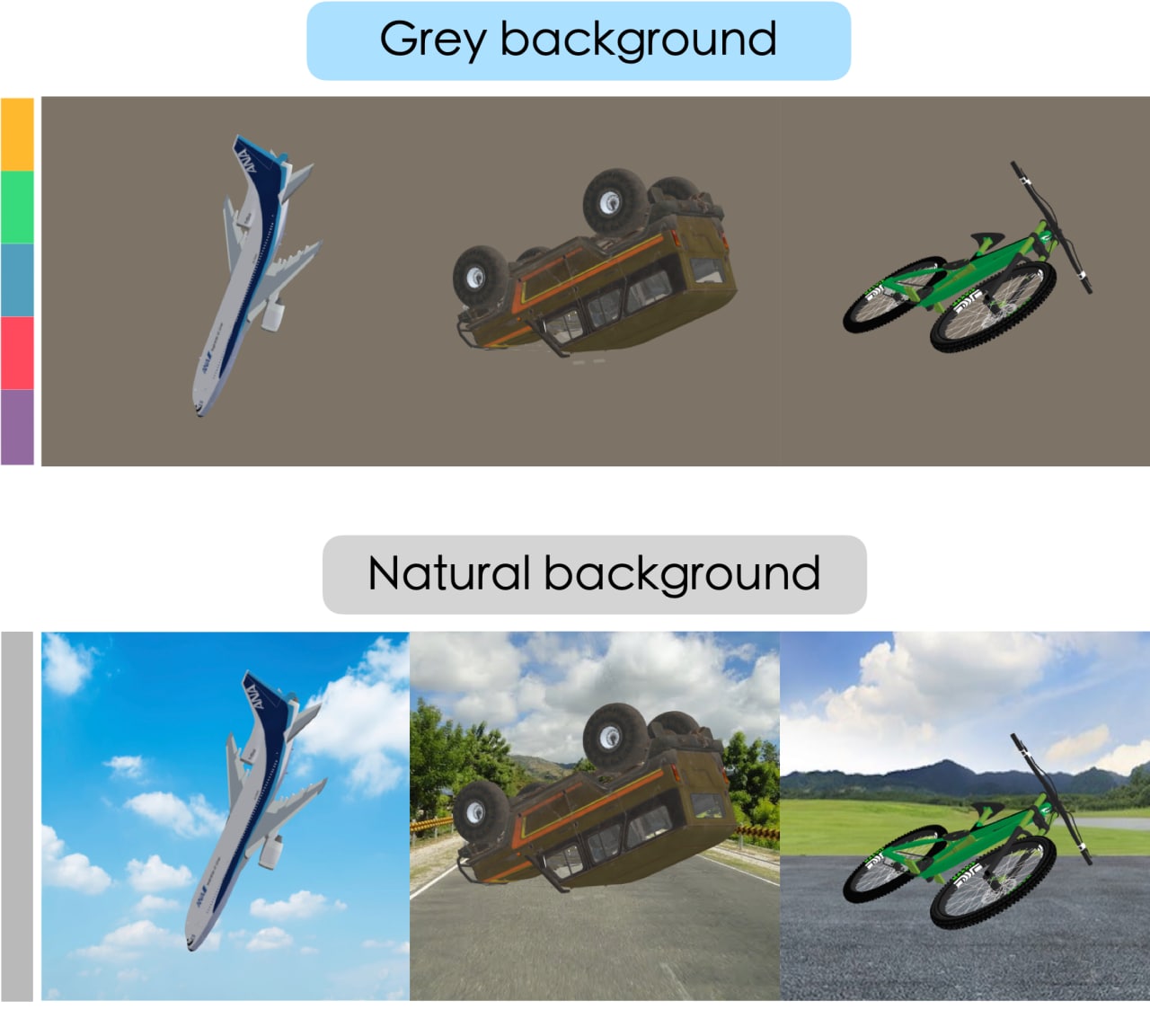}}}$
    \hspace{0em}
    $\vcenter{\hbox{\includegraphics[width=.66\textwidth]{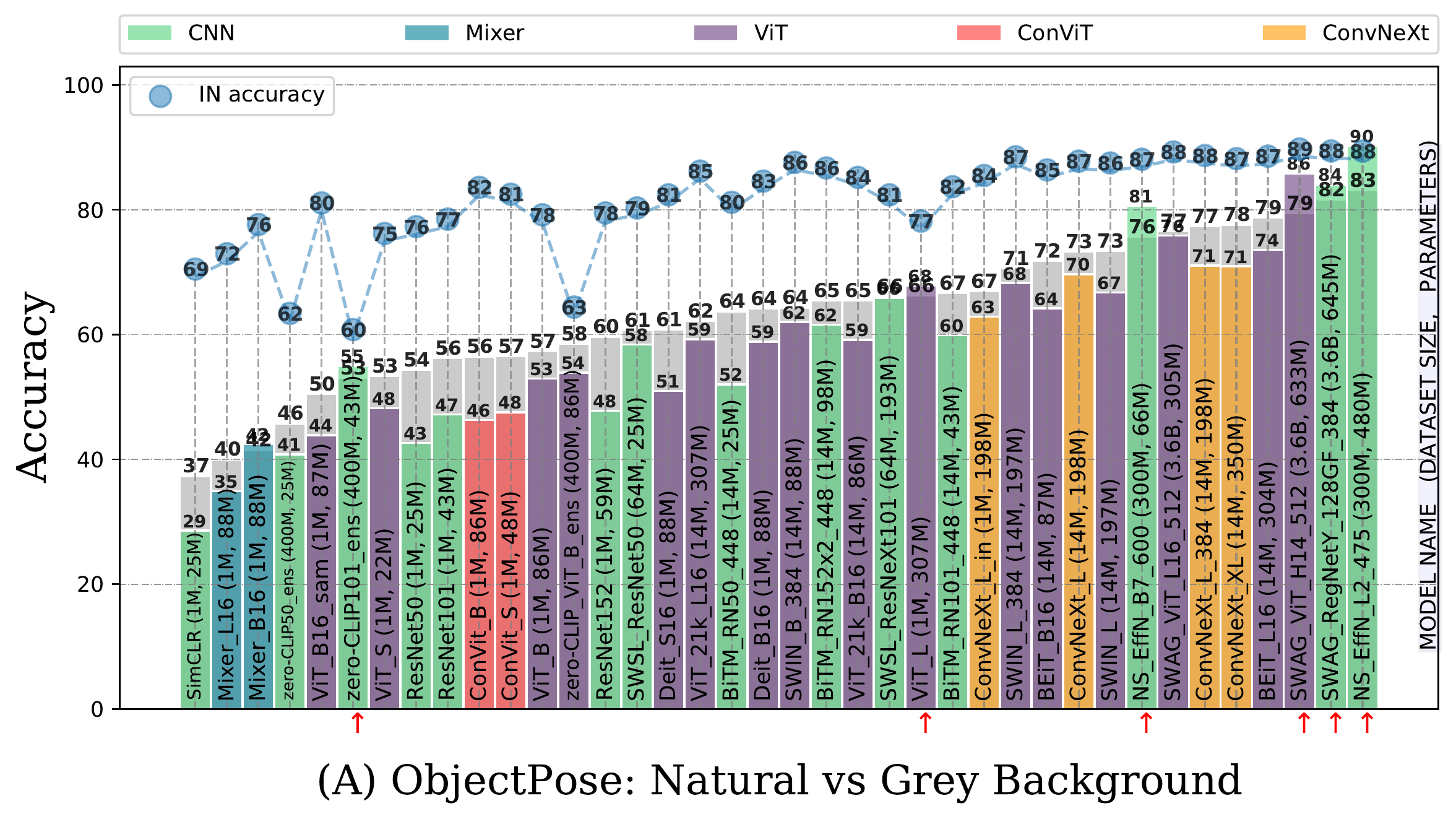}}}$
    $\vcenter{\hbox{\includegraphics[width=.325\textwidth]{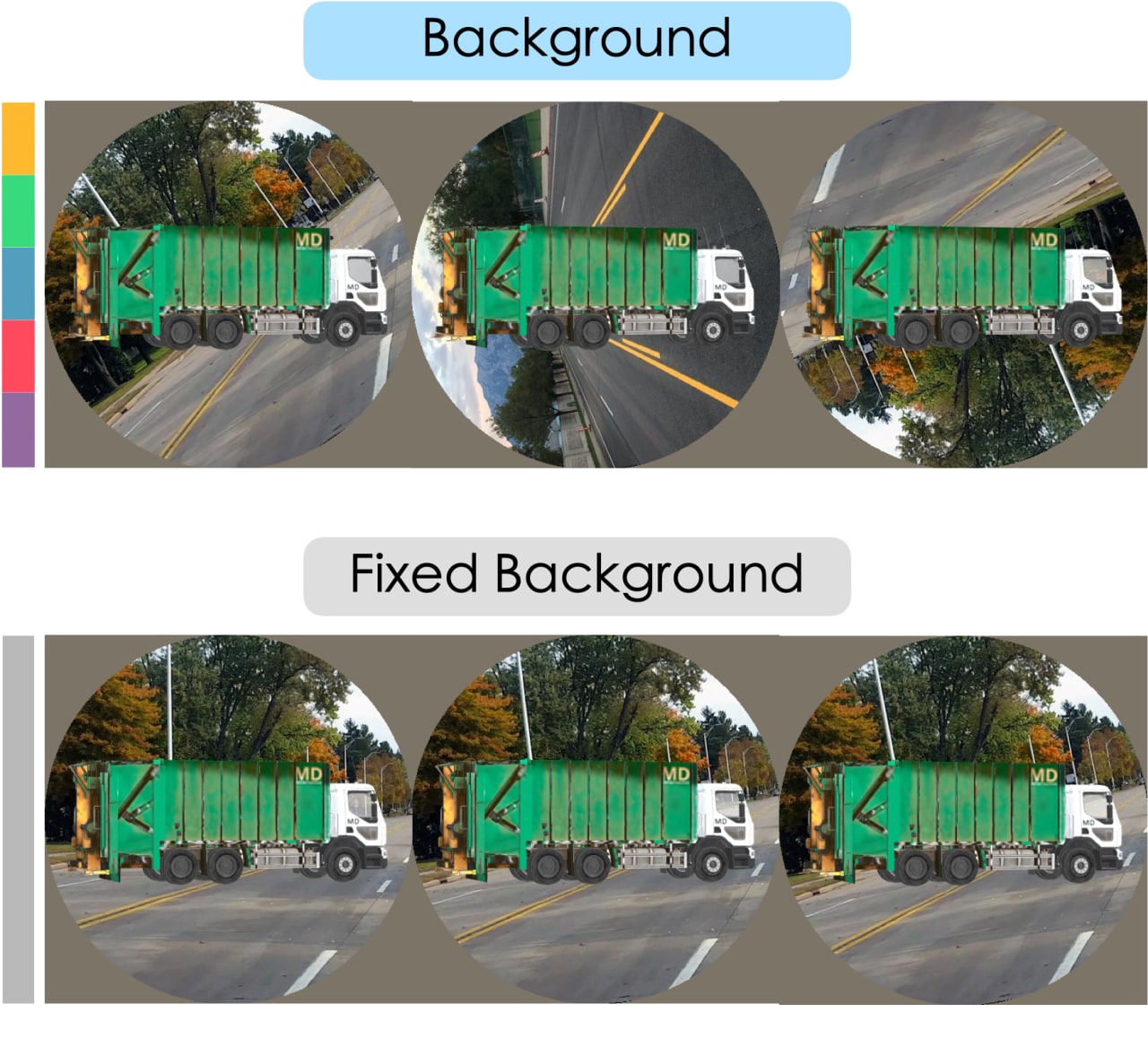}}}$
    \hspace{0em}
    $\vcenter{\hbox{\includegraphics[width=.66\textwidth]{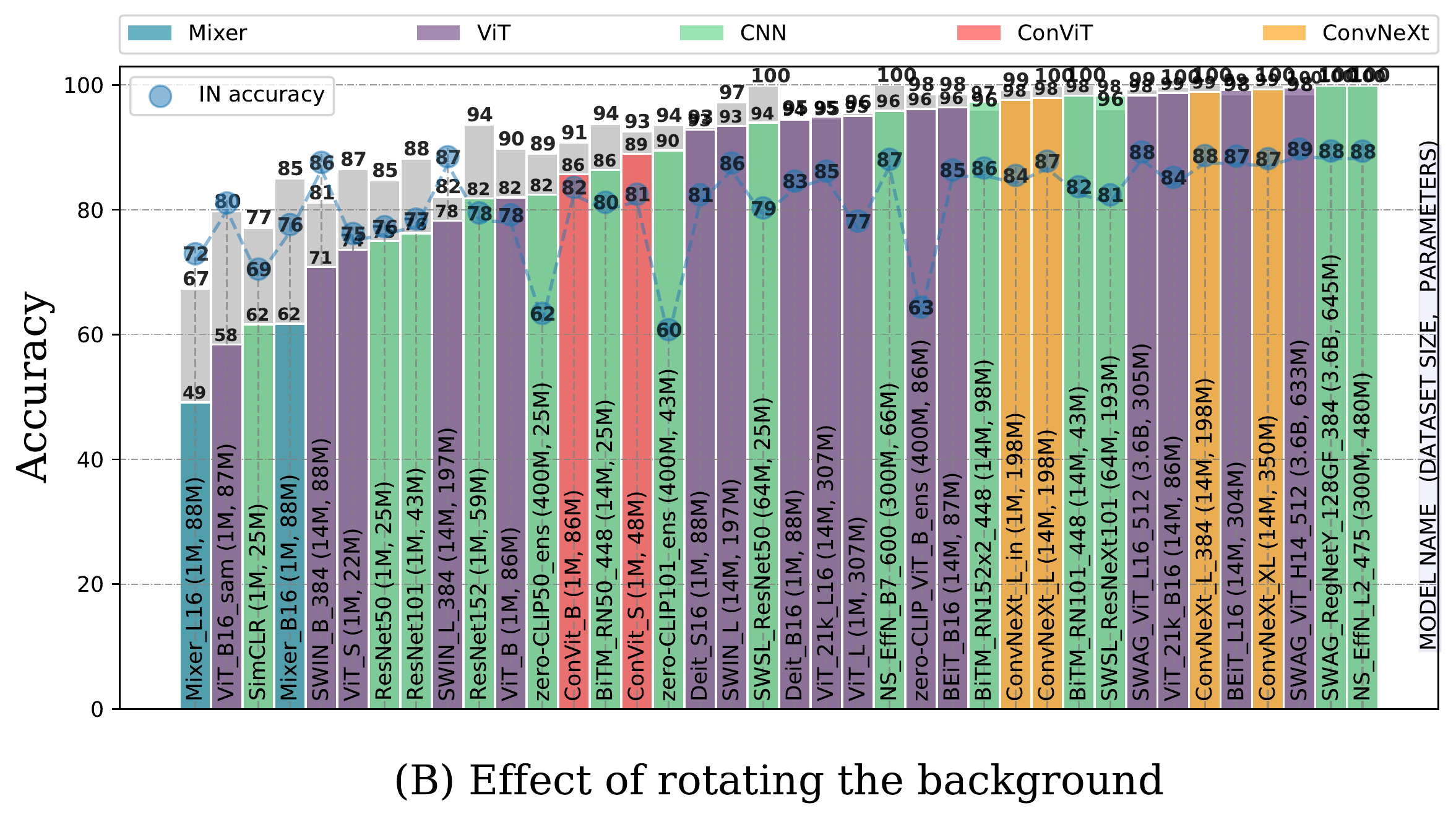}}}$
    \caption{\textbf{(A)} Effect of removing the background image on accuracy. Models are sorted by their accuracy on a natural background. Removing the background impairs performance for most models, but it increases performance for the best models, revealing a different strategy used by these models. \textbf{(B)} Effect of rotating the background image on accuracy, when keeping the object upright. Models are sorted by their accuracy on a rotated background. The best models are unperturbed by this manipulation, again revealing a difference in strategy with the smaller models.}
    \label{fig:bg_rot}
\end{figure}

To better understand the reliance of different networks on the background image in taking decision, we conduct additional analysis. We find that removing the background and replacing it with grey background harms most of the models on ObjectPose (Fig. \ref{fig:bg_rot}.A). However, for the strong models this results in improving the model's performance.
In addition, we study the effect of rotating the background image while keeping the foreground object upright (Fig. \ref{fig:bg_rot}.B). We observe that the strong models on ObjectPose are not affected by rotating the background when the object is upright, while the weak models show some performance degradation. This supports the result mentioned in section \ref{sec:results} that the weak models rely on different strategies than the strong models regarding exploitation of the background image.

\subsection{Additional Experiments with CLIP}\label{sec:clipexperiments}

\begin{figure}[h]
    \centering
    \subfigure{\includegraphics[width=1\textwidth]{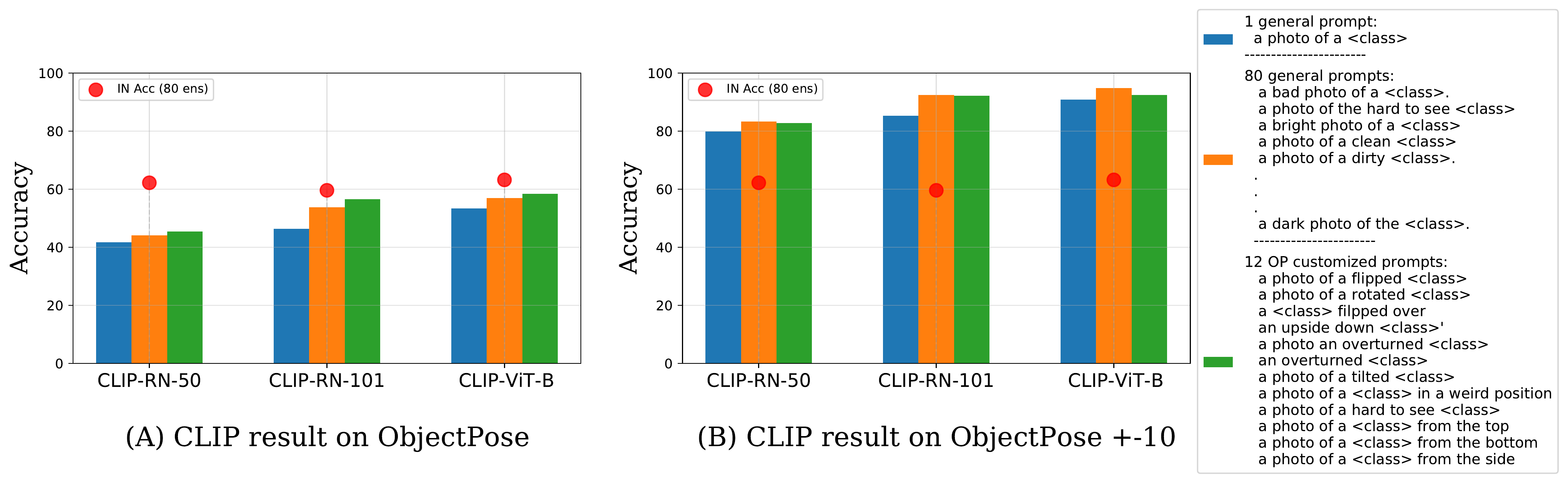}
    } 
    \caption{\textbf{How prompt engineering and ensembling affect CLIP performance for both ObjectPose and ObjectPose +-10 (upright objects).} For the ObjectPose customized prompts, we inject the information that the object is rotated through the prompts. The legend shows all 12 customized prompts. \textbf{(A)} On ObjectPose, the customized prompts outperform both the single prompt and the ensemble of the 80 general prompts. \textbf{(B)} When the object is upright (ObjectPose +-10) the customized prompts become misleading and the general prompts perform better. }
    \label{fig:clipanalysis}
\end{figure}

The natural language supervision of CLIP using a very large dataset (400M) gives zero-shot CLIP the ability to generalize to many datasets. This helps CLIP-ResNet-101 and CLIP-ViT-B/16 achieve accuracy close to their ImageNet accuracy on ImageNetV2, CO3D dataset, and ObjectPose +-10 datasets. However, on ObjectPose, the performance of these models drops as shown in Fig. \ref{fig:co3d_results} and Fig. \ref{fig:accvsacc}.  To measure the ability of CLIP models to benefit from the information in the prompt to detect objects in unusual poses (ObjectPose), we tested three CLIP models (CLIP-ResNet-50, CLIP-ResNet-101, and CLIP-ViT-B/16) with three different prompt choices: (1) single general prompt, this is a single prompt proposed by CLIP paper ("a photo of a <class name>"), (2) 80 general ImageNet prompts, these are 80 prompts proposed by the paper for the ensembling over multiple zero-shot predictions for ImageNet (results reported in the main text), (3) 12 prompts we customized for ObjectPose by adding a word that indicates that the object is rotated such as "flipped", "rotated", "upside-down", etc. We find that ensembling ObjectPose-customized prompts improves the performance on ObjectPose. The result in Fig. \ref{fig:clipanalysis} shows that the 12 ObjectPose-customized prompts are better than the 80 general prompts on ObjectPose. This indicates that CLIP benefits from the information provided in the prompts telling that the object in the images is flipped. This is in contrast to ObjectPose +-10 where the ObjectPose-customized prompts become misleading and give lower accuracy than the general prompts but still better than the single prompt. 

\subsection{The Relationship Between ImageNet Accuracy and OOD Accuracy: The OOD robustness on ImageNetV2 and CO3D does not always transfer to robustness on ObjectPose}
\begin{figure}[t]
    \centering
    \subfigure{\includegraphics[width=0.30\textwidth]{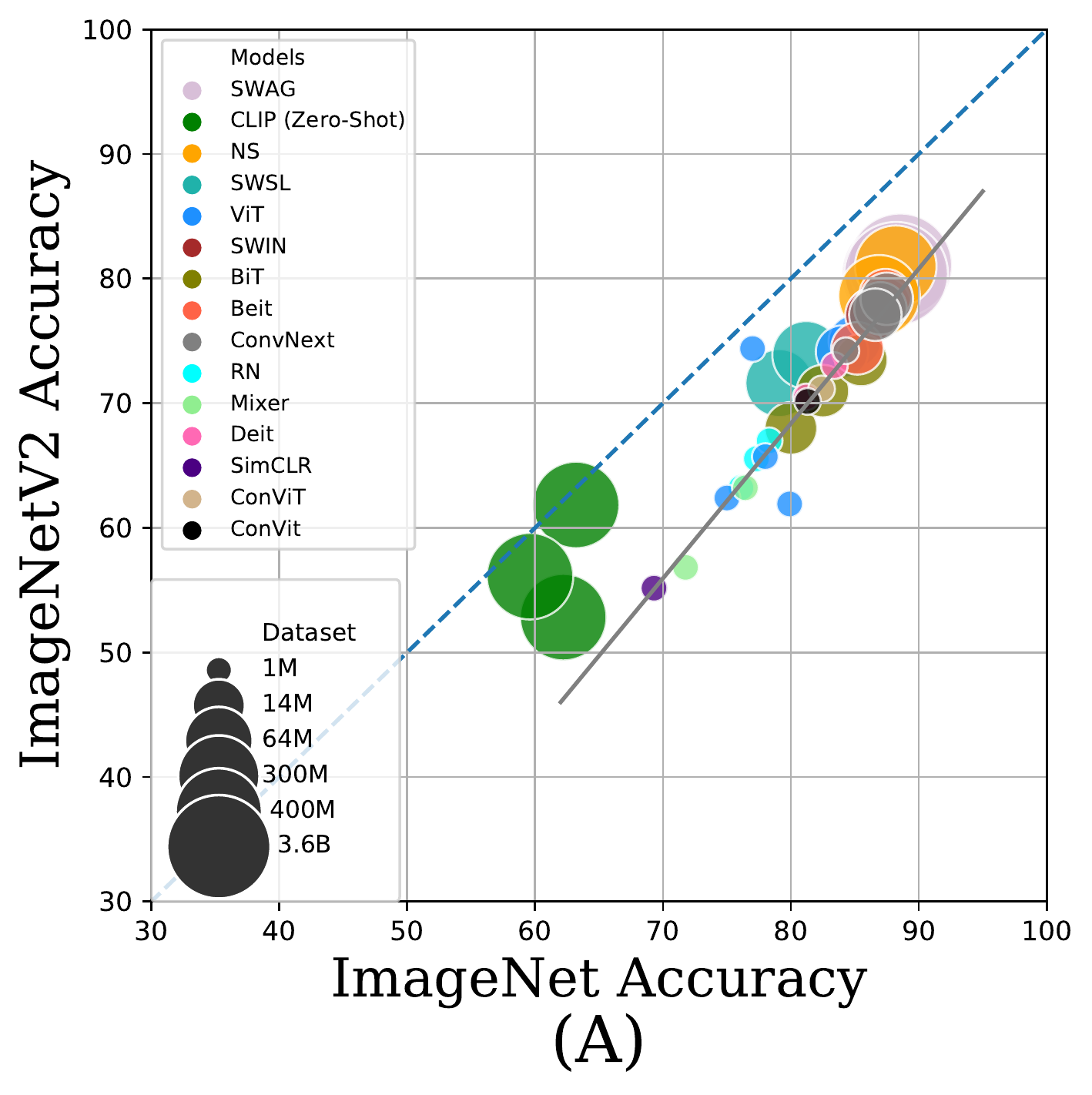}
    } 
    \subfigure{\includegraphics[width=0.30\textwidth]{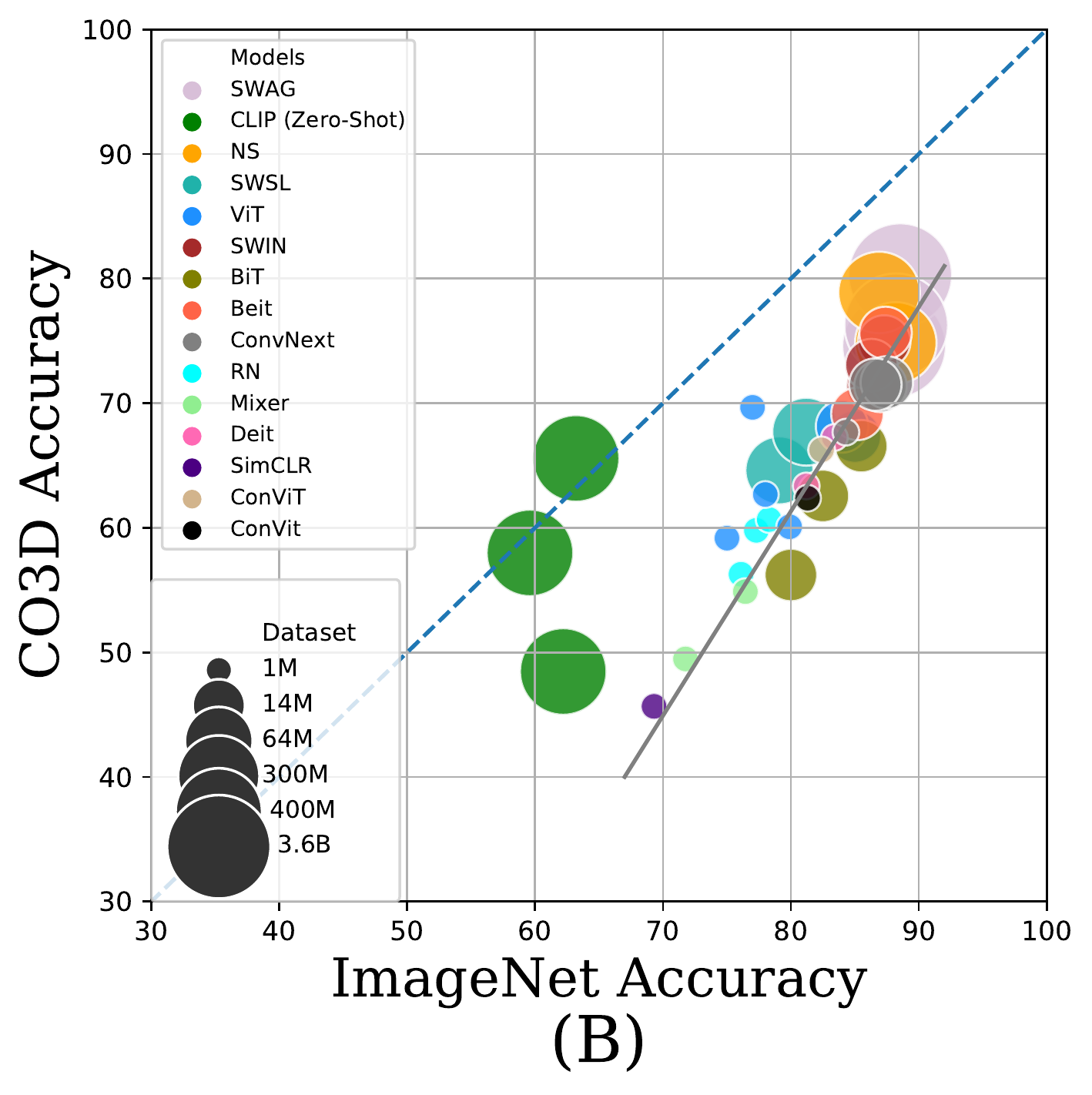}
    } 
  \subfigure{\includegraphics[width=0.30\textwidth]{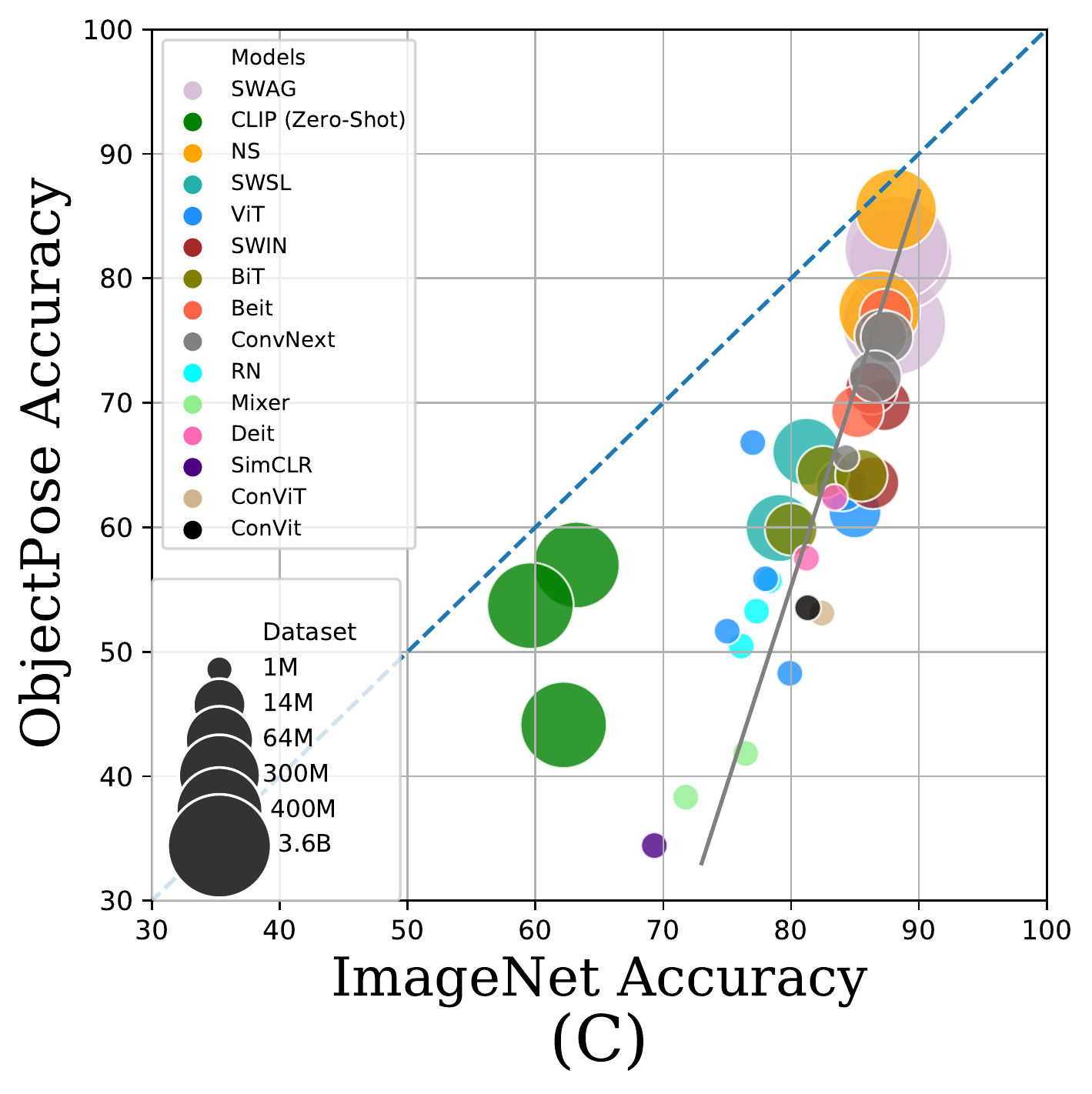}
    } 

    \caption{Networks' accuracies on ImageNet vs ImageNetV2, vs CO3D and vs ObjectPose. The dotted line corresponds to the ideal case when the OOD performance equals the in-distribution performance. CLIP models almost fit the dotted line on ImageNet and CO3D, but their performance degrade on ObjectPose. \textbf{(A)} On \textbf{ImageNetV2}, all the models lie below the dotted line, but they follow a line parallel to it (the grey line). The percentage of progress on ImageNet results in the same percentage of progress on ImagenetV2 for all the models. \textbf{(B)} On \textbf{CO3D}, the grey line starts to deviate from the dotted line. The weak models' accuracy transfers to OOD accuracy with less percentage than the stronger models. \textbf{(C)} On \textbf{ObjectPose}, the grey line deviates further from the ideal line. The best ImageNet models are less affected by the object poses than the weak models. Fig. \ref{fig:allaccvsacc} provides more comparisons.}
    \label{fig:accvsacc}
\end{figure}

Prior studies \citep{taori2020measuring} \citep{accuracyontheLine} show that in-distribution performance correlates strongly with out-of-distribution generalization performance for a wide range of models, tasks, and distribution shifts. Fig. \ref{fig:accvsacc} shows the relationship between the accuracy on ImageNet and the accuracy on the OOD datasets we are testing (ImageNetV2, CO3D, and ObjectPose). The figure shows that as we move from ImageNetV2, to CO3D, to ObjectPose, the accuracy relationship starts to deviate from the ideal correlation (when the OOD accuracy equals the in-distribution accuracy) for most of the models except the ones pretrained on large datasets (EfficientNet-L2, and SWAG). This proves that OOD robustness on ImageNetV2 and CO3D does not always transfer to robustness in detecting objects in unusual poses (ObjectPose). We observe also in Fig. \ref{fig:accvsacc} that the weak ImageNet models suffer from a higher accuracy drop on CO3D and ObjectPose compared to the stronger ImageNet models.

\begin{figure}[h]
    \centering
    \subfigure{\includegraphics[width=0.30\textwidth]{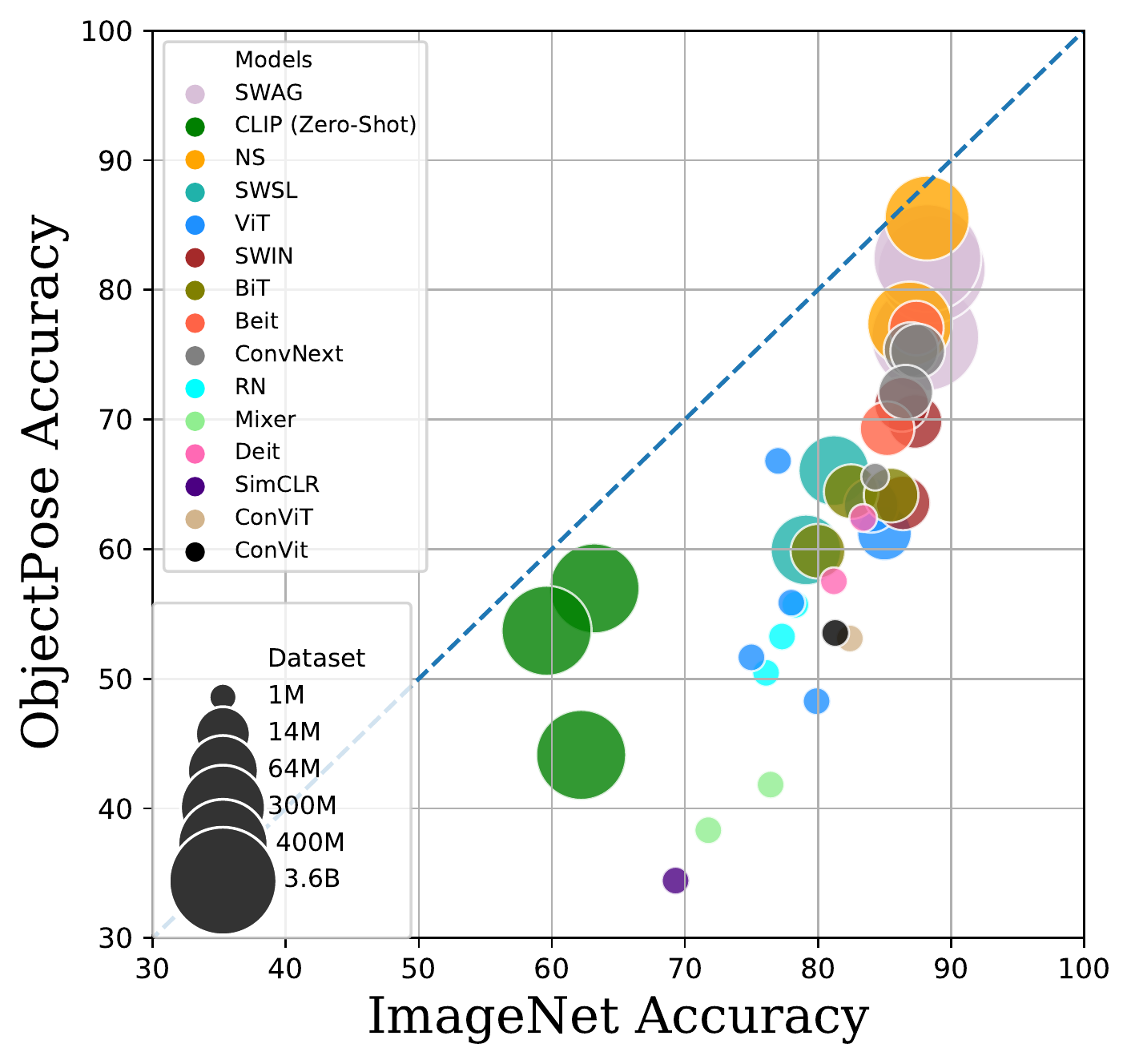}
    } 
    \subfigure{\includegraphics[width=0.30\textwidth]{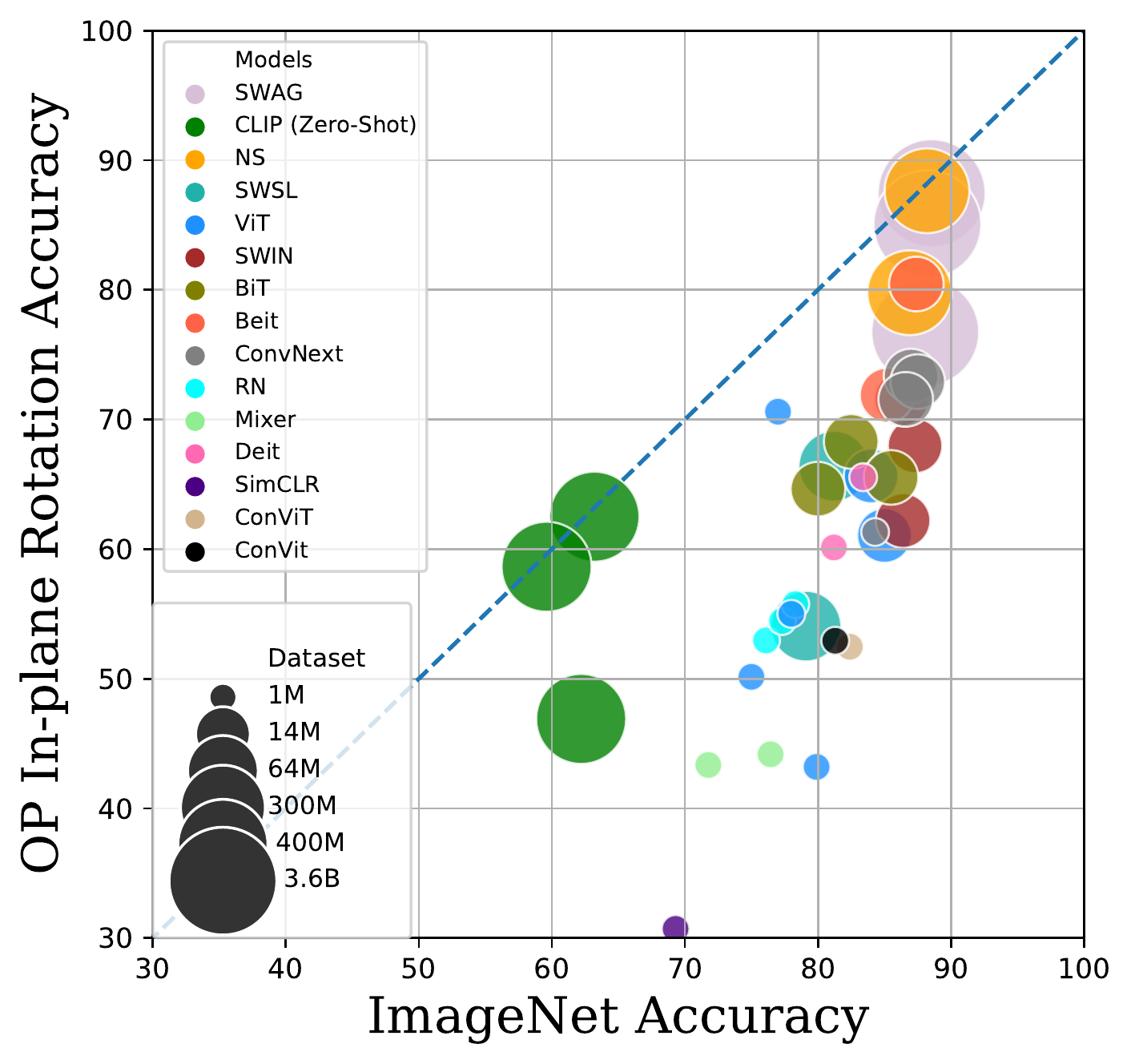}
    } 
    \subfigure{\includegraphics[width=0.30\textwidth]{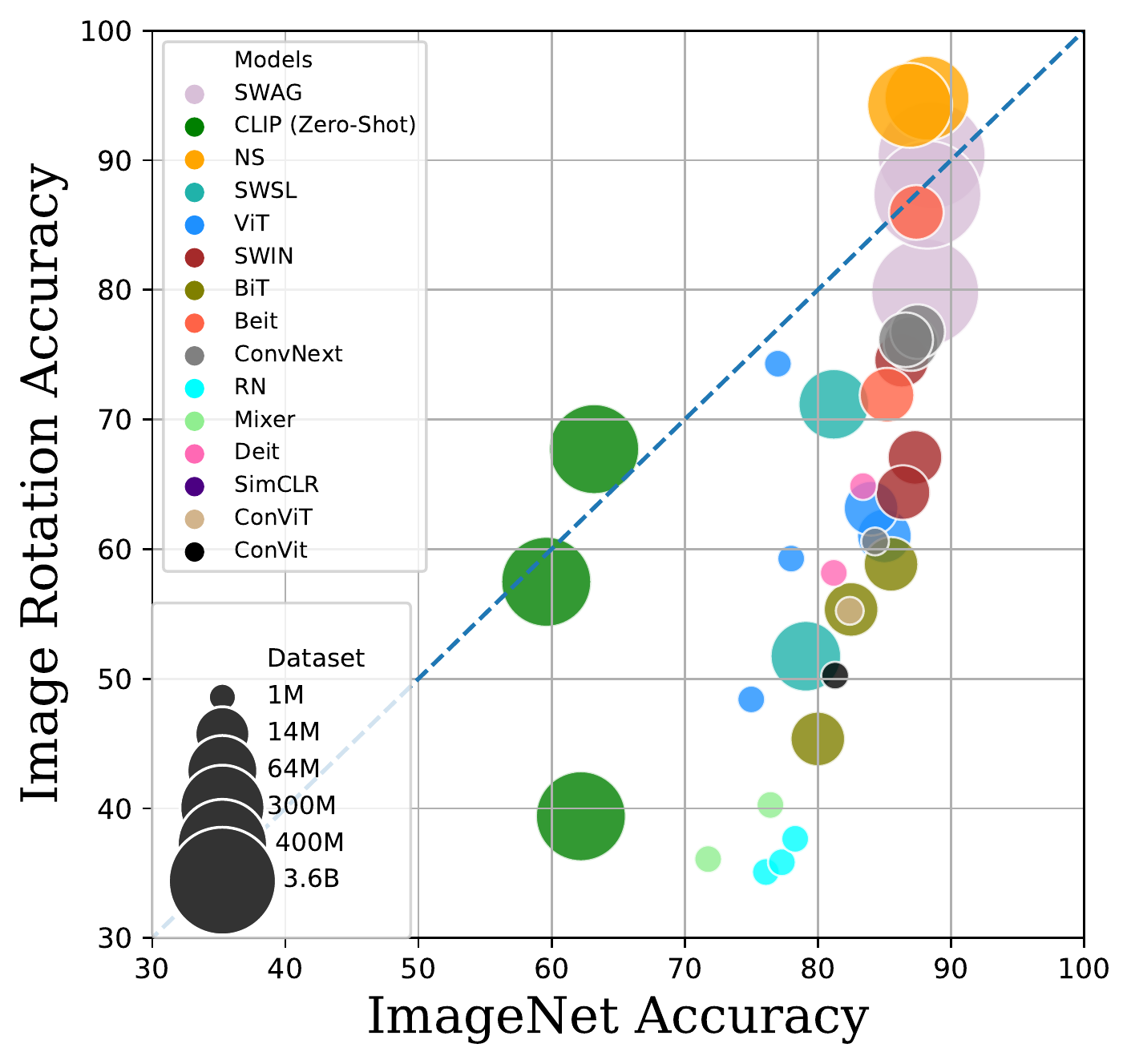}
    } 
    \subfigure{\includegraphics[width=0.30\textwidth]{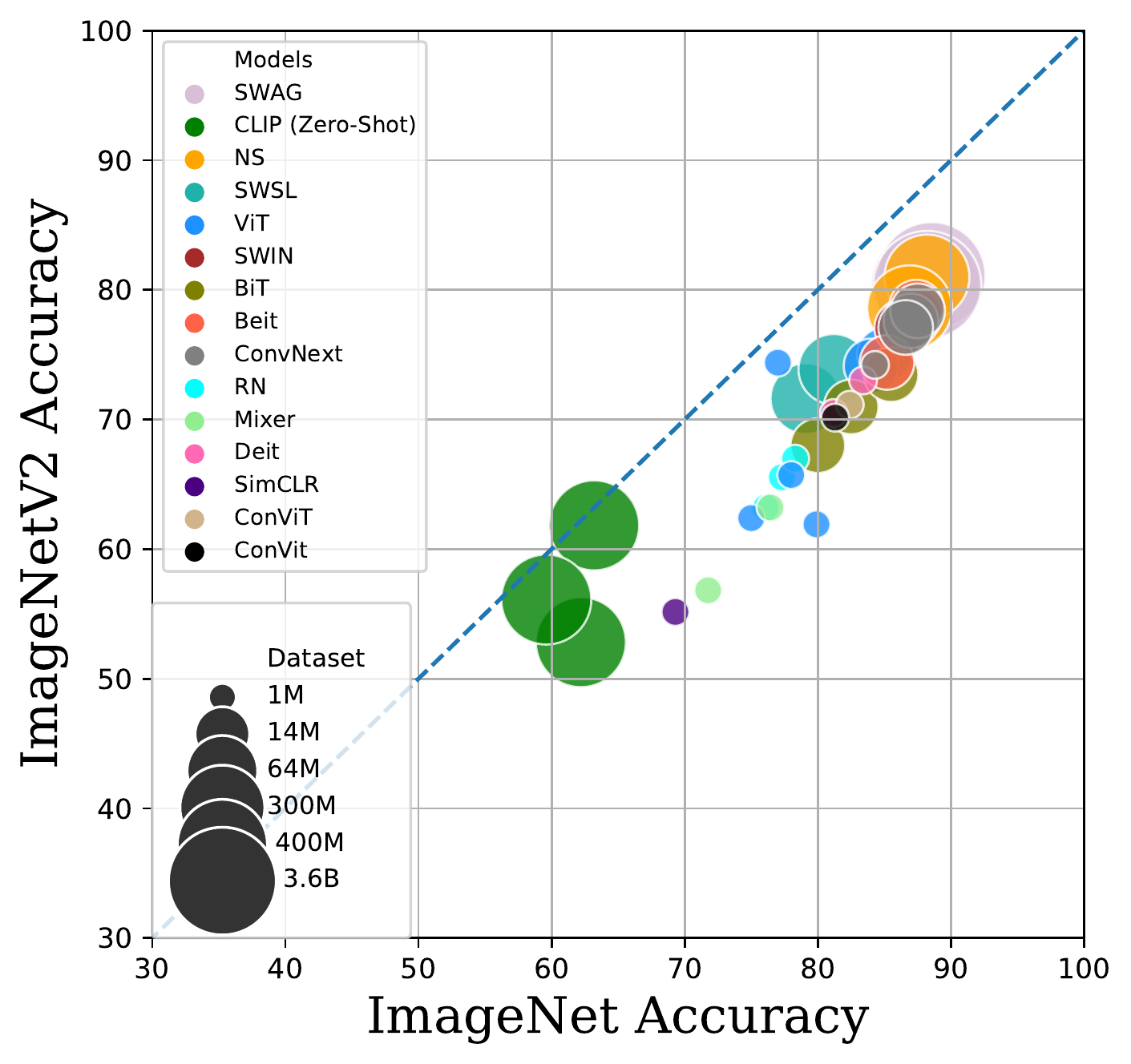}
    } 
    \subfigure{\includegraphics[width=0.30\textwidth]{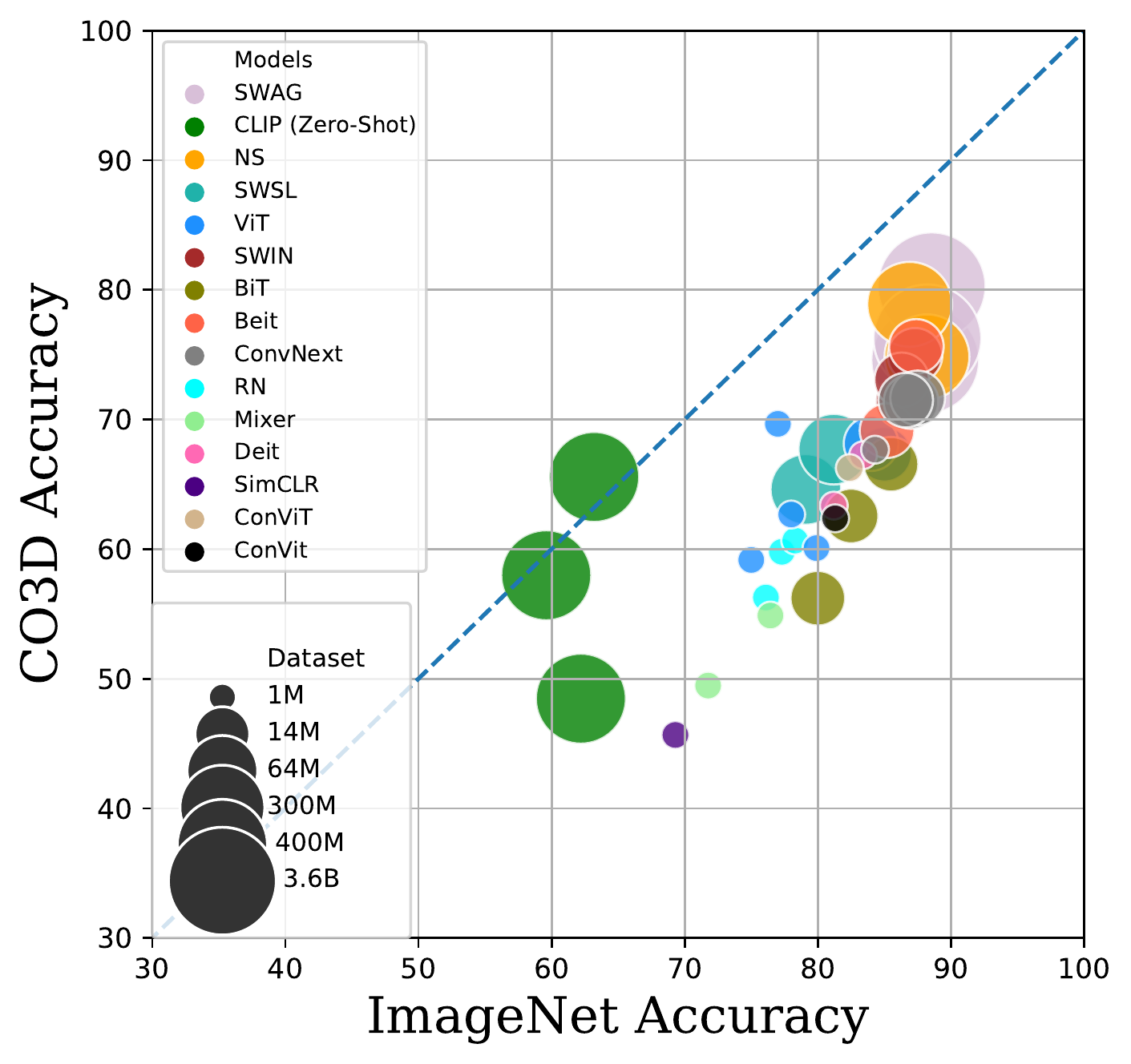}
    } 
    \subfigure{\includegraphics[width=0.30\textwidth]{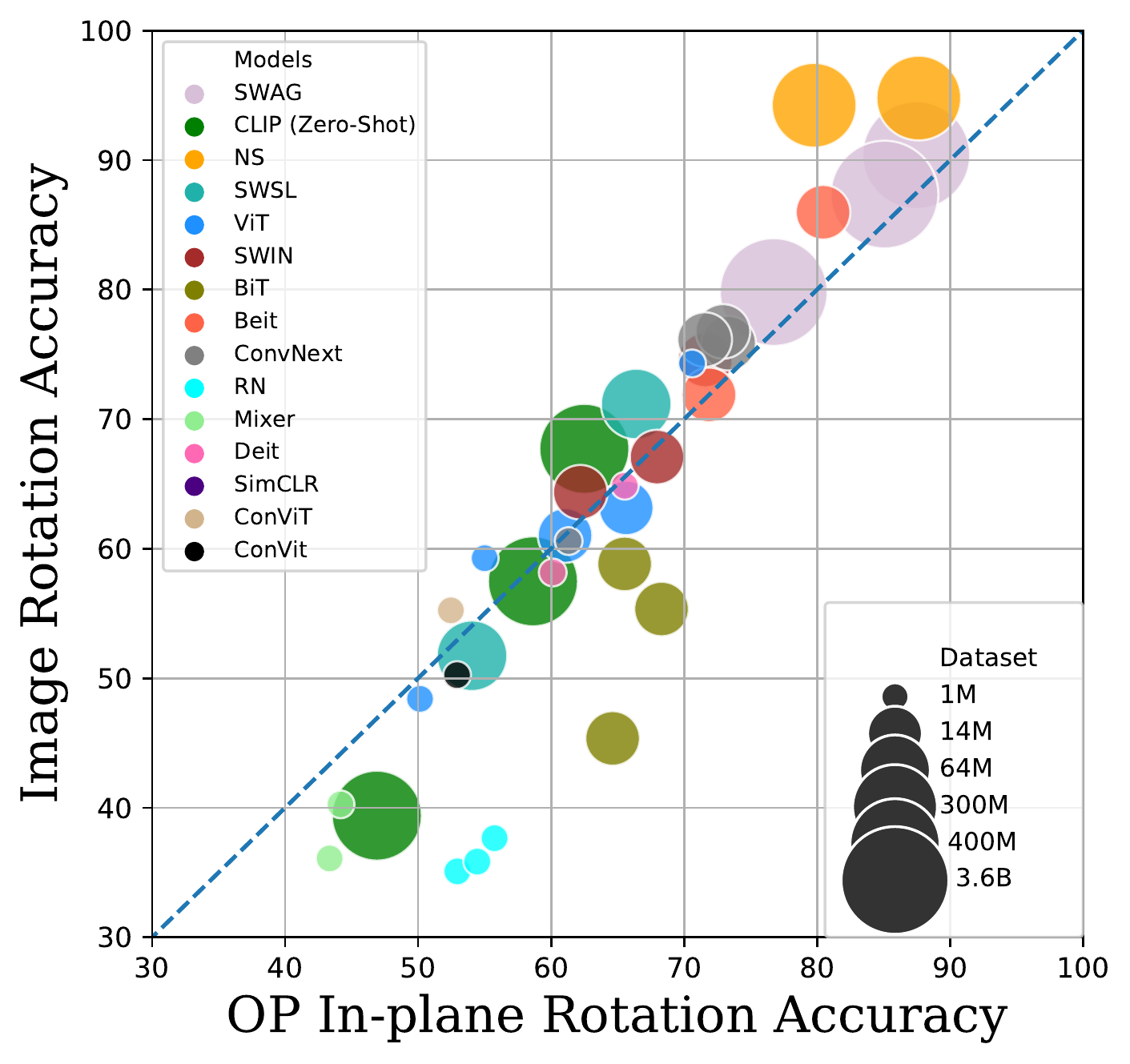}
    }      
    \subfigure{\includegraphics[width=0.30\textwidth]{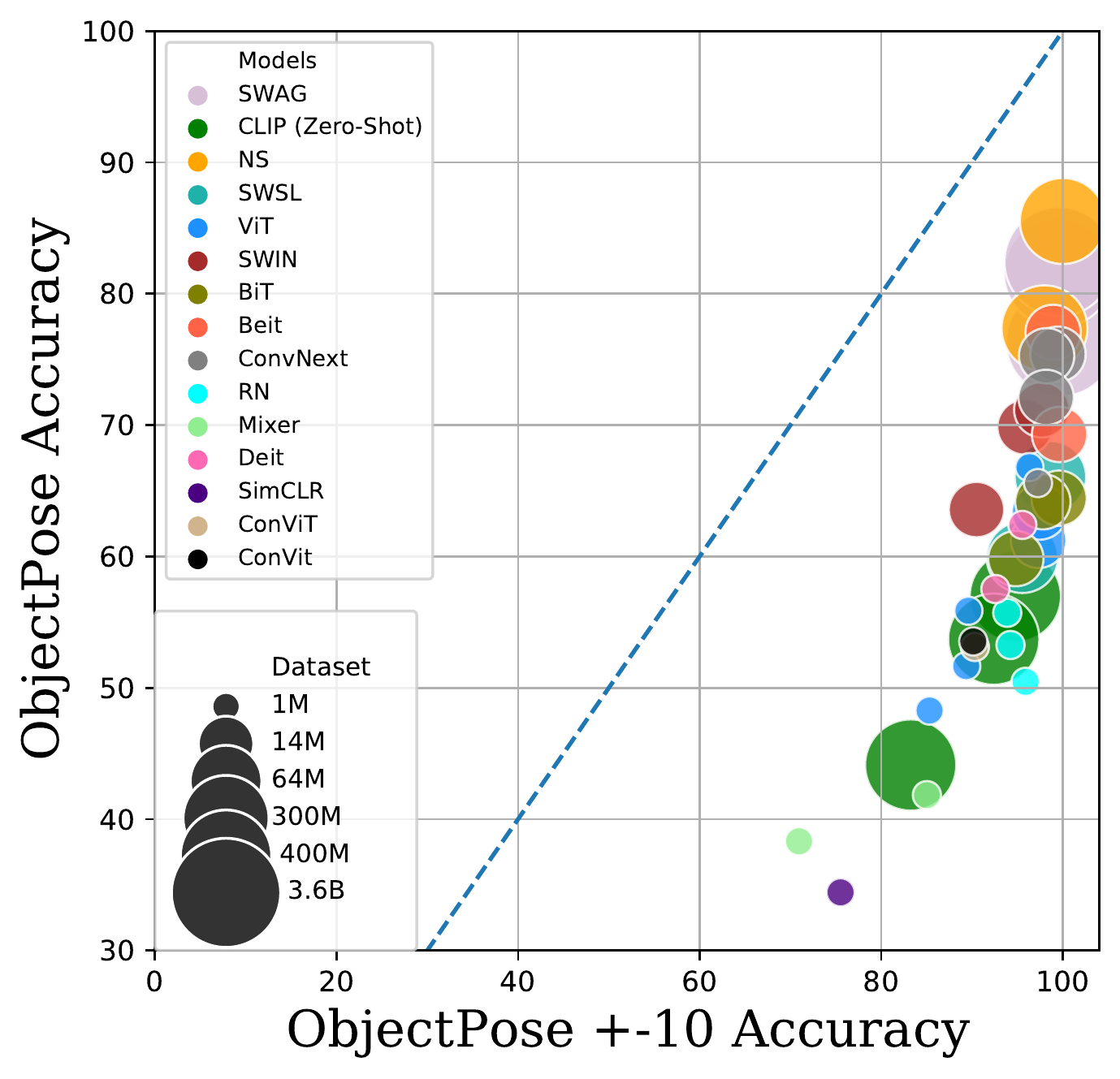}
    } 
    \subfigure{\includegraphics[width=0.30\textwidth]{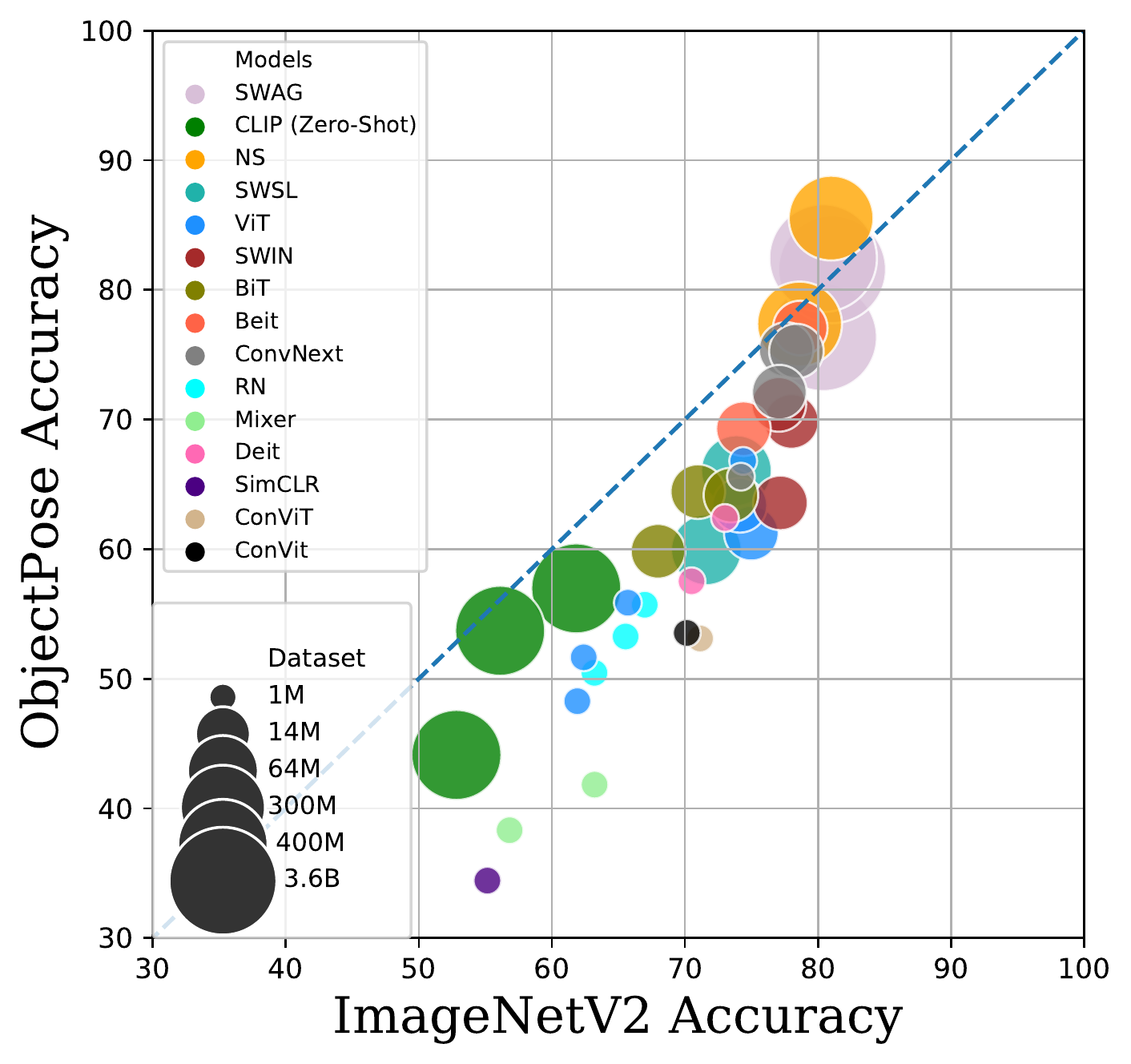}
    }      
    \subfigure{\includegraphics[width=0.30\textwidth]{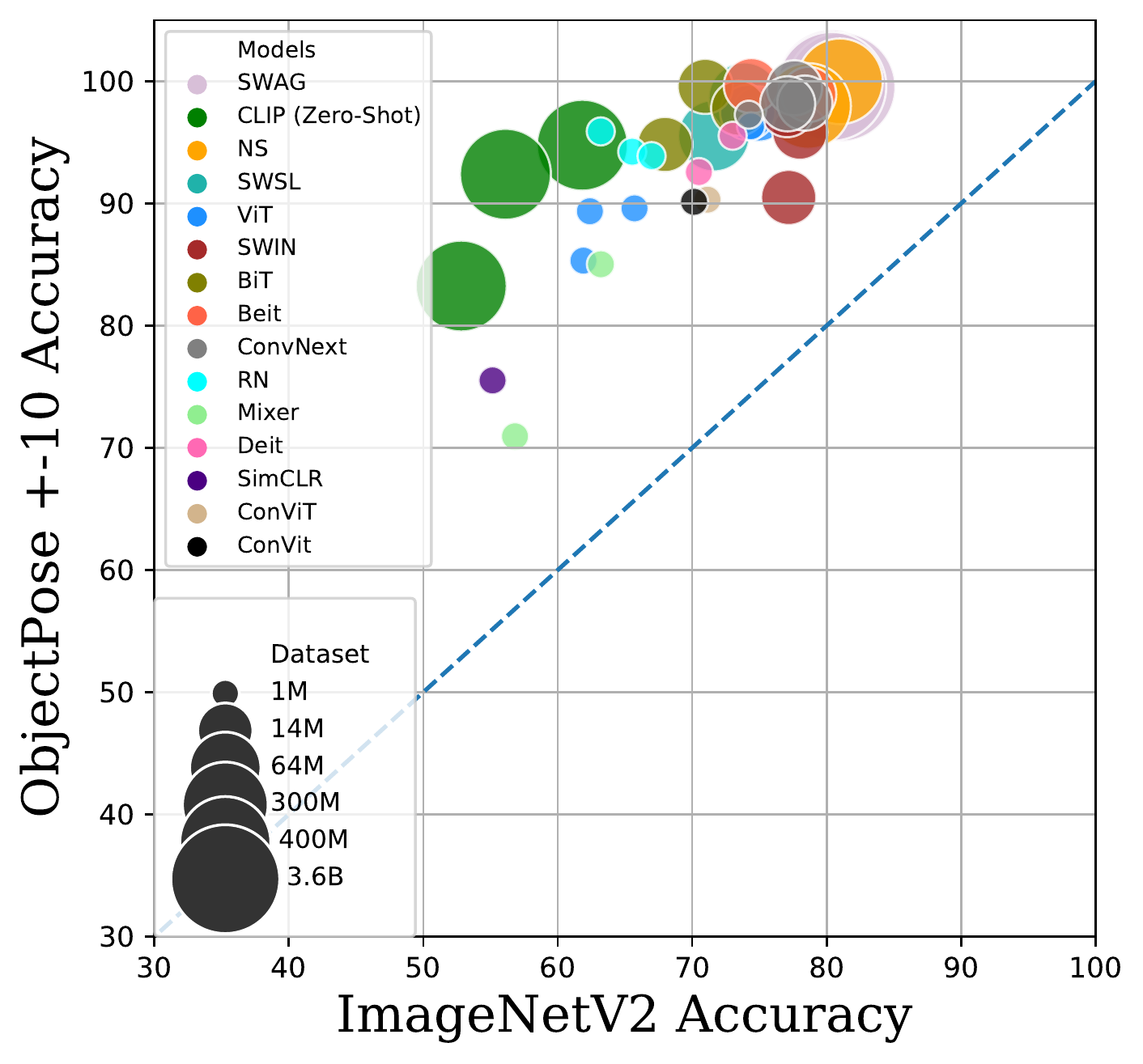}
    } 
    \subfigure{\includegraphics[width=0.30\textwidth]{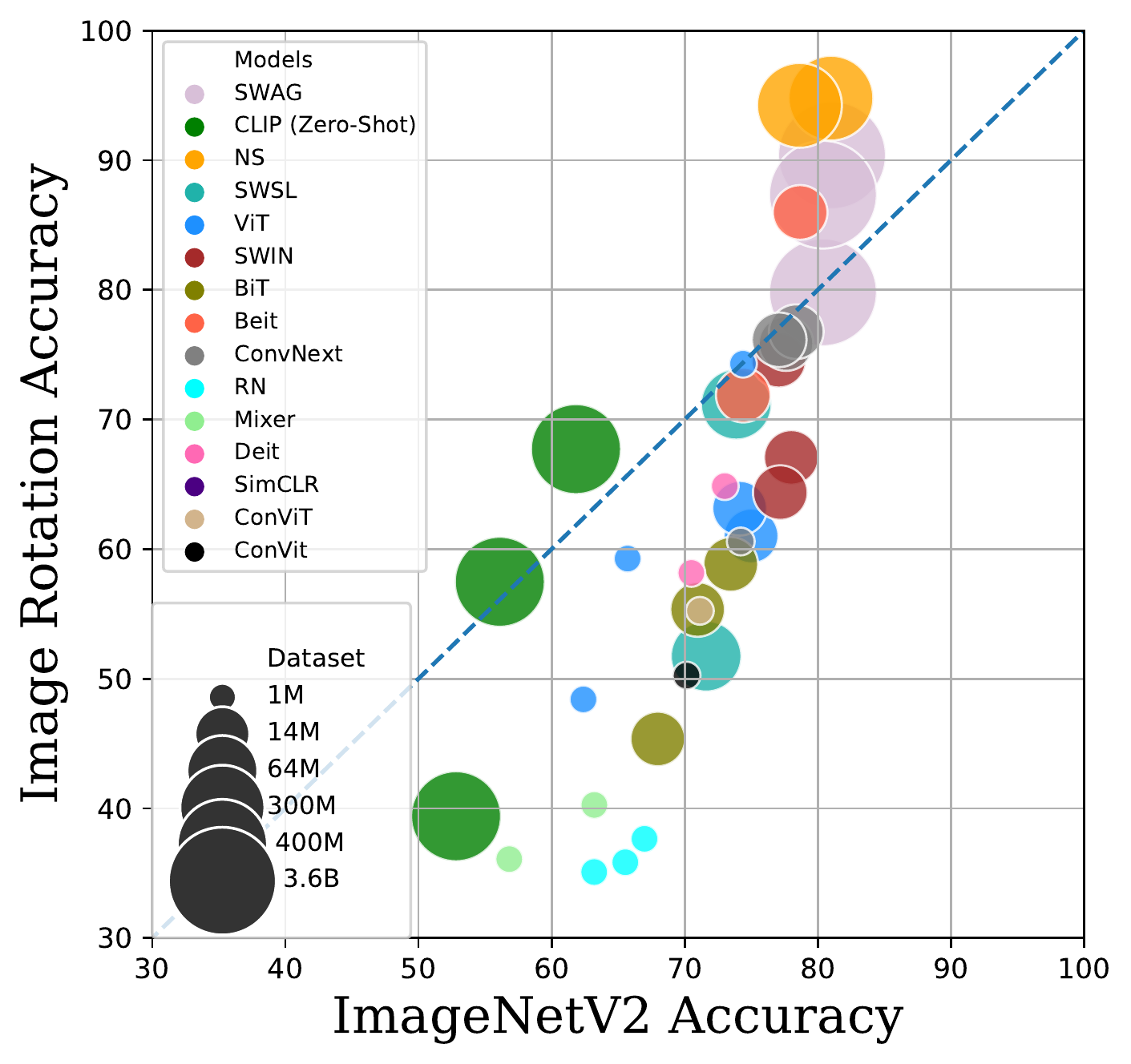}
    } 
    \subfigure{\includegraphics[width=0.30\textwidth]{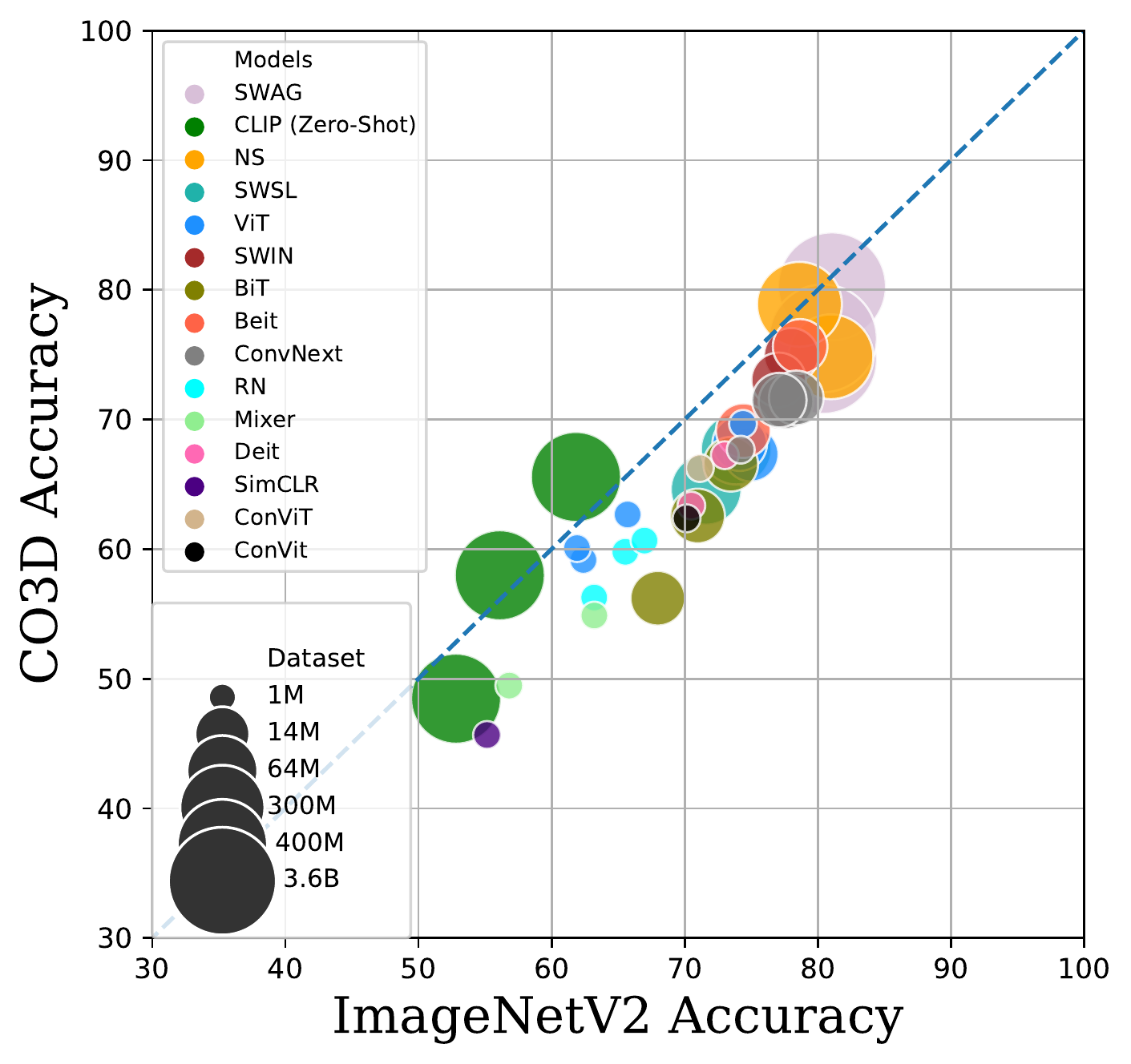}
    } 
    \subfigure{\includegraphics[width=0.30\textwidth]{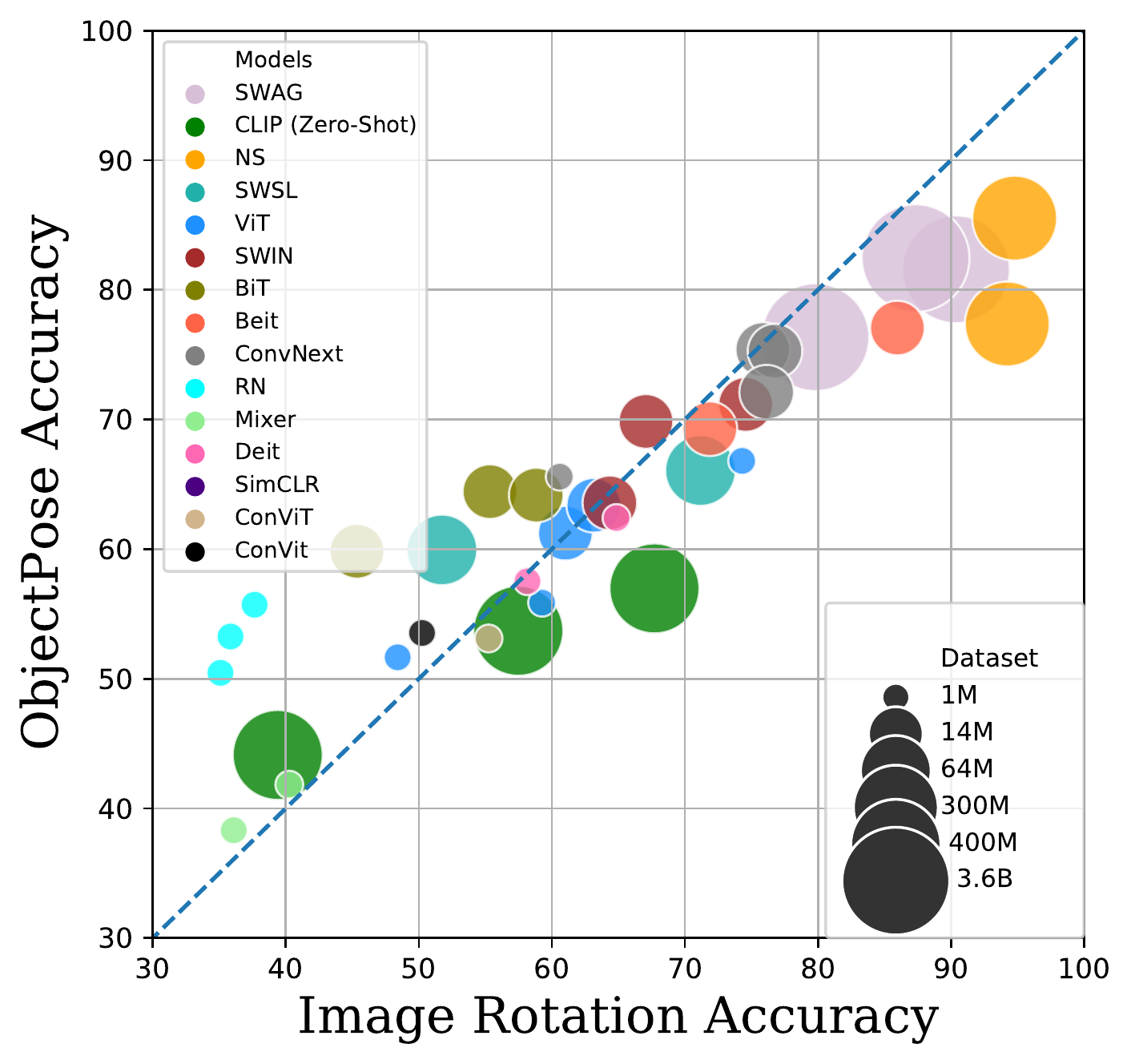}
    } 
    \subfigure{\includegraphics[width=0.30\textwidth]{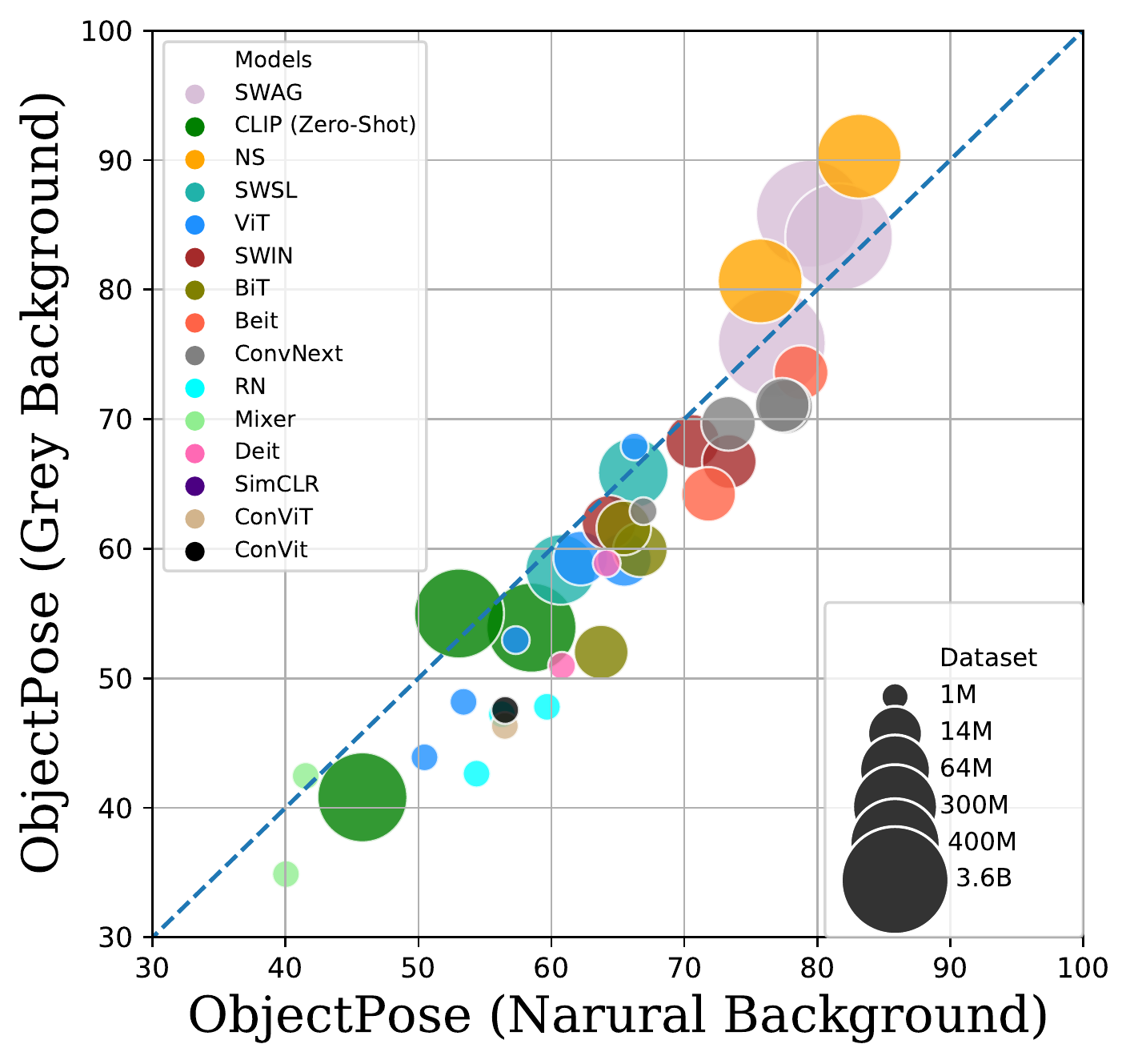}
    } 
    \subfigure{\includegraphics[width=0.30\textwidth]{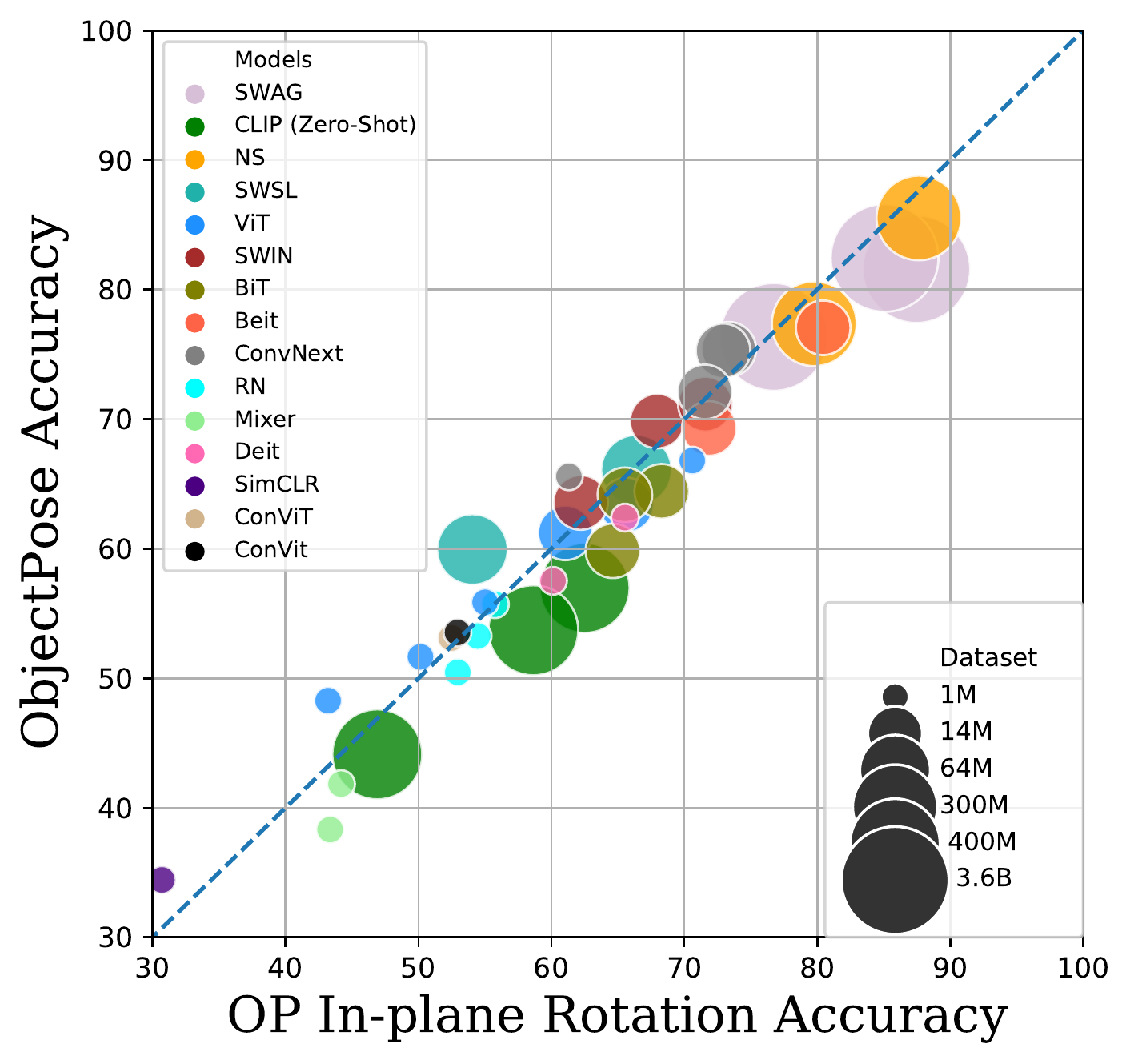}
    }     
    \subfigure{\includegraphics[width=0.30\textwidth]{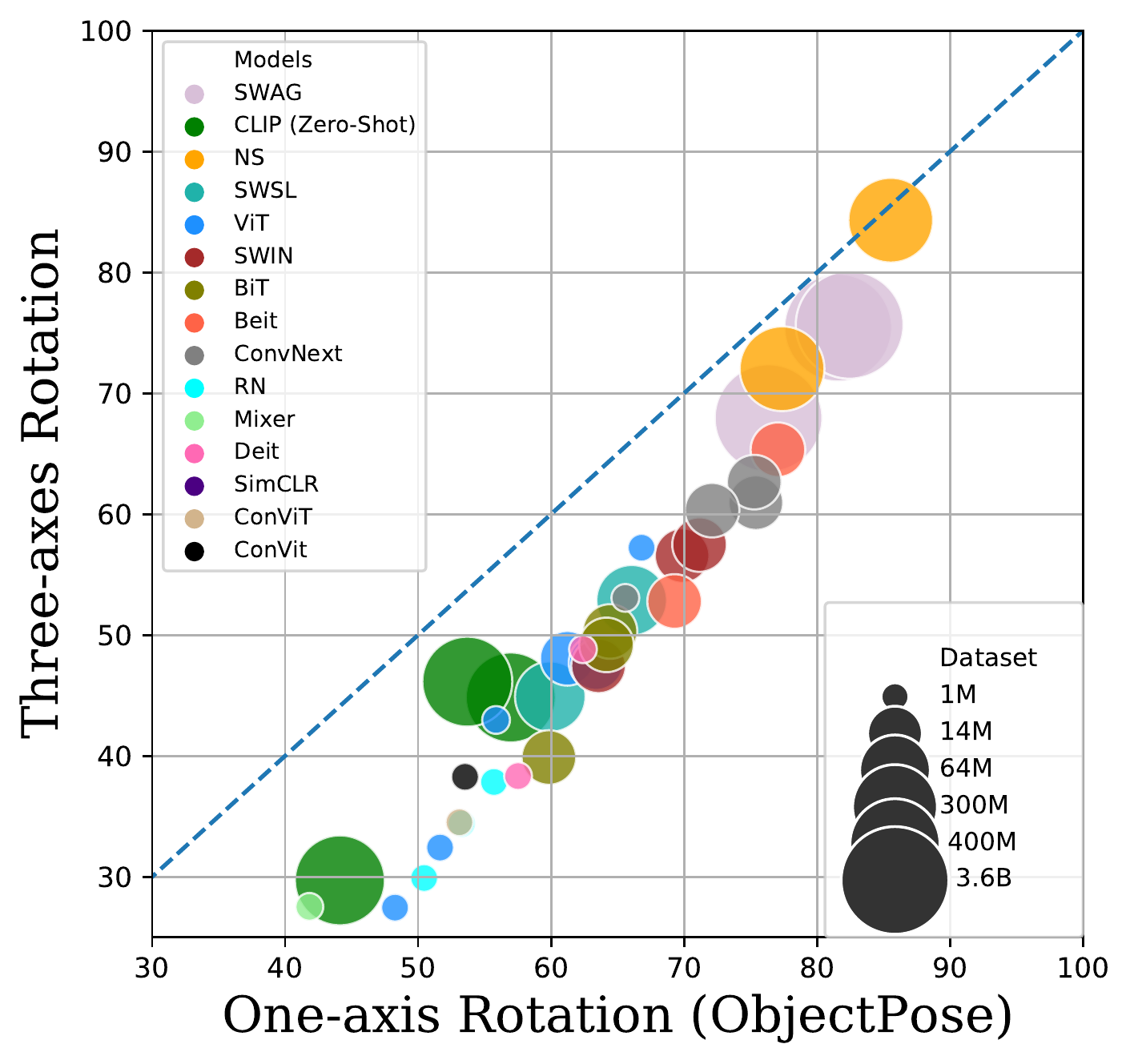}
    } 
    \caption{Correlations between the performance of \nummodels models on the different datasets (ImageNet, ObjectPose, ObjectPose +-10, ImageNetV2, and CO3D).}
    \label{fig:allaccvsacc}
\end{figure}


\begin{figure}[h]
    \centering
    \includegraphics[width=0.245\textwidth]{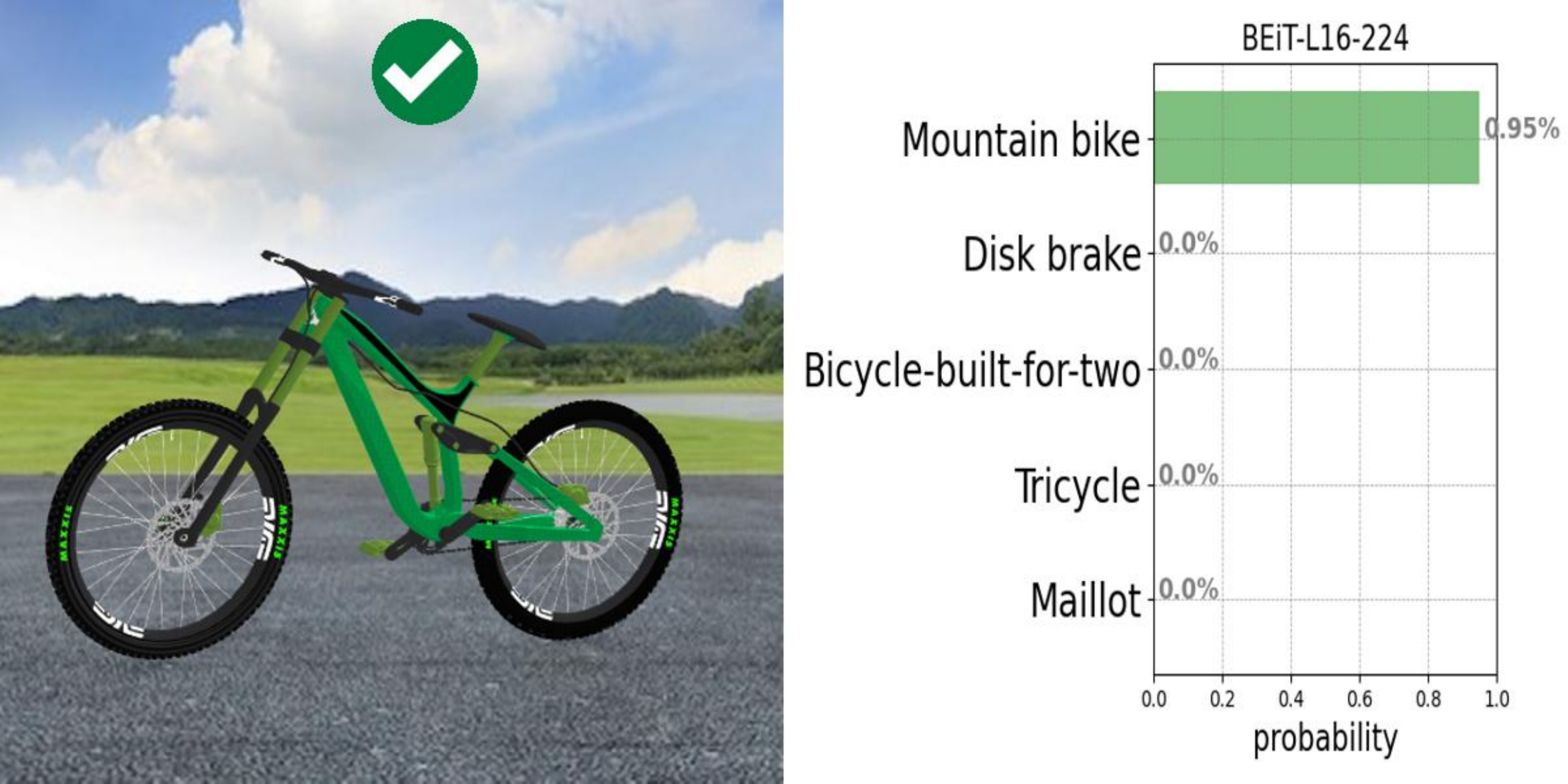}
    \includegraphics[width=0.245\textwidth]{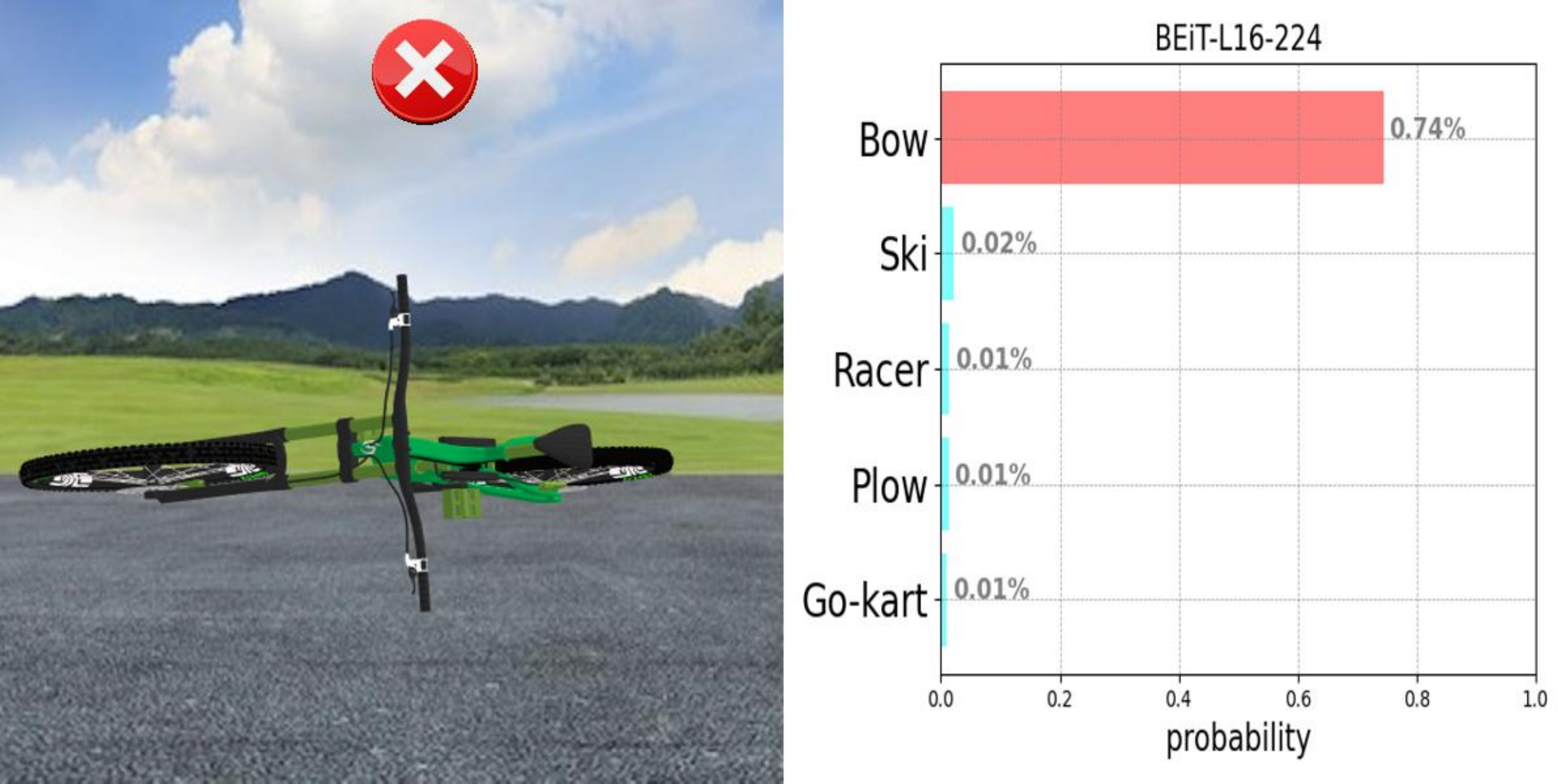}
    \includegraphics[width=0.245\textwidth]{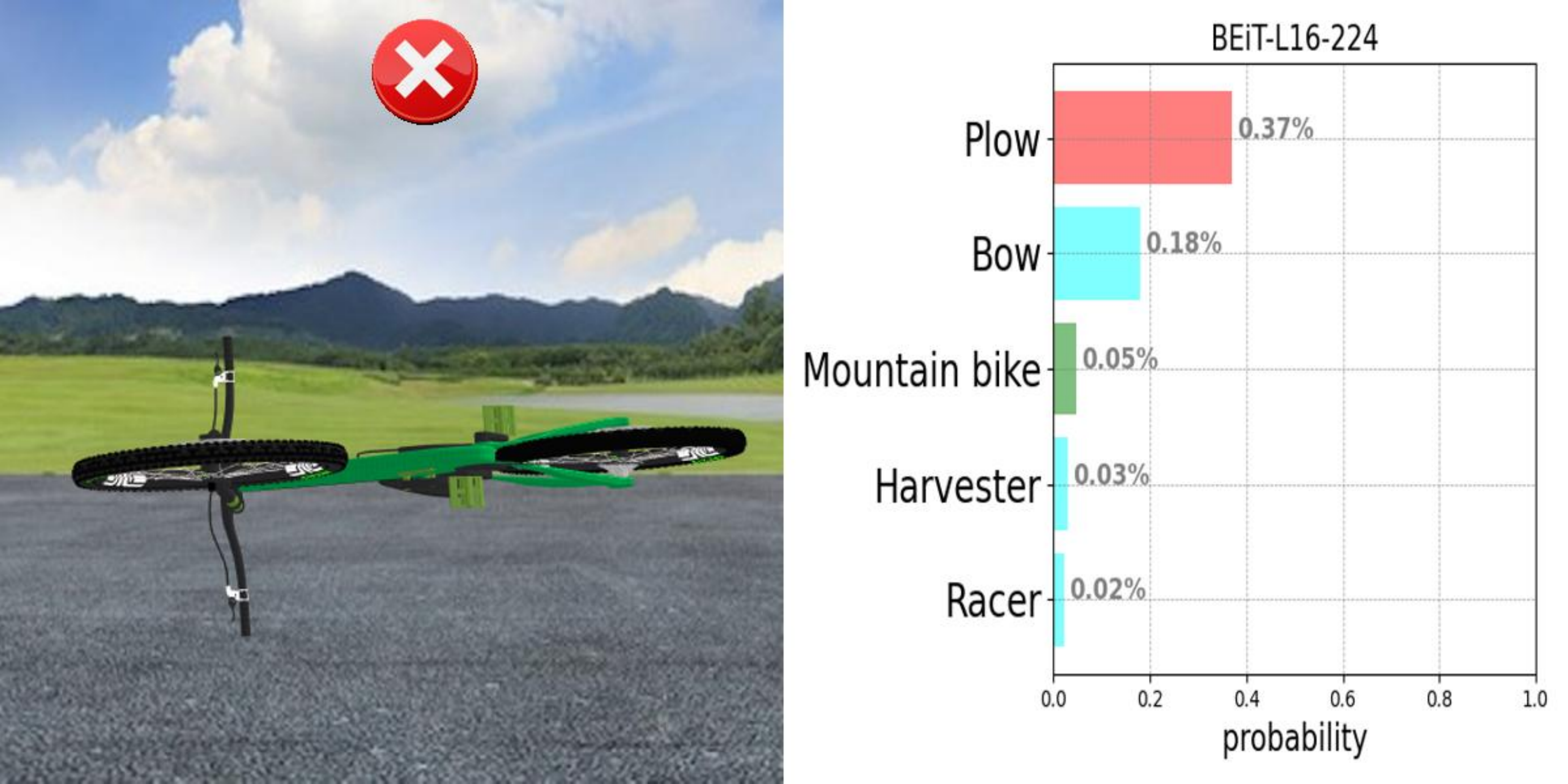}
    \includegraphics[width=0.245\textwidth]{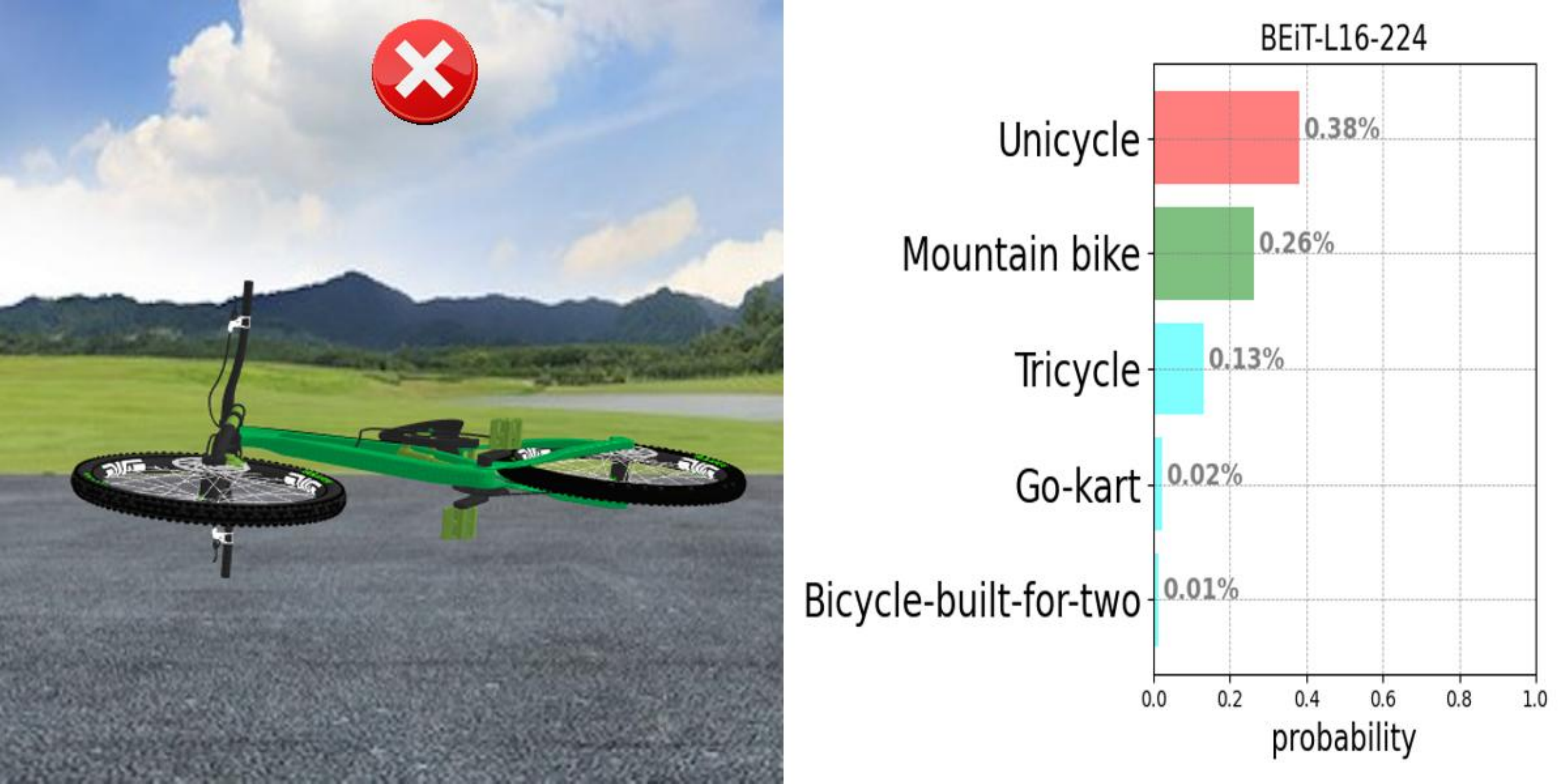}

    \includegraphics[width=0.245\textwidth]{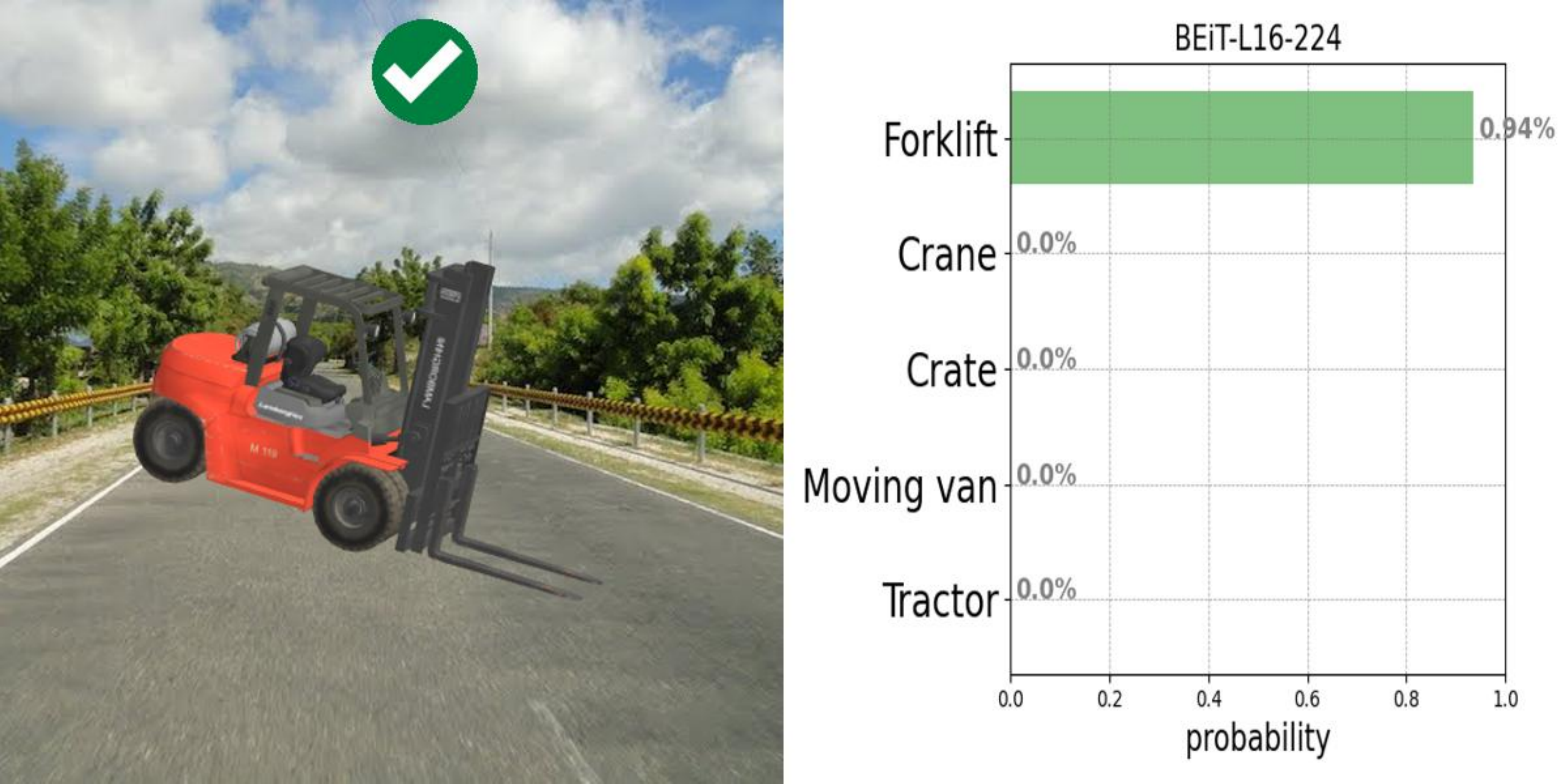}
    \includegraphics[width=0.245\textwidth]{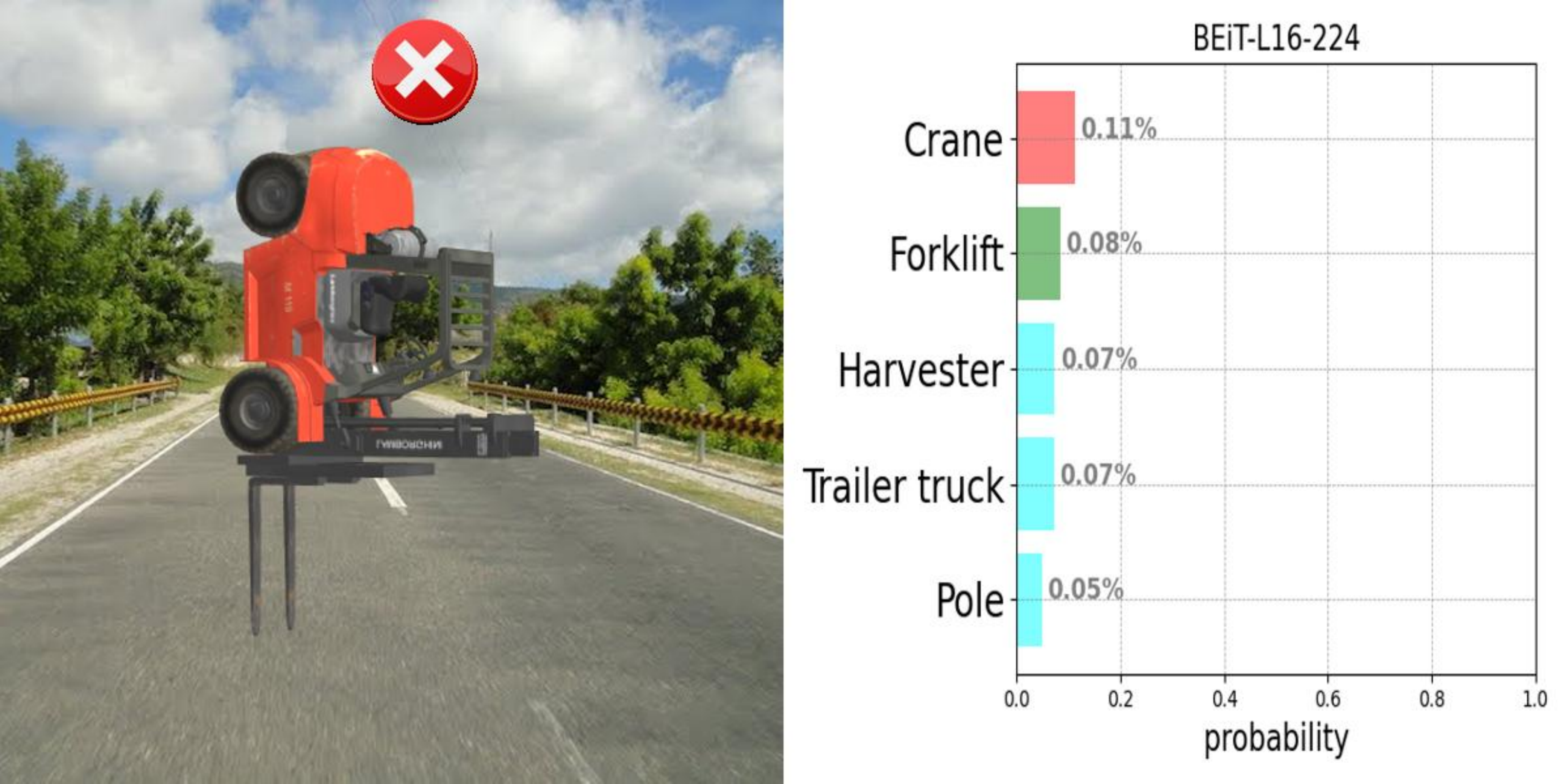}
    \includegraphics[width=0.245\textwidth]{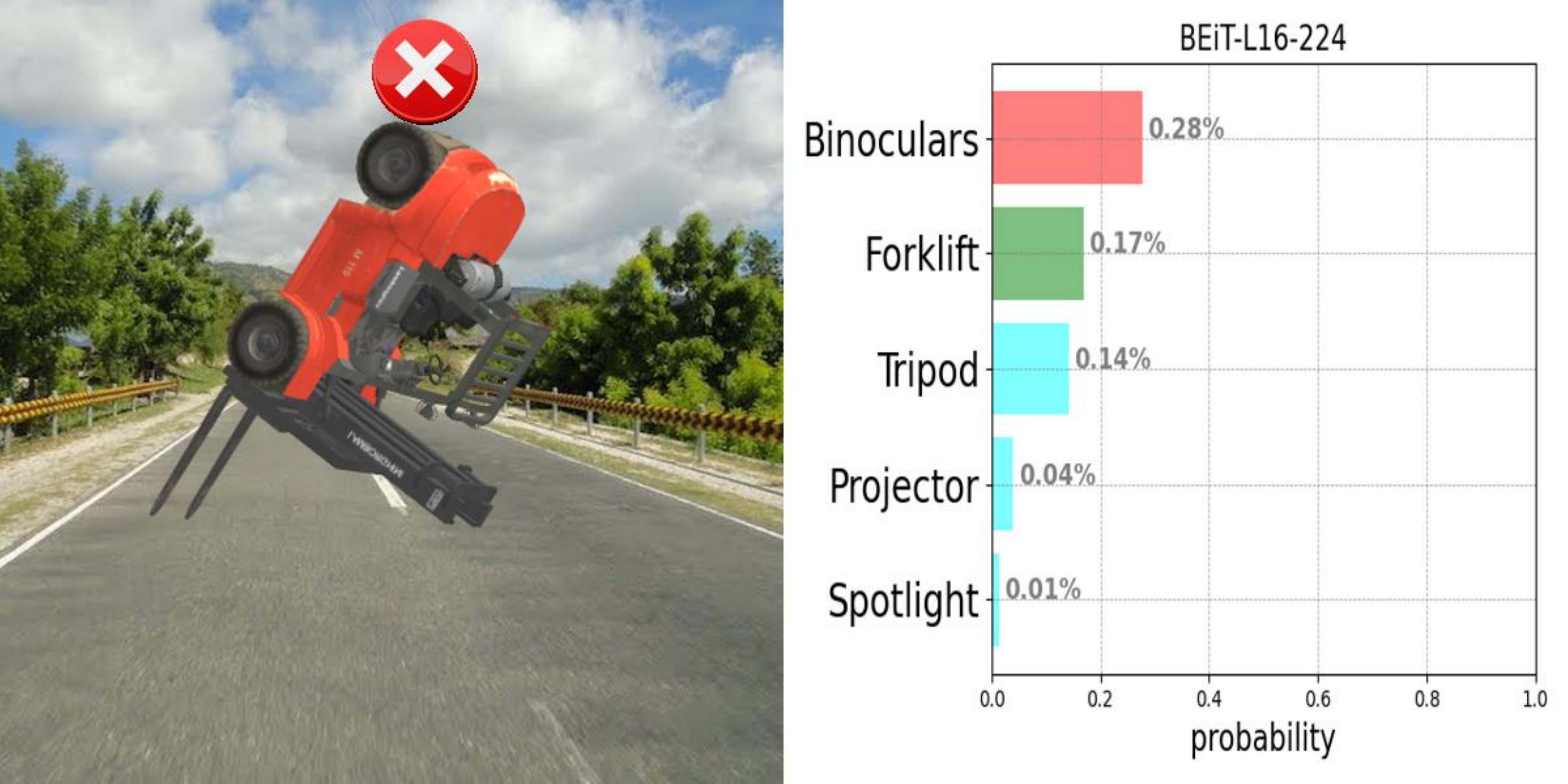}
    \includegraphics[width=0.245\textwidth]{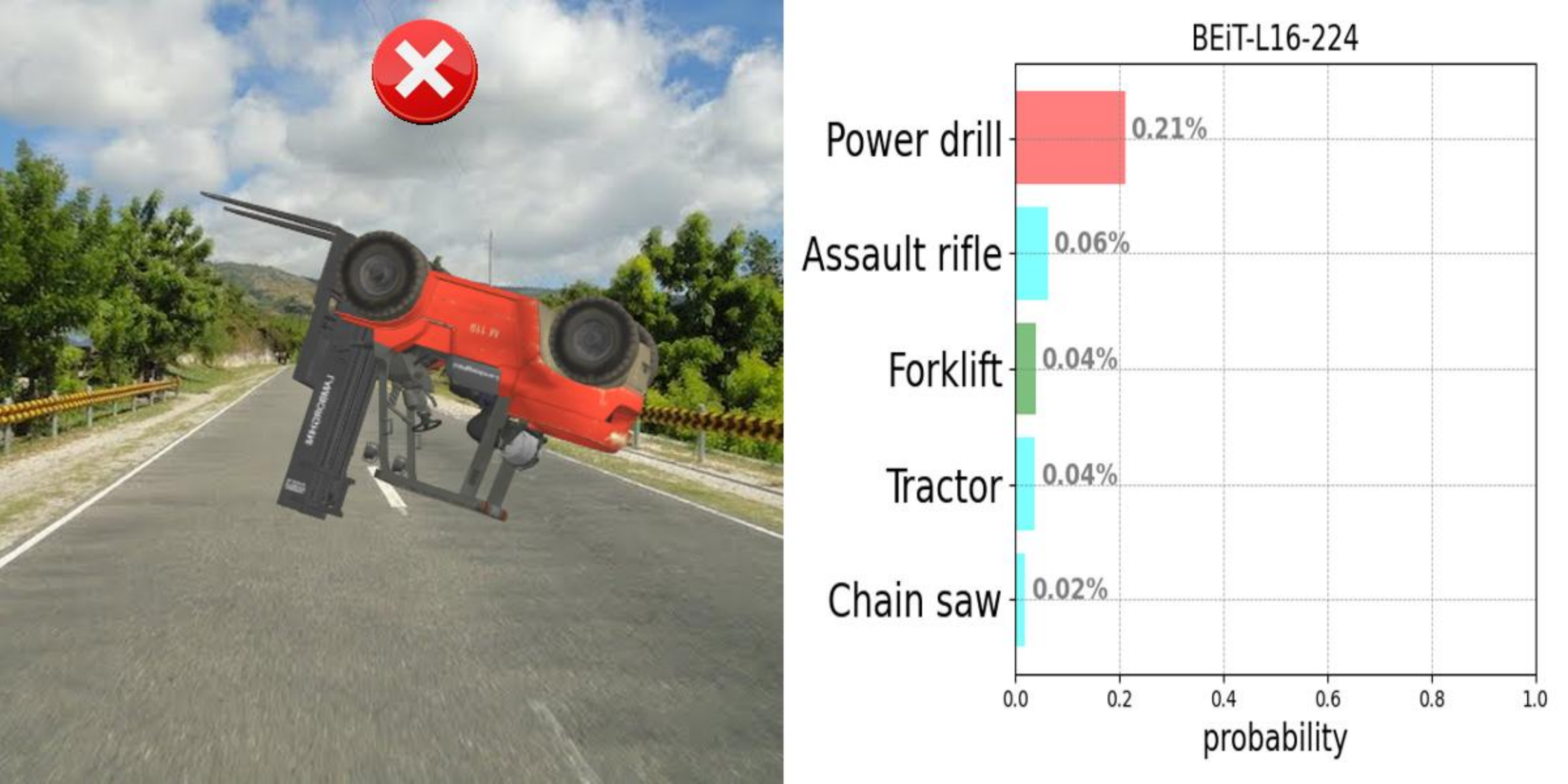}
    
    \includegraphics[width=0.245\textwidth]{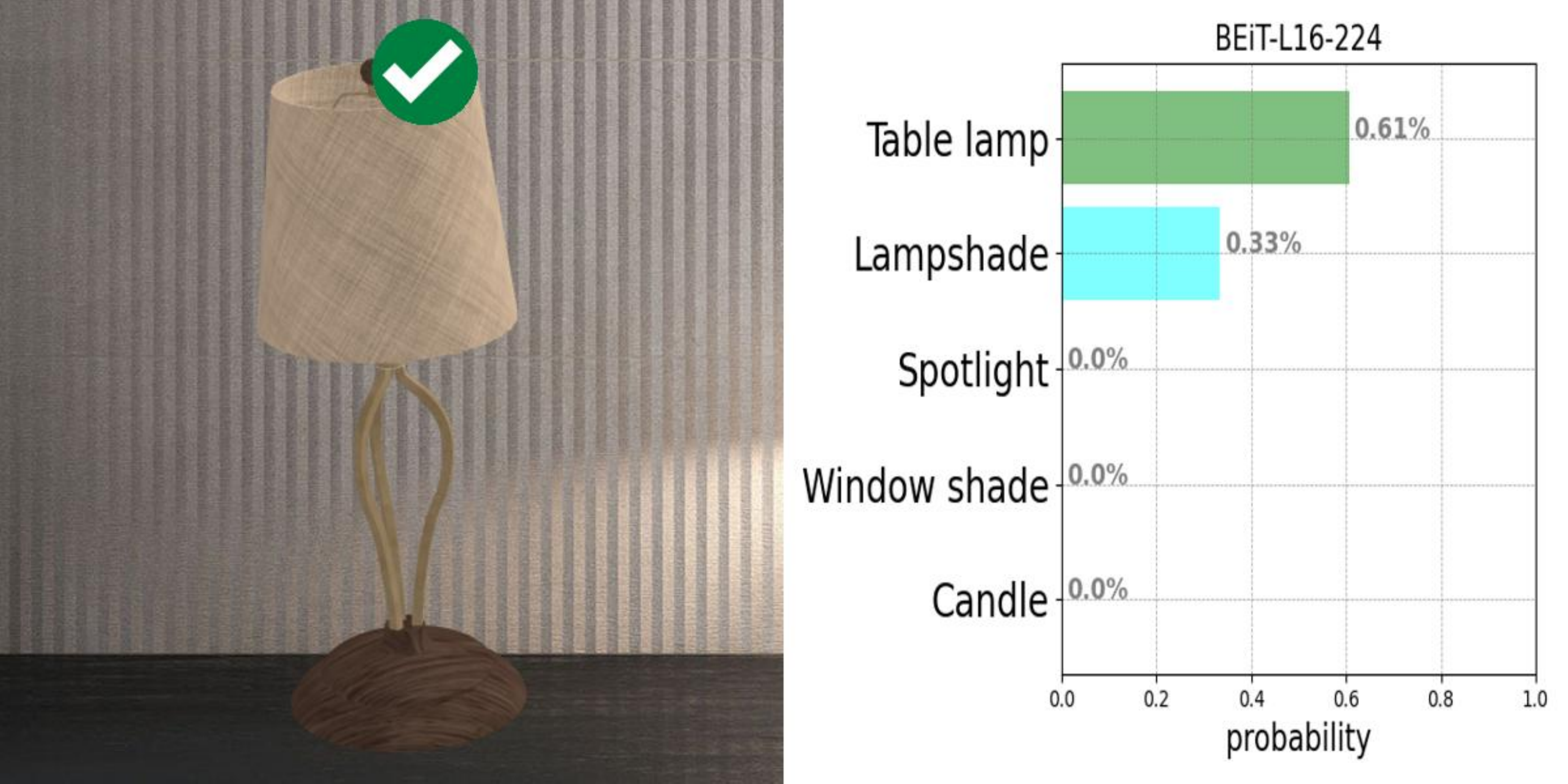}
    \includegraphics[width=0.245\textwidth]{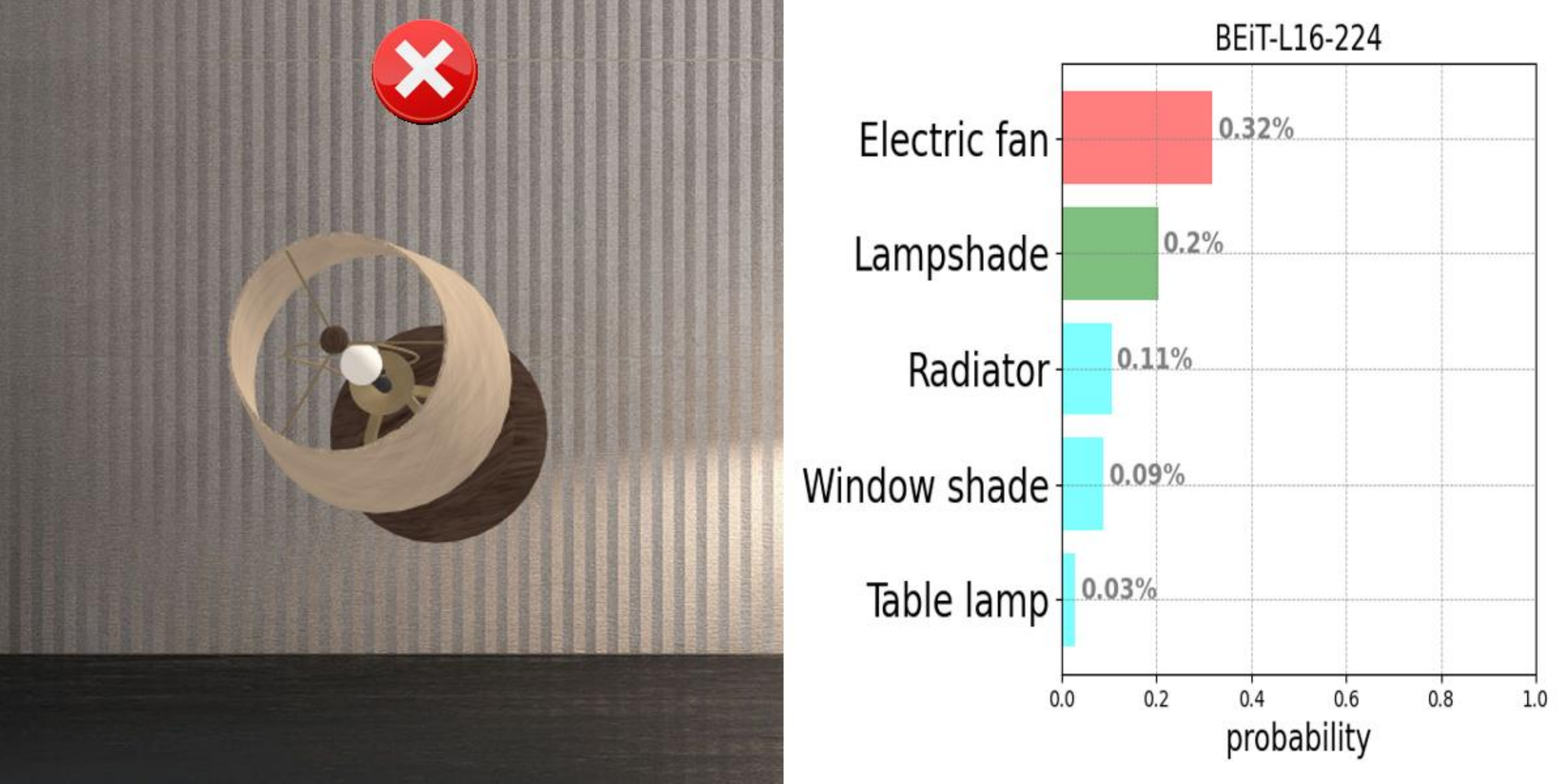}
    \includegraphics[width=0.245\textwidth]{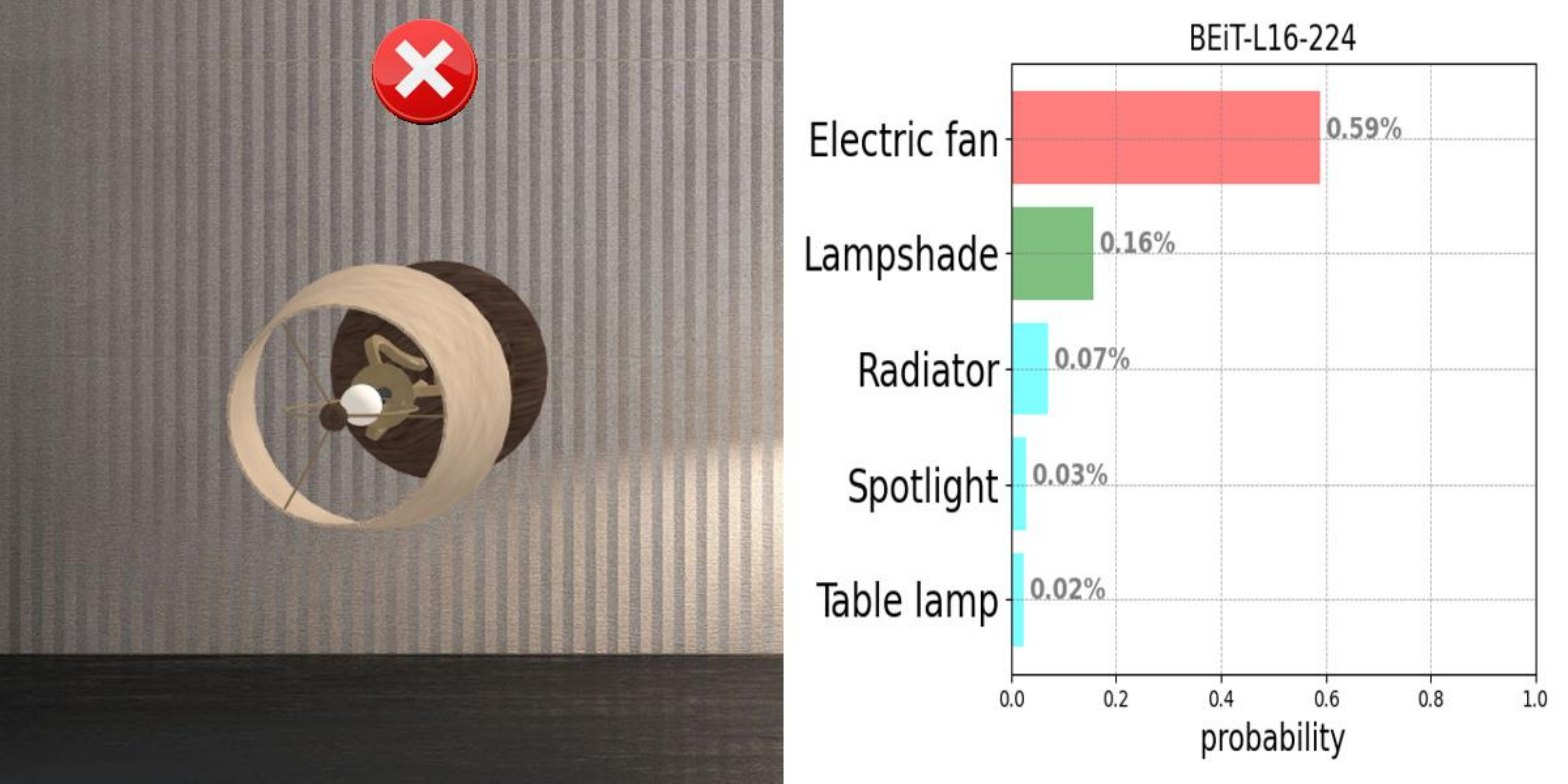}
    \includegraphics[width=0.245\textwidth]{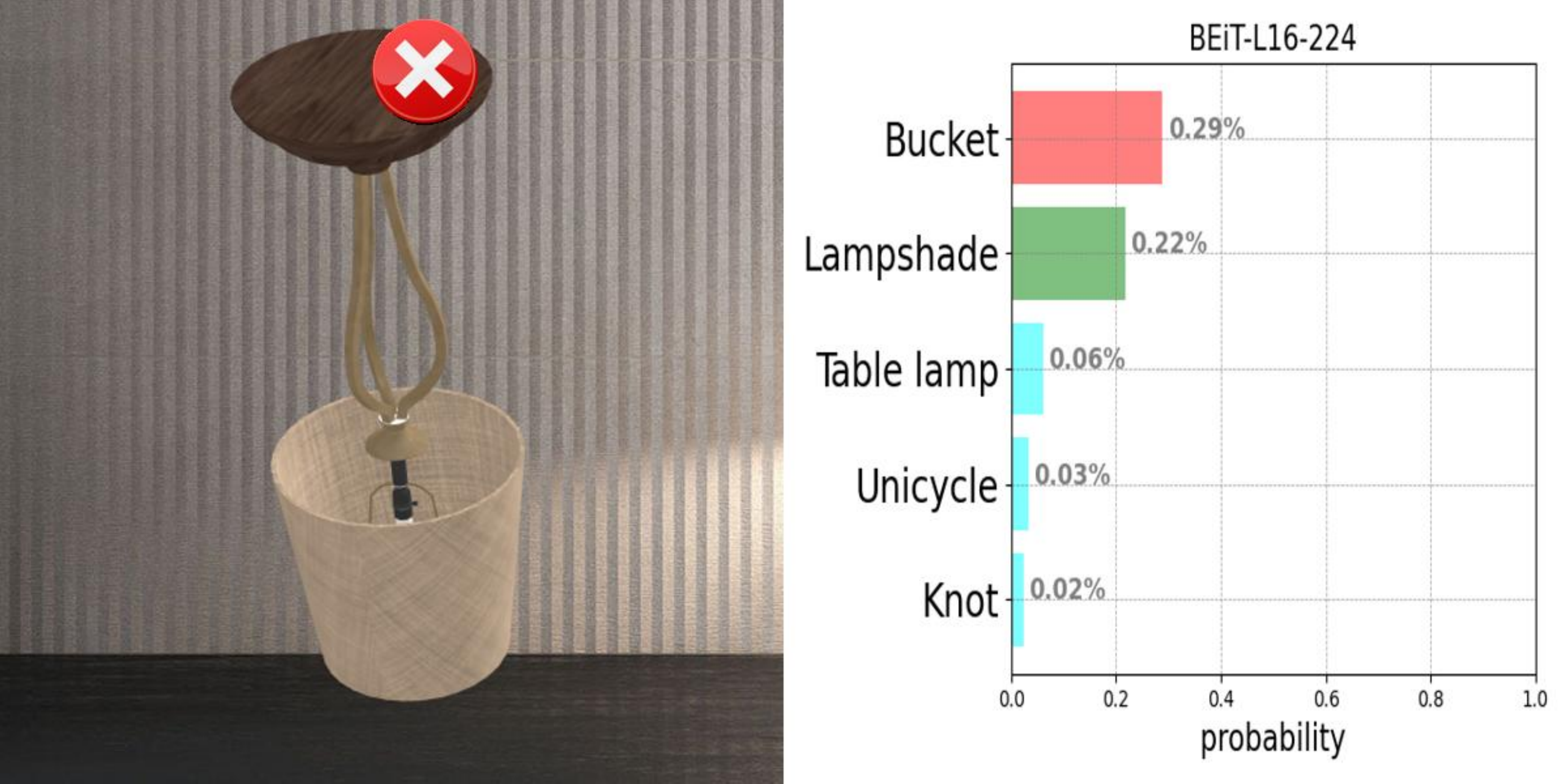}
    
    \includegraphics[width=0.245\textwidth]{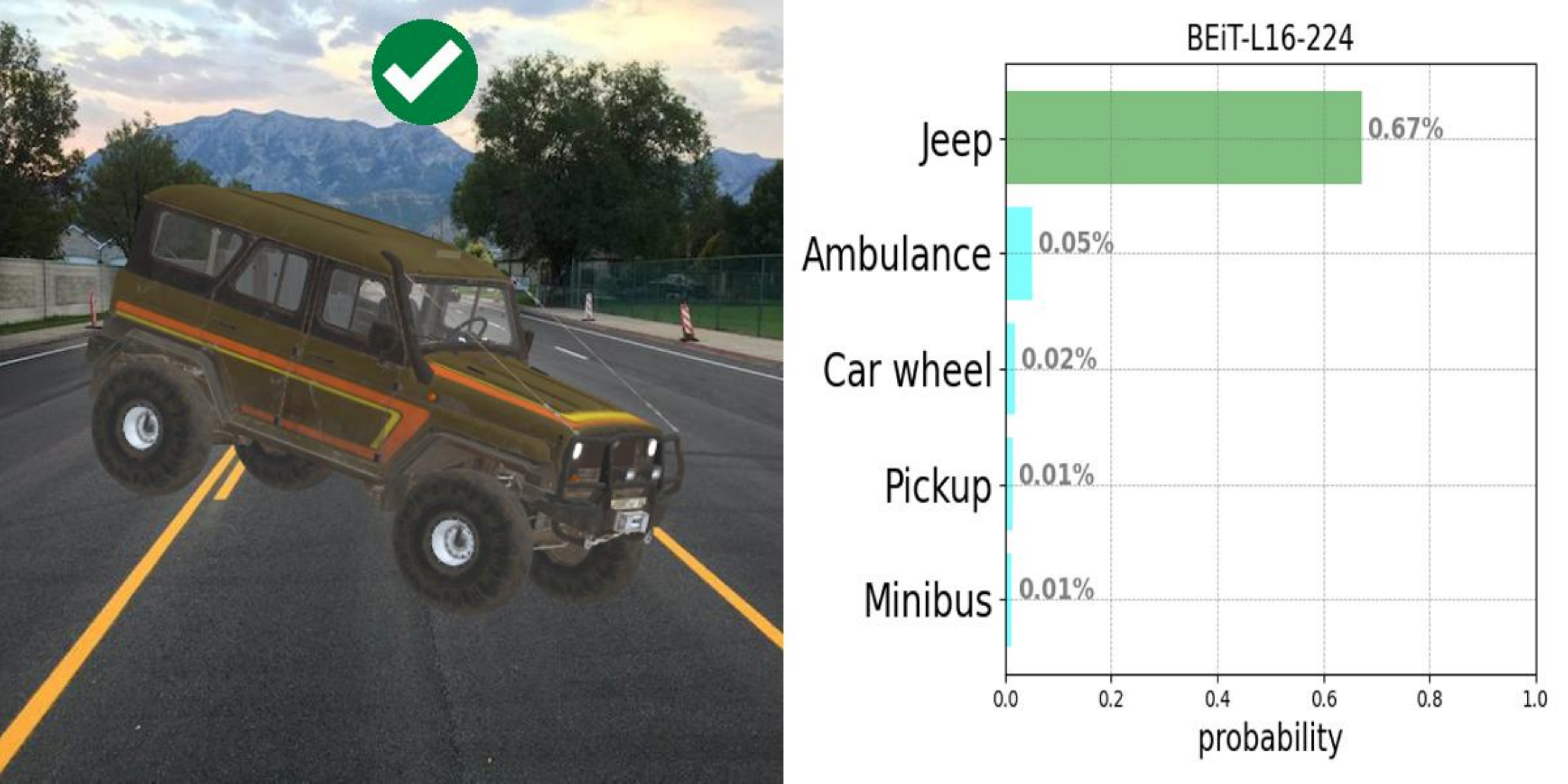}
    \includegraphics[width=0.245\textwidth]{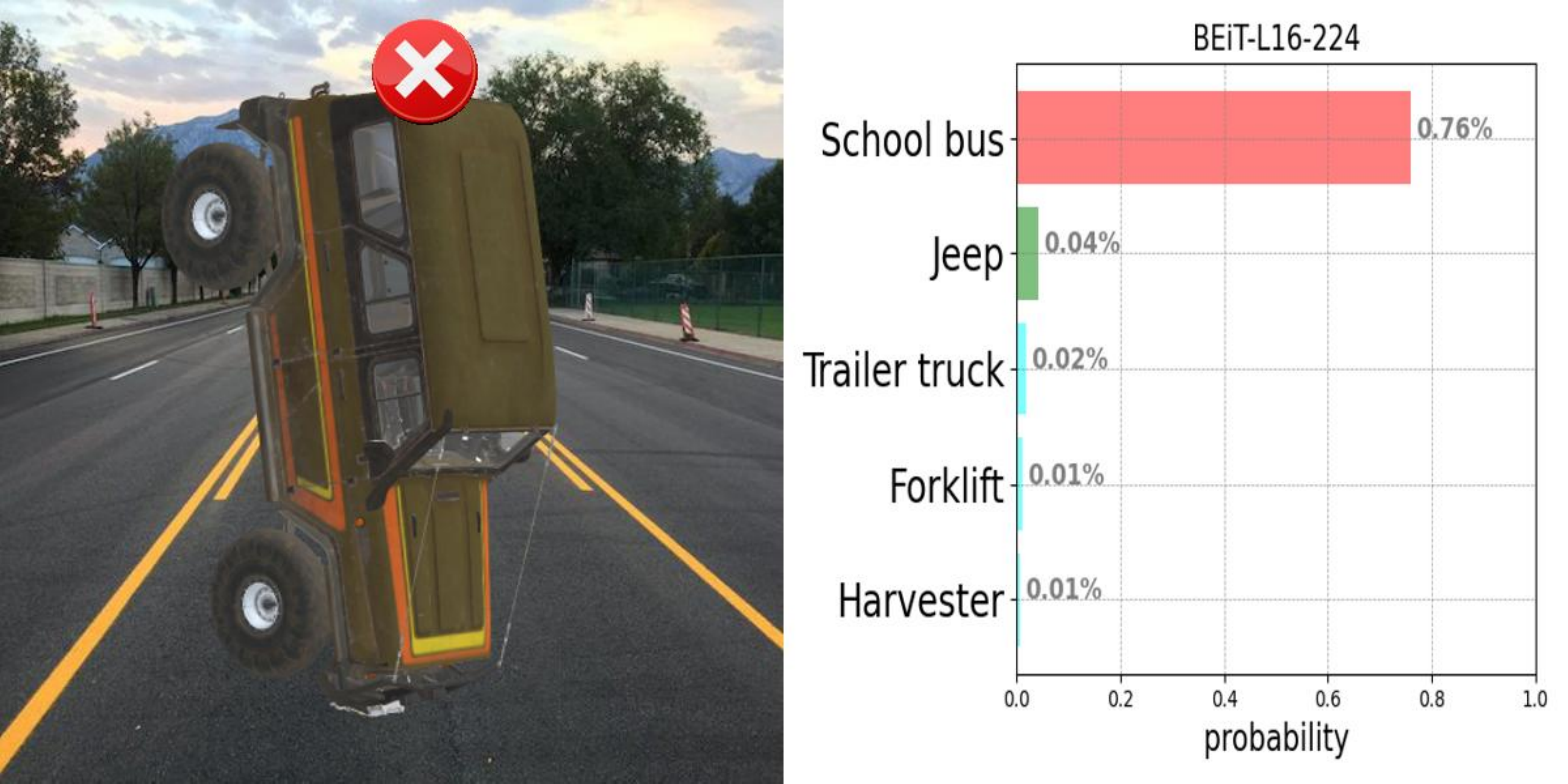}
    \includegraphics[width=0.245\textwidth]{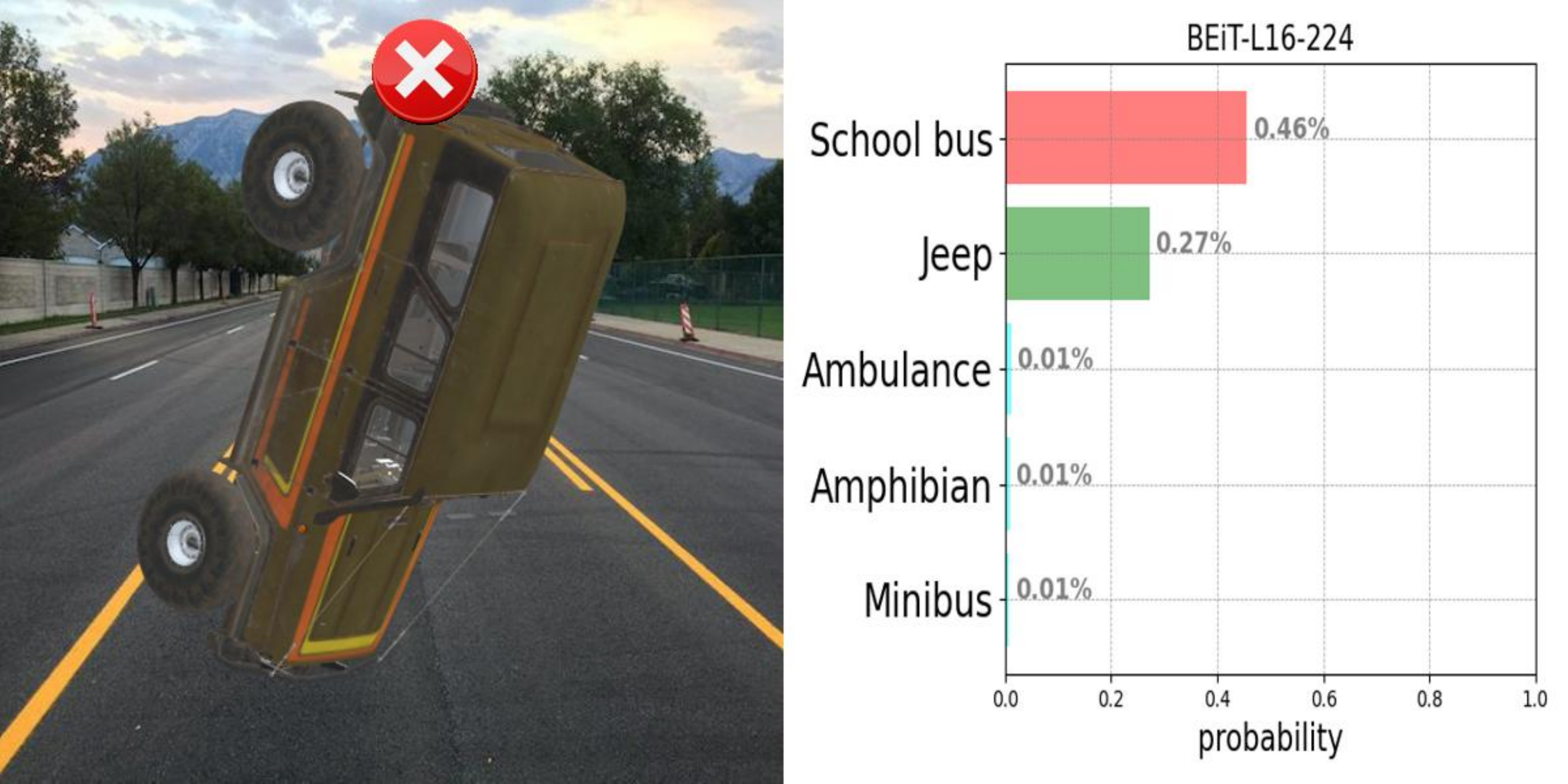}
    \includegraphics[width=0.245\textwidth]{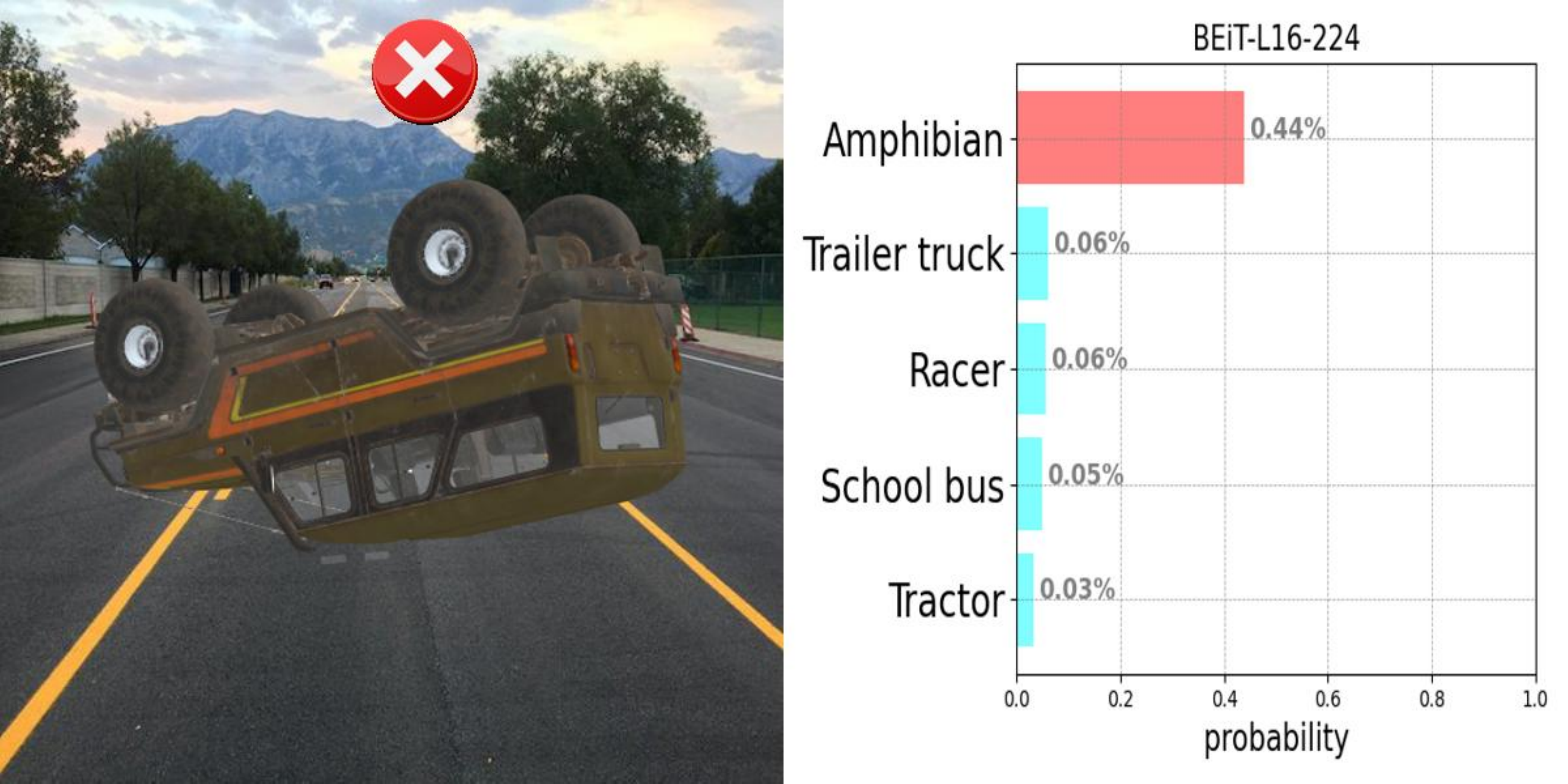}
    
    \includegraphics[width=0.245\textwidth]{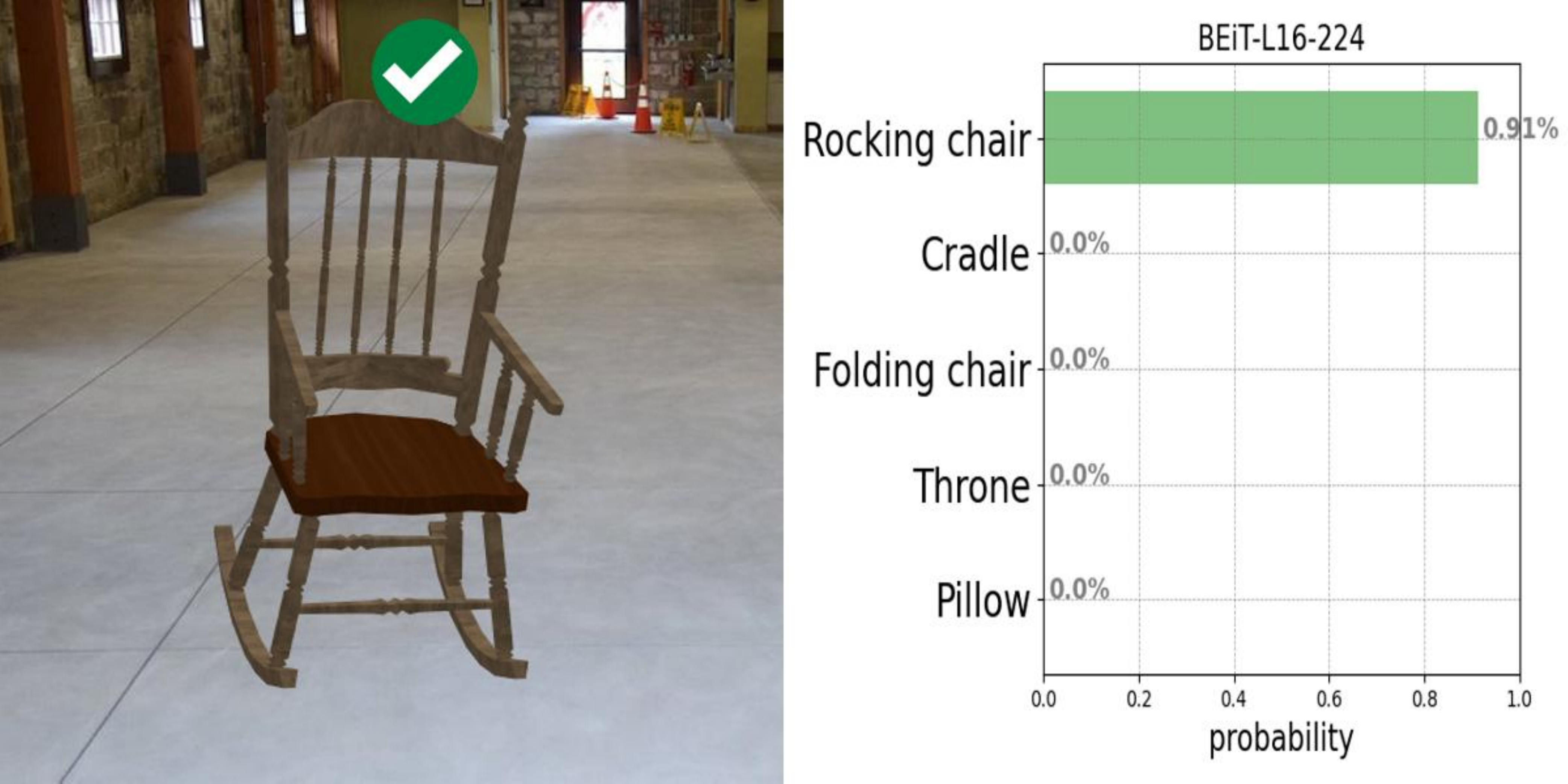}
    \includegraphics[width=0.245\textwidth]{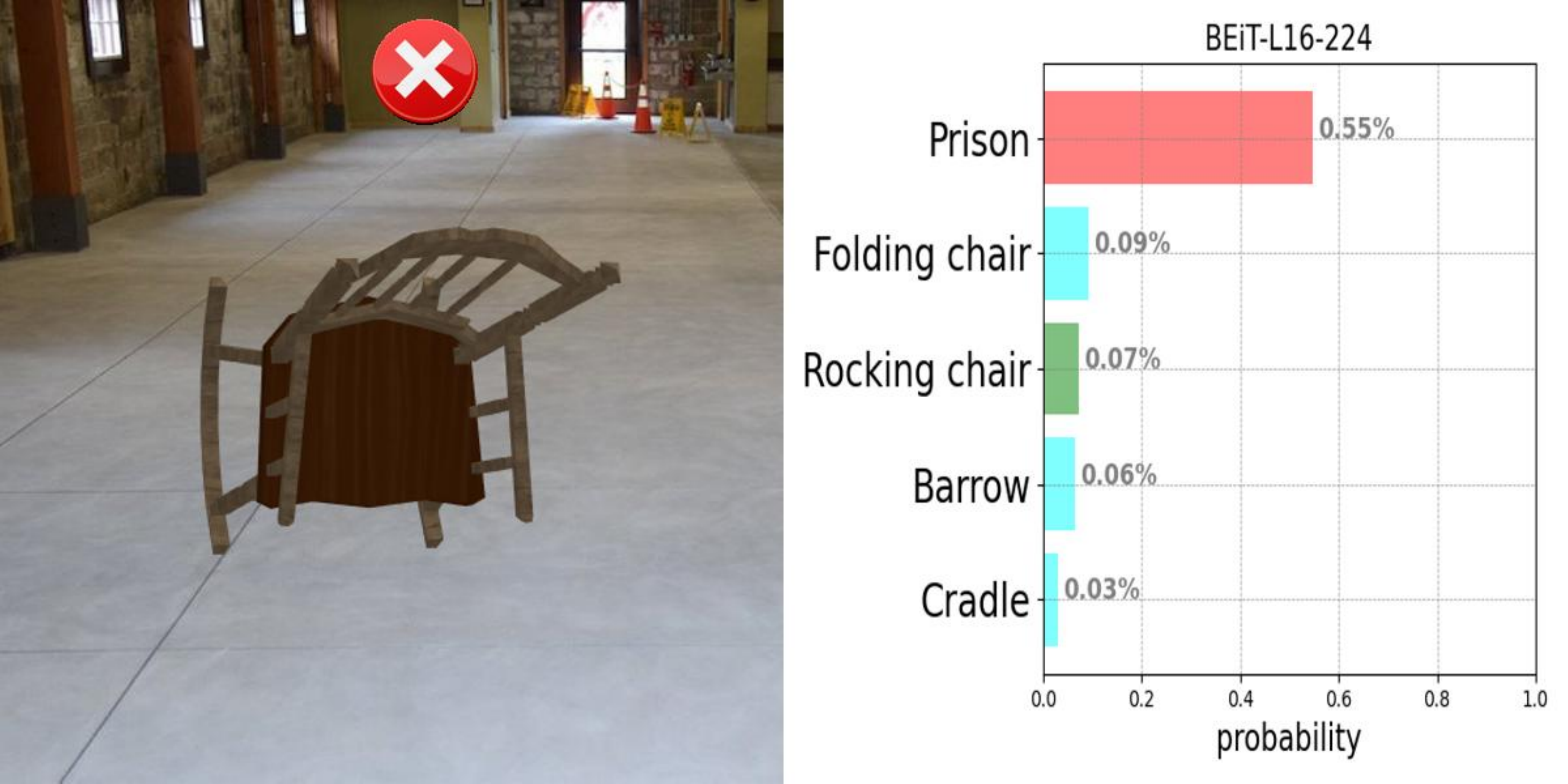}
    \includegraphics[width=0.245\textwidth]{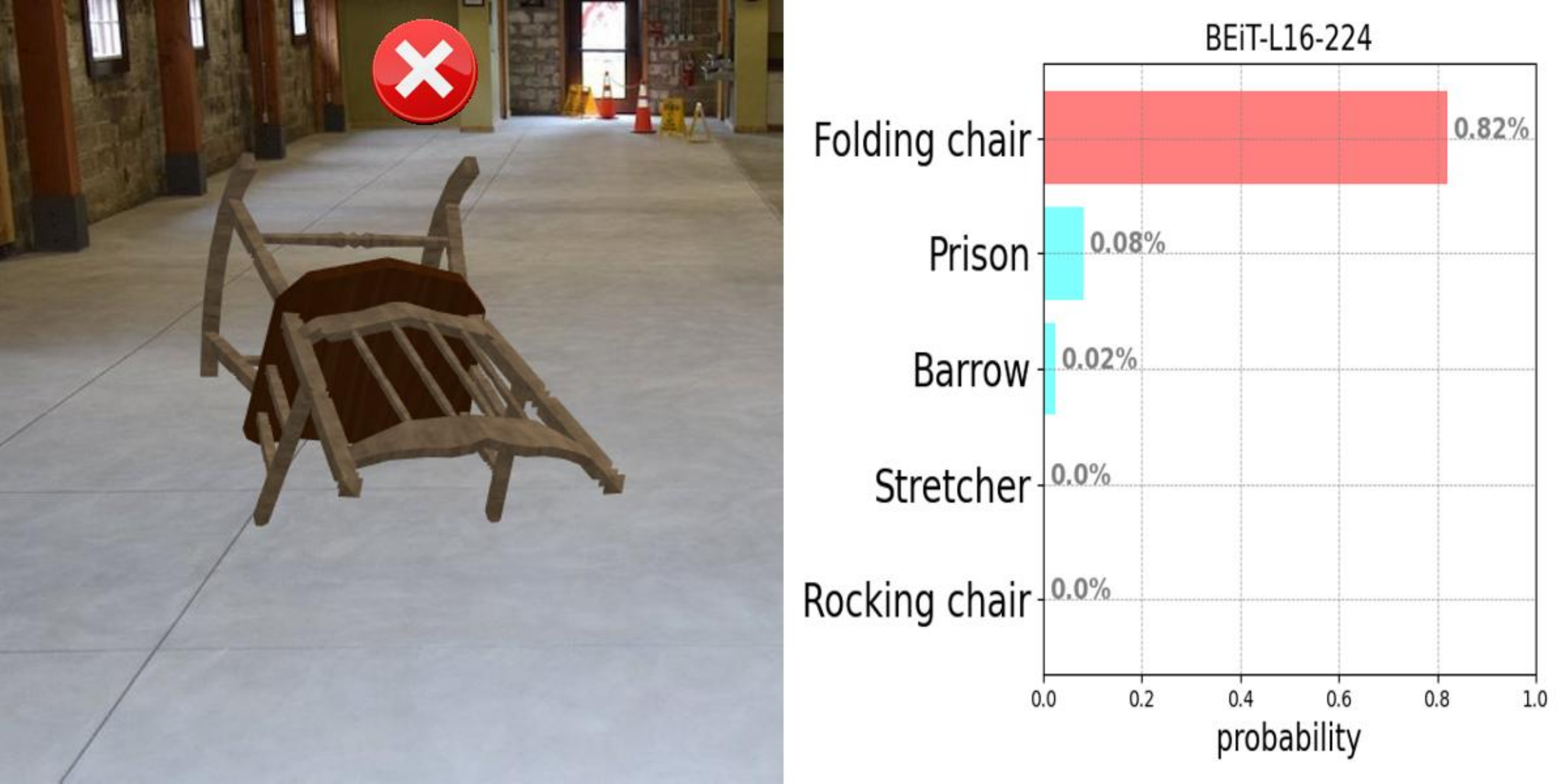}
    \includegraphics[width=0.245\textwidth]{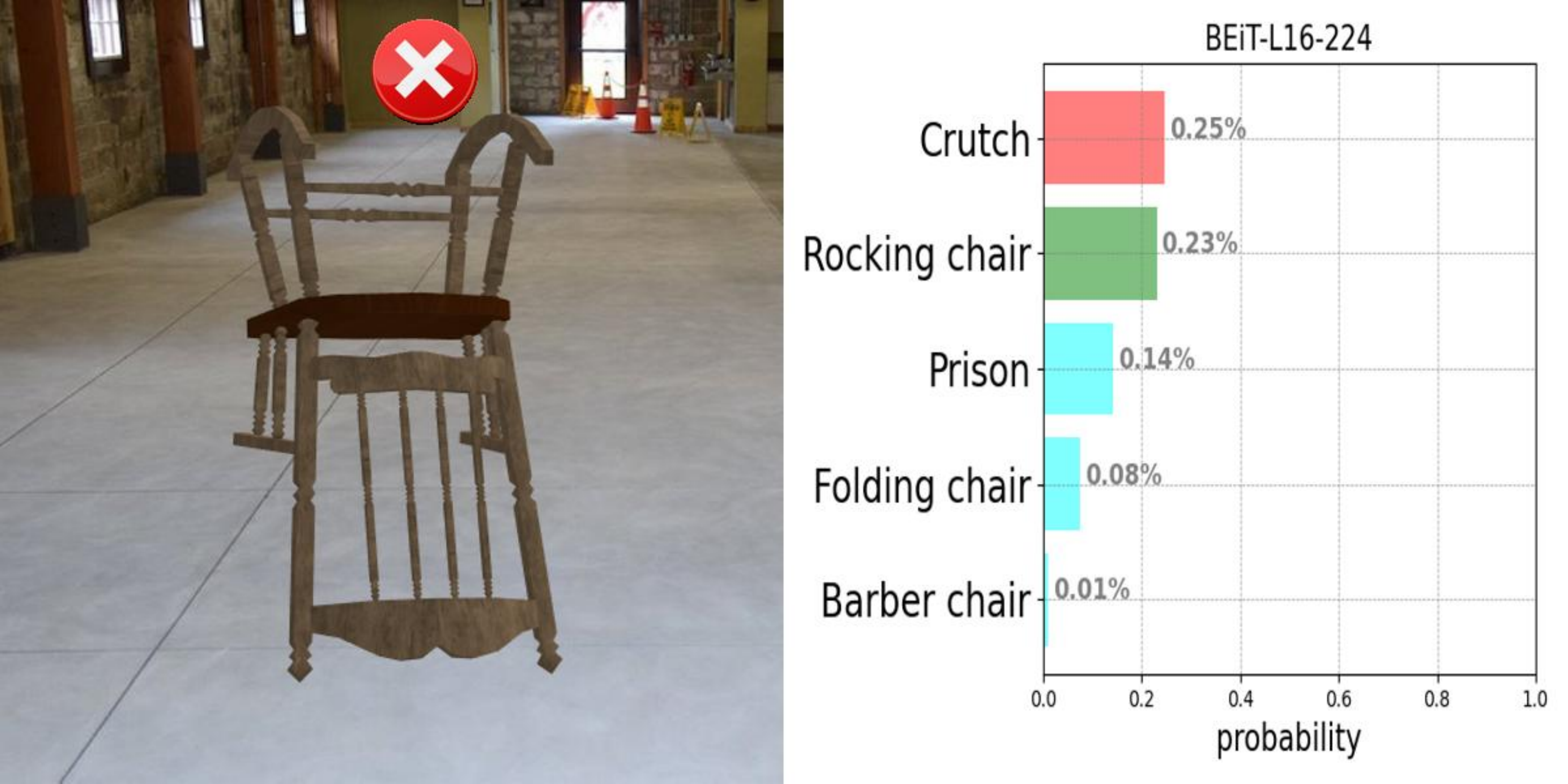}

    \includegraphics[width=0.245\textwidth]{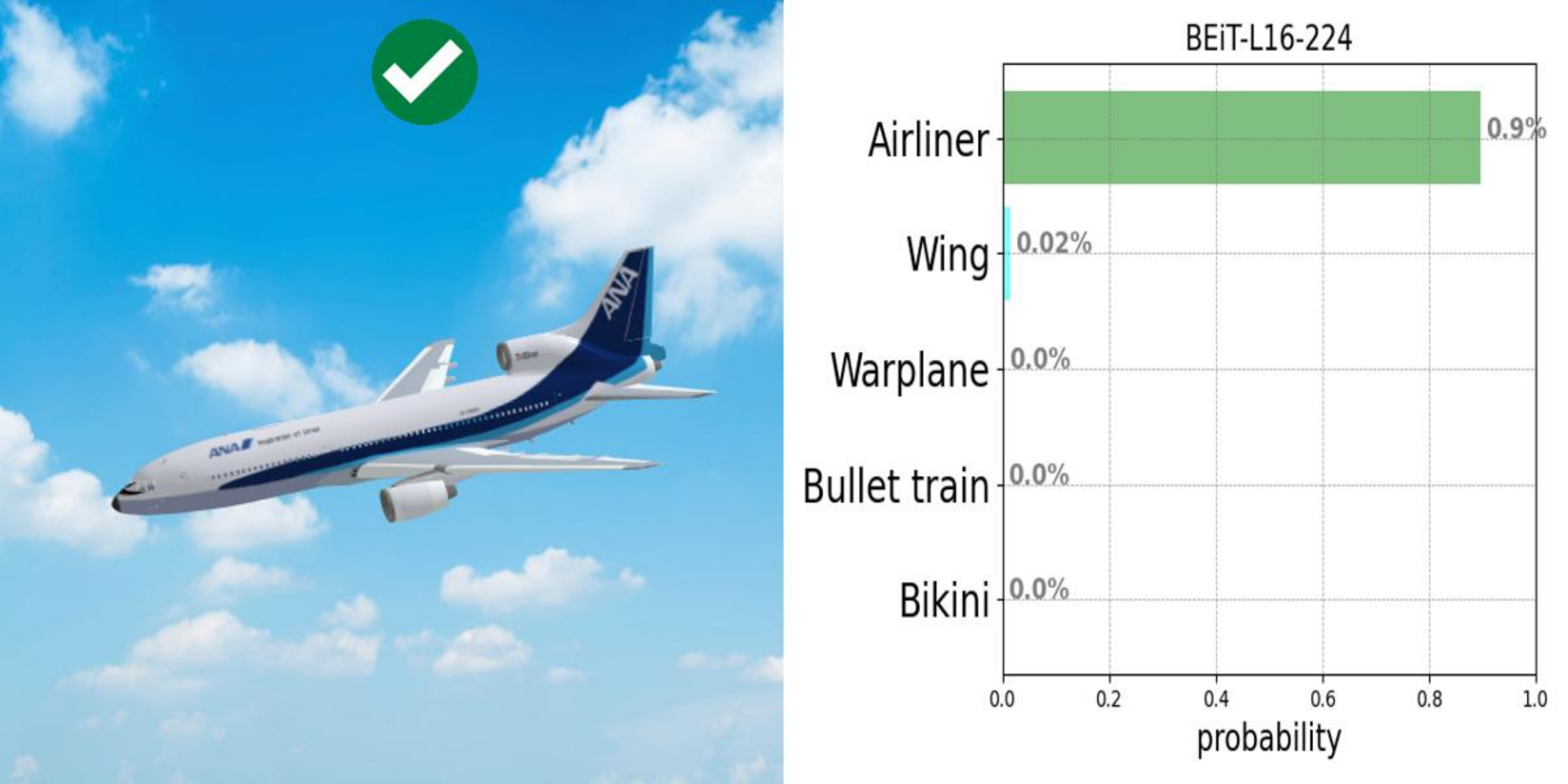}
    \includegraphics[width=0.245\textwidth]{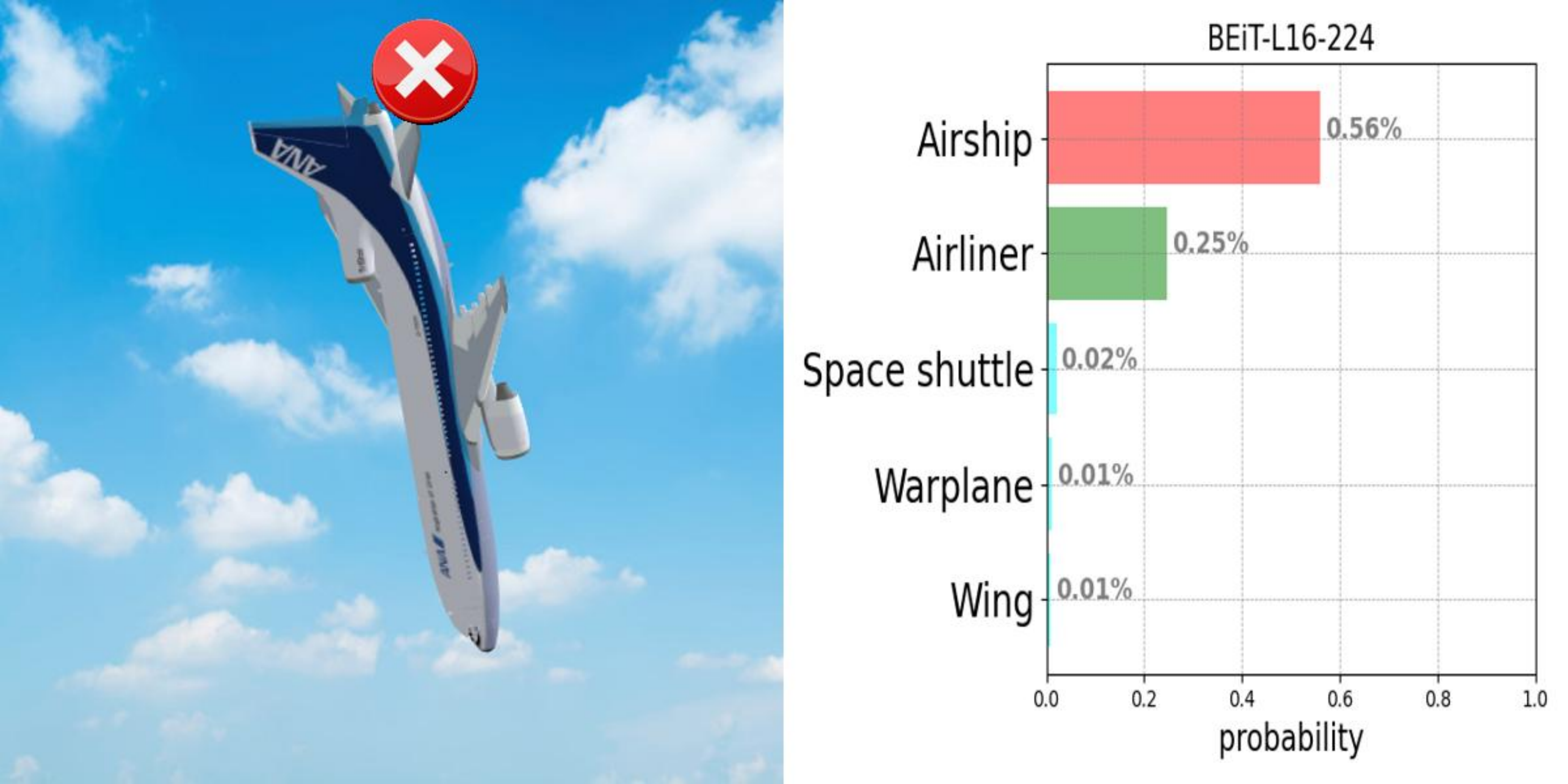}
    \includegraphics[width=0.245\textwidth]{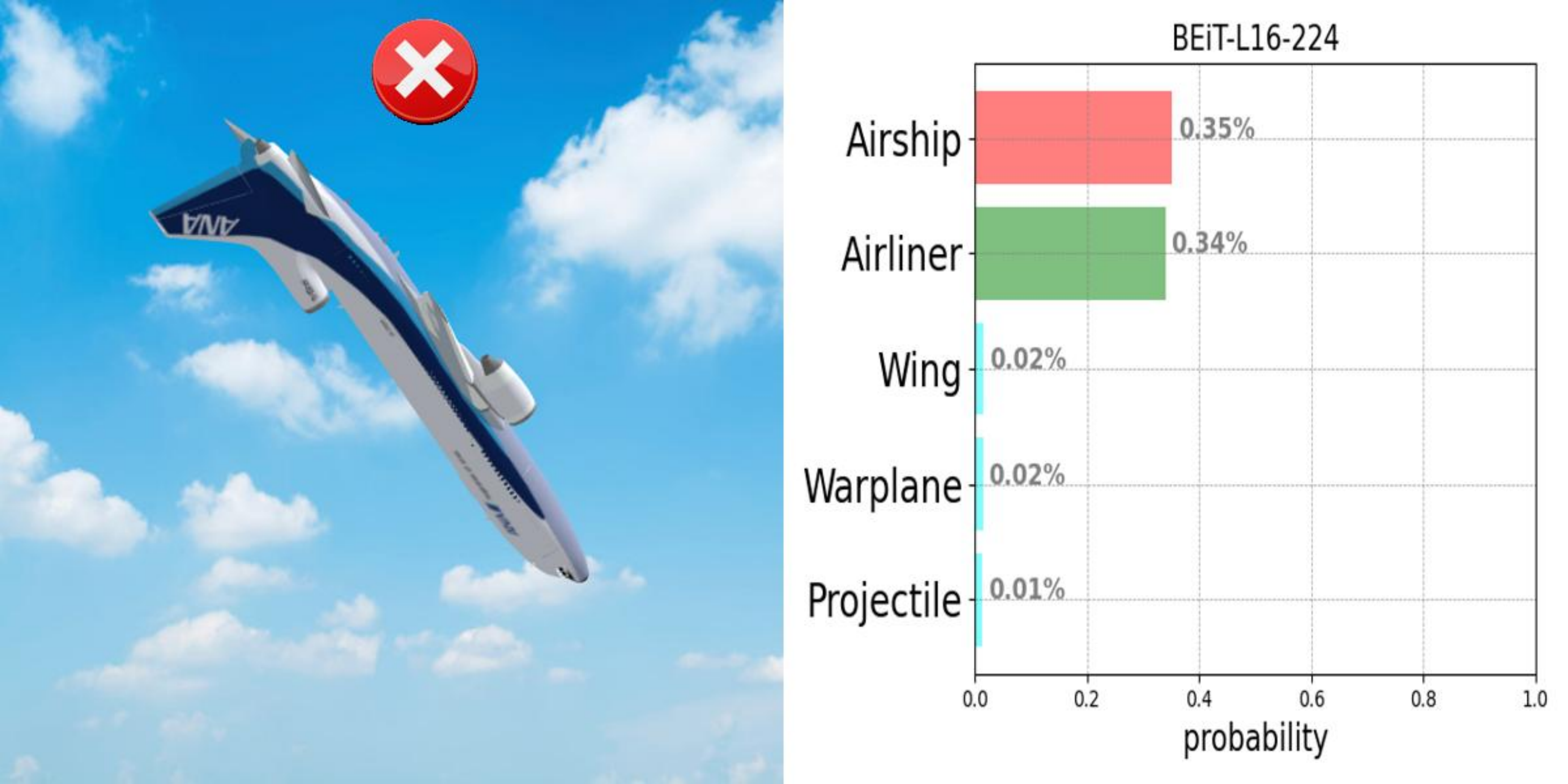}
    \includegraphics[width=0.245\textwidth]{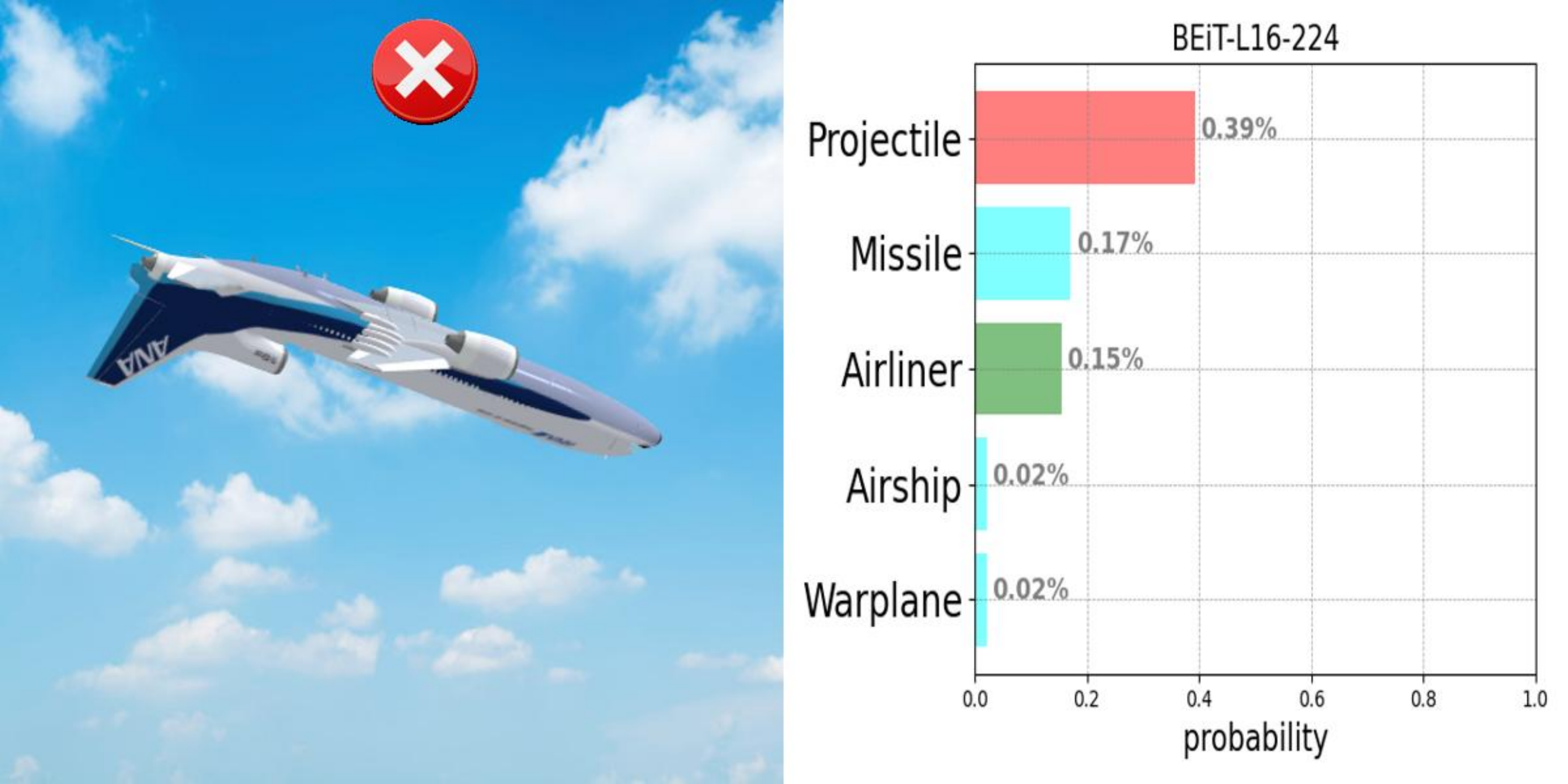}
 
    \caption{Selected errors made by BEiT-L/16 on the ObjectPose dataset.}
    \label{fig:beitfailureexamples}
\end{figure}

\subsection{Two-axes Rotation}
\begin{figure}[h]
    \centering
    \subfigure{\includegraphics[width=1\textwidth]{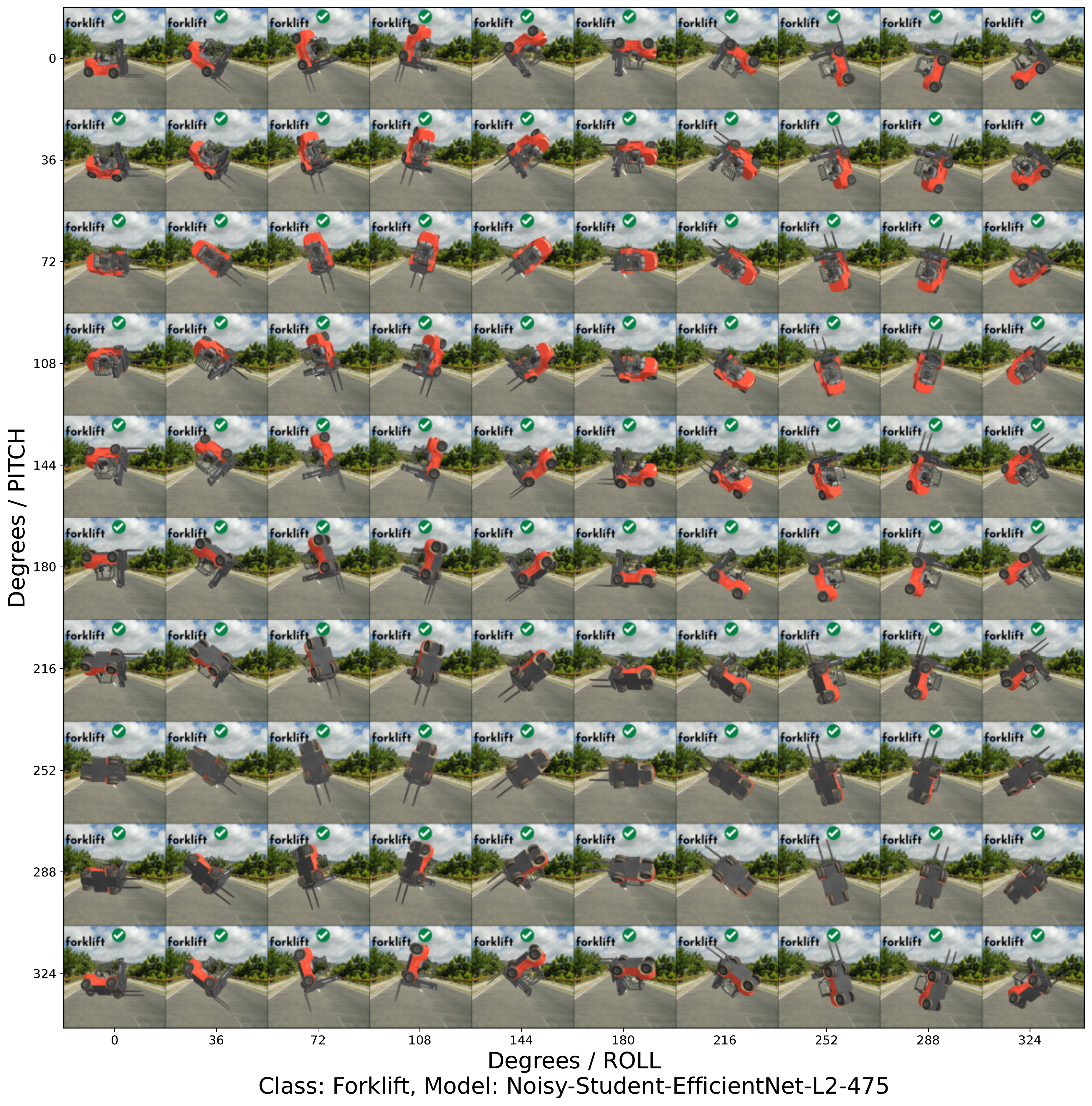}
    } 
    \caption{Noisy Student EfficientNet-L2 predictions on two-axes rotations of a forklift. Impressively, the model is able to classify all images perfectly.}
    \label{fig:engrid1}
\end{figure}

\begin{figure}[h]
    \centering
    \subfigure{\includegraphics[width=1\textwidth]{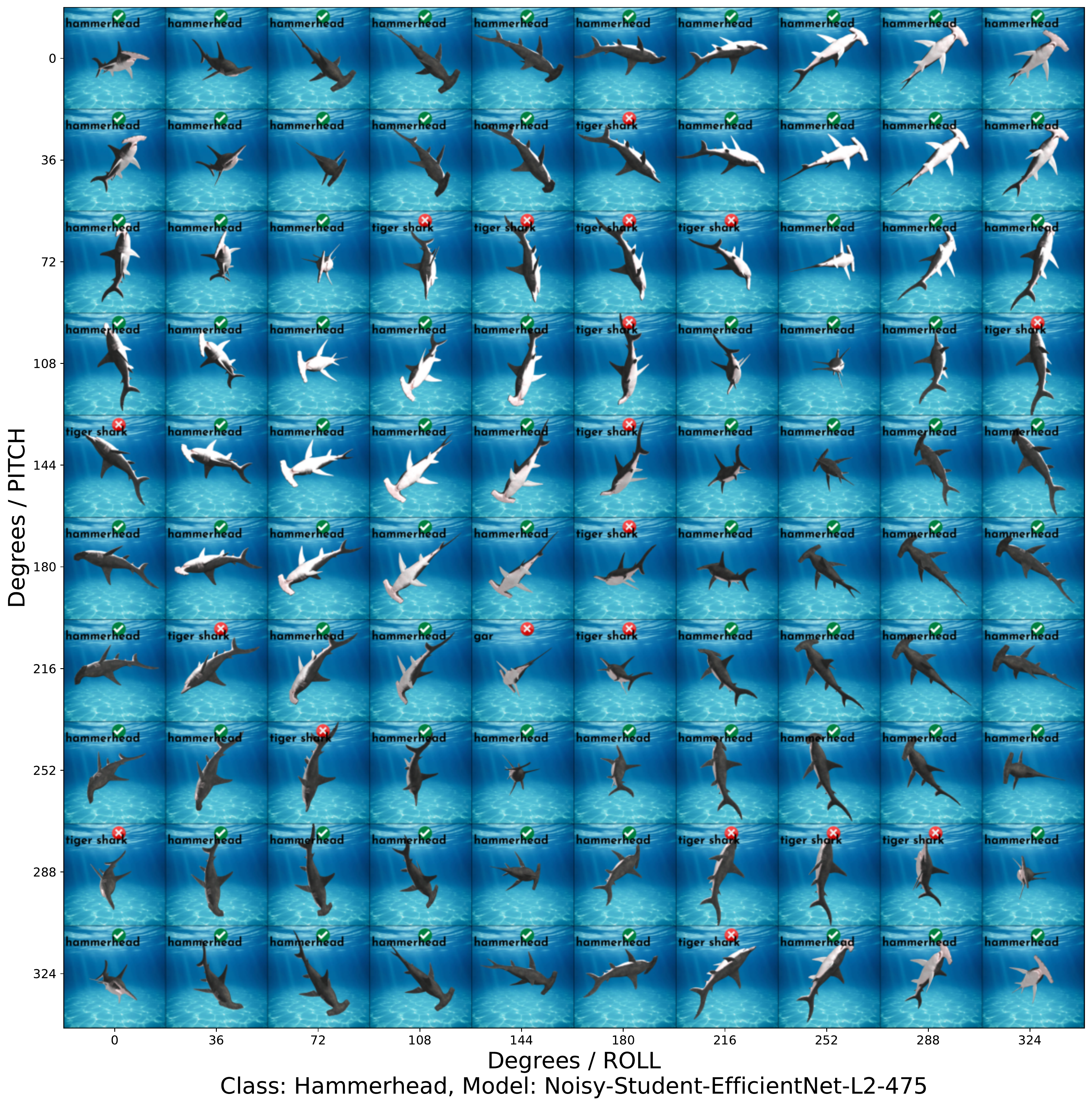}} 
    \caption{Noisy Student EfficientNet-L2 confuses a hammerhead shark with a tiger shark for poses where the hammerhead is invisible or barely visible. This is an error a human could easily do.}
    \label{fig:engrid2}
\end{figure}

\begin{figure}[h]
    \centering
    \subfigure{\includegraphics[width=1\textwidth]{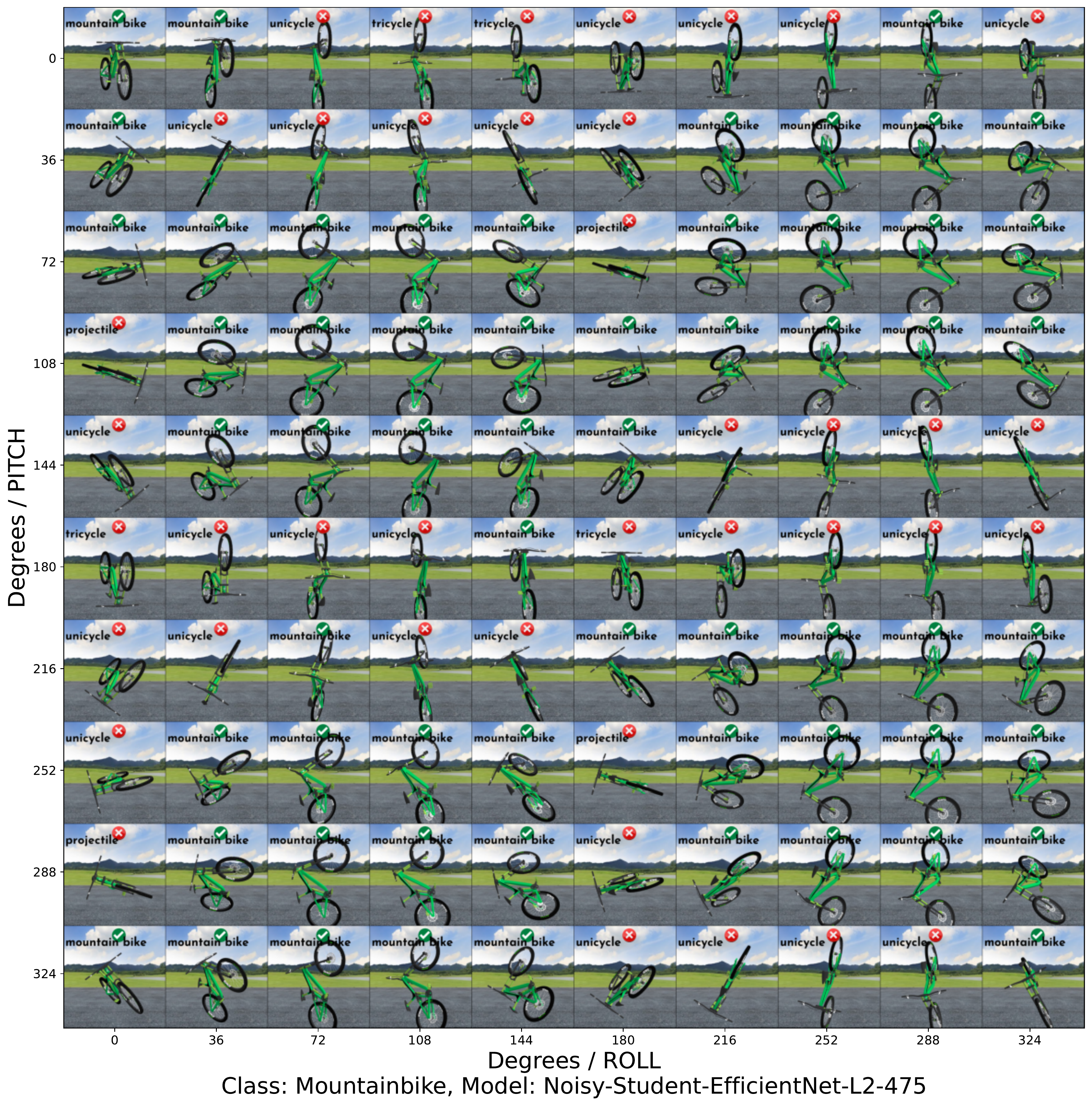}
    } 
    \caption{Noisy Student EfficientNet-L2 predictions on the two-axes rotation images of a mountain bike. The model confuses the bicycle with a tricyle or a unicycle for some unusual poses.}
    \label{fig:engrid3}
\end{figure}

\begin{figure}[h]
    \centering
    \subfigure{\includegraphics[width=1\textwidth]{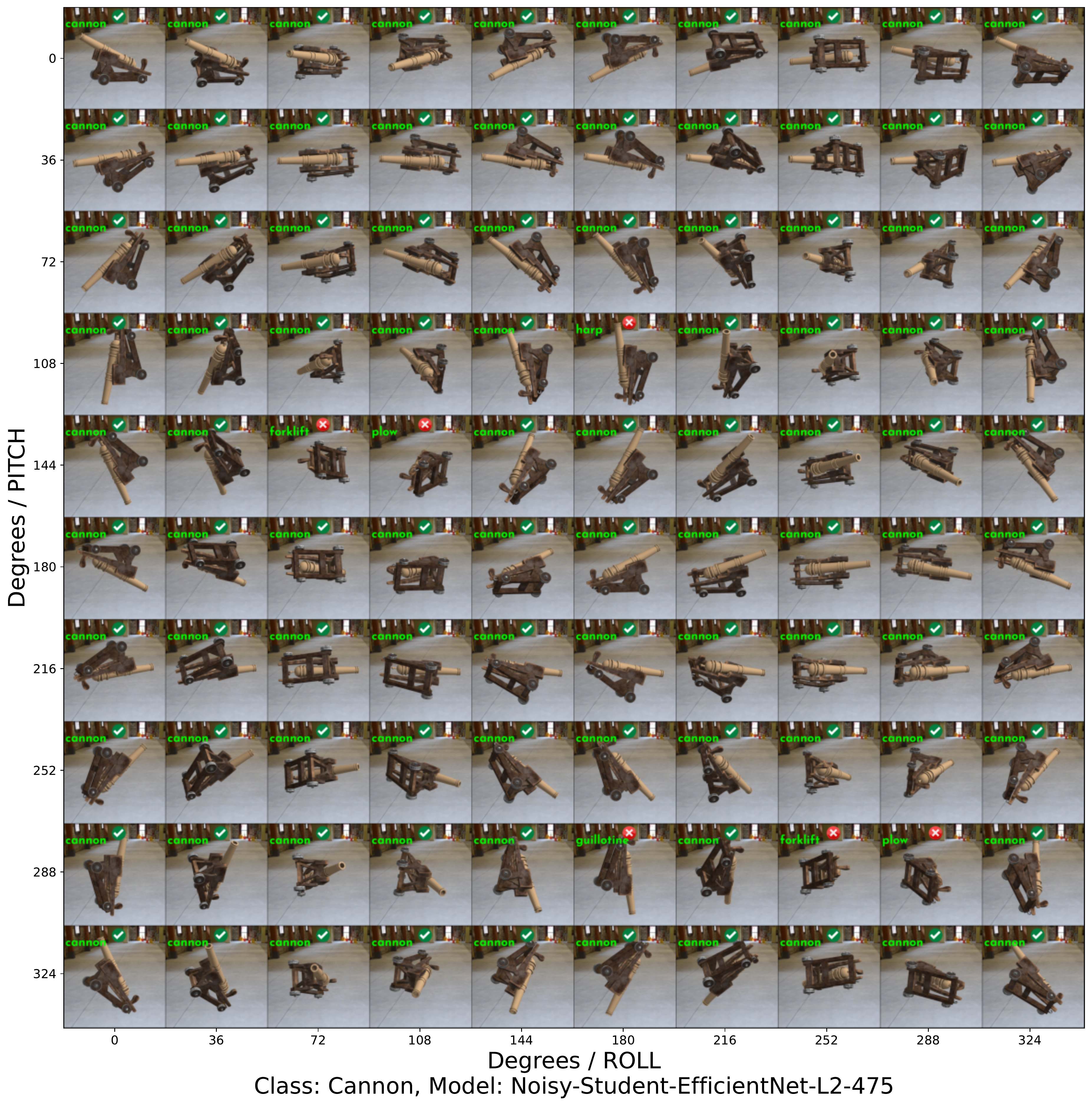}
    } 
    \caption{Noisy Student EfficientNet-L2 confuses a canon with a harp or a guillotine when it is at 90° from the upright position. This is an error that a human would probably not do.}
    \label{fig:engrid4}
\end{figure}

\begin{figure}[h]
    \subfigure{\includegraphics[width=1\textwidth]{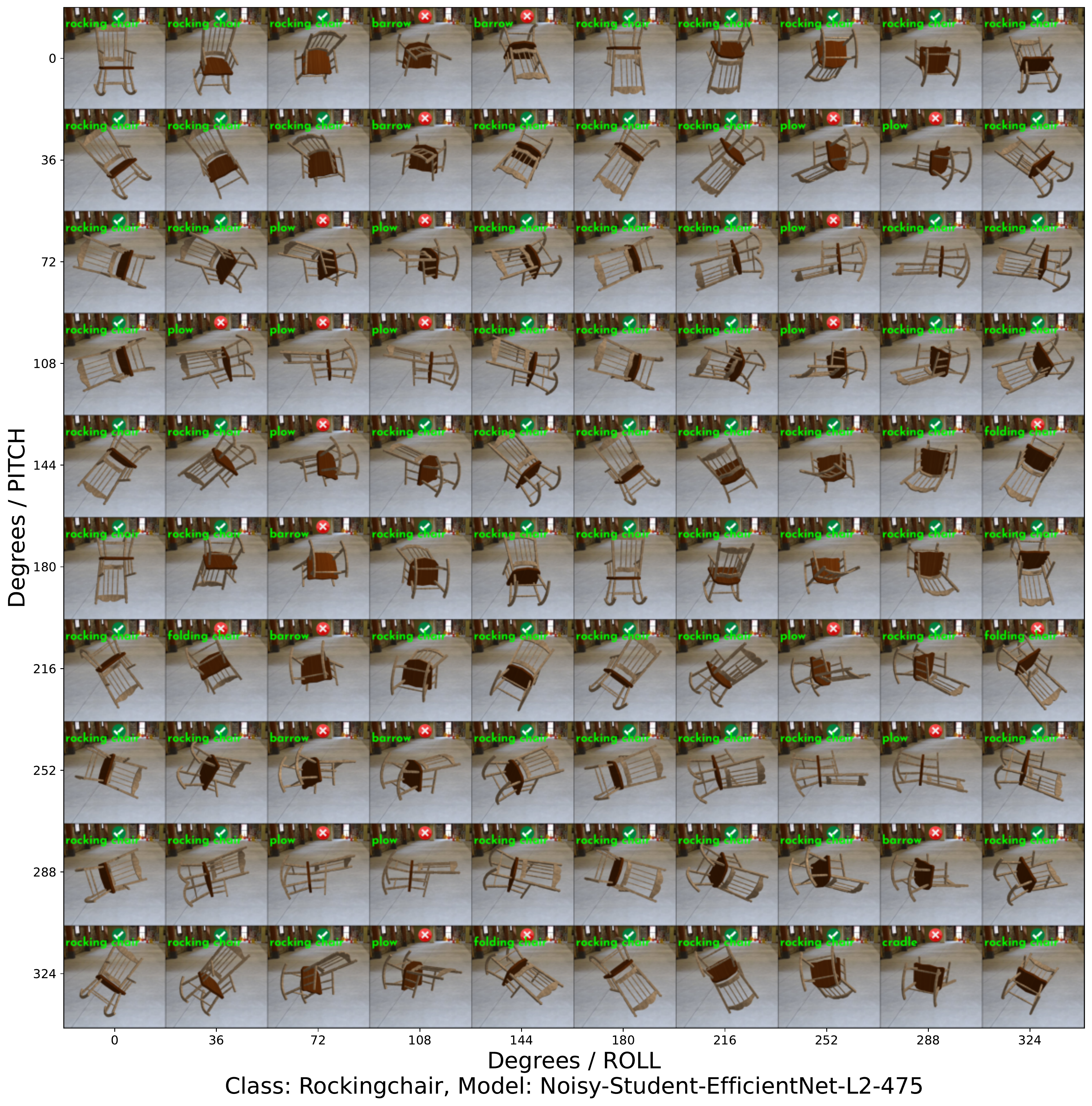}
    } 
    \caption{Noisy Student EfficientNet-L2 predictions on two-axes rotations of a rocking chair. The model makes some mistakes in classifying some of the rocking chair images.}
    \label{fig:engrid5}
\end{figure}

\begin{figure}[h]
    \centering
    \subfigure{\includegraphics[width=1\textwidth]{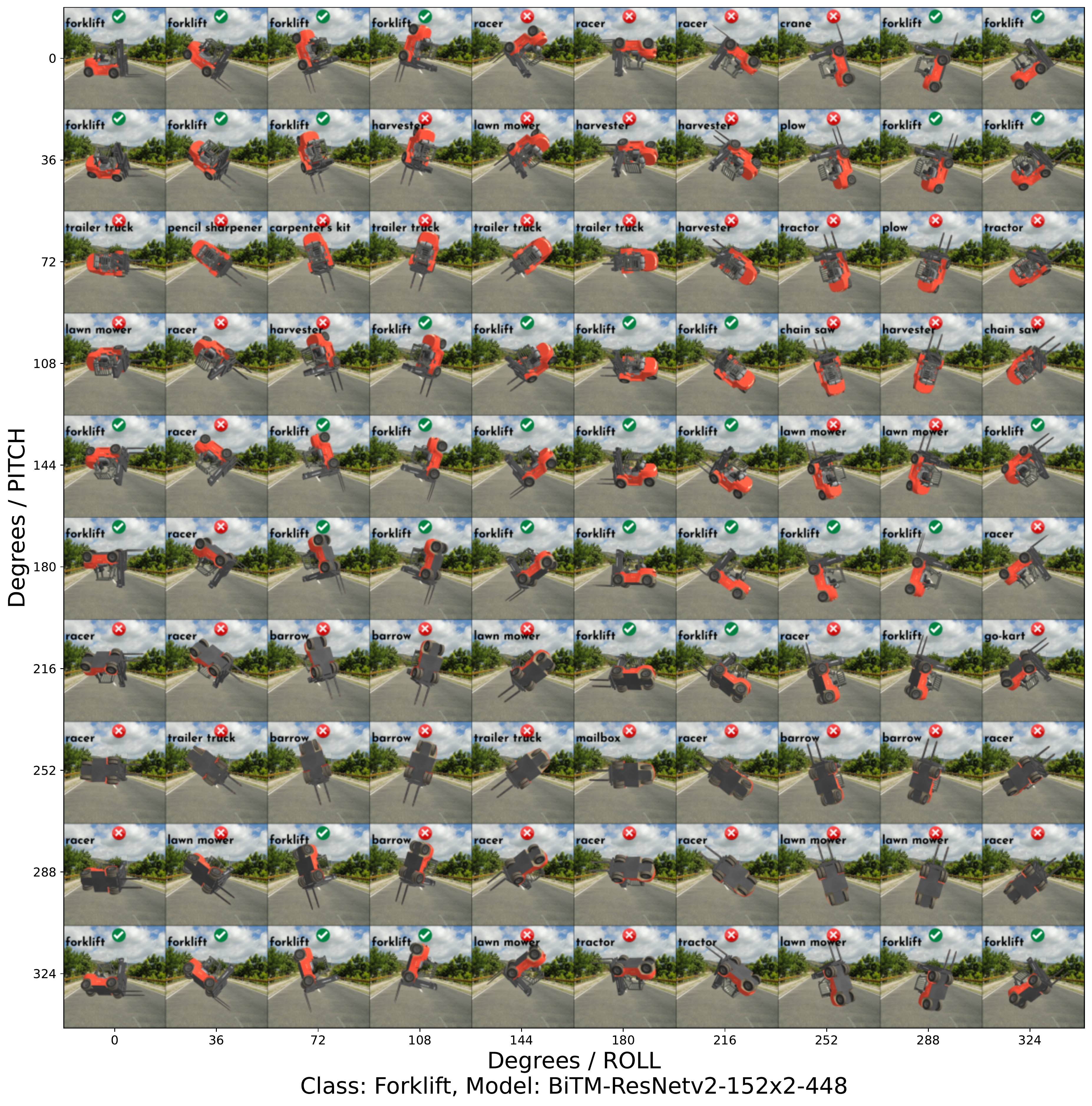}
    } 
    \caption{BiTM predictions on two-axes rotations of a forklift. This model is much less robust than larger models such as Noisy Student EfficientNet-L2.}
    \label{fig:bitgrid1}
\end{figure}

\begin{figure}[h]
    \centering
    \subfigure{\includegraphics[width=1\textwidth]{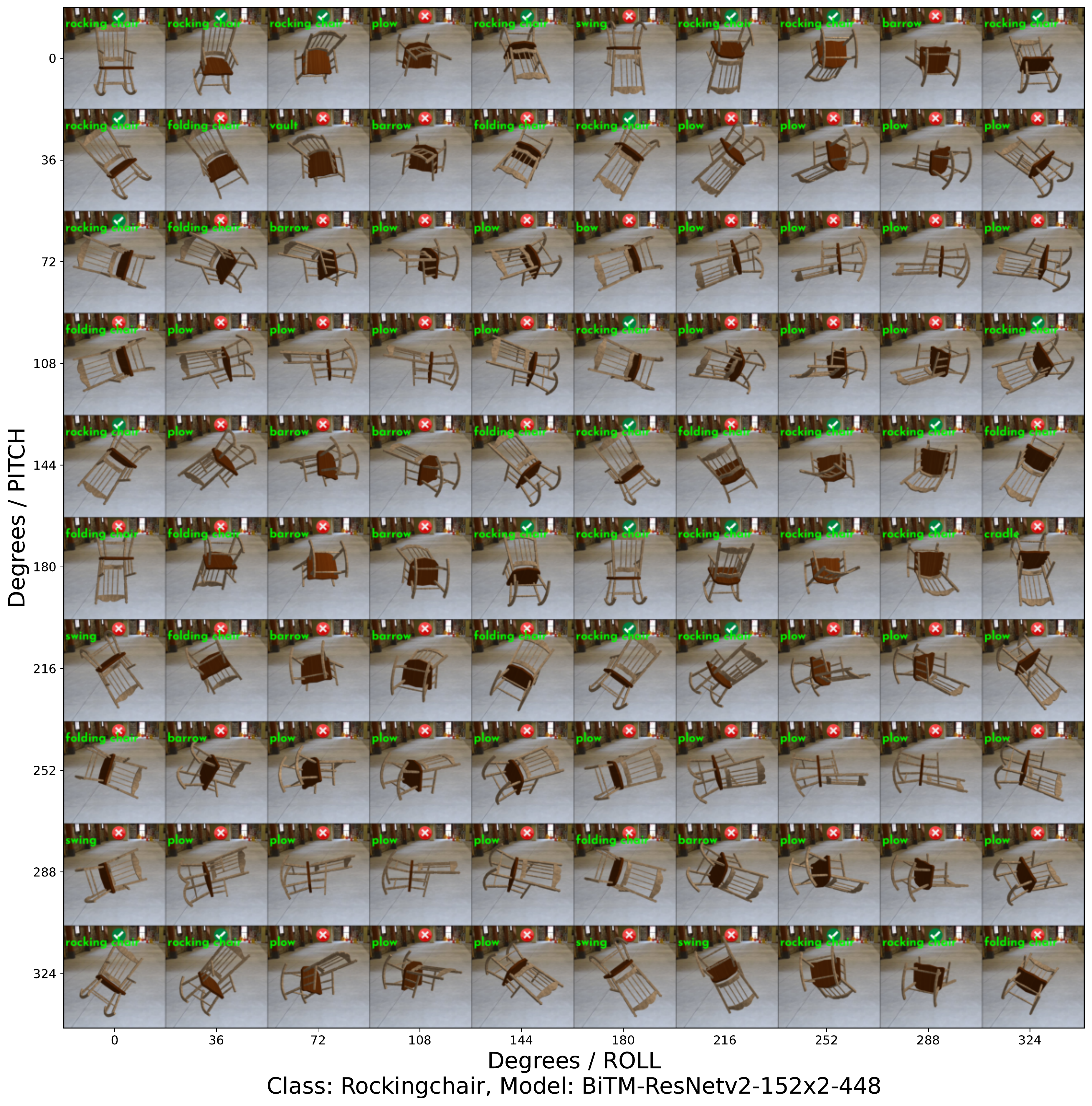}
    } 
    
    \caption{BiTM predictions on two-axes rotations of a rocking char. This model is much less robust than larger models such as Noisy Student EfficientNet-L2.}
    \label{fig:bitgrid2}
\end{figure}

To further assess the models' robustness to unusual poses we created a set of images by rotating the 3D object along two axes. We picked the two axes to be the $ROLL$ and $PITCH$ axes orthogonal to the 3D object. We rotate the object 10 times along each axis by increasing the rotation angle by 36° each time. This generates 100 images in total for each 3D object. Figures \ref{fig:engrid1}-\ref{fig:bitgrid2} shows the predictions on these 100 images by two models (EfficientNet and BiT) for some objects. Although these models achieve high in-distribution performance, they are still making many mistakes on images that seems easy for the human.

\end{document}